\documentclass[lettersize,journal]{IEEEtran}
\usepackage{amsmath,amsfonts}
\usepackage{amsthm}

\usepackage{algorithmic}
\usepackage{algorithm}
\usepackage{array}
\usepackage{textcomp}
\usepackage{stfloats}
\usepackage{url}
\usepackage{verbatim}
\usepackage{graphicx}
\usepackage{orcidlink}
\usepackage{booktabs}
\usepackage[caption=false,font=normalsize]{subfig}
\usepackage{enumitem}
\usepackage{float}
\usepackage{multirow}
\usepackage{tabularx}
\usepackage{siunitx}
\graphicspath{{./figs/}}
\usepackage{cite}

\begin{document}

\title{Behavioral Safety Assessment towards Large-scale Deployment of Autonomous Vehicles, Part II: Assessment Results}

\author{Henry X. Liu\textsuperscript{\#}\,\orcidlink{0000-0002-3685-9920},~\IEEEmembership{Fellow,~IEEE,}
        Tinghan Wang\textsuperscript{\#}\,\orcidlink{0000-0002-5660-8889},
        Xintao Yan\textsuperscript{\#}\,\orcidlink{0000-0002-0569-5628},
        Haowei Sun\textsuperscript{\#}\,\orcidlink{0000-0002-0232-117X},
        Zhijie Qiao,
        Kenneth Boyd,
        Shuo Feng\,\orcidlink{0000-0002-2117-4427},
        Greg Stevens,
        Greg McGuire
\thanks{Henry X. Liu (e-mail: henryliu@umich.edu), Kenneth Boyd (e-mail: kjboyd@umich.edu), Greg Stevens (e-mail: gregstvn@umich.edu), and Greg McGuire (e-mail: mcguireg@umich.edu) are with the University of Michigan Transportation Research Institute, Ann Arbor, MI 48109 USA.\par
Henry X. Liu, Tinghan Wang (e-mail: tinghanw@umich.edu), Zhijie Qiao (e-mail: zhijieq@umich.edu) are with the Department of Civil and Environmental Engineering, University of Michigan, Ann Arbor, MI 48109 USA.\par
Xintao Yan (e-mail: xintaoy@hku.hk) is with the Department of Civil Engineering, The University of Hong Kong, Hong Kong 999077, China.\par
Haowei Sun (e-mail: howie@inchor.ai) is with Laplace Intelligence, Ann Arbor, MI 48109 USA.\par
Shuo Feng (e-mail: fshuo@tsinghua.edu.cn) is with the Department of Automation, Tsinghua University, Beijing 100084, China.\par
Corresponding author: Henry X. Liu (e-mail: henryliu@umich.edu).\par
\# Equal Contribution.}%
}

\maketitle

\begin{abstract}
Third-party evaluations of autonomous vehicle (AV) safety can play a vital role in improving public acceptance, building consumer confidence, and establishing effective safety standards. In Part I of this study~\cite{Liu2026partI}, we propose a dedicated third-party testing initiative for systematically evaluating AV behavioral safety. In this paper, we validate our proposed framework using Autoware.Universe, an open-source Level 4 Automated Driving System (ADS), tested both in simulated environments and on the physical test track at the University of Michigan's Mcity Testing Facility\footnote{\url{https://mcity.umich.edu/}}. The results indicate that Autoware.Universe possesses 6 out of 14 behavioral competencies and exhibited a crash rate of $3.01\times 10^{-3}$ crashes per mile, approximately 1,000 times higher than the average human driver crash rate. During the tests, we also uncovered a number of unknown unsafe scenarios for Autoware.Universe. These findings underscore the necessity of behavioral safety evaluations for improving AV safety performance prior to widespread public deployment.
\end{abstract}


\section{Introduction}
\IEEEPARstart{T}{hird-party} evaluations of autonomous vehicle (AV) safety are essential for improving public acceptance, gaining consumer confidence, and establishing effective safety standards.
The existing well-established testing programs for conventional vehicles, such as NCAP~\cite{nhtsa_ncap} and IIHS~\cite{iihs_ratings}, and notable standards, including ISO 26262~\cite{iso26262}, ISO 21448~\cite{iso21448}, ISO/PAS 8800~\cite{iso8800}, ISO 34502~\cite{iso34502}, and UL 4600~\cite{ul4600}, are inadequate for Level 4--5 AVs~\cite{sae2021j3016} for two key reasons: (1) they adopt a vehicle-centric perspective focused on passive and active safety features or system-level malfunctions, rather than evaluating an AV's holistic driving behavior within complex traffic environments, and (2) they rely on deterministic assessment of predetermined scenarios, whereas modern AVs exhibit inherently non-deterministic behavior due to their reliance on deep learning models, necessitating probabilistic evaluation. These limitations underscore the necessity of shifting the AV evaluation paradigm from functional safety to behavioral safety, which evaluates whether a functionally sufficient AV can make safe decisions in reacting to and interacting with the traffic environment, without creating or exacerbating safety-critical situations.

While the need for behavioral safety evaluation is clear, conducting third-party evaluations presents significant challenges. First, modern AVs operate as black-box systems with limited interpretability, making failures unpredictable and necessitating holistic evaluation of basic behavioral competency rather than just extreme-case testing. Second, the naturalistic driving environment involves intricate spatial-temporal interactions among diverse road users, across varying road geometries, and under dynamic traffic conditions, making AV assessment highly complex. Third, safety-critical events occur with extreme rarity in real-world traffic, requiring billions of miles for meaningful evaluation~\cite{feng2023dense} and rendering brute-force mileage-based testing impractical even in simulation.

\IEEEpubidadjcol

To overcome these challenges, in Part I of this study~\cite{Liu2026partI}, we propose a dedicated third-party testing initiative for systematically evaluating AV behavioral safety.
The framework consists of two key components: the Behavioral Competency Test (BCT) and the Driving Intelligence Test (DIT).
The BCT evaluates AV performance in predefined scenarios of varying risk levels, akin to human driver's license tests, to verify basic behavioral competency in both normal and safety-critical situations.
The DIT goes beyond known scenarios by constructing a Naturalistic and Adversarial Driving Environment (NADE) that leverages importance sampling to efficiently uncover unknown failure modes and statistically measure safety-critical event rates without bias. Together, the BCT ensures baseline competency in expected driving situations, while the DIT reveals long-tail risks that may only emerge at scale.

Part I establishes the evaluation pipeline and methodological foundation. In this paper (Part II), we first present a systematic evaluation of an open-source Level 4 Automated Driving System (ADS), demonstrating how the proposed framework is applied. We then report quantitative evaluation results, including pass/fail outcomes of the BCT across multiple behavioral competencies and unbiased statistical metric estimates from the DIT. Finally, we present a field test conducted on a real test track, validating the practical applicability of the framework beyond simulation.

More specifically, to demonstrate the effectiveness of our approach, we applied it to evaluate a Level 4 AV system running open-source automated driving software, Autoware.Universe~\cite{autoware}, in a simulated testing environment at the University of Michigan’s Mcity Test Facility.
We assessed the AV’s performance within an urban driving ODD, including intersections, roundabouts, and urban arterials, under normal weather conditions. High-fidelity testing scenarios and environments were constructed using large-scale real-world datasets that model human driving behavior.
For the BCT, we selected 14 representative behavioral competencies, comprising both single-agent and multi-agent scenarios, that span all major collision configurations (head-on, angle, rear-end, and sideswipe crashes, as well as crashes involving pedestrians), which together account for 63.3\% of all reported crashes in the study area~\cite{wats2023crash}.
The DIT was developed based on state-of-the-art intelligent testing methodologies~\cite{feng2021intelligent} to obtain statistical performance of the AV accurately and efficiently.

Simulation results reveal that Autoware.Universe, the latest version of Autoware, possesses 6 out of 14 behavioral competencies in the BCT.
The DIT estimated a crash rate of $3.01\times 10^{-3}$ crashes per mile, about 1,000 times higher than that of an average human driver~\cite{mtcf2024} ($3.00\times 10^{-6}$ crash per mile).
These testing results successfully identified unknown unsafe events, exposing potential weaknesses and bugs in the AV’s software stack, such as its failure to navigate certain roundabouts, leading to unexpected stops.
This issue was reported to the developers, who were able to address and resolve it based on our feedback.
These findings highlight the framework’s effectiveness in detecting and diagnosing system vulnerabilities, particularly those unknown unsafe behaviors, which are essential for refining AV systems and supporting a closed-loop development process for large-scale deployment.

To further validate the applicability of our framework in real-world settings, we conducted physical tests using Mcity’s closed test tracks with a Tesla Model 3 (equipped with Full Self-Driving Version 12.5.5) and a prototyping vehicle equipped with Autoware.Universe on a Lincoln MKZ platform.
Due to the resource constraints, the goal was not comprehensive testing but to demonstrate feasibility.
For the BCT, we evaluated two behavioral competencies with the Tesla, one involving a vehicle and the other involving a vulnerable road user (VRU), both of which were successfully passed across multiple risk levels.
For the DIT, testing with the Autoware-equipped Lincoln MKZ revealed that the system could benefit from further improvements to proactively reduce risk and limit exposure to safety-critical situations.
These results highlight the framework’s versatility and effectiveness in evaluating AV systems across both simulated and physical environments, reinforcing its potential as a systematic, standardized, and third-party safety validation tool for AV deployment.

\section{Experimental Setup of the Behavioral Safety Assessment}

This section introduces the experimental setup, including the development process of BCT and DIT, and the AV under test.
Consistent with Part I~\cite{Liu2026partI}, scenarios in the NDE and NADE are referred to as ``environments'' throughout this paper.

\subsection{Development process of the Behavioral Competency Test}

\begin{figure*}[t]
  \centering
  \includegraphics[width=\textwidth]{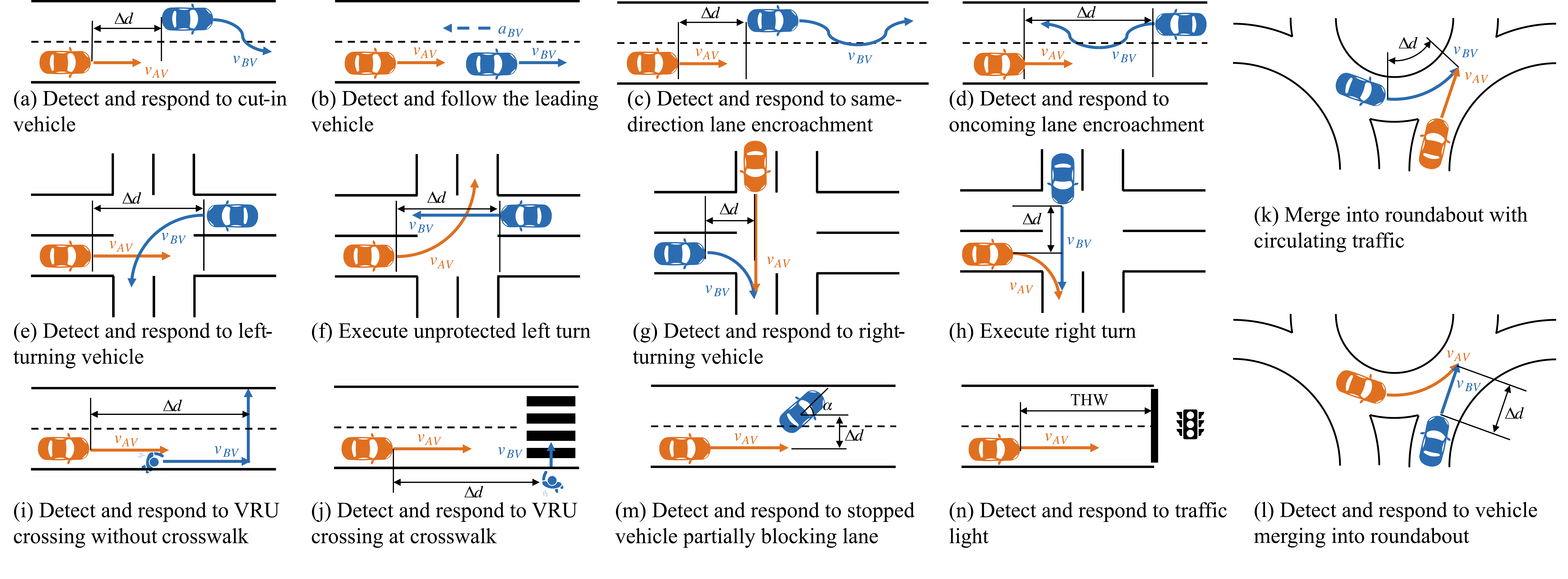}
  \caption{Sketches of the behavioral competencies tested in this study}
  \label{fig: behavioral competency sketch}
\end{figure*}

This section details the development process of the BCT. The initial step involves abstracting basic behavioral competencies and building a testing behavioral competency library grounded in real-world dynamic driving tasks.
Several existing studies, standards, and best practices contribute to the creation of the libraries~\cite{swanson2019statistics, degelder2020scenario, thorn2018framework, waymo2021safety}.
Based on this foundation, we compiled our testing competency set, considering the typical driving scenarios within the urban ODD.
Specifically, we leveraged the scenario library from Ref.~\cite{peng2019mcity} to guide our selection and curated 14 representative behavioral competencies, as shown in Fig.~\ref{fig: behavioral competency sketch}, where the orange vehicle represents the AV under test, the blue vehicle represents the BV, and the blue human figure represents the VRU.
The selected competencies encompass both single-agent situations, such as detecting and responding to the traffic light, as well as multi-agent situations featuring various road user types.
Notably, these competencies cover every multi-vehicle collision configuration recorded in local crash statistics (head-on, angle or left-turn, rear-end, and sideswipe), as well as crashes involving pedestrians. Together, these crash types account for 63.3\% of all crashes reported in Washtenaw County, Michigan, in 2021~\cite{wats2023crash}, offering reasonable coverage.
While the selected 14 competencies may not fully represent the entirety of the urban ODD, they were chosen to demonstrate our proposed methodology within the constraints of resources.
The methodology is designed to be flexible, allowing for the addition of more scenarios as needed.

\begin{figure*}[t]
  \centering
  \includegraphics[width=0.8\textwidth]{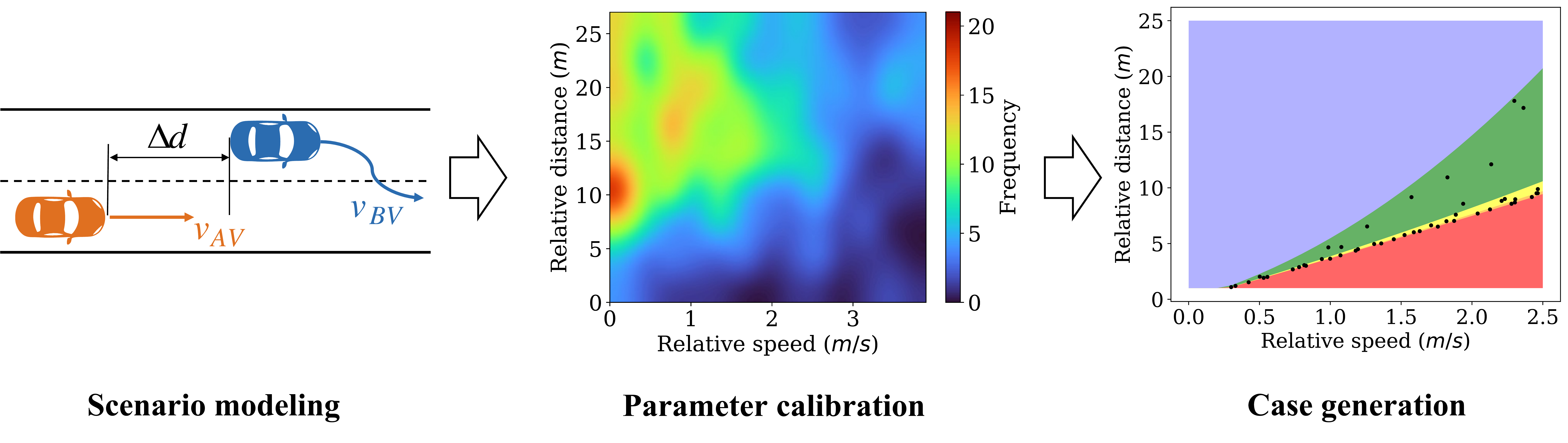}
  \caption{Demonstration of the cut-in scenario development pipeline}
  \label{fig: bct cut in development pipeline}
\end{figure*}

Each behavioral competency follows a three-step development process: scenario modeling (parameterization in state space), parameter calibration (constructing the parameter’s NDE distribution), and case generation (building the NADE distribution for test case sampling).
As shown in Fig.~\ref{fig: bct cut in development pipeline}, we demonstrate the three-step process using behavioral competency (a), which is evaluated in cut-in scenario.
Since the AV is treated as a black box, the tester cannot control its speed during testing.
Therefore, for a cut-in scenario, the controllable variables are the relative speed and the relative longitudinal distance between the AV and the BV, and the BV will execute lane changes at a constant speed, following a sine-function trajectory.
As a result, it is modeled as a two-dimensional problem.
The constraints include the initial relative speed and the longitudinal distance between AV and BV equal to the given value, which means these two parameters uniquely determine the test case.

In the parameter calibration step, we aim to derive real-world distributions of constraint parameters.
Using the Argoverse 2 Motion Forecasting Dataset~\cite{wilson2021argoverse2}, we extracted 11701 cut-in segments and developed the NDE distribution for this scenario, as shown in Fig.~\ref{fig: bct cut in development pipeline} (middle).
Rather than randomly sampling from the naturalistic distribution, we balance coverage and difficulty in the case generation process to account for different risk levels.
For each test case, the risk is calculated based on the minimum deceleration required by the AV to avoid a collision.
By referencing human drivers' deceleration distributions in cut-in events from the NDD, we categorize the scenario state space into five risk levels (trivial, low, medium, high, and infeasible) based on the required deceleration.
These levels are color-coded in Fig.~\ref{fig: bct cut in development pipeline} (right): blue (trivial), green (low), yellow (medium), orange (high), and red (infeasible), corresponding to different percentiles of human deceleration: above 80\%, 80\%-10\%, 10\%-1\%, below 1\%, and beyond observed human capabilities.
For example, if the required deceleration falls between the 1st and 10th percentiles of the cut-in event data, the test case is categorized as medium-risk.
We will generate test cases from low-, medium-, and high-risk levels, with a higher sampling density in high-risk regions, and 95\% confidence level.
Fig.~\ref{fig: bct cut in development pipeline} (right) illustrates the sampling results, where each black dot represents a test case.
The number of generated test cases for all scenarios is presented in the second column of Table~\ref{tab: bct overall results}.
More development and implementation details of different scenarios can be found in Appendix~\ref{sec:bct}.
Each test case is run 3 times to account for stochastic variability in the AV and testing systems.
A test case is considered a failure if a collision occurs in any run, and a success if no collisions occur.
By analyzing the results across different scenarios, we can assess the AV's basic behavioral competency.
The AV is deemed to have failed the BCT if it encounters a failure in any of the testing scenarios.
An illustration video of the BCT is provided in Supplementary Movie 1.

\subsection{Development process of the Driving Intelligence Test}

\begin{figure*}[ht]
    \centering
    \subfloat[Validation of the safety-critical events statistics\label{subfig: dit statistics}]{%
        \includegraphics[width=.7\textwidth]{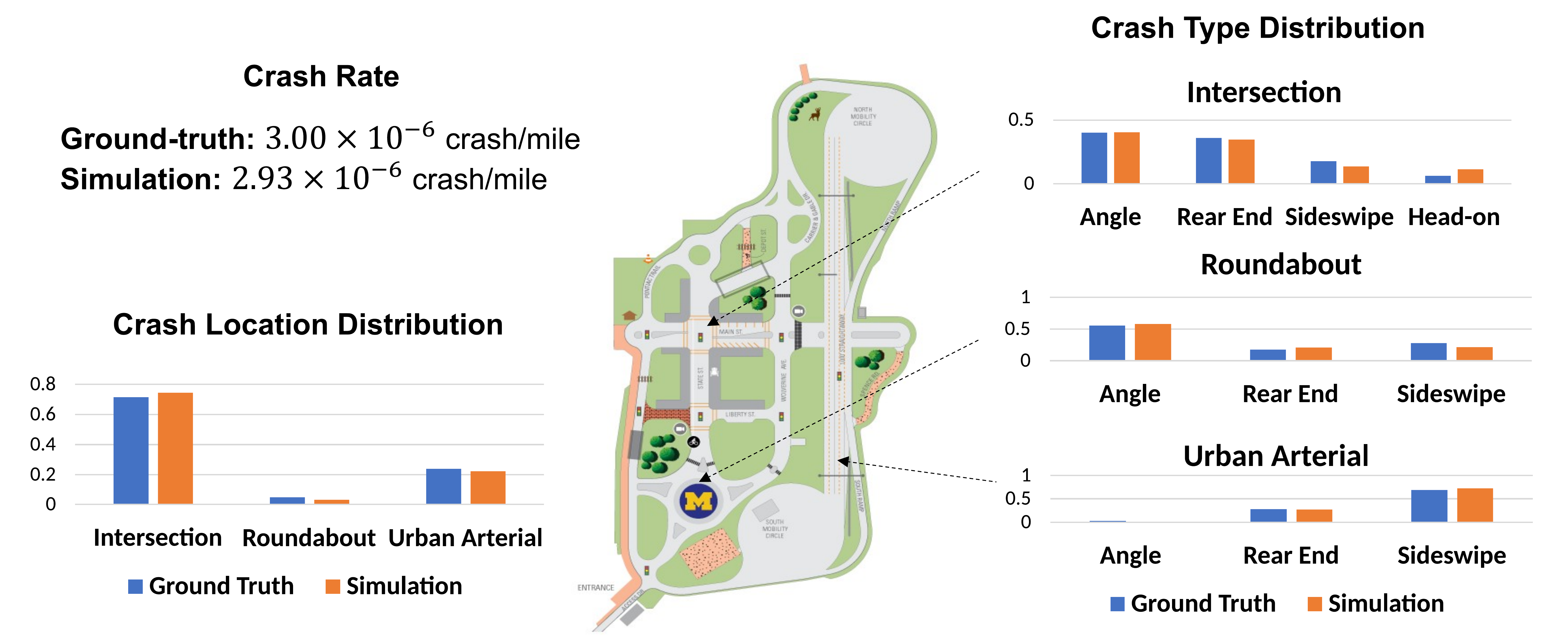}}
    \hfill
    \subfloat[Illustration of the normal driving events statistics\label{subfig: ndd statistics}]{%
        \includegraphics[width=.7\textwidth]{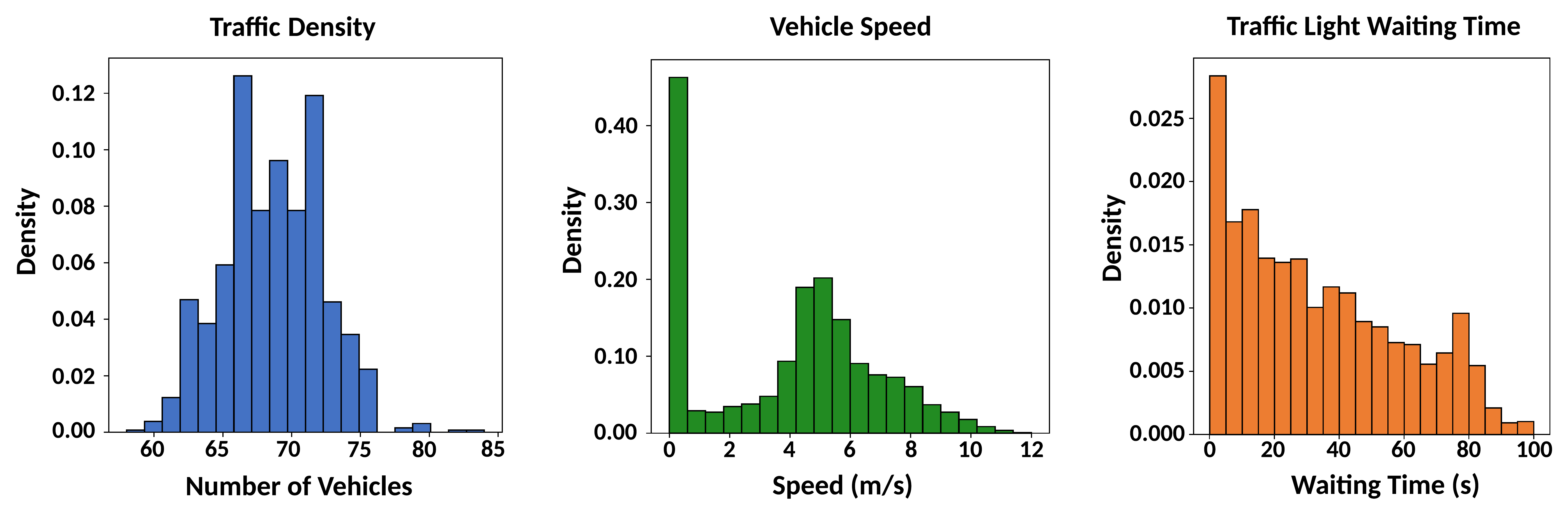}}
    \caption{Demonstration of the simulated Mcity environment for the Driving Intelligence Test}
    \label{fig: dit development}
\end{figure*}

This section details the development process of DIT.
Unlike the BCT, which models individual scenarios independently, the DIT begins by constructing a complete traffic environment through high-fidelity simulation of human driving behavior.
This simulated NDE encompasses all relevant components of the AV's ODD, such as intersections and roundabouts, when evaluating performance in urban settings.
The DIT enables the evaluation of an AV’s capabilities over entire trips.
To achieve this, we have developed TeraSim~\cite{sun2025terasim}, an open-source, high-fidelity traffic simulation platform designed specifically for AV safety testing.
TeraSim employs a generative paradigm to construct traffic environments using two core models. The NDE~\cite{yan2023learning, yan2025distributionally-nde} reconstructs realistic traffic conditions with statistical fidelity, ensuring that generated environments follow real-world distributions, particularly for those safety-critical events.
Building on NDE, the NADE~\cite{feng2023dense, feng2021intelligent} amplifies rare but critical events, systematically exposing AVs to safety-critical conditions to accelerate the evaluation of AV safety performance.

Our method enables high-fidelity simulation of the Mcity environment, which includes a 1,000-foot urban arterial, six signalized intersections (one with an unprotected left turn), three roundabouts, and various single and multi-lane roads, effectively replicating a realistic urban driving context.
A key requirement for AV testing in the NDE is the accurate reproduction of safety-critical events based on real-world data~\cite{yan2023learning, feng2023dense}.
To validate this, we compared the simulated crash rate in Mcity with ground-truth crash data from five years (2016--2021) in Michigan, US~\cite{mtcf2024}.
The ground-truth crash rate is $3.00\times 10^{-6}$ crashes per mile, while our simulation yielded $2.93\times 10^{-6}$ crashes per mile, confirming the fidelity of the NDE.
Additionally, as shown in Fig.~\ref{subfig: dit statistics}, the simulation successfully replicates the composition of crashes across various road geometries and crash types, closely mirroring real-world distributions from the city of Ann Arbor, Michigan~\cite{mtcf2024}.
Moreover, our simulations capture a broad range of urban traffic conditions, including variations in traffic density, speed, and signalized travel delay, as illustrated in Fig.~\ref{subfig: ndd statistics}.
Traffic density represents the number of vehicles present within the simulation environment, vehicle speed reflects the instantaneous speed of each vehicle, and traffic light waiting time captures the duration vehicles spent at signalized intersections.
While we do not have real-world data for direct comparison, these results demonstrate the simulation’s ability to generate diverse urban traffic patterns, highlighting its potential for comprehensive traffic representation. 

Testing AVs directly within the NDE can be time-consuming due to the rarity of safety-critical events~\cite{liu2024curse}.
Studies suggest that hundreds of millions of miles of testing might be required to demonstrate AV safety performance at a human driver level~\cite{kalra2016driving}.
The NADE amplifies the likelihood of safety-critical events by training BVs to execute strategic adversarial maneuvers that challenge the AV under test.
This strategic approach significantly accelerates testing efficiency by orders of magnitude while ensuring that safety assessments remain unbiased~\cite{feng2021intelligent}.
By deploying the AV in the NADE and conducting a large number of runs, the DIT provides a comprehensive statistical assessment of AV safety, including crash rates, crash types, and crash severity statistics.
Beyond efficiency, DIT offers a unique advantage in uncovering unknown unsafe behaviors.
Its interactive, large-scale, and dynamic testing environment continuously generates safety-critical situations tailored to the specific AV under test.
These emergent interactions naturally reveal failure modes that may remain hidden in static or scenario-based evaluations.
As a result, DIT not only provides rigorous statistical validation but also serves as a powerful tool for exposing latent vulnerabilities, delivering a comprehensive assessment of AV driving intelligence.

\begin{table*}[b]
\centering
\footnotesize
\setlength{\tabcolsep}{4pt}
\renewcommand{\tabularxcolumn}[1]{m{#1}}
\caption{Overall evaluation results of BCT}
\label{tab: bct overall results}
\begin{tabularx}{\textwidth}{@{} >{\raggedright\arraybackslash}X c c r c *{5}{S[table-format=3.2]} @{}}
\toprule
\multirow{2}{*}{Behavioral Competency} & \multirow{2}{*}{Case number} & \multirow{2}{*}{Result} & \multicolumn{1}{c}{Collision} & \multirow{2}{*}{Phase}
 & \multicolumn{5}{c}{Behavioral category (\%)} \\
\cmidrule(lr){6-10}
 & & & \multicolumn{1}{c}{(\%)} & & {Aggressive} & {Assertive} & {Normal} & {Conservative} & {Ultra-conservative} \\
\midrule
\multirow{2}{=}{(a) Detect and respond to cut-in vehicle} & \multirow{2}{*}{36} & \multirow{2}{*}{Pass} & \multirow{2}{*}{0.00} & Pre  & 2.78  & 13.89 & 1.85  & 2.78  & 78.70 \\
 & & & & Post & 0.00  & 0.00  & 100.00 & 0.00  & 0.00  \\
\midrule
\multirow{3}{=}{(b) Detect and follow the leading vehicle} & \multirow{3}{*}{40} & \multirow{3}{*}{Pass} & \multirow{3}{*}{0.00} & Acc    & 0.00 & 0.00 & 13.33 & 86.67 & 0.00   \\
 & & & & Cruise & 0.00 & 0.00 & 0.00  & 0.00  & 100.00 \\
 & & & & Dec    & 0.00 & 0.00 & 0.00  & 0.00  & 100.00 \\
\midrule
\multirow{2}{=}{(c) Detect and respond to same-direction lane encroachment} & \multirow{2}{*}{36} & \multirow{2}{*}{Pass} & \multirow{2}{*}{0.00} & Pre  & 50.93 & 14.81 & 0.00  & 1.85  & 32.41 \\
 & & & & Post & 0.00  & 4.63  & 95.37 & 0.00  & 0.00  \\
\midrule
\multirow{2}{=}{(d) Detect and respond to oncoming lane encroachment} & \multirow{2}{*}{51} & \multirow{2}{*}{Fail} & \multirow{2}{*}{71.24} & Pre  & 15.69 & 7.19 & 0.00  & 0.00  & 5.88 \\
 & & & & Post & 0.00  & 0.00 & 9.80  & 15.03 & 3.92 \\
\midrule
\multirow{2}{=}{(e) Detect and respond to left-turning vehicle} & \multirow{2}{*}{99} & \multirow{2}{*}{Fail} & \multirow{2}{*}{55.56} & Pre  & 0.00 & 8.42 & 0.00 & 0.34  & 35.69 \\
 & & & & Post & 0.00  & 0.00  & 8.08  & 36.36 & 0.00 \\
\midrule
\multirow{2}{=}{(f) Execute unprotected left turn} & \multirow{2}{*}{91} & \multirow{2}{*}{Fail} & \multirow{2}{*}{2.56} & Pre  & 10.26 & 7.69 & 10.26 & 2.93  & 66.30  \\
 & & & & Post & 0.00  & 0.00  & 28.21 & 65.93 & 3.30 \\
\midrule
\multirow{2}{=}{(g) Detect and respond to right-turning vehicle} & \multirow{2}{*}{72} & \multirow{2}{*}{Fail} & \multirow{2}{*}{2.78} & Pre  & 45.83 & 30.09 & 5.09 & 9.72 & 6.48  \\
 & & & & Post & 26.85 & 3.70 & 0.00 & 5.09 & 61.57 \\
\midrule
\multirow{2}{=}{(h) Execute right turn} & \multirow{2}{*}{37} & \multirow{2}{*}{Fail} & \multirow{2}{*}{32.43} & Pre  & 18.02 & 13.51 & 2.70  & 8.11 & 25.23  \\
 & & & & Post & 0.00  & 2.70 & 20.72 & 14.41 & 29.73 \\
\midrule
\multirow{2}{=}{(i) Detect and respond to VRU crossing without crosswalk} & \multirow{2}{*}{33} & \multirow{2}{*}{Fail} & \multirow{2}{*}{78.79} & Pre  & 21.21 & 0.00 & 0.00 & 0.00 & 0.00 \\
 & & & & Post & 15.15 & 0.00 & 0.00 & 2.02 & 4.04 \\
\midrule
\multirow{2}{=}{(j) Detect and respond to VRU crossing at crosswalk} & \multirow{2}{*}{50} & \multirow{2}{*}{Pass} & \multirow{2}{*}{0.00} & Pre  & 0.00 & 1.33 & 60.67 & 38.00  & 0.00  \\
 & & & & Post & 4.00 & 0.67  & 24.67 & 1.33  & 69.33 \\
\midrule
\multirow{2}{=}{(k) Merge into roundabout with circulating traffic} & \multirow{2}{*}{40} & \multirow{2}{*}{Fail} & \multirow{2}{*}{6.67} & Pre  & 0.00 & 7.50 & 7.50  & 30.83 & 47.50  \\
 & & & & Post & 0.00  & 0.00  & 69.17 & 10.83 & 13.33 \\
\midrule
\multirow{2}{=}{(l) Detect and respond to vehicle merging into roundabout} & \multirow{2}{*}{74} & \multirow{2}{*}{Fail} & \multirow{2}{*}{45.95} & Pre  & 53.15 & 0.45 & 0.00 & 0.00 & 0.45 \\
 & & & & Post & 46.85 & 0.00 & 0.45 & 0.00 & 6.76 \\
\midrule
(m) Detect and respond to stopped vehicle partially blocking lane & 84 & Pass & 0.00 & N/A & 0.00 & 0.00 & 33.33 & 1.19 & 65.48 \\
\midrule
(n) Detect and respond to traffic light & 40 & Pass & 0.00 & N/A & 0.00 & 0.00 & 100.00 & 0.00 & 0.00 \\
\bottomrule
\end{tabularx}
\end{table*}

\subsection{AV under test}

In this study, we aim to validate the proposed methodology by assessing the behavioral safety of AVs in urban driving environments.
The AV's ODD is typical urban conditions in U.S. cities, with a speed limit of 30 mph, during daytime hours, and in sunny or cloudy weather.
For our evaluation, we utilize an open-source Level 4 AV system: Autoware.Universe~\cite{autoware}, the latest version of the widely adopted Autoware platform.
Although its performance may not yet match that of commercial systems like Waymo, open-source platforms such as Autoware.Universe are among the most accessible and widely used Automated Driving System (ADS) available to the research community.

The evaluation was conducted at the Mcity testing facility at the University of Michigan, Ann Arbor.
We began by assessing the performance of Autoware.Universe within the simulated Mcity environment.
To further demonstrate the real-world applicability of our framework, we extended the evaluation to two real vehicles: a Tesla Model 3 equipped with Full Self-Driving (version: v12.5.5) and a Lincoln MKZ equipped with Autoware.Universe.
Compared to simulations, real-vehicle testing provides higher fidelity by incorporating factors such as vehicle dynamics and other physical effects that are often challenging to replicate in simulation.

\section{AV testing results in simulation}

\subsection{Results of the Behavioral Competency Test}

The overall BCT results for Autoware.Universe are summarized in Table~\ref{tab: bct overall results}.
For each behavioral competency, the table reports the number of test cases ($n$), the pass/fail outcome, and the percentage of cases that ended in a collision.
As collisions are determined at the case level rather than per phase, the collision rate is identical across phases and is therefore listed once per competency.
The remaining columns report, for each case phase, the percentage of cases exhibiting each of the five behavioral categories (aggressive, assertive, normal, conservative, and ultra-conservative, ordered from least to most cautious), so that the collision rate and these five categories together sum to 100\% within each phase.
Most competencies are assessed over the pre- and post-encroachment phases, whereas car-following (b) is divided into the acceleration, cruising, and deceleration phases.
Of the 14 competencies evaluated, Autoware.Universe passed 6.
Examples of each competency are provided in Supplementary Movie 2.

Table~\ref{tab: bct overall results} enables a more detailed assessment of Autoware.Universe's performance.
The AV failed in eight competencies, namely (d)--(i), (k), and (l).
In the six passed competencies, the AV tends to behave conservatively.
It over-reacts to a cut-in vehicle (a) in 81\% of cases, behaves conservatively in all three phases of car following (b), keeps an overly large distance from a stopped vehicle partially blocking the lane (m) in 67\% of cases, and waits several seconds before proceeding after a VRU has crossed at a crosswalk (j) in 71\% of cases.
The exception is same-direction lane encroachment (c), where 51\% of cases are aggressive in the pre-encroachment phase.
After a cut-in or lane encroachment, the AV resumes normal driving within a human-like recovery time in almost all cases, and it makes the correct stop-or-go decision at the traffic light (n) in every case.

\begin{figure}[t]
  \centering
  \includegraphics[width=0.5\textwidth]{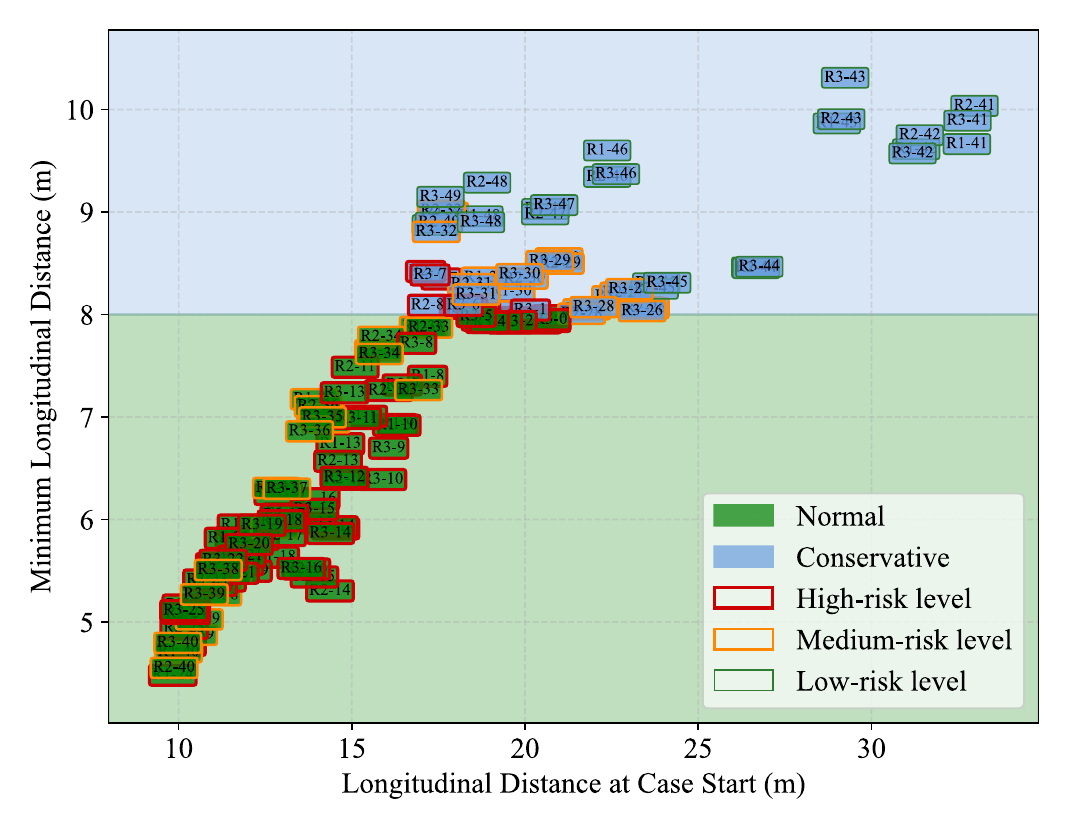}
  \caption{Analysis of pre-encroachment phase}
  \label{fig: analysis pre}
\end{figure}

We take ``Detect and respond to VRU crossing at crosswalk'' (competency j) as a representative example to illustrate how the pre- and post-encroachment phases are assessed, and why the AV's behavior in this competency is undesirable.
The pre-encroachment phase is analyzed in Fig.~\ref{fig: analysis pre}, which plots the minimum longitudinal clearance the AV maintains from the VRU against the longitudinal distance available at case start. The green and blue background zones denote the Normal and Conservative regions, respectively.
Each test case is rendered as a box whose outline color encodes its risk level (red, orange, and green for high, medium, and low) and whose fill color encodes its measured behavioral category.
Because every case is executed three times, each case is labeled by its round and case number (e.g., R2-40 denotes case 40 in round 2).
As shown in Fig.~\ref{fig: analysis pre}, the minimum longitudinal clearance increases with the distance available at case start. In 58 of the 150 cases (39\%), the minimum clearance exceeds 8~m and falls in the Conservative zone, and all remaining cases fall in the Normal zone (3--8~m).

\begin{figure}[t]
  \centering
  \includegraphics[width=0.5\textwidth]{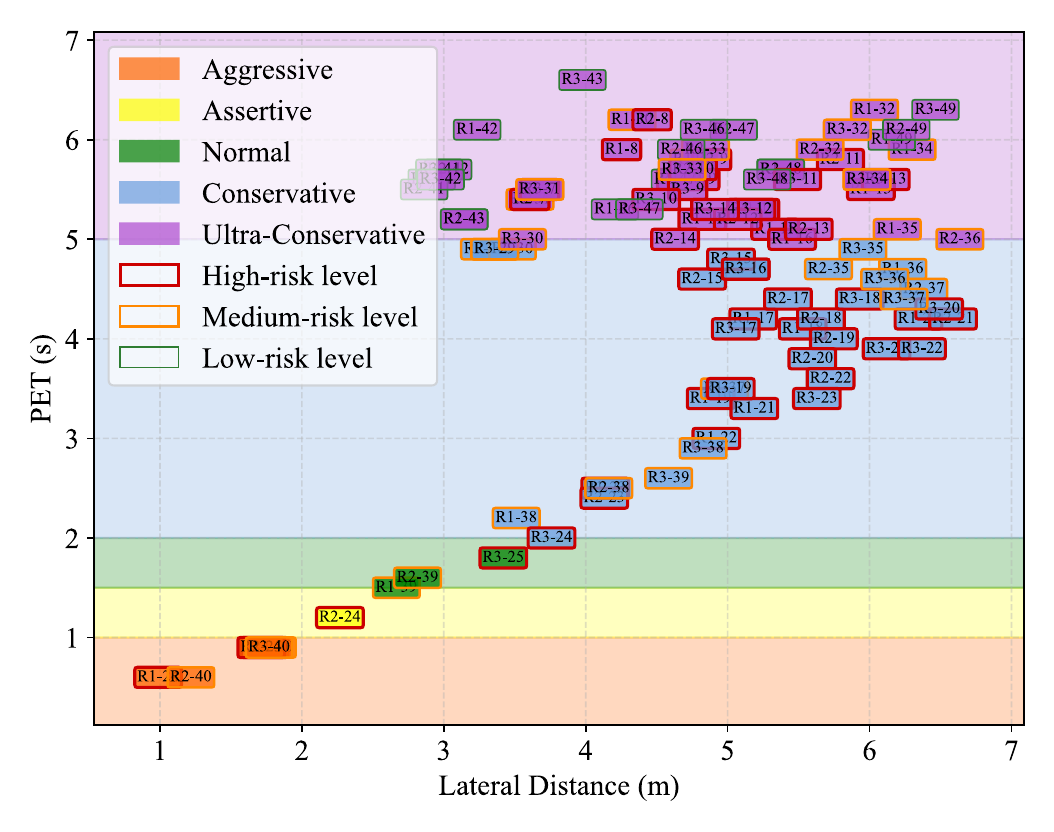}
  \caption{Analysis of post-encroachment phase}
  \label{fig: analysis post}
\end{figure}

For the post-encroachment phase, we use the Post-Encroachment Time (PET) to capture whether the two road users nearly occupy the same space at the same time, together with the lateral distance, i.e., the edge-to-edge distance between the VRU and the AV's path while the AV occupies the conflict zone.
Fig.~\ref{fig: analysis post} plots PET against this lateral distance. The orange, yellow, green, blue, and purple background zones denote the Aggressive, Assertive, Normal, Conservative, and Ultra-Conservative PET regions, respectively.
In 140 of the 150 cases (93\%), PET falls in the Conservative or Ultra-Conservative zone, i.e., the AV waits several seconds after the VRU has left the conflict zone before entering it. Six cases at the lower-left of the figure have a PET below 1~s and fall in the Aggressive zone, i.e., the AV enters the conflict zone almost immediately after the VRU has cleared it.

\begin{figure}[t]
  \centering
  \includegraphics[width=0.5\textwidth]{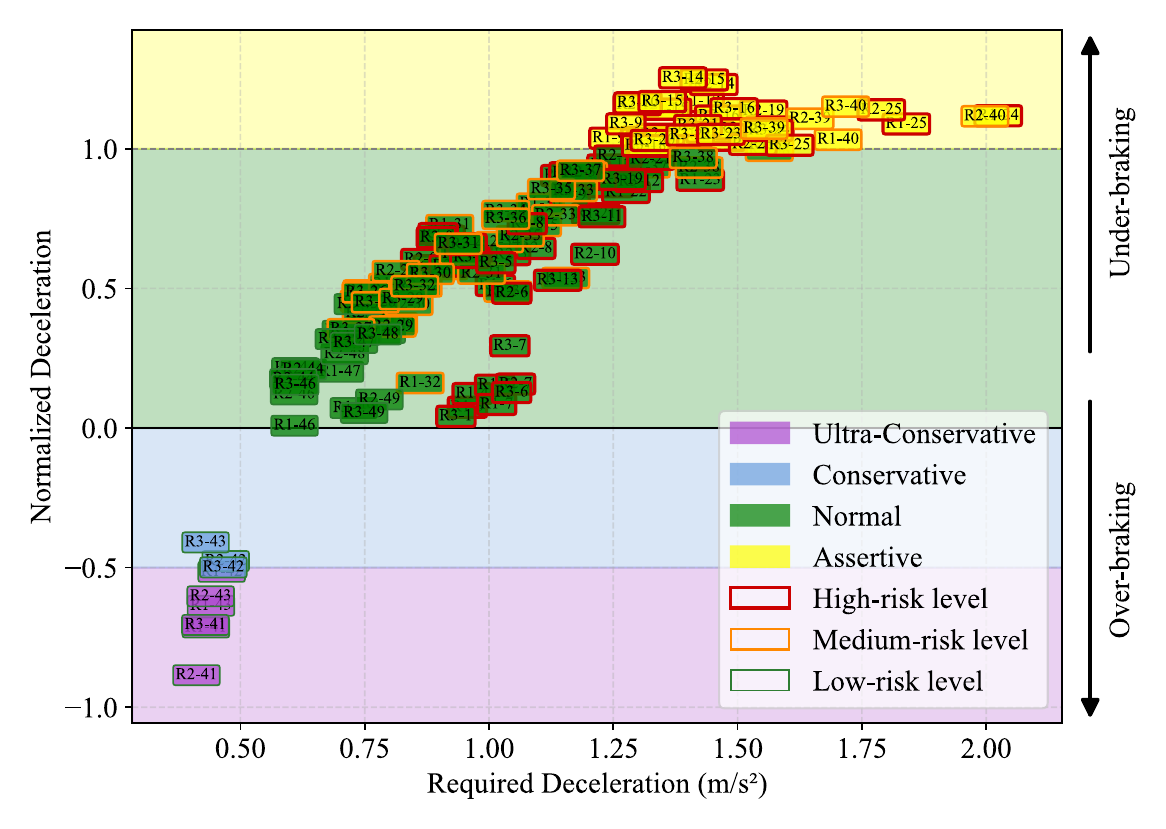}
  \caption{Behavior diagnosis}
  \label{fig: behavior diagnosis}
\end{figure}

Finally, we conduct a behavioral diagnosis to identify the root cause of this behavior: Fig.~\ref{fig: behavior diagnosis} plots the AV's normalized deceleration, which rescales its actual average deceleration from the instant it begins to brake so that 0 corresponds to the constant deceleration needed to keep a 4~m clearance from the VRU and 1 to the deceleration needed to keep only a 0.75~m clearance (higher values thus indicate under-braking), against the required deceleration, i.e., the constant deceleration needed to keep the 0.75~m clearance until the VRU has cleared the AV's path.
Most cases (100 of 150, 67\%) fall in the Normal zone.
Case R1-24, which has the largest required deceleration, falls in the Assertive zone (under-braking) and has the smallest minimum longitudinal clearance in Fig.~\ref{fig: analysis pre} (4.48~m).
Case R2-41, which has the smallest required deceleration, falls in the Ultra-Conservative zone (over-braking) and has one of the largest minimum longitudinal clearances in Fig.~\ref{fig: analysis pre} (10.03~m, Conservative zone).
Analyses of the remaining competencies are provided in Appendix~\ref{sec:results}.

\subsection{Results of the Driving Intelligence Test}

\begin{figure*}[p]
  \centering
  \includegraphics[width=0.9\textwidth]{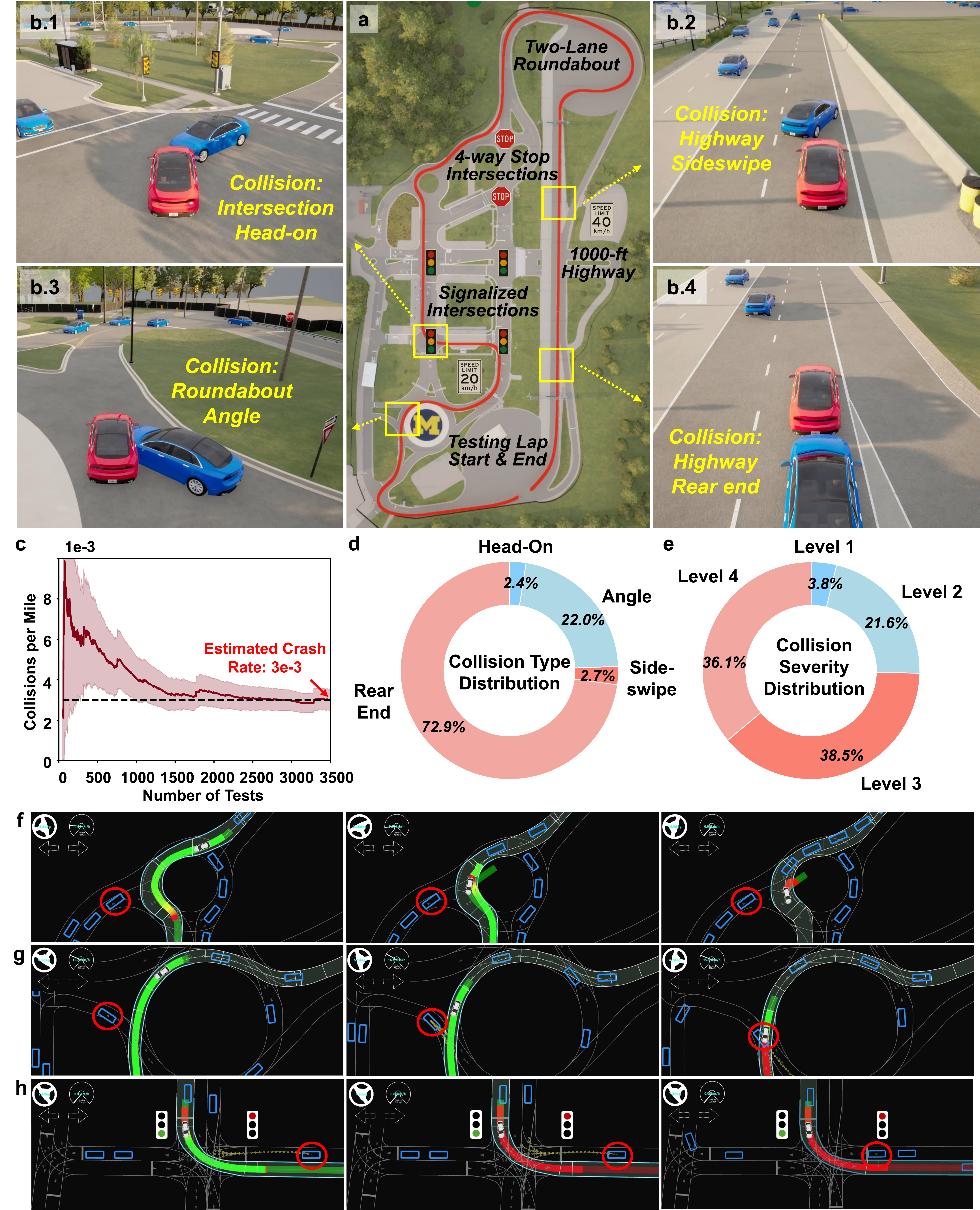}
  \caption{The Driving Intelligence Test results of Autoware.Universe. a, Demonstration of the Mcity testing environment and testing route of AV. b, Examples of different crash types experienced by Autoware.Universe: head-on (b.1), sideswipe (b.2), angle (b.3), and rear-end (b.4) collisions. The red vehicle represents the AV under test, while the blue vehicles represent background traffic. c, Crash rate estimation. d, Crash type estimation. e, Crash severity estimation. f-h, Demonstration of identified issues in Autoware.Universe: obstacle avoidance replanning error (f), right-of-way yielding issue (g), and trajectory prediction error (h). The white vehicle represents the AV, blue vehicles indicate background traffic, and the vehicle circled in red highlights the BV that is in conflict with the AV. A video demonstration can be found in Supplementary Movie 4.}
  \label{fig: dit simulation results}
\end{figure*}

The Mcity testing environment and the AV's route for the DIT are depicted in Fig.~\ref{fig: dit simulation results}a.
The route incorporates various urban driving layouts, including urban arterials, signalized intersections (covering straight paths, unprotected left turns, and right turns), as well as single-lane and double-lane roundabouts.
A full lap around the route constitutes a testing episode, which is reset if the AV either crashes or successfully completes the lap. 
The environment is stochastic, with randomized behavior and initialization of BVs in each lap, ensuring the AV encounters a wide range of real-world driving situations, as validated in previous sections.
This dynamic setup mitigates the risk of gaming the test, ensuring a robust and reliable evaluation of the AV's performance.
To evaluate AV’s behavioral safety, we conducted around 3,500 episodes of testing around 936 wall-time hours with multiple AWS cloud server EC2 C5.9xlarge instances (each instance includes 36 cores and 72 GB RAM).
Various types of crashes were observed throughout the testing route, examples of which are shown in Fig.~\ref{fig: dit simulation results}b, including head-on (Fig.~\ref{fig: dit simulation results}b.1), sideswipe (Fig.~\ref{fig: dit simulation results}b.2), angle (Fig.~\ref{fig: dit simulation results}b.3), and rear-end (Fig.~\ref{fig: dit simulation results}b.4) collisions.
The crash rate for Autoware.Universe was estimated at $3.01\times 10^{-3}$ crashes per mile (Fig.~\ref{fig: dit simulation results}c), about 1,000 times higher than that of an average human driver~\cite{mtcf2024} ($3.00\times 10^{-6}$ crash per mile).
Detailed analyses of specific Autoware.Universe issues identified during the tests are discussed in the following paragraphs.
Figs.~\ref{fig: dit simulation results}d-e show the distribution of crash types and crash severity, which is classified into four levels based on the speed difference at the moment of impact, following existing literature~\cite{feng2023dense}: level 1 (0-5 mph), level 2 (5-10 mph), level 3 (10-15 mph), and level 4 (above 15 mph).
These results highlight the DIT's capacity to statistically evaluate the frequency of safety-critical events, providing a comprehensive assessment of the AV's safety performance under real-world conditions.
A demonstration of DIT procedure can be found in Supplementary Movie 3.

A key advantage of environment-level testing in the DIT is its continuity in time and space, with a large number of background agents, exposing the AV to a virtually infinite variety of potential driving situations.
In this dynamic environment, the DIT can uncover hidden issues that are difficult to detect through scenario-based testing. We present three examples to demonstrate DIT's capability in identifying potential issues in Autoware.Universe.
Notably, the AV is treated as a black box, meaning no prior knowledge of the AV is used in designing the testing environment or identifying problems.
A video demonstration can be found in Supplementary Movie 4.
The first example, illustrated in Fig.~\ref{fig: dit simulation results}f, involves the AV navigating a small roundabout in the northern section of Mcity (Fig.~\ref{fig: dit simulation results}f left).
When a BV slightly encroaches while entering the roundabout (as indicated by the red-circled vehicle), the AV attempts to replan its path, adjusting its trajectory to avoid close proximity to the BV (Fig.~\ref{fig: dit simulation results}f middle).
However, this replanning triggers an unknown error, causing the AV to become stuck at the replanned position (Fig.~\ref{fig: dit simulation results}f right). This issue only occurs when the AV and BV occupy a specific, subtle relative position (not every time a BV merges), making it challenging to detect through modular testing or scenario-based approaches. This highlights the advantage of DIT’s continuous, dynamic environment in uncovering unknown unsafe events. We reported this issue to the Autoware development team, who subsequently proposed a fix that successfully resolved the problem\footnote{\url{https://github.com/autowarefoundation/autoware.universe/issues/5687}}. This example highlights the strength of our methodology in identifying hidden bugs in AV systems and supporting their development. 

The second example highlights a right-of-way yielding issue observed in Autoware.Universe.
When the AV has the right of way, it fails to proactively yield to aggressive BVs approaching from lower-priority lanes.
Instead, the AV only responds once the BV enters its path.
This behavior was especially problematic in scenarios like roundabouts (see Fig.~\ref{fig: dit simulation results}g), where the AV was navigating while a BV attempted to merge.
In Fig.~\ref{fig: dit simulation results}g, the AV is shown driving in the roundabout (Fig.~\ref{fig: dit simulation results}g left), the BV attempting to enter (Fig.~\ref{fig: dit simulation results}g middle), and the resulting collision (Fig.~\ref{fig: dit simulation results}g right).
While it's understandable that the AV might choose not to yield in these cases as it has the right-of-way, this rigid behavior led to multiple crashes during testing.
When faced with aggressive BVs showing no intention to stop, the AV should exercise greater caution to reduce the risk of collision.
This example underscores the DIT’s strength in identifying areas for improvement, providing valuable insights for enhancing AV models.
The third example illustrates a trajectory prediction module issue, shown in Fig.~\ref{fig: dit simulation results}h.
The AV is driving southbound at an intersection with a green light for a left turn (Fig.~\ref{fig: dit simulation results}h left).
However, it begins to decelerate (Fig.~\ref{fig: dit simulation results}h middle) and comes to a stop, obstructing southbound traffic at the intersection (Fig.~\ref{fig: dit simulation results}h right).
Upon further analysis, this behavior was due to the AV yielding to a BV traveling eastbound (circled in red) from a perpendicular lane, predicting that the BV would continue at its current speed to pass through the intersection, despite the BV facing a red light.
This issue stems from the trajectory prediction module not incorporating traffic light information, leading to an incorrect conflict prediction.
This issue has been acknowledged as a known limitation in the Autoware.Universe manual\footnote{\url{https://autowarefoundation.github.io/autoware.universe/main/planning/behavior_velocity_planner/autoware_behavior_velocity_intersection_module/\#how-towhy-set-rightofway-tag}}, reinforcing the DIT’s ability to uncover critical design flaws in AV behavior.

\section{Demonstration of Behavioral Safety Assessment on Physical Test Track}

After thoroughly evaluating Autoware.Universe in simulation, we extended the experiment to assess real AV systems, specifically a Tesla Model 3 and a Lincoln MKZ, on Mcity physical test tracks.
Given the resource constraints, the objective of these physical tests was not to generate comprehensive testing results but to validate the applicability of our framework in real-world closed testing facilities.
By testing real vehicles, we aimed to demonstrate how the framework can be applied in actual testing procedures, facilitating third-party evaluation of AV systems.
Demonstrations of testing events can be found in Supplementary Movie 5.

\subsection{Demonstration of the Behavioral Competency Test}

\begin{figure*}[t]
  \centering
  \includegraphics[width=0.9\textwidth]{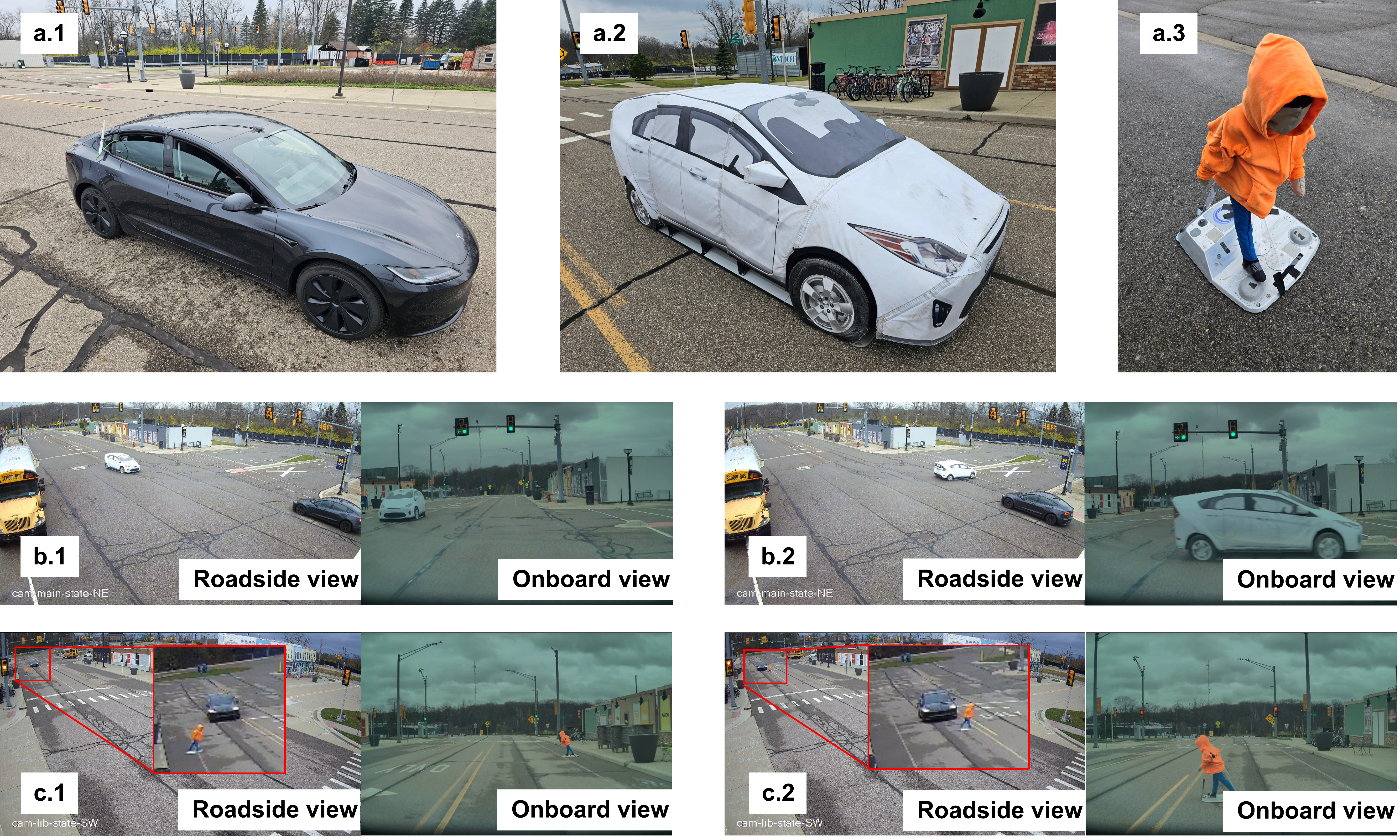}
  \caption{The field experiment of the BCT. a.1-a.3, The test system. a.1, The vehicle under test equipped with the RTK system. a.2, The Humanetics UFO Pro platform installed with the dummy vehicle. a.3, The Humanetics UFO Nano platform installed with the dummy child. b.1-b.2, Testing process for the left turn (AV goes straight) scenario. b.1, Images captured by the roadside camera and Tesla’s front-facing camera at the scenario start moment, defined as the point when the relative distance and speed satisfy the conditions specified in the test case. b.2, Images captured by the roadside camera and Tesla’s front-facing camera when the AV comes to a stop. c.1-c.2, Testing process for the VRU crossing the street without the crosswalk scenario. c.1, Images captured by the roadside camera and Tesla’s front-facing camera at the scenario start moment. c.2, Images captured by the roadside camera and Tesla’s front-facing camera when the AV comes to a stop.}
  \label{fig: bct field test}
\end{figure*}

The Behavioral Competency Test was conducted at the Mcity test facility to evaluate Tesla’s Full Self-Driving (FSD) feature, as shown in Fig.~\ref{fig: bct field test}a.1. Two competencies were tested: ``detect and respond to left-turning vehicle'' and ``detect and respond to VRU crossing without crosswalk''.
To ensure safety during the testing process, Humanetics robotic platforms\footnote{\url{https://www.humaneticsgroup.com/products/active-safety-test-systems}}, specifically UFO Pro for BV (Fig.~\ref{fig: bct field test}a.2) and UFO Nano for VRU (Fig.~\ref{fig: bct field test}a.3), equipped with a dummy vehicle and child were used instead of a real vehicle and pedestrian.
These platforms are specifically designed for vehicle testing, featuring a chassis that is sufficiently low to allow real vehicles to drive over them during collisions, while being robust enough to withstand the impact without sustaining damage.
Each platform is equipped with a real-time kinematic (RTK) global positioning system (GPS).
Similarly, an RTK system was installed on the AV to collect real-time state information during the test.
The platform is controlled adaptively to satisfy the initial conditions of the testing case according to the AV and BV/VRU state information.
Importantly, the RTK GPS is the only device attached to the AV under test to collect its state information for triggering the test scenario, and the testing system did not transmit any information to the AV.
The AV relied solely on its own sensors and algorithms to perform driving tasks, serving as a black-box to the testers.

The measured free-driving speed of Tesla’s FSD system was 8.94 m/s (20 mph) on the test road.
Based on this speed, risk levels and test cases were generated.
Considering time constraints and platform capabilities, five test cases were conducted for each scenario with at least one case for each risk level. Tesla’s FSD system successfully passed all ten sampled test cases.
For each scenario, we choose one test case and illustrate it in Fig.~\ref{fig: bct field test}b and Fig.~\ref{fig: bct field test}c, respectively. Fig.~\ref{fig: bct field test}b.1 and \ref{fig: bct field test}b.2 demonstrate time moments when the case started and AV successfully stopped in the left-turn scenario, taken by the roadside and AV front-facing cameras.
It can be found that Tesla can successfully identify potential risk and apply reasonable behavior to stop and let the aggressive BV pass first.
Similarly, Fig.~\ref{fig: bct field test}c.1 and \ref{fig: bct field test}c.2 show Tesla can properly react to the VRU jaywalking situation, demonstrating behavior competency in such safety-critical scenarios that commonly occur in urban ODD.
These experiments demonstrate that the proposed BCT enables third-party testing of commercial vehicles in physical test tracks.
During the test, the AV is treated as a black box, with no reliance on any data application programming interfaces (APIs), demonstrating its effectiveness for third-party AV safety evaluation.

\subsection{Demonstration of the Driving Intelligence Test}

\begin{figure*}[t]
  \centering
  \includegraphics[width=0.9\textwidth]{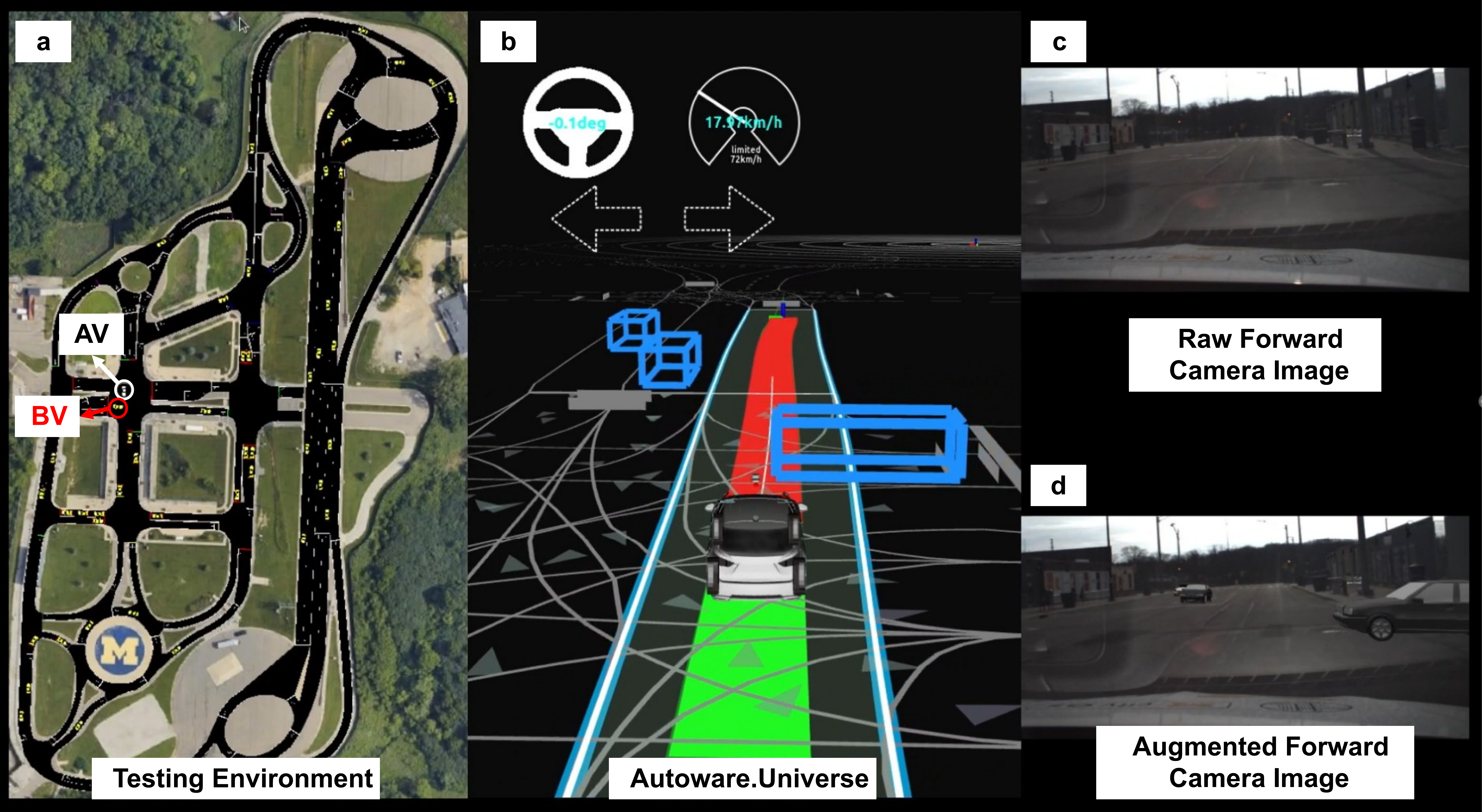}
  \caption{A snapshot of the field experiment of the Driving Intelligence Test. a, Testing environment view showing the complete traffic environment of Mcity, with the white vehicle representing the AV and all yellow vehicles representing the BVs. The red circled vehicle denotes the conflicting BV. b, Autoware.Universe’s view, with the white vehicle representing the AV and surrounding BVs indicated by blue boxes. c, Raw forward-facing camera image from the AV. d, Augmented forward-facing camera image, with BVs integrated into the scene.}
  \label{fig: dit field test}
\end{figure*}

The Driving Intelligence Test was conducted at the Mcity test facility to assess the performance of Autoware.Universe using a real vehicle.
The vehicle under test was a retrofitted Lincoln MKZ, equipped with the Dataspeed drive-by-wire system and a suite of sensors, including LiDAR, radar, camera, RTK GPS, and IMU.
The vehicle’s planned trajectory, generated by Autoware.Universe, was executed using Preview Control method~\cite{xu2019design}, which provides control commands for throttle, brake, and steering angle.
To streamline the testing process, we bypassed the perception module and fed ground-truth object states directly into the AV stack.

Unlike the BCT, the DIT involves a larger number of background agents, making it difficult to use physical proxy robots as in the BCT.
To address this challenge, we leveraged the Augmented Reality (AR) platform at the Mcity facility, where a simulated traffic environment is synchronized with the physical world, including the AV and traffic infrastructure (e.g., traffic signals).
This setup enables bidirectional information flow between the virtual BVs and the real AV, allowing interaction between the two with only the AV present on the physical test track.
This approach facilitates testing in a realistic traffic environment without requiring large numbers of real vehicles, reducing both costs and safety risks.

The testing route and conditions were consistent with the simulation tests described earlier.
We conducted multiple test laps and observed a variety of crash events, consistent with the simulation testing results.
Due to space constraints, we present one example in Fig.~\ref{fig: dit field test}.
This figure shows a snapshot from the field experiment, where Fig.~\ref{fig: dit field test}a depicts the complete traffic environment at Mcity, with the white vehicle representing the AV and the yellow vehicles representing BVs.
Fig.~\ref{fig: dit field test}b shows the AV’s view, with surrounding BVs highlighted by blue boxes.
Fig.~\ref{fig: dit field test}c and \ref{fig: dit field test}d display raw and augmented images from the AV’s forward-facing camera, respectively.
In this example, the AV is traveling southbound through a signalized intersection with a green light, while a BV runs a red light from the west.
The AV decelerates but ultimately collides with the BV.
Although the AV had the right-of-way, a more proactive safety maneuver could have been executed to avoid the collision, resulting in better overall safety performance.
Using the AR platform, we can either feed the augmented images directly into the AV’s sensing system to evaluate the full AV pipeline, or send object information to focus specifically on testing the decision-making component of the AV.
A video version of this scenario, along with additional testing scenarios, can be found in Supplementary Movie 5.
These results demonstrate the capability of the proposed method for evaluating AVs in real-world physical test tracks.

\section{Conclusion}
In this study, we present a systematic and standardized third-party testing framework for assessing AV behavioral safety, operating under a black-box approach that requires no prior knowledge of the AV’s internal workings. The framework aims to create a comprehensive blueprint that unifies testing objectives, procedures, and measurements with executable methodologies that can be refined and expanded to support large-scale AV deployment. This evaluation framework addresses both minimum performance requirements, such as behavioral competency, and comprehensive performance assessment, such as the occurrence rate of measurable safety-related events, like crash rates, to comprehensively evaluate AV behavior performance. The testing scenarios and environments are designed to be both naturalistic and adversarial, substantially improving testing efficiency without sacrificing unbiasedness. The proposed method is versatile, applicable to both simulation and physical test tracks, and has the potential to empower testing facilities worldwide in advancing AV behavioral safety testing and validation.

One of the most significant challenges in evaluating AV behavioral safety is bridging the sim-to-real gap, particularly for the DIT. This gap stems from two primary sources: the fidelity of simulated sensor inputs for the AV stack and the realism of human agents’ behavior in simulations. In our current case studies with Autoware.Universe, we simplified the testing process by bypassing the perception module and feeding object ground-truth states directly into the AV stack. As end-to-end AV stacks become increasingly prevalent, our testing framework can be extended to incorporate state-of-the-art sensor simulation packages, such as NVIDIA’s Omniverse~\cite{nvidia_omniverse} and Cosmos~\cite{nvidia2025cosmos}. Additionally, achieving high-fidelity simulations that accurately replicate human driving behavior over extended time period remains a significant challenge. Recent advancements, such as the development of world models~\cite{hu2023gaia1, nvidia2025cosmos}, have shown potential in addressing these issues, but continued research and innovation are necessary to fully close these gaps and ensure the reliability of AV testing.

Behavioral compliance with traffic regulations is fundamental to AV safety and should be evaluated. In the BCT, diverse scenarios should be designed to assess whether AVs adhere to traffic laws. In this study, we only demonstrate it with a single case with traffic light, evaluating whether AV can properly stop at red lights. More scenarios should be designed and evaluated according to the local traffic laws and regulations. Additionally, in the DIT, compliance can be monitored dynamically, for example, using trigger-based violation detection methods proposed in literature~\cite{Yu2024} to identify infractions in naturalistic driving environments. Beyond safety, AV behavior also impacts broader traffic dynamics, including mobility and congestion. For instance, Cruise AVs in San Francisco were reported to have blocked intersections, causing severe traffic disruptions~\cite{wayland2022selfdriving}. Additionally, AVs may adopt different car-following headways compared to human drivers, potentially impacting overall traffic flow and efficiency. While this paper focuses on safety and mobility and congestion effects are out of the scope of this study, dedicated evaluation frameworks for these secondary effects should be developed in the future, as they also play a crucial role in the large-scale deployment and public acceptance of AVs.

\section*{Supplementary Videos}
Supplementary videos are available at \url{https://zenodo.org/records/23176547}.\\
\noindent Supplementary Movie 1: Illustration of the BCT.\\
Supplementary Movie 2: Demonstration of passed and failed testing scenarios in the BCT.\\
Supplementary Movie 3: Illustration of the DIT.\\
Supplementary Movie 4: Demonstration of three safety-critical cases identified in the DIT.\\
Supplementary Movie 5: Demonstration of AV testing at the Mcity physical test track.\\

\renewcommand{\dbltopfraction}{0.9}
\renewcommand{\dblfloatpagefraction}{0.6}
\appendices
\section{Implementation of the Behavioral Competency Test}
\label{sec:bct}

\subsection{Competency Selection}

Based on typical scenarios within the urban operational design domain (ODD) of autonomous vehicles (AVs), the Behavioral Competency Test (BCT) comprises 14 behavioral competencies to evaluate AV performance.
For all competencies except the traffic light competency, the test ends when: (i) a collision occurs, or (ii) a collision is no longer possible.
For the traffic light competency, the test concludes once the AV has passed through the intersection.
The detailed definitions of these 14 competencies are provided below:

\begin{enumerate}[label=\textbf{\alph*}.]

    \item \textbf{Detect and respond to cut-in vehicle:} The AV travels straight, while the background
    vehicle (BV) initially travels straight in the adjacent lane. The BV then
    changes lanes in front of the AV. The competency begins when the BV's
    lateral speed becomes greater than~0. The initial conditions consist of the
    initial relative longitudinal distance and initial relative speed between
    the AV and the BV.

    \item \textbf{Detect and follow the leading vehicle:} The AV and BV both travel straight in the
    same lane, with the BV initially stopped in front of the AV. The competency
    begins when the AV stops behind the BV. The BV then accelerates until it
    reaches a given speed, maintains that speed until the distance between the
    AV and BV stabilizes, and finally decelerates to a complete stop. The
    initial conditions consist of the BV acceleration, constant speed, and
    deceleration.

    \item \textbf{Detect and respond to same-direction lane encroachment:} Both the AV and BV travel
    straight, with the BV in the adjacent lane. The BV departs from its lane in
    front of the AV. The competency begins when the BV's lateral speed becomes
    greater than~0. The initial conditions consist of the initial relative
    longitudinal distance and initial relative speed between the AV and the BV.

    \item \textbf{Detect and respond to oncoming lane encroachment:} The AV travels straight
    while the BV begins in the adjacent lane traveling in the opposite
    direction. The BV departs from its lane in front of the AV. The competency
    begins when the BV's lateral speed becomes greater than~0. The initial
    conditions consist of the initial relative longitudinal distance and initial
    relative speed between the AV and the BV.

    \item \textbf{Detect and respond to left-turning vehicle:} The AV travels straight
    through an intersection while the BV turns left in front of it. The
    competency begins when the BV starts to turn. The initial conditions consist
    of the initial relative longitudinal distance between the AV and the BV in
    the AV heading direction and the initial BV speed.

    \item \textbf{Execute unprotected left turn:} The AV turns left at an
    intersection while the BV travels straight. The competency begins when the
    BV starts to enter the intersection. The initial conditions consist of the
    initial relative longitudinal distance between the AV and the BV in the AV
    heading direction and the initial BV speed.

    \item \textbf{Detect and respond to right-turning vehicle:} The AV travels straight
    through an intersection while the BV turns right in front of it. The
    competency begins when the BV starts to turn. The initial conditions consist
    of the initial relative longitudinal distance between the AV and the BV in
    the AV heading direction and the initial BV speed.

    \item \textbf{Execute right turn:} The AV turns right at an
    intersection while the BV travels straight. The competency begins when the
    BV's front bumper aligns with the stop line. The initial conditions consist
    of the initial relative longitudinal distance between the AV and the BV in
    the AV heading direction and the initial BV speed.

    \item \textbf{Detect and respond to VRU crossing without crosswalk:} The
    AV travels straight while the VRU initially walks along the roadside and
    suddenly begins crossing the road. The competency begins when the VRU starts
    to cross. The initial conditions consist of the initial relative longitudinal
    distance between the AV and the VRU in the AV heading direction and the
    initial VRU speed.

    \item \textbf{Detect and respond to VRU crossing at crosswalk:} The AV travels straight toward a
    crosswalk while the VRU crosses at the same time. The competency begins when
    the VRU starts to cross. The initial conditions consist of the initial
    relative longitudinal distance between the AV and the VRU in the AV heading
    direction and the initial VRU speed.

    \item \textbf{Merge into roundabout with circulating traffic:} The AV merges into the
    roundabout while the BV is traveling inside it. The competency begins when
    the BV passes the previous entrance of the roundabout. The initial
    conditions consist of the initial relative distance between the AV and the
    conflict point along the AV route and the initial BV speed.

    \item \textbf{Detect and respond to vehicle merging into roundabout:} The AV travels inside the
    roundabout while a BV merges in. The competency begins when the BV passes
    the yield line. The initial conditions consist of the initial relative
    distance between the AV and the conflict point along the AV route and the
    initial BV speed.

    \item \textbf{Detect and respond to stopped vehicle partially blocking lane:} The AV travels in a straight lane
    while a BV is stopped in front of it at an oblique angle. The competency
    begins when the AV starts to move. The initial conditions consist of the
    distance between the BV center and the lane center, and the BV heading.

    \item \textbf{Detect and respond to traffic light:} The AV travels toward an intersection while
    the traffic light changes from green to yellow. The competency begins when
    the AV starts to move. The initial condition is the time headway (THW)
    between the AV's front bumper and the stop line at the moment the light
    changes.

\end{enumerate}

\subsection{Parameter Calibration}

The parameters of the AV, BV, VRU, and road geometry for each competency are shown in Table~\ref{tab:participants}.

\begin{table}[!t]
    \caption{Parameters of road participants and road geometry}
    \label{tab:participants}
    \begin{center}
        \begin{tabular}{l c l c}
            Parameter   & Value  & Parameter   & Value  \\ \hline
            AV length   & 5\,m   & VRU length  & 0.5\,m \\
            AV width    & 2\,m   & VRU width   & 0.5\,m \\
            BV length   & 5\,m   & Lane width  & 3.8\,m \\
            BV width    & 2\,m   &             &        \\ \hline
        \end{tabular}
    \end{center}
\end{table}

The AV's free-flow speed in each competency (recorded in an empty environment without other road participants) is listed in Table~\ref{tab:av_speed}.
These values serve as the reference AV initial speed for computing risk level boundaries.

\begin{table}[!t]
    \caption{AV free-flow speed per competency}
    \label{tab:av_speed}
    \begin{center}
        \begin{tabular}{l c}
            Competency & AV speed \\ \hline
            Detect and respond to cut-in vehicle                          & 8.3\,m/s \\
            Detect and follow the leading vehicle                         & 8.3\,m/s \\
            Detect and respond to same-direction lane encroachment        & 8.3\,m/s \\
            Detect and respond to oncoming lane encroachment              & 8.3\,m/s \\
            Detect and respond to left-turning vehicle                    & 5.0\,m/s \\
            Execute unprotected left turn                                 & 3.61\,m/s \\
            Detect and respond to right-turning vehicle                   & 5.0\,m/s \\
            Execute right turn                                            & 3.6\,m/s \\
            Detect and respond to VRU crossing without crosswalk          & 5.0\,m/s \\
            Detect and respond to VRU crossing at crosswalk               & 5.0\,m/s \\
            Merge into roundabout with circulating traffic                & 5.0\,m/s \\
            Detect and respond to vehicle merging into roundabout         & 4.45\,m/s \\
            Detect and respond to stopped vehicle partially blocking lane & 8.3\,m/s \\
            Detect and respond to traffic light                           & 5.0\,m/s \\ \hline
        \end{tabular}
    \end{center}
\end{table}

The controllable variables for each competency are the BV/VRU behavior parameters.
Since the AV is treated as a black box, we do not control its speed, acceleration, or reactions.
For most competencies the controllable variables are the initial relative distance~($\Delta d$) and either the initial relative speed (cut-in and lane encroachment competencies) or the initial BV speed~($v_\text{BV}$, all other competencies).
In the car-following competency the controllable variables are BV speed and BV acceleration~($a_\text{BV}$).
In the stopped vehicle partially blocking lane competency they are the BV heading and the lateral offset of the BV center from the lane center.
In the traffic light competency the controllable variable is the THW between the AV front bumper and the stop line at the moment the light changes.

To classify test cases by difficulty, we compute the minimum AV deceleration or acceleration required to avoid a collision in each competency.
The greater the required maneuver, the higher the risk level.
Five levels are defined: trivial, low-risk, medium-risk, high-risk, and infeasible.
In the lane departure (opposite direction) competency the AV must change lanes rather than brake or accelerate, so lane-changing time is used instead.
In the four turn competencies the AV can avoid collision by either braking or accelerating, and both options are evaluated independently before being combined into a single risk level.
Risk levels are not defined for the stopped vehicle partially blocking lane and traffic light competencies because the AV's braking onset distance is not externally controlled.

Boundary thresholds are derived from a naturalistic driving dataset (NDD).
The trivial boundary is the 80th percentile of AV deceleration/acceleration in the NDD: cases below this threshold require only routine maneuvers.
The infeasible boundary is the lower of the 1st NDD percentile and the physical limit. The acceleration limit is set to 2.87\,m/s$^2$~\cite{bokare2017acceleration} and the deceleration limit to $-7.06$\,m/s$^2$ (road friction coefficient $\approx 0.72$ on dry roads~\cite{mumford1997inverted}).
The maximum cut-in time is 8.5\,s based on lane-change time distributions in Argoverse 2~\cite{wilson2023argoverse}.
The low-, medium-, and high-risk boundaries correspond approximately to the 1st, 10th, and 80th NDD percentiles.
The NDD sources are the rounD dataset~\cite{rounDdataset} for roundabout competencies and the Argoverse~2 Motion Forecasting Dataset~\cite{wilson2023argoverse} for all others.

The resulting boundaries are summarized in Table~\ref{tab:risk_boundaries}, where the Mitigation column gives the maneuver used to grade each competency:
``Dec.''\ denotes deceleration, ``Acc.''\ denotes acceleration, and ``Lane chg''\ denotes lane change.

\begin{table*}[!t]
    \caption{Risk level boundaries for each competency.}
    \label{tab:risk_boundaries}
    \begin{center}
    \begin{tabular}{m{3.6cm} m{1.2cm} m{2cm} m{2.1cm} m{2.1cm} m{2.1cm} m{2.1cm}}
        \multirow{2}{*}{Competency}
            & \multirow{2}{*}{\centering Mitigation}
            & \multicolumn{5}{c}{Boundary} \\
        \cline{3-7}
            & & Trivial & Low-risk & Medium-risk & High-risk & Infeasible \\ \hline

        a.\ Detect and respond to cut-in vehicle
            & Dec.
            & $[-0.27,0)$ m/s$^2$
            & $[-1.96,-0.27)$ m/s$^2$
            & $[-4.41,-1.96)$ m/s$^2$
            & $[-7.06,-4.41)$ m/s$^2$
            & $(-\infty,-7.06)$ m/s$^2$ \\ \hline

        b.\ Detect and follow the leading vehicle
            & Dec.
            & $[-0.26,0)$ m/s$^2$
            & $[-1.91,-0.26)$ m/s$^2$
            & $[-4.03,-1.91)$ m/s$^2$
            & $[-7.06,-4.03)$ m/s$^2$
            & $(-\infty,-7.06)$ m/s$^2$ \\ \hline

        c.\ Detect and respond to same-direction lane encroachment
            & Dec.
            & $[-0.27,0)$ m/s$^2$
            & $[-1.96,-0.27)$ m/s$^2$
            & $[-4.41,-1.96)$ m/s$^2$
            & $[-7.06,-4.41)$ m/s$^2$
            & $(-\infty,-7.06)$ m/s$^2$ \\ \hline

        d.\ Detect and respond to oncoming lane encroachment
            & Lane chg
            & $(8.4,+\infty)$ s
            & $(2.3,8.4]$ s
            & $(1.3,2.3]$ s
            & $(0.5,1.3)$ s
            & $(0,0.5]$ s \\ \hline

        \multirow{2}{4.0cm}{e.\ Detect and respond to left-turning vehicle}
            & Acc.
            & $(0,0.34]$ m/s$^2$
            & $(0.34,1.85]$ m/s$^2$
            & $(1.85,2.73]$ m/s$^2$
            & $(2.73,2.87]$ m/s$^2$
            & $(2.87,+\infty)$ m/s$^2$ \\
        \cline{2-7}
            & Dec.
            & $[-0.42,0)$ m/s$^2$
            & $[-2.31,-0.42)$ m/s$^2$
            & $[-4.24,-2.31)$ m/s$^2$
            & $[-7.06,-4.24)$ m/s$^2$
            & $(-\infty,-7.06)$ m/s$^2$ \\ \hline

        \multirow{2}{4.0cm}{f.\ Execute unprotected left turn}
            & Acc.
            & $(0,0.38]$ m/s$^2$
            & $(0.38,1.96]$ m/s$^2$
            & $(1.96,2.73]$ m/s$^2$
            & $(2.73,2.87]$ m/s$^2$
            & $(2.87,+\infty)$ m/s$^2$ \\
        \cline{2-7}
            & Dec.
            & $[-0.33,0)$ m/s$^2$
            & $[-2.37,-0.33)$ m/s$^2$
            & $[-4.18,-2.37)$ m/s$^2$
            & $[-7.06,-4.18)$ m/s$^2$
            & $(-\infty,-7.06)$ m/s$^2$ \\ \hline

        \multirow{2}{4.0cm}{g.\ Detect and respond to right-turning vehicle}
            & Acc.
            & $(0,0.26]$ m/s$^2$
            & $(0.26,1.66]$ m/s$^2$
            & $(1.66,2.61]$ m/s$^2$
            & $(2.61,2.87]$ m/s$^2$
            & $(2.87,+\infty)$ m/s$^2$ \\
        \cline{2-7}
            & Dec.
            & $[-0.26,0)$ m/s$^2$
            & $[-1.89,-0.26)$ m/s$^2$
            & $[-3.64,-1.89)$ m/s$^2$
            & $[-7.06,-3.64)$ m/s$^2$
            & $(-\infty,-7.06)$ m/s$^2$ \\ \hline

        \multirow{2}{4.0cm}{h.\ Execute right turn}
            & Acc.
            & $(0,0.34]$ m/s$^2$
            & $(0.34,1.89]$ m/s$^2$
            & $(1.89,2.71]$ m/s$^2$
            & $(2.71,2.87]$ m/s$^2$
            & $(2.87,+\infty)$ m/s$^2$ \\
        \cline{2-7}
            & Dec.
            & $[-0.23,0)$ m/s$^2$
            & $[-2.06,-0.23)$ m/s$^2$
            & $[-4.53,-2.06)$ m/s$^2$
            & $[-7.06,-4.53)$ m/s$^2$
            & $(-\infty,-7.06)$ m/s$^2$ \\ \hline

        i.\ Detect and respond to VRU crossing without crosswalk
            & Dec.
            & $[-0.33,0)$ m/s$^2$
            & $[-1.96,-0.33)$ m/s$^2$
            & $[-3.78,-1.96)$ m/s$^2$
            & $[-7.06,-3.78)$ m/s$^2$
            & $(-\infty,-7.06)$ m/s$^2$ \\ \hline

        j.\ Detect and respond to VRU crossing at crosswalk
            & Dec.
            & $[-0.33,0)$ m/s$^2$
            & $[-1.96,-0.33)$ m/s$^2$
            & $[-3.78,-1.96)$ m/s$^2$
            & $[-7.06,-3.78)$ m/s$^2$
            & $(-\infty,-7.06)$ m/s$^2$ \\ \hline

        \multirow{2}{4.0cm}{k.\ Merge into roundabout with circulating traffic}
            & Acc.
            & $(0,0.51]$ m/s$^2$
            & $(0.51,2.04]$ m/s$^2$
            & $(2.04,2.72]$ m/s$^2$
            & $(2.72,2.87]$ m/s$^2$
            & $(2.87,+\infty)$ m/s$^2$ \\
        \cline{2-7}
            & Dec.
            & $[-0.33,0)$ m/s$^2$
            & $[-2.48,-0.33)$ m/s$^2$
            & $[-5.59,-2.48)$ m/s$^2$
            & $[-7.06,-5.59)$ m/s$^2$
            & $(-\infty,-7.06)$ m/s$^2$ \\ \hline

        \multirow{2}{4.0cm}{l.\ Detect and respond to vehicle merging into roundabout}
            & Acc.
            & $(0,0.40]$ m/s$^2$
            & $(0.40,2.08]$ m/s$^2$
            & $(2.08,2.73]$ m/s$^2$
            & $(2.73,2.87]$ m/s$^2$
            & $(2.87,+\infty)$ m/s$^2$ \\
        \cline{2-7}
            & Dec.
            & $[-0.36,0)$ m/s$^2$
            & $[-2.45,-0.36)$ m/s$^2$
            & $[-5.09,-2.45)$ m/s$^2$
            & $[-7.06,-5.09)$ m/s$^2$
            & $(-\infty,-7.06)$ m/s$^2$ \\ \hline
    \end{tabular}
    \end{center}
\end{table*}

\subsection{Case Generation}

\begin{figure*}[!t]
    \centering
    \subfloat[Cut-In Vehicle\label{subfig:cases_cut_in}]{%
        \begin{minipage}[t]{.22\textwidth}\centering
            \includegraphics[height=3.3cm]{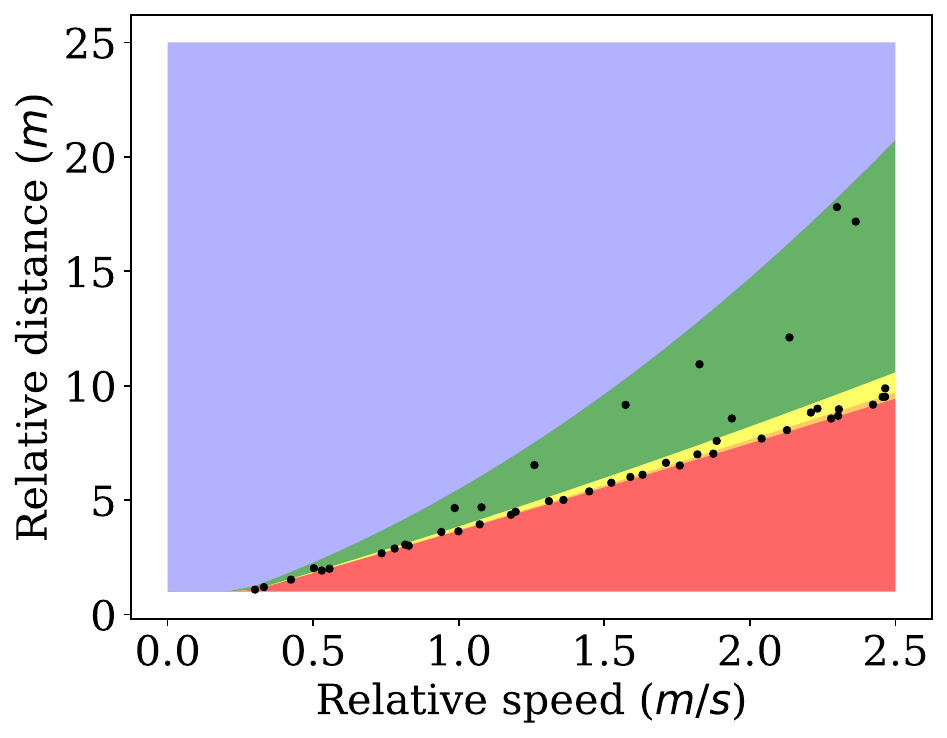}
        \end{minipage}}\quad
    \subfloat[Follow Leading Vehicle\label{subfig:cases_car_following}]{%
        \begin{minipage}[t]{.22\textwidth}\centering
            \includegraphics[height=3.3cm]{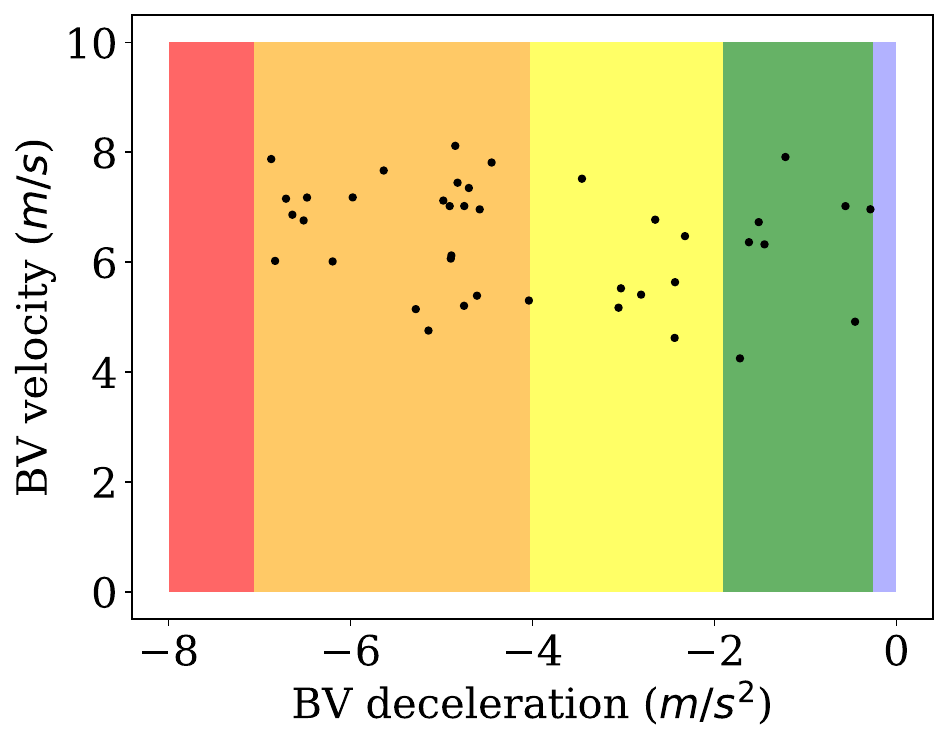}
        \end{minipage}}\quad
    \subfloat[Same-Dir.\ Lane Encr.\label{subfig:cases_lane_dep_same}]{%
        \begin{minipage}[t]{.22\textwidth}\centering
            \includegraphics[height=3.3cm]{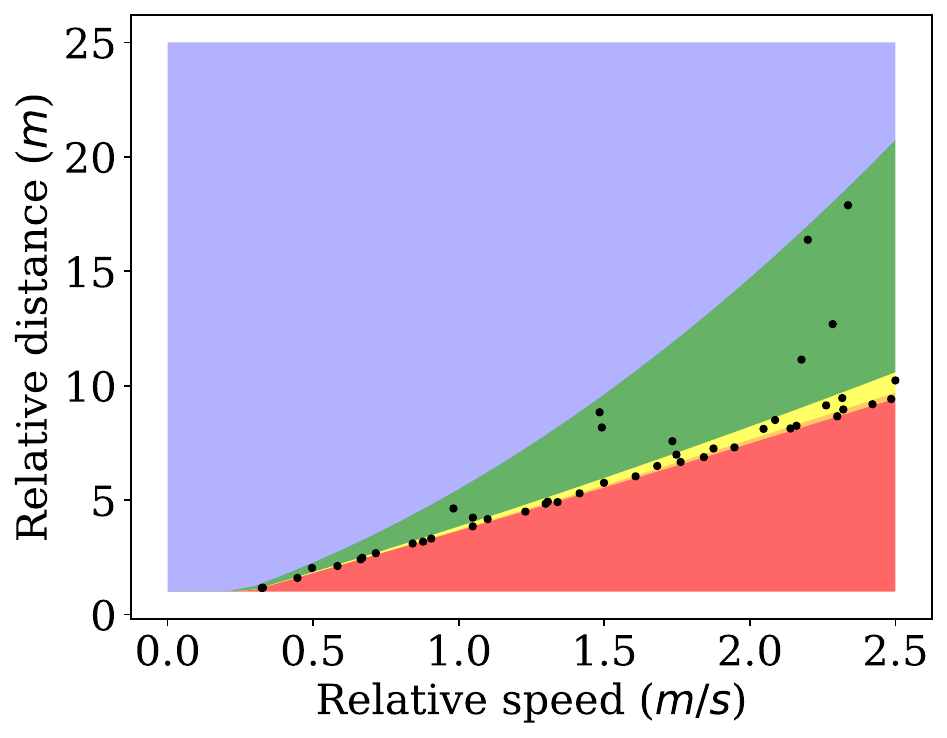}
        \end{minipage}}\quad
    \subfloat[Oncoming Lane Encr.\label{subfig:cases_lane_dep_opp}]{%
        \begin{minipage}[t]{.22\textwidth}\centering
            \includegraphics[height=3.3cm]{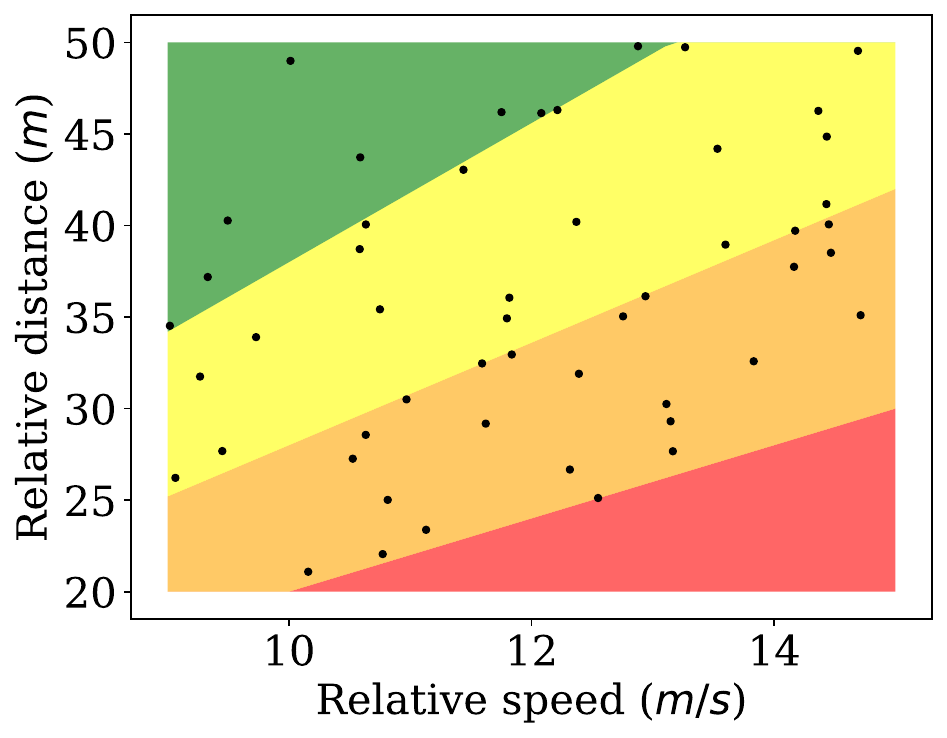}
        \end{minipage}}\quad
    \subfloat[Left-Turning Vehicle\label{subfig:cases_left_turn_straight}]{%
        \begin{minipage}[t]{.22\textwidth}\centering
            \includegraphics[height=3.3cm]{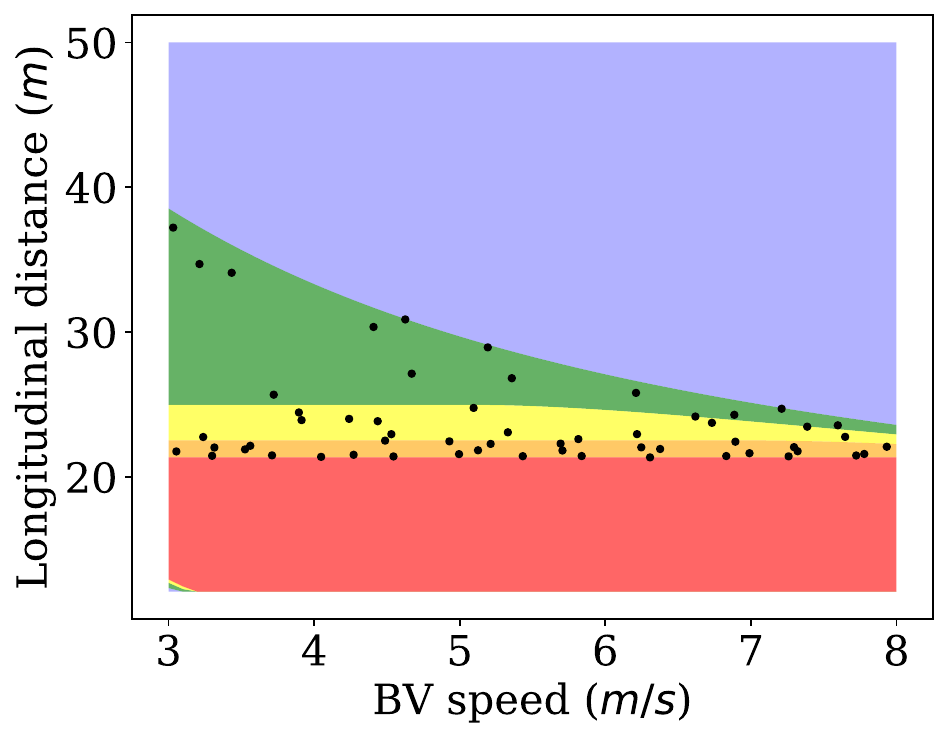}
        \end{minipage}}\quad
    \subfloat[Unprotected Left Turn\label{subfig:cases_left_turn_turn}]{%
        \begin{minipage}[t]{.22\textwidth}\centering
            \includegraphics[height=3.3cm]{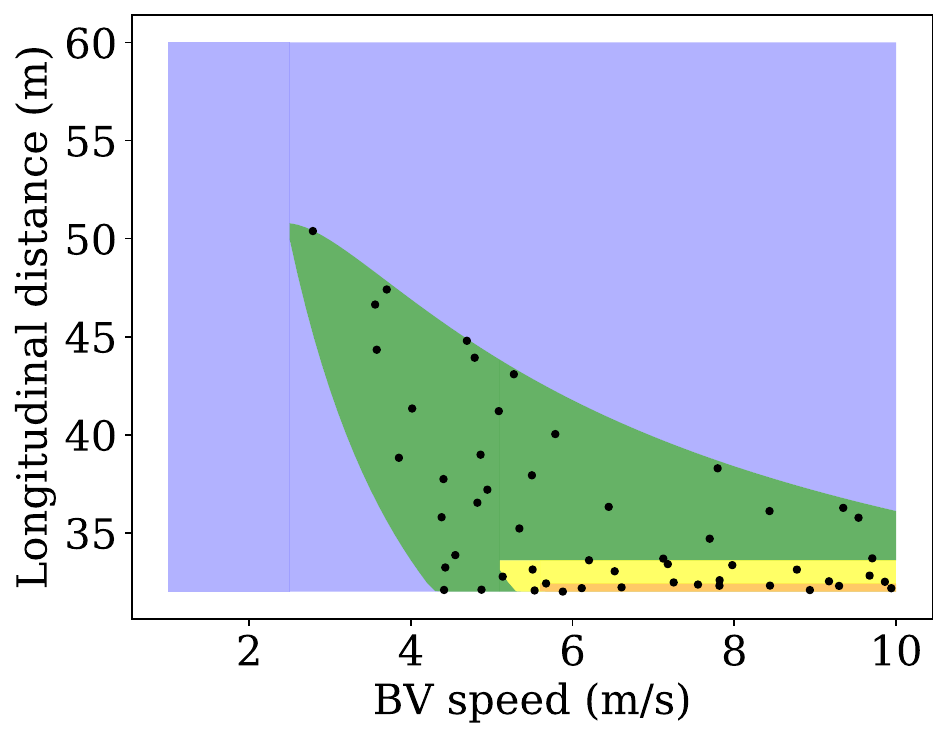}
        \end{minipage}}\quad
    \subfloat[Right-Turning Vehicle\label{subfig:cases_right_turn_straight}]{%
        \begin{minipage}[t]{.22\textwidth}\centering
            \includegraphics[height=3.3cm]{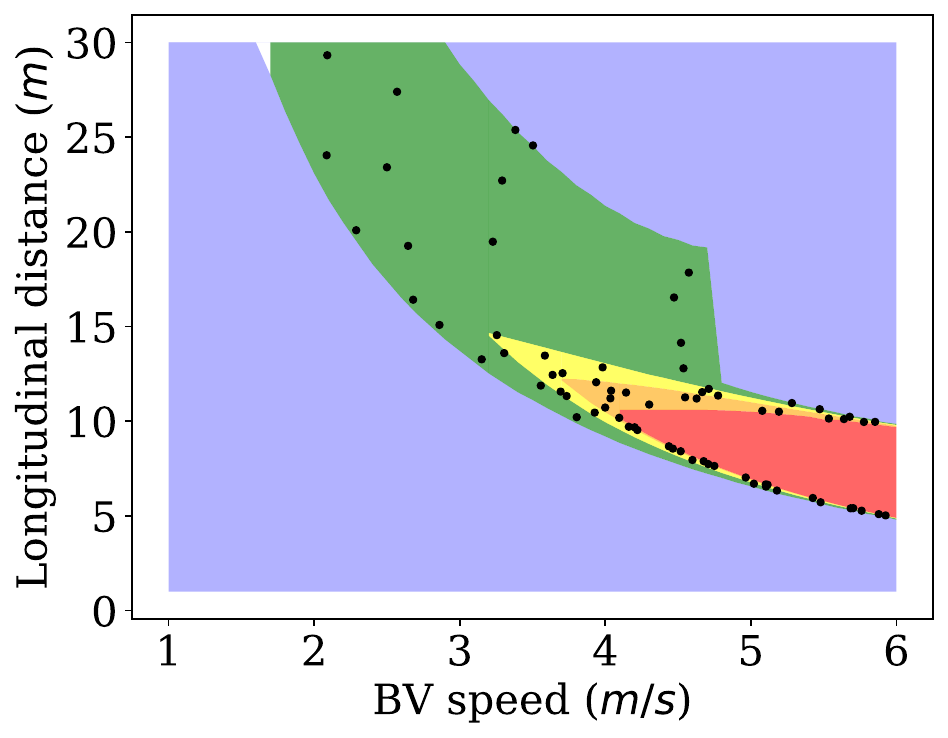}
        \end{minipage}}\quad
    \subfloat[Execute Right Turn\label{subfig:cases_right_turn_turn}]{%
        \begin{minipage}[t]{.22\textwidth}\centering
            \includegraphics[height=3.3cm]{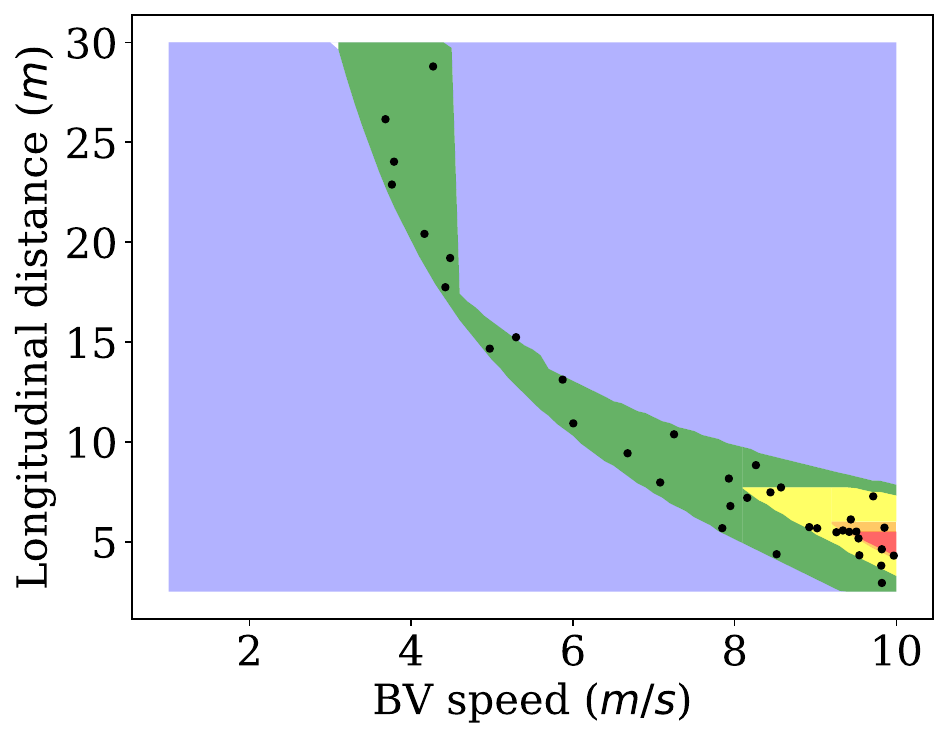}
        \end{minipage}}\quad
    \subfloat[VRU w/o Crosswalk\label{subfig:cases_vru_no_crosswalk}]{%
        \begin{minipage}[t]{.22\textwidth}\centering
            \includegraphics[height=3.3cm]{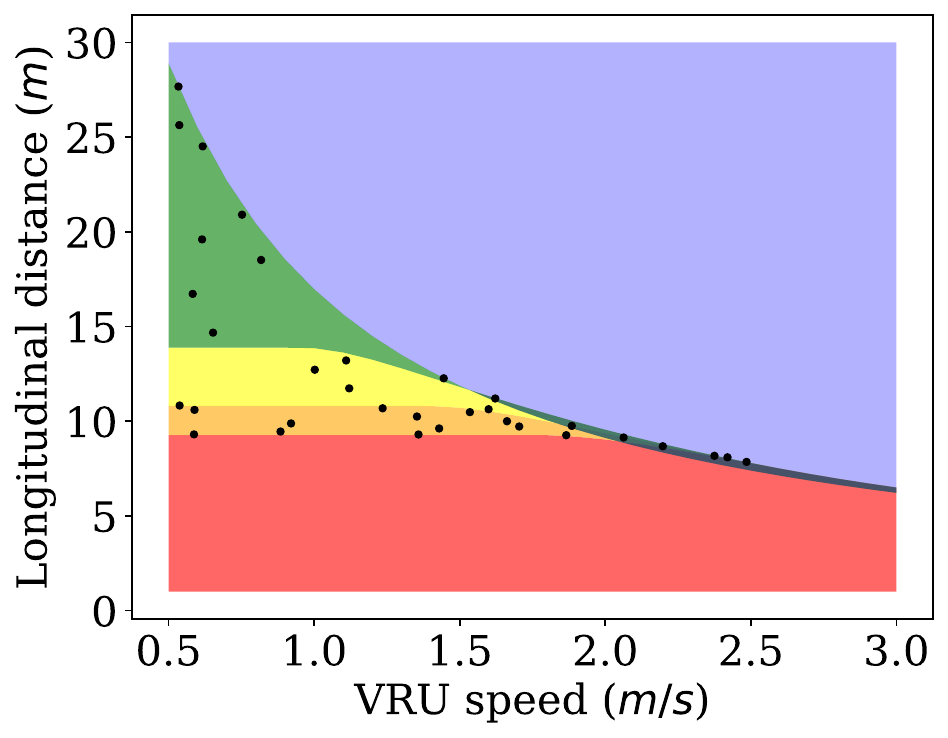}
        \end{minipage}}\quad
    \subfloat[VRU at Crosswalk\label{subfig:cases_vru_crosswalk}]{%
        \begin{minipage}[t]{.22\textwidth}\centering
            \includegraphics[height=3.3cm]{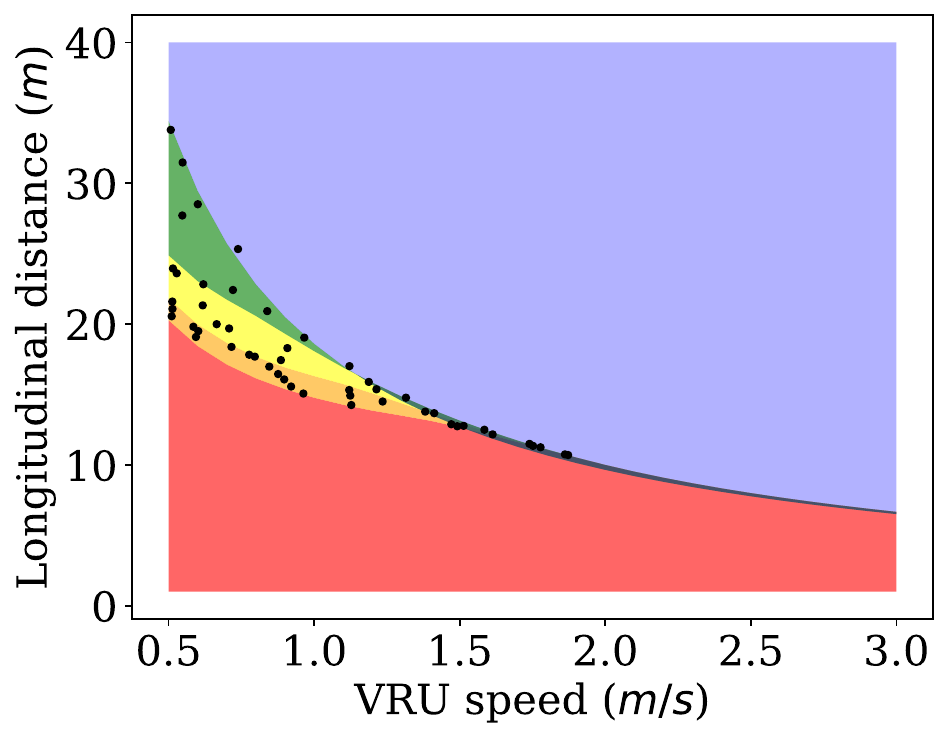}
        \end{minipage}}\quad
    \subfloat[Merge into Roundabout\label{subfig:cases_roundabout_outside}]{%
        \begin{minipage}[t]{.22\textwidth}\centering
            \includegraphics[height=3.3cm]{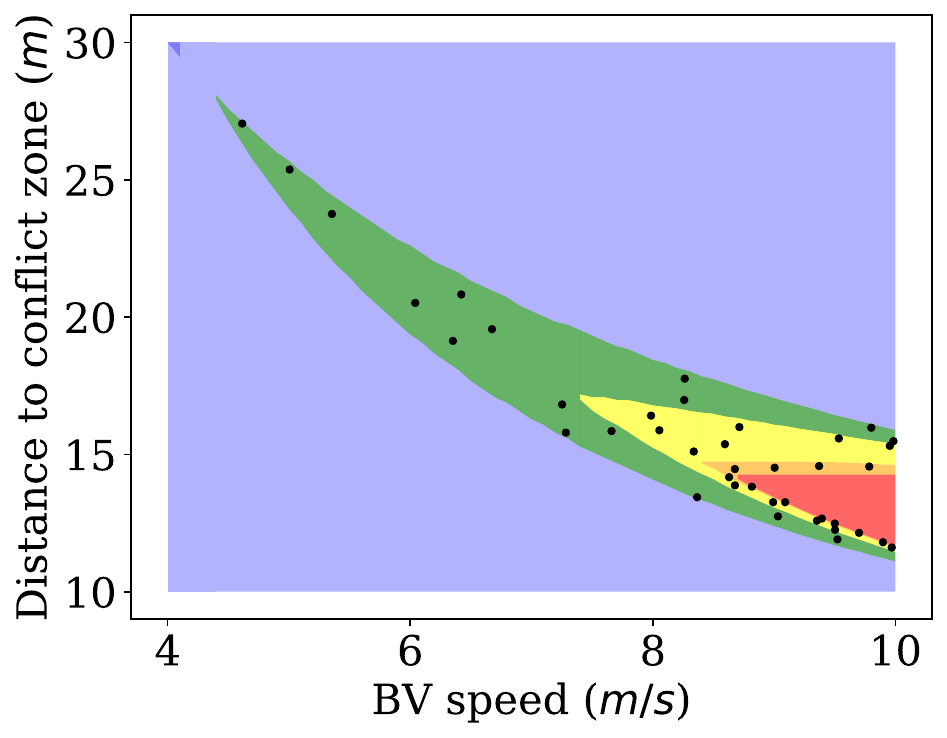}
        \end{minipage}}\quad
    \subfloat[Vehicle into Roundabout\label{subfig:cases_roundabout_inside}]{%
        \begin{minipage}[t]{.22\textwidth}\centering
            \includegraphics[height=3.3cm]{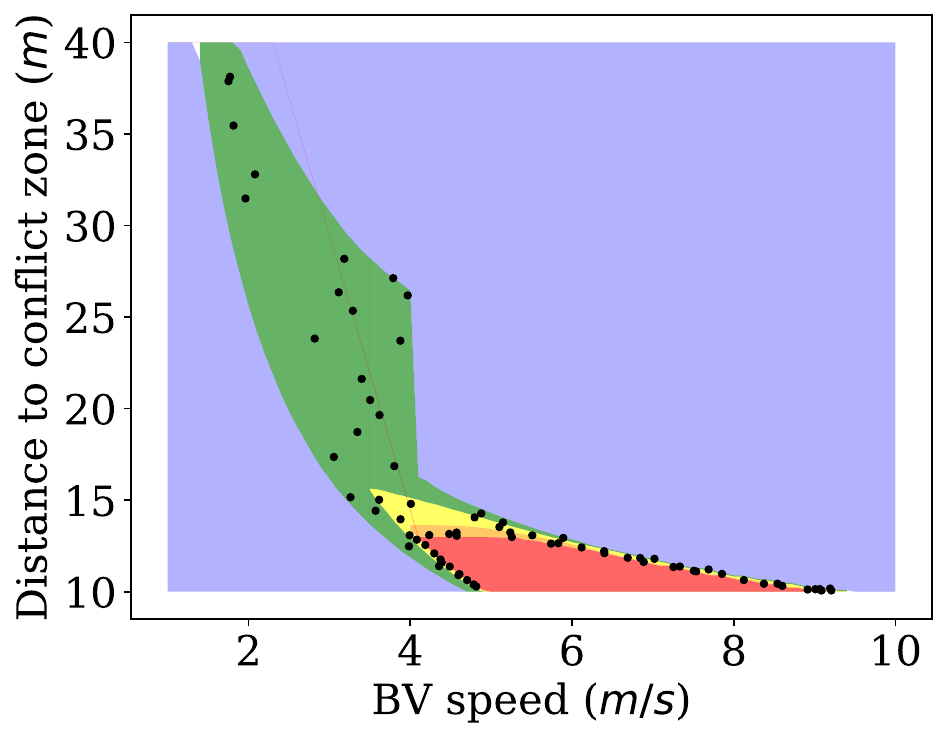}
        \end{minipage}}\quad
    \subfloat[Stopped Vehicle Blocking Lane\label{subfig:cases_vehicle_encroachment}]{%
        \begin{minipage}[t]{.44\textwidth}\centering
            \includegraphics[height=3.3cm]{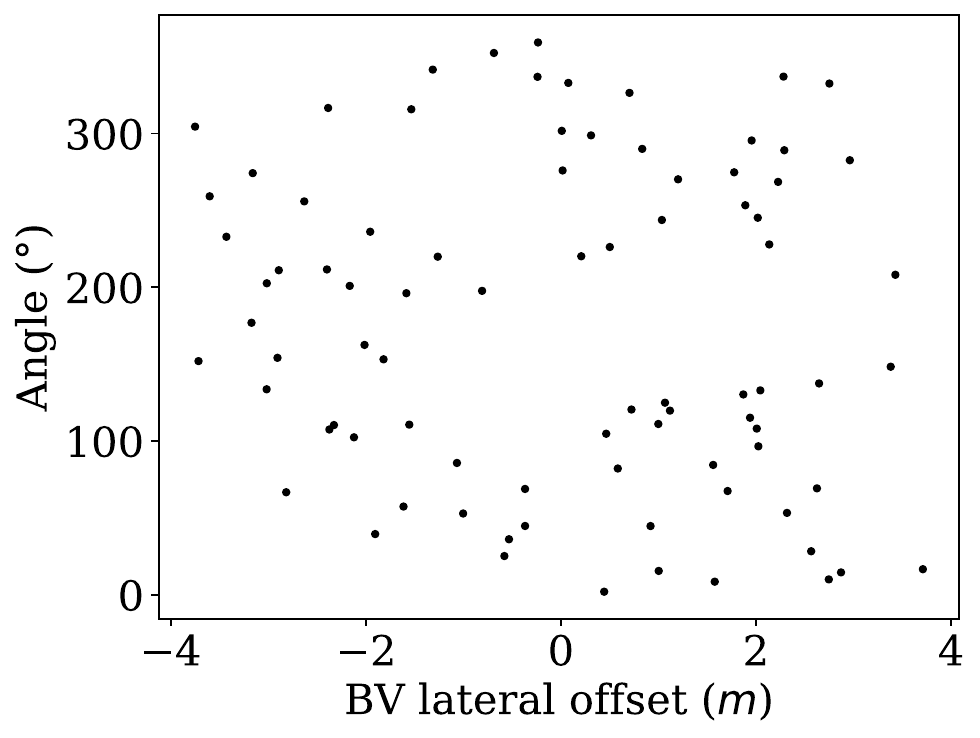}
        \end{minipage}}\quad
    \subfloat[Traffic Light\label{subfig:cases_traffic_signal}]{%
        \begin{minipage}[t]{.44\textwidth}\centering
            \includegraphics[height=3.3cm]{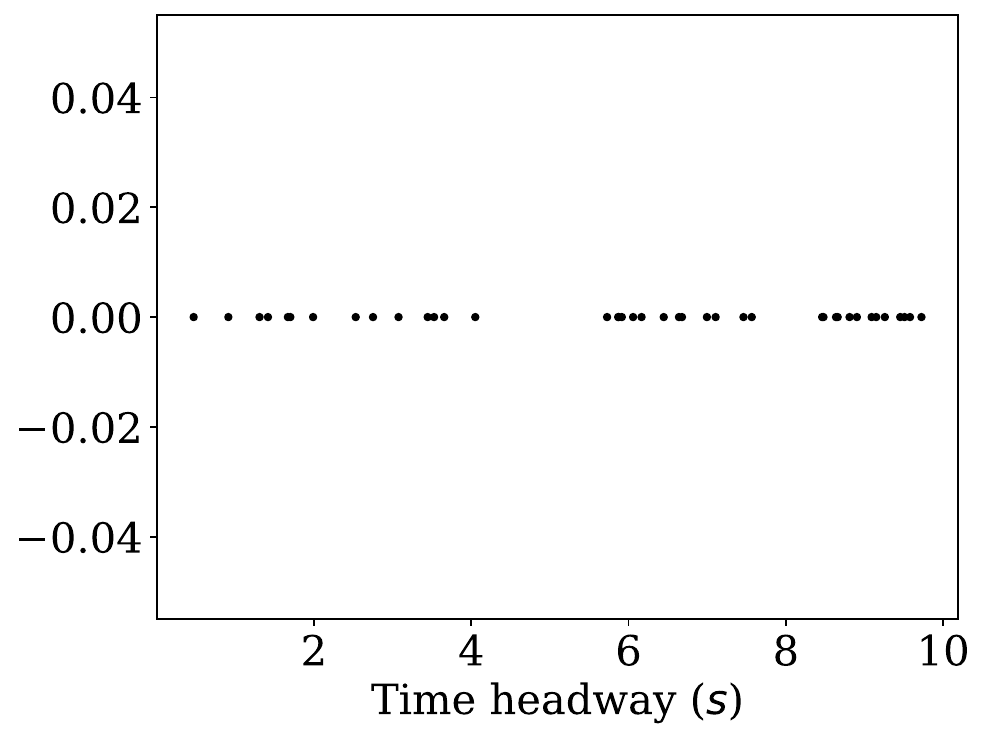}
        \end{minipage}}
    \caption{Risk levels and generated test cases for all 14 competencies.}
    \label{fig:cases}
\end{figure*}

With the risk level boundaries established, each test case is categorized by computing the required minimum AV acceleration, deceleration, or lane-changing time.
The feasibility analysis results for Autoware.Universe are visualized in Fig.~\ref{fig:cases}, where blue, green, yellow, orange, and red areas represent the trivial, low-risk, medium-risk, high-risk, and infeasible levels, respectively.
Risk level visualization is omitted for the stopped vehicle partially blocking lane and traffic light competencies (see Appendix~\ref{sec:bct}-B).

Test cases are then drawn from the low-, medium-, and high-risk regions, shown as black points in Fig.~\ref{fig:cases}.
Trivial cases offer insufficient challenge, and infeasible cases are by definition unsolvable, so both extremes are excluded from sampling.
The number of test cases generated per competency is listed in Table~\ref{tab:case_numbers}.

\begin{table}[!t]
    \caption{Number of test cases per competency}
    \label{tab:case_numbers}
    \begin{center}
        \begin{tabular}{l c}
            Competency & Cases \\ \hline
            Detect and respond to cut-in vehicle                          &  36 \\
            Detect and follow the leading vehicle                         &  40 \\
            Detect and respond to same-direction lane encroachment        &  36 \\
            Detect and respond to oncoming lane encroachment              &  51 \\
            Detect and respond to left-turning vehicle                    &  99 \\
            Execute unprotected left turn                                 &  91 \\
            Detect and respond to right-turning vehicle                   &  72 \\
            Execute right turn                                            &  37 \\
            Detect and respond to VRU crossing without crosswalk          &  33 \\
            Detect and respond to VRU crossing at crosswalk               &  50 \\
            Merge into roundabout with circulating traffic                &  40 \\
            Detect and respond to vehicle merging into roundabout         &  74 \\
            Detect and respond to stopped vehicle partially blocking lane &  84 \\
            Detect and respond to traffic light                           &  40 \\ \hline
            Total                                                         & 783 \\ \hline
        \end{tabular}
    \end{center}
\end{table}

\subsection{Evaluation Metrics}
\label{subsec:metrics}

\subsubsection{Pass/Fail Criterion}

A competency is considered \emph{passed} if no collision occurs in any test case and \emph{failed} if at least one collision is recorded.
The five behavior zones provide a finer-grained characterization of how the AV behaved in cases where it avoided a collision.

\subsubsection{Behavior Zone Classification}

Each test case is classified into one of six mutually exclusive behavioral categories based on the AV's response.
A case is labeled \emph{Collision} if any physical overlap between the AV and the other road user is detected.
Otherwise, the case is assigned to one of five behavior zones drawn from the driving-style taxonomy used in the naturalistic driving literature~\cite{wilson2023argoverse}: \emph{Aggressive}, \emph{Assertive},
\emph{Normal}, \emph{Conservative}, or \emph{Ultra-Conservative}. The zone
assignment is determined by the worst-performing metric across all metrics computed for that competency: if any metric falls into the Aggressive zone, the case is classified as Aggressive regardless of other metrics, and so on in priority order.

The zone boundaries of each metric are listed below.
Where a boundary is supported by standards, regulations, or naturalistic driving studies, the source is cited.
The remaining boundaries are engineering calibrations chosen for the vehicle geometry and case ranges of this study.

\subsubsection{General Metrics}

The following metrics are computed for most competencies (all except b.~Follow Leading Vehicle, m.~Stopped Vehicle Partially Blocking Lane, and n.~Traffic Light) and form the basis of the scorecard zone classification:

\begin{enumerate}

    \item \textbf{Normalized Minimum TTC (Norm Min TTC):} The time-to-collision
    (TTC)~\cite{hayward1972near} is computed continuously as the ratio of the longitudinal gap between
    the AV and the other road user to their closing speed. The minimum TTC
    observed during the encounter, $\text{TTC}_{\min}$, is then normalized
    against two reference minimum TTCs, $\text{TTC}_{\text{ref}}(T_r)$:
    the minimum TTC that would result from a reference braking
    maneuver targeting a $4.0\,\text{m}$ residual
    clearance~\cite{tusimple2025patent} ($0\,\text{m}$ in the cut-in and
    same-direction lane encroachment competencies) if the AV braked after a
    delay $T_r$ from case start, with $T_r=0$ (instant reaction) and
    $T_r=1.5\,\text{s}$, the human-typical reaction-delay
    reference~\cite{green2000long,johansson1971drivers}. Both
    reference values depend on the case's initial speed and clearance and
    are recomputed per case rather than being fixed constants. Because
    $\text{TTC}_{\min}$ is bounded below by $0$ (contact) but unbounded
    above (no interaction), the two references are interpolated on a
    logarithmic time scale rather than linearly, so that a genuine
    collision ($\text{TTC}_{\min}=0$) diverges to $+\infty$ and an
    arbitrarily large clearance diverges to $-\infty$, instead of both
    saturating at a finite, case-dependent value as a linear interpolation
    would:
    \[
        \text{Norm Min TTC} = \frac{\ln\!\big(\text{TTC}_{\text{ref}}(0\,\text{s})\big) - \ln(\text{TTC}_{\min})}
        {\ln\!\big(\text{TTC}_{\text{ref}}(0\,\text{s})\big) - \ln\!\big(\text{TTC}_{\text{ref}}(1.5\,\text{s})\big)}
    \]
    A value of 0 means the AV's actual minimum TTC matched the instant-reaction
    ($T_r=0$) reference maneuver, and a value of 1 means it matched the $1.5\,\text{s}$
    human-reaction-delay reference. Values greater than 1 indicate a minimum TTC
    more critical than even a $1.5\,\text{s}$-delayed reference response and diverge
    to $+\infty$ as $\text{TTC}_{\min}\to 0$. Negative values indicate a minimum TTC
    less critical than the instant-reaction reference and diverge to $-\infty$ as
    $\text{TTC}_{\min}\to\infty$. In short, increasing Norm Min TTC means a
    smaller minimum TTC relative to the references, i.e.\ more aggressive (less
    safety margin), and decreasing Norm Min TTC means a larger minimum TTC,
    i.e.\ more conservative. Values in $[0, 1)$ are classified as Normal, i.e.,
    between the instant-reaction and the human-reaction references. The outer
    boundaries are placed half a reference interval beyond these anchors:
    Assertive for $[1, 1.5)$, Aggressive for $1.5$ and above, Conservative for
    $[-0.5, 0)$, and Ultra-Conservative below $-0.5$.

    \item \textbf{Minimum Longitudinal Clearance (Min Long Clr):} The minimum
    edge-to-edge longitudinal distance between the AV and the conflict zone
    while the other road user occupies it. A negative value means the AV had
    already passed the conflict-zone boundary at that moment without a
    collision, and such cases are classified as Aggressive. Zone boundaries of
    $0.5$/$3.0$/$8.0$/$15.0\,\text{m}$ (Aggressive/Assertive/Normal/Conservative/
    Ultra-Conservative) are used. The upper boundary is consistent with the
    $6.1$--$15.2\,\text{m}$ (20--50\,ft) yield-line setback recommended before
    midblock crosswalks~\cite{mutcd2009}, and the remaining boundaries are
    engineering calibrations.

    \item \textbf{Lateral Distance:} The distance between the other road user
    and the conflict zone while the AV occupies it. For the two VRU
    competencies, it is measured edge-to-edge between the VRU and the AV's path,
    and its zone boundaries of $0.9$/$1.5$/$2.0$/$3.0\,\text{m}$ follow
    lateral passing-clearance rules and observations for vulnerable road users:
    the 3-ft ($0.9\,\text{m}$) passing laws common in the United
    States~\cite{nhtsa_passing}, the $1.5\,\text{m}$ minimum overtaking clearance
    in built-up areas required in Germany~\cite{stvo2013}, and the mean minimum
    clearances of $0.9$--$2.0\,\text{m}$ observed when drivers overtake
    pedestrians~\cite{rasch2020overtake}. For the vehicle-vehicle competencies,
    the boundaries of the Minimum Longitudinal Clearance are used.

    \item \textbf{Post-Encroachment Time (PET):} Used for intersection,
    roundabout, and VRU competencies where the AV and the other road user share a
    conflict zone. PET is the time gap between the moment the first road user
    exits the conflict zone and the moment the second one enters it. A positive
    value means the other road user crossed first (the AV yielded), and a
    negative value means the AV crossed first. Collision cases are assigned
    $\text{PET}=0$. The zones are Aggressive below $1.0\,\text{s}$, Assertive for
    $1.0$--$1.5\,\text{s}$, Normal for $1.5$--$2.0\,\text{s}$, Conservative for
    $2.0$--$5.0\,\text{s}$, and Ultra-Conservative at $5.0\,\text{s}$ or more.
    A PET below $1\,\text{s}$ has been treated as a serious conflict and a PET
    above $2\,\text{s}$ as normal interaction~\cite{varhelyi1998drivers},
    $1.5\,\text{s}$ has been used as a critical PET threshold~\cite{peesapati2018can},
    and $5.0\,\text{s}$ is the default PET threshold of the Surrogate Safety
    Assessment Model~\cite{gettman2008surrogate}.

\end{enumerate}

Two additional behavior-diagnosis plots accompany each competency.
\emph{Reaction Time} is the time from case start to AV braking onset, with
Fast/Good/Slow boundaries at $0.7\,\text{s}$ and $1.5\,\text{s}$. These values
correspond to the perception-brake time of drivers who expect to brake and of
drivers surprised by an unexpected event~\cite{green2000long}, and $0.7\,\text{s}$
also matches the 15th-percentile brake-response time observed at yellow
onset~\cite{gates2007analysis}.
\emph{Deceleration Adequacy} plots the required deceleration against the
\emph{Normalized Deceleration}, i.e., the AV's average deceleration normalized,
per case, between two required-deceleration anchors: the deceleration needed to
preserve the conservative target clearance (value 0) and the deceleration needed
to keep only the aggressive, minimal clearance (value 1). The two clearances are
$4.0\,\text{m}$ and $0.75\,\text{m}$~\cite{tusimple2025patent}, and $0.5\,\text{m}$
and $0\,\text{m}$ in the cut-in and same-direction lane encroachment competencies.
Required decelerations are bounded by the $2.87\,\text{m/s}^2$
acceleration~\cite{bokare2017acceleration} and $-7.06\,\text{m/s}^2$
deceleration~\cite{mumford1997inverted} limits.
Increasing Normalized Deceleration means the AV braked more softly than
the anchors require (under-braking), and decreasing it means the AV braked
harder than required (over-braking). The zones use the same convention as Norm
Min TTC: Normal for $[0, 1)$, Assertive for $[1, 1.5)$, Aggressive for $1.5$ and
above, Conservative for $[-0.5, 0)$, and Ultra-Conservative below $-0.5$.
These plots do not feed into the zone scorecard but help diagnose why the AV was classified into a particular zone.

Some competencies need different metrics. The cut-in competency replaces Minimum Longitudinal Clearance and PET with two metrics that better characterize the AV's response to a sudden lane change:

\begin{enumerate}

    \item \textbf{Minimum Gap Ratio:} The minimum longitudinal gap between the
    AV and the cut-in vehicle during the maneuver, divided by the longitudinal
    gap at the start of the case. A ratio near zero indicates the AV almost
    reached the BV, and a ratio near one indicates the AV maintained the original
    spacing throughout. Zone boundaries of $0.2$/$0.4$/$0.6$/$0.8$
    (Aggressive/Assertive/Normal/Conservative/Ultra-Conservative) are engineering
    calibrations.

    \item \textbf{Speed Recovery Time:} The time from the end of the BV's lane
    change until the AV recovers to 90\% of its pre-cut-in speed. A short
    recovery time indicates little or no braking, and a long recovery time
    indicates the AV braked more than necessary and took time to re-accelerate.
    The zones are Aggressive below $4\,\text{s}$, Assertive for $4$--$8\,\text{s}$,
    Normal for $8$--$16\,\text{s}$, Conservative for $16$--$30\,\text{s}$, and
    Ultra-Conservative at $30\,\text{s}$ or more. The Normal range spans the
    relaxation period after which human drivers who have been cut in on return to
    normal car following, reported as about $8$--$16\,\text{s}$ in naturalistic
    trajectory data~\cite{ma2008comparisons,duret2011passing}. The upper boundary
    of $30\,\text{s}$ is about twice this human value.

\end{enumerate}

In terms of Detect and Respond to Oncoming Lane Encroachment, because the required response is a lateral evasion rather than braking, the following metrics replace Norm Min TTC, Minimum Longitudinal Clearance and PET:

\begin{enumerate}

    \item \textbf{Normalized Pre-Lateral TTC:} The minimum TTC observed
    before the AV begins its lateral movement, normalized on the same log
    scale as Norm Min TTC, with the references set to the TTC at case start
    (value 0, instant evasion) and that TTC minus $1.5\,\text{s}$ (value 1,
    human-typical delay). Increasing values mean a later
    evasion, i.e.\ more aggressive, and decreasing values mean an earlier
    evasion, i.e.\ more conservative. The zone boundaries are the same as for
    Norm Min TTC.

    \item \textbf{Lateral Distance at Center Alignment:} The center-to-center
    lateral offset between the AV and the oncoming BV at the moment their
    longitudinal centers are aligned (i.e., the point of closest lateral
    approach). Zone boundaries of $0.3$/$0.8$/$1.5$/$2.5\,\text{m}$
    (Aggressive/Assertive/Normal/Conservative/Ultra-Conservative) are
    engineering calibrations based on the lane geometry.

    \item \textbf{Lane Return Time:} The time from case start until the AV's
    center returns to within 0.2\,m of the original lane center. Return times
    of $3$--$8\,\text{s}$ are classified as Normal, consistent with the 30th to
    70th percentiles ($3.6$--$8.1\,\text{s}$) of human lane-change
    durations~\cite{huang2020developing}. Times of $8$--$13\,\text{s}$ and above
    $13\,\text{s}$ (about the 90th percentile, $13.1\,\text{s}$) are Assertive and
    Aggressive, and times of $1$--$3\,\text{s}$ and below $1\,\text{s}$ (a minimal
    evasion) are Conservative and Ultra-Conservative.

\end{enumerate}

The car-following competency is divided into three stages matching the BV's motion profile, each with dedicated metrics:

\begin{enumerate}

    \item \textbf{Startup Delay} (Stage~1, Acceleration): Time from the
    moment the BV begins to accelerate from rest until the AV starts to
    accelerate. The zones are Aggressive below $0.3\,\text{s}$, Assertive for
    $0.3$--$0.8\,\text{s}$, Normal for $0.8$--$2.0\,\text{s}$, Conservative for
    $2.0$--$3.5\,\text{s}$, and Ultra-Conservative above $3.5\,\text{s}$. The
    Normal range covers the $1.1$--$1.8\,\text{s}$ increments in response time
    between successive queued drivers at signalized intersections~\cite{yang2012driver}
    and the $2\,\text{s}$ start-up lost time commonly assumed in signal
    timing~\cite{koonce2008traffic}.

    \item \textbf{Minimum Time Headway} (Stage~2, Cruise): The minimum
    time headway (THW $= d/v_\text{AV}$, where $d$ is the edge-to-edge gap)
    observed while the BV travels at constant speed. The zones are Aggressive
    below $0.5\,\text{s}$, Assertive for $0.5$--$0.9\,\text{s}$, Normal for
    $0.9$--$2.2\,\text{s}$, Conservative for $2.2$--$3.0\,\text{s}$, and
    Ultra-Conservative above $3.0\,\text{s}$. The Normal range covers the time-gap
    settings of adaptive cruise control systems~\cite{iso15622} and the mean
    naturalistic highway headway of $1.42\,\text{s}$~\cite{huang2020developing}.
    A headway below $0.9\,\text{s}$ is treated as an endangering distance in
    German traffic law~\cite{kruber2019highway}, $0.5\,\text{s}$ is close to the
    2.5th percentile of human headways~\cite{huang2020developing}, and $3\,\text{s}$
    is the headway recommended in driving manuals~\cite{ayres2001preferred}.

    \item \textbf{Gap at AV Braking Onset} (Stage~3, Deceleration): The
    edge-to-edge clearance between the AV and the stopped BV at the moment the
    AV begins to brake. A small gap indicates the AV detected the BV's stop
    late, and a large gap indicates an early, conservative response. Zone
    boundaries of $2$/$5$/$10$/$15\,\text{m}$ are engineering calibrations.

    \item \textbf{Final Stop Gap} (Stage~3, Deceleration): The
    edge-to-edge clearance once both vehicles have come to rest. A negative
    value indicates a collision. The zones are Aggressive below $0\,\text{m}$,
    Assertive for $0$--$1\,\text{m}$, Normal for $1$--$3\,\text{m}$, Conservative
    for $3$--$5\,\text{m}$, and Ultra-Conservative above $5\,\text{m}$. The Normal
    range covers the measured standstill spacing of about $1.4\,\text{m}$ between
    queued vehicles~\cite{jin2009modeling} and the $2\,\text{m}$ jam distance of
    the Intelligent Driver Model~\cite{treiber2000congested}.

\end{enumerate}

For detecting and responding to traffic light competency, each traffic light case is either a \emph{stop case} (the AV decelerates to stop before the line) or a \emph{pass case} (the AV accelerates to clear the line while the light is yellow).
The zone boundaries below are expressed through the constant deceleration or acceleration that a given THW requires at the AV speed of $5\,\text{m/s}$ and a yellow phase of $4\,\text{s}$.
Each case type has a dedicated metric:

\begin{enumerate}

    \item \textbf{THW at Yellow Onset (Stop Cases):} The time headway between
    the AV's front bumper and the stop line at the moment the light turns yellow.
    A stop case is classified as Aggressive if stopping before the line requires
    more than $4.5\,\text{m/s}^2$, Assertive if it requires $3.0$--$4.5\,\text{m/s}^2$,
    and Normal otherwise. $3.0\,\text{m/s}^2$ is the deceleration commonly assumed
    in the yellow-interval formula~\cite{koonce2008traffic,rakha2011development},
    and $4.5\,\text{m/s}^2$ is the upper end of the values assumed in
    practice~\cite{rakha2011development}.

    \item \textbf{THW at Yellow Onset (Pass Cases):} The same measurement for
    cases where the AV crosses the stop line. A pass case is classified as
    Normal if the AV can reach the stop line within the yellow phase with an
    acceleration of at most $0.5\,\text{m/s}^2$, Assertive for
    $0.5$--$2.0\,\text{m/s}^2$, and Aggressive above $2.0\,\text{m/s}^2$. Drivers who
    proceed through a yellow light typically maintain their approach speed
    rather than accelerate~\cite{rakha2011development}.

    \item \textbf{Stop Distance Relative to Stop Line:} The AV's final
    longitudinal position minus the stop line position. A negative value means
    the AV stopped before the line, and a positive value means it overshot.
    Stops more than $1\,\text{m}$ before the line are Ultra-Conservative and stops
    $0$--$1\,\text{m}$ before it are Conservative. An overshoot of up to
    $1.2\,\text{m}$ is Normal, $1.2$--$3.0\,\text{m}$ is Assertive, and more than
    $3.0\,\text{m}$ is Aggressive. These boundaries follow the stop-line placement
    of at least 4\,ft ($1.2\,\text{m}$) before a crosswalk and the minimum crosswalk
    width of 6\,ft ($1.8\,\text{m}$)~\cite{mutcd2009}, so that an Assertive stop
    ends inside the crosswalk and an Aggressive stop ends beyond it.

    \item \textbf{Braking THW:} The THW between the AV's front bumper and the
    stop line at the moment the AV initiates braking. A shorter braking THW
    requires a harder deceleration to avoid crossing the line. The zone
    boundaries correspond to required decelerations of $3.9$, $3.0$, and
    $2.2\,\text{m/s}^2$, i.e., the 85th, 50th, and 15th percentiles of driver
    deceleration at yellow onset~\cite{gates2007analysis}, and $1.0\,\text{m/s}^2$
    (engineering calibration): Aggressive above $3.9\,\text{m/s}^2$, Assertive for
    $3.0$--$3.9\,\text{m/s}^2$, Normal for $2.2$--$3.0\,\text{m/s}^2$, Conservative
    for $1.0$--$2.2\,\text{m/s}^2$, and Ultra-Conservative below $1.0\,\text{m/s}^2$.

\end{enumerate}

\subsubsection{Detect and Respond to Stopped Vehicle Partially Blocking Lane: Specific Metric}

Because the BV is stationary and risk is fully determined by the AV's response, only one metric is used:

\begin{enumerate}

    \item \textbf{Stop Distance:} The edge-to-edge clearance between the AV and
    the obliquely parked BV when the AV comes to rest. A negative value
    indicates a collision. The zones are the same as for the Final Stop Gap in
    the car-following competency (Normal for $1$--$3\,\text{m}$), based on the
    same standstill-spacing evidence~\cite{jin2009modeling,treiber2000congested}.

\end{enumerate}

\section{Evaluation Results of the Behavioral Competency Test}
\label{sec:results}

\begin{figure*}[!p]
    \centering
    \includegraphics[width=\textwidth]{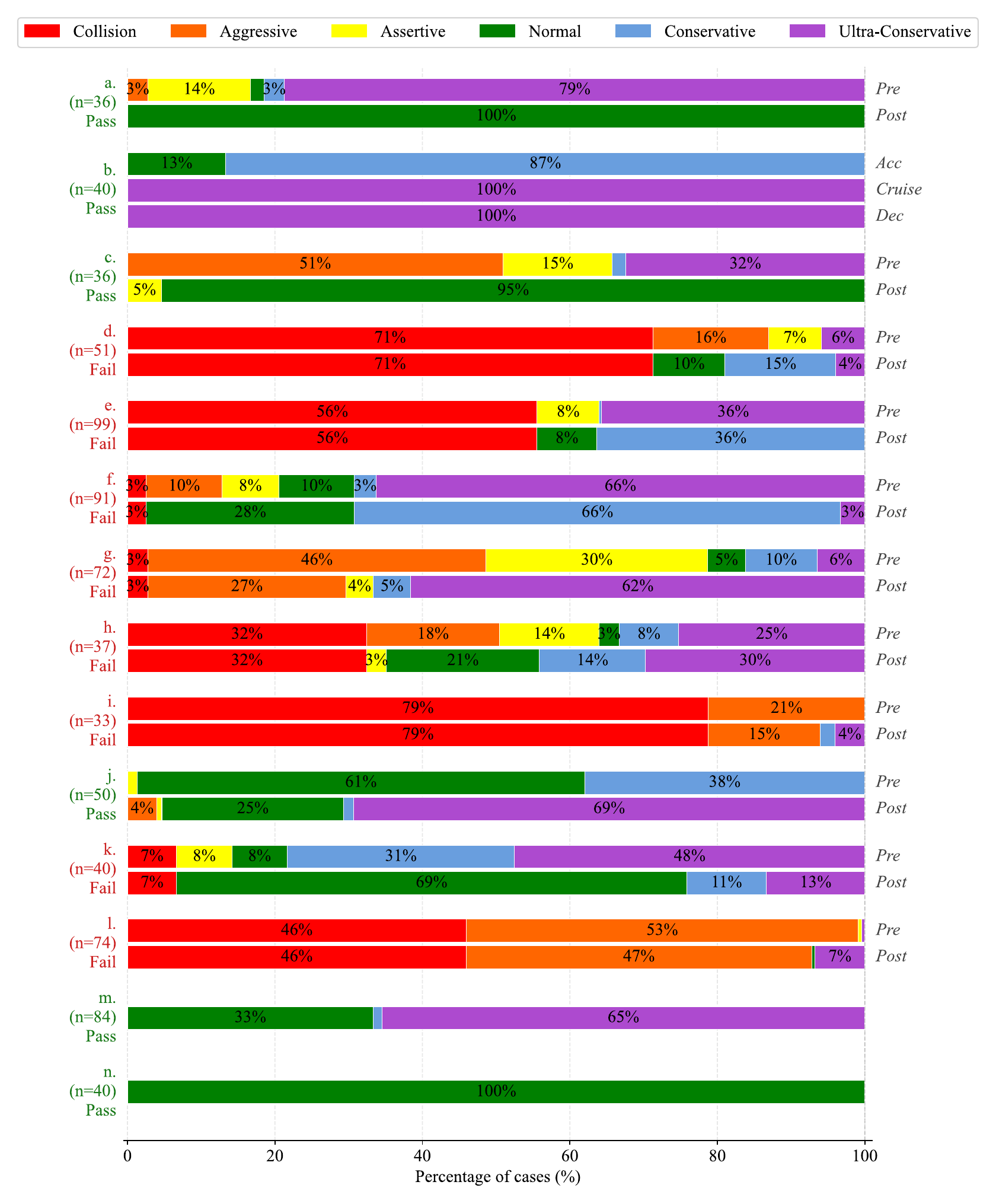}
    \caption{Behavioral zone distribution across all 14 competencies.}
    \label{fig:summary}
\end{figure*}

This section presents a detailed analysis of each of the 14 behavioral competencies.
For each competency, we report the overall pass/fail verdict, examine the pre- and post-encroachment stages with the corresponding scatter plots, analyze the AV's braking timing and intensity, and report the behavioral zone distribution for each metric.
Fig.~\ref{fig:summary} provides an overview of the behavioral zone distribution across all competencies and phases. Each row pair shows the pre- and post-encroachment phase distributions of one competency, with the pass/fail verdict indicated on the left.
In the per-competency figures, each marker represents one test case, filled with the color of its behavioral zone and outlined by its risk level (red, orange, and green for high, medium, and low risk). Collision cases are shown in red, and each case is labeled by its round and case number (e.g., R2-40 denotes case 40 in round 2).

Across the 14 competencies, Autoware.Universe passes 6 and fails 8, as summarized in Fig.~\ref{fig:summary}.
The six passing competencies are a.~Detect and respond to cut-in vehicle ($n=36$), b.~Detect and follow the leading vehicle ($n=40$), c.~Detect and respond to same-direction lane encroachment ($n=36$), j.~Detect and respond to VRU crossing at crosswalk ($n=50$), m.~Detect and respond to stopped vehicle partially blocking lane ($n=84$), and n.~Detect and respond to traffic light ($n=40$).
Among the eight failing competencies, i.~Detect and respond to VRU crossing without crosswalk records the highest collision rate at $78.8\%$ ($n=33$), followed by d.~Detect and respond to oncoming lane encroachment at $71.2\%$ ($n=51$), e.~Detect and respond to left-turning vehicle at $55.6\%$ ($n=99$), l.~Detect and respond to vehicle merging into roundabout at $45.9\%$ ($n=74$), and h.~Execute right turn at $32.4\%$ ($n=37$).
f.~Execute unprotected left turn, g.~Detect and respond to right-turning vehicle, and k.~Merge into roundabout with circulating traffic fail with lower collision rates of $2.6\%$, $2.8\%$, and $6.7\%$, respectively.
Among the passing competencies, the AV mostly behaves conservatively.
In a.~Detect and respond to cut-in vehicle, $81.5\%$ of cases are Conservative or Ultra-Conservative in the pre-encroachment phase.
In b.~Detect and follow the leading vehicle, the cruise and deceleration phases are each $100\%$ Ultra-Conservative and the acceleration phase is $86.7\%$ Conservative.
In j.~Detect and respond to VRU crossing at crosswalk, $69.3\%$ of post-encroachment cases are Ultra-Conservative, and in m.~Detect and respond to stopped vehicle partially blocking lane, $65.5\%$ of cases are Ultra-Conservative.
The exception is c.~Detect and respond to same-direction lane encroachment, where $50.9\%$ of pre-encroachment cases fall in the Aggressive zone.
After a cut-in or same-direction lane encroachment, the AV resumes normal driving within a human-like recovery time in $100\%$ and $95.4\%$ of cases, respectively.
n.~Detect and respond to traffic light is the only competency in which every case is classified as Normal ($n=40$).

\subsection{Detect and Respond to Cut-In Vehicle}
\label{subsec:cut_in}

\begin{figure*}[!p]
    \centering
    \subfloat[Scorecard.\label{fig:ci_scorecard}]{%
        \includegraphics[width=0.48\textwidth]{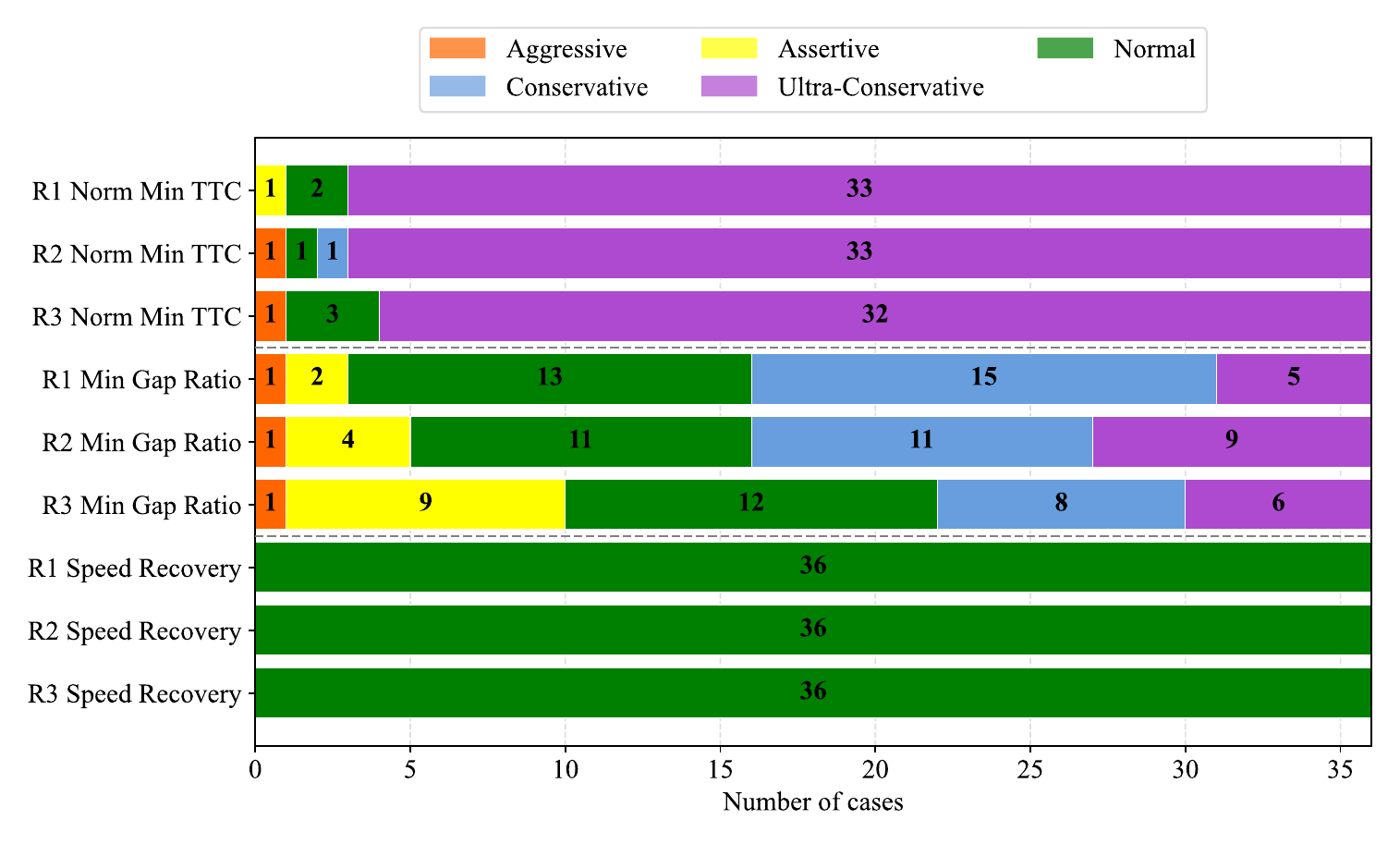}%
    }
    \hfill
    \subfloat[Normalized Minimum TTC vs.\ TTC at case start.\label{fig:ci_stage1}]{%
        \includegraphics[width=0.48\textwidth]{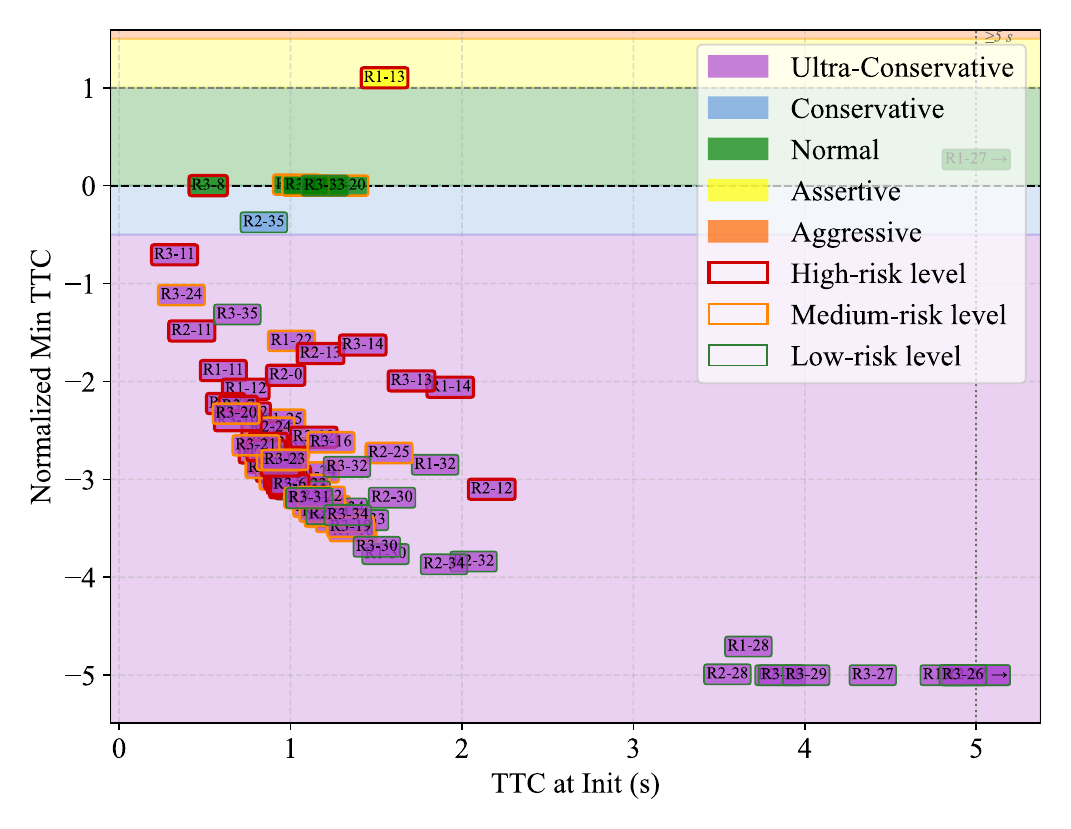}%
    }\\[4pt]
    \subfloat[Minimum Gap Ratio vs.\ Longitudinal Gap at Case Start.\label{fig:ci_stage2}]{%
        \includegraphics[width=0.48\textwidth]{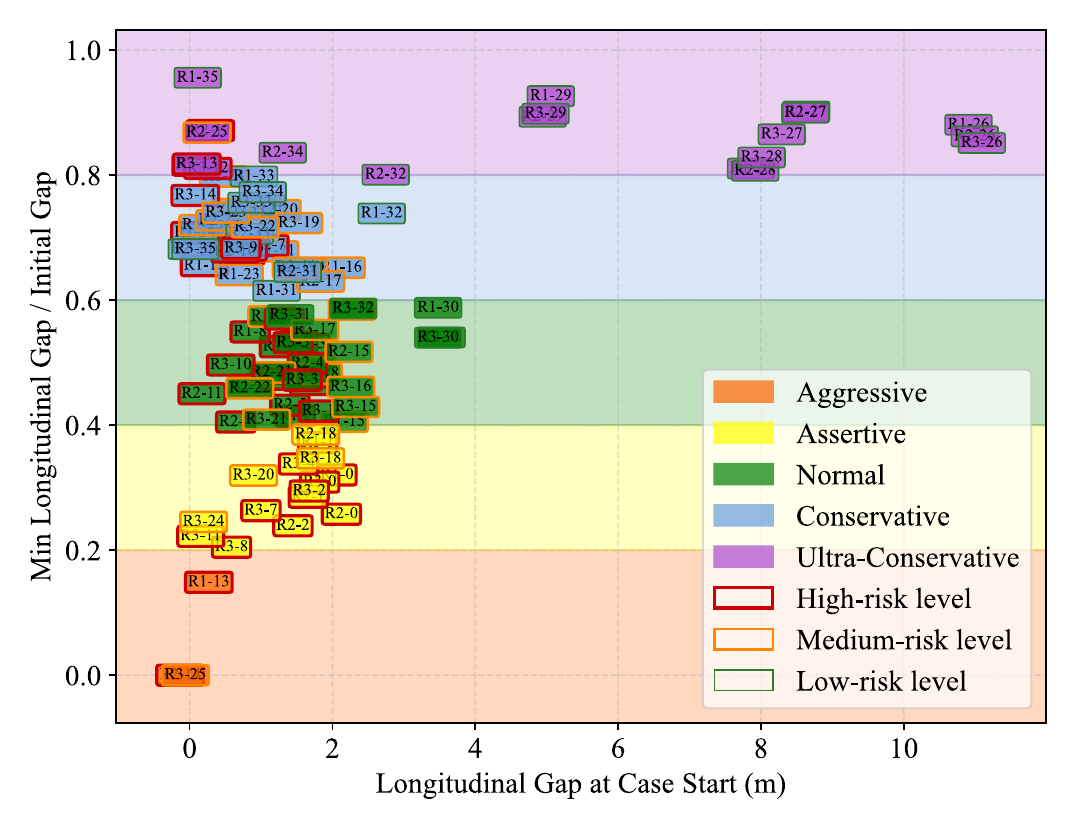}%
    }
    \hfill
    \subfloat[Speed Recovery Time vs.\ Minimum Gap Ratio.\label{fig:ci_stage3}]{%
        \includegraphics[width=0.48\textwidth]{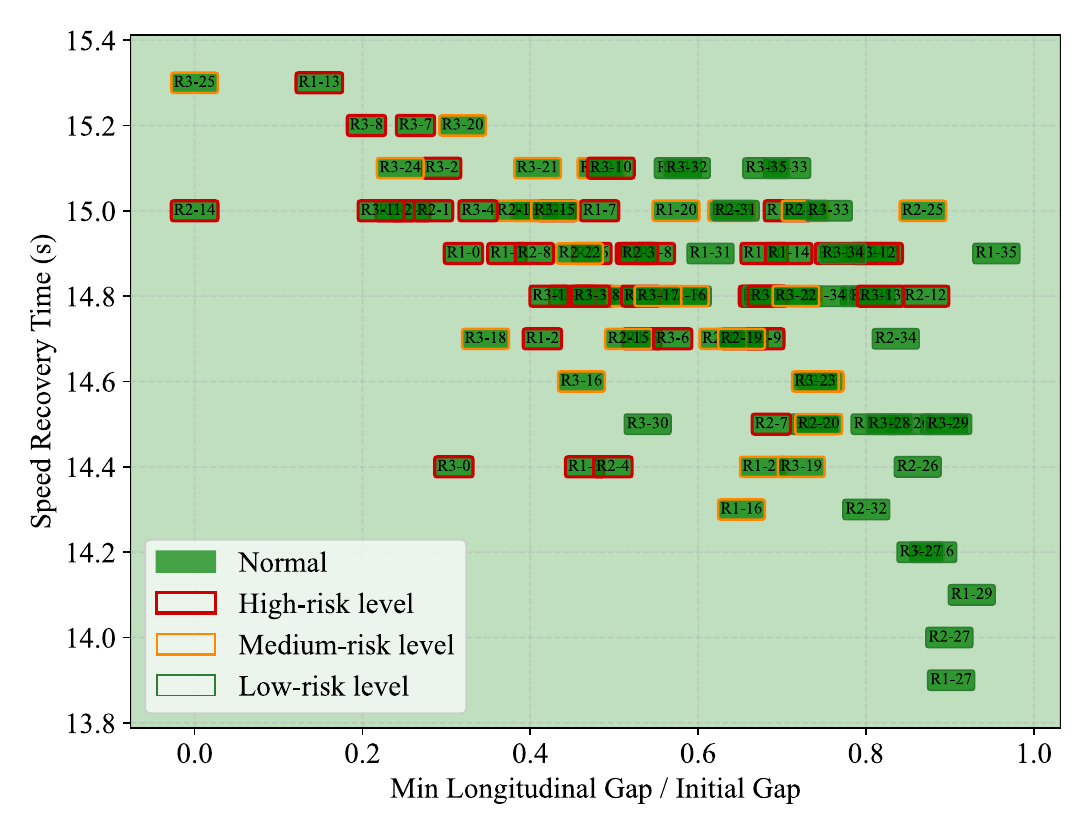}%
    }\\[4pt]
    \subfloat[Reaction time vs.\ TTC at case start.\label{fig:ci_reaction}]{%
        \includegraphics[width=0.48\textwidth]{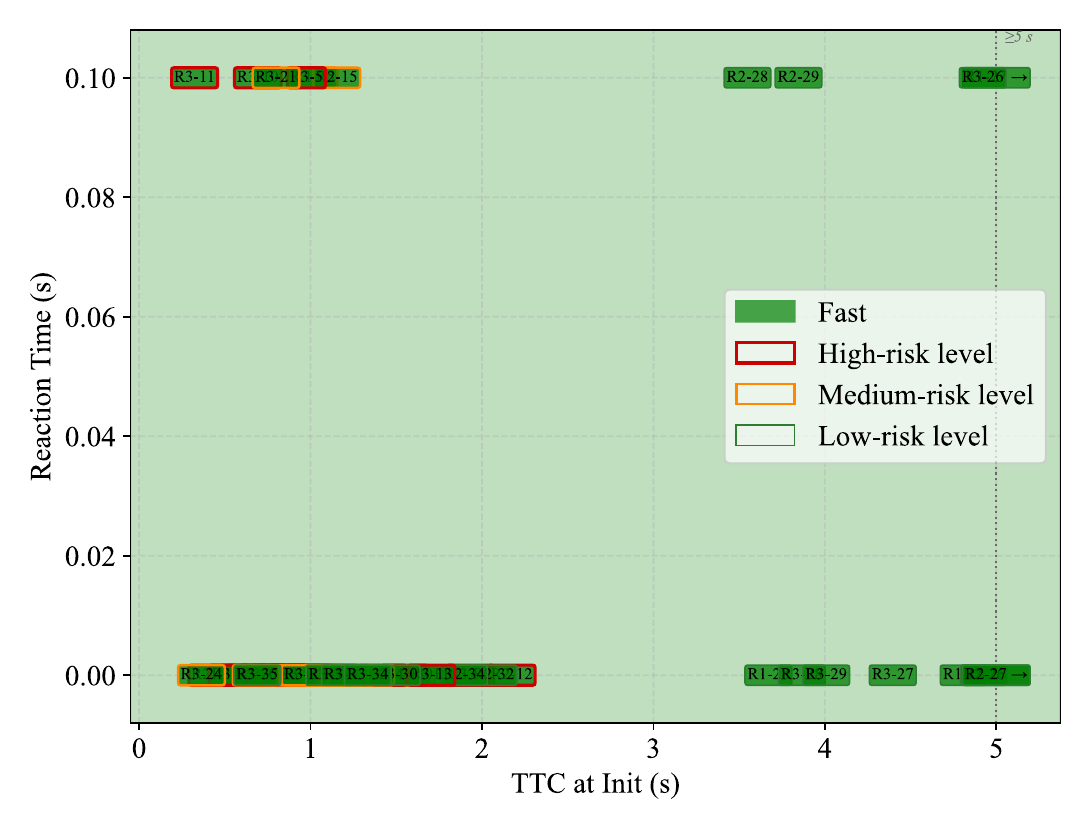}%
    }
    \hfill
    \subfloat[Normalized Deceleration vs.\ required deceleration.\label{fig:ci_decel}]{%
        \includegraphics[width=0.48\textwidth]{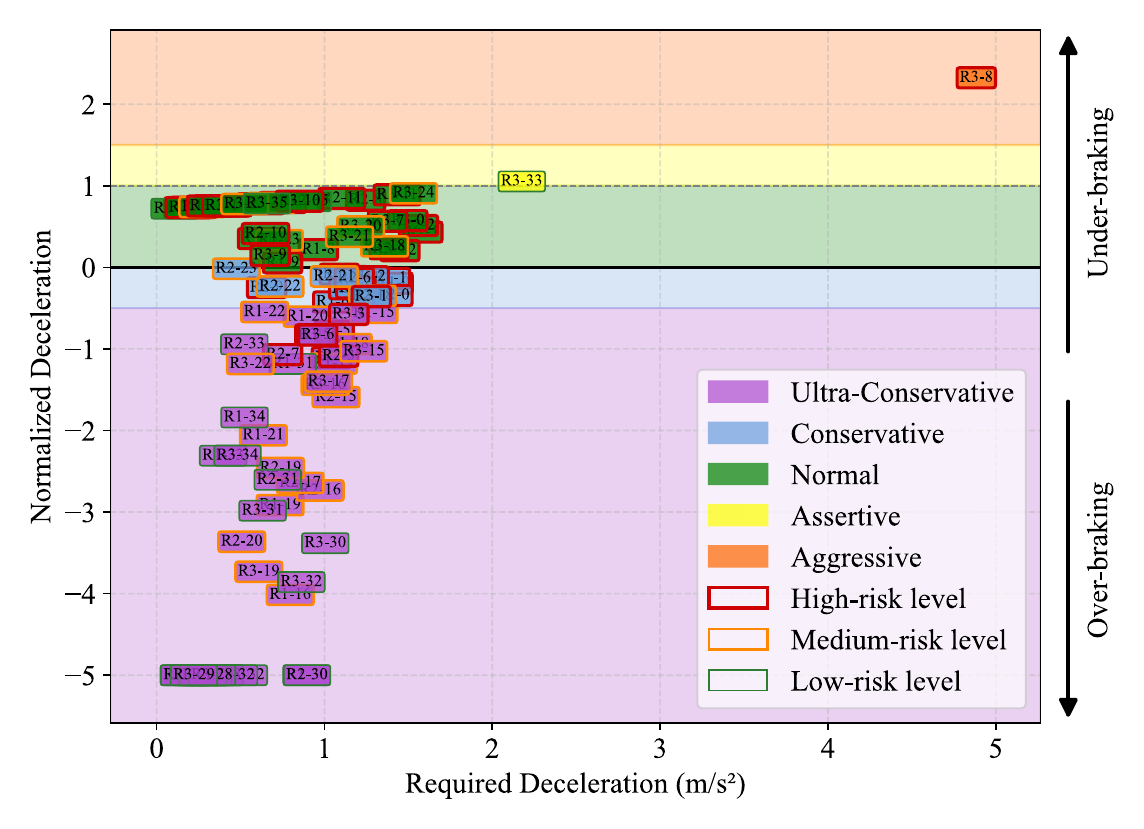}%
    }
    \caption{Analysis results for the detect and respond to cut-in vehicle competency.}
    \label{fig:ci_all}
\end{figure*}

\subsubsection*{Competency Certification}

Autoware.Universe \textbf{passes} this competency with no collision in any of the $108$ runs ($n = 36$ per round, collision rate $= 0\%$).
No rear-end contact with the cutting-in vehicle occurs in any run.

\subsubsection*{Behavioral Cautiousness Assessment}

\textbf{Stage 1: Normalized Minimum TTC.}
In the scorecard (Fig.~\ref{fig:ci_scorecard}), $98$ of $108$ runs ($91\%$) are Ultra-Conservative.
The remaining runs are $2$ Aggressive, $1$ Assertive, $6$ Normal, and $1$ Conservative.
The two Aggressive runs (R2-14 and R3-25) start with the two vehicles already overlapping longitudinally (initial gap $-0.13$ and $-0.06\,\text{m}$).
In the scatter plot (Fig.~\ref{fig:ci_stage1}), $86$ of the $93$ runs with TTC at case start of $0.32$--$2.17\,\text{s}$ are Ultra-Conservative (norm min TTC $-3.87$ to $-0.71$, median $-2.87$).
Of the other seven, R1-13 is Assertive, R2-35 is Conservative, and the remaining five are Normal.
The $12$ low-risk runs of cases~26--29 have TTC at case start of $3.5$--$5.2\,\text{s}$.
Nine of them sit at the $-5$ floor, R1-28 and R2-28 are at $-4.71$ and $-4.99$, and R1-27 falls in the Normal zone ($0.27$).
Norm min TTC decreases with TTC at case start ($r = -0.53$).

\textbf{Stage 2: Minimum Gap Ratio.}
In the scorecard, the counts (Aggressive / Assertive / Normal / Conservative / Ultra-Conservative) are $3/15/36/34/20$.
In the scatter plot (Fig.~\ref{fig:ci_stage2}), the gap ratio depends on the initial longitudinal gap.
The $82$ runs with initial gaps below $2\,\text{m}$ ($43$ high-, $27$ medium-, and $12$ low-risk) span all five zones ($32$ Conservative, $27$ Normal, $13$ Assertive, $7$ Ultra-Conservative, $3$ Aggressive), with a median ratio of $0.57$.
The three Aggressive runs are R1-13 (ratio $0.15$, high-risk) and the two runs that start overlapping, R2-14 (high-risk) and R3-25 (medium-risk), both with a ratio of $0$.
The $14$ runs with initial gaps of $2$--$4\,\text{m}$ are mostly Normal ($9$), with $2$ Assertive, $2$ Conservative, and $1$ Ultra-Conservative.
All $12$ runs with initial gaps of $4.9$--$11.1\,\text{m}$ (low-risk cases~26--29) are Ultra-Conservative (ratio $0.81$--$0.93$).
Over all runs, the median ratio is $0.60$.

\textbf{Stage 3: Speed Recovery Time.}
In the scorecard, speed recovery is Normal ($8$--$16\,\text{s}$) in all $108$ runs.
In the scatter plot (Fig.~\ref{fig:ci_stage3}), recovery time lies between $13.9$ and $15.3\,\text{s}$ (median $14.8\,\text{s}$), a spread of $1.4\,\text{s}$ in the upper half of the Normal zone.
Recovery time decreases as the Min Gap Ratio increases ($r = -0.55$).
The longest recoveries ($15.3\,\text{s}$) belong to R1-13 and R3-25, two of the three lowest gap ratios, and the shortest ($13.9\,\text{s}$) to R1-27 (ratio $0.90$).

\subsubsection*{Behavior Diagnosis}

\textbf{Reaction Time.}
All $108$ runs fall in the Fast zone ($<0.7\,\text{s}$) of the reaction time plot (Fig.~\ref{fig:ci_reaction}).
In $98$ runs the reaction time is $0\,\text{s}$, with braking onset at the case start.
The remaining $10$ runs react within $0.1\,\text{s}$: six at TTC at case start of $0.32$--$1.15\,\text{s}$ (R1-15, R2-15, R3-5, R3-10, R3-11, R3-21) and four low-risk runs at $3.5$--$5.1\,\text{s}$ (R1-27, R2-28, R2-29, R3-26).

\textbf{Deceleration Adequacy.}
In the deceleration adequacy plot (Fig.~\ref{fig:ci_decel}), the AV's average deceleration is between $1.65$ and $2.07\,\text{m/s}^2$ in every run, with a median of $2.01\,\text{m/s}^2$ and an interquartile range of $1.97$--$2.03\,\text{m/s}^2$, while the required deceleration spans $0.11$--$4.88\,\text{m/s}^2$.
Overall, $51$ runs are Ultra-Conservative, $16$ Conservative, $37$ Normal, $1$ Assertive, and $1$ Aggressive.
For $a_{\text{req}} < 0.5\,\text{m/s}^2$ ($n = 28$), $16$ runs are Ultra-Conservative, $1$ Conservative, and $11$ Normal.
For $0.5$--$1\,\text{m/s}^2$ ($n = 38$) the split is $23$ Ultra-Conservative, $2$ Conservative, and $13$ Normal, and for $1$--$2\,\text{m/s}^2$ ($n = 38$) it is $12$ Ultra-Conservative, $13$ Conservative, and $13$ Normal.
Seventeen low-risk runs (cases~26--30, 32, and 34) sit at the $-5$ floor with $a_{\text{req}}$ of only $0.16$--$0.90\,\text{m/s}^2$, the strongest over-braking.
Only two runs require more than $2\,\text{m/s}^2$: R3-33 ($a_{\text{req}} = 2.18\,\text{m/s}^2$, Assertive, norm decel $1.05$) and R3-8 ($a_{\text{req}} = 4.88\,\text{m/s}^2$, the highest in the dataset, Aggressive, norm decel $2.32$).
Both brake at the same ${\approx}2.0\,\text{m/s}^2$ as the other runs.
R3-8 is high-risk, ends with a Min Gap Ratio of $0.20$ (Assertive), and does not collide.

\subsection{Detect and Follow the Leading Vehicle}
\label{subsec:car_following}

\begin{figure*}[!t]
    \centering
    \subfloat[AV startup delay vs.\ BV acceleration.\label{fig:cf_stage1}]{%
        \includegraphics[width=0.48\textwidth]{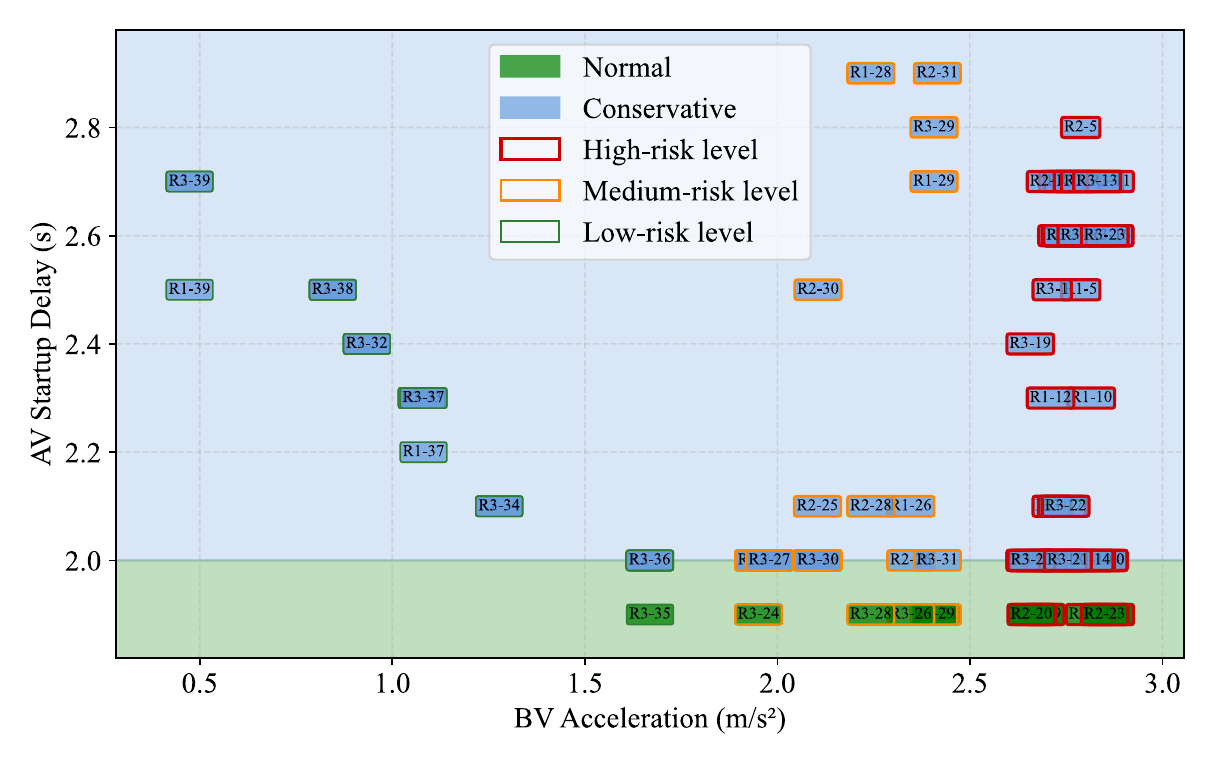}%
    }
    \hfill
    \subfloat[Minimum time headway vs.\ BV cruise speed.\label{fig:cf_stage2}]{%
        \includegraphics[width=0.48\textwidth]{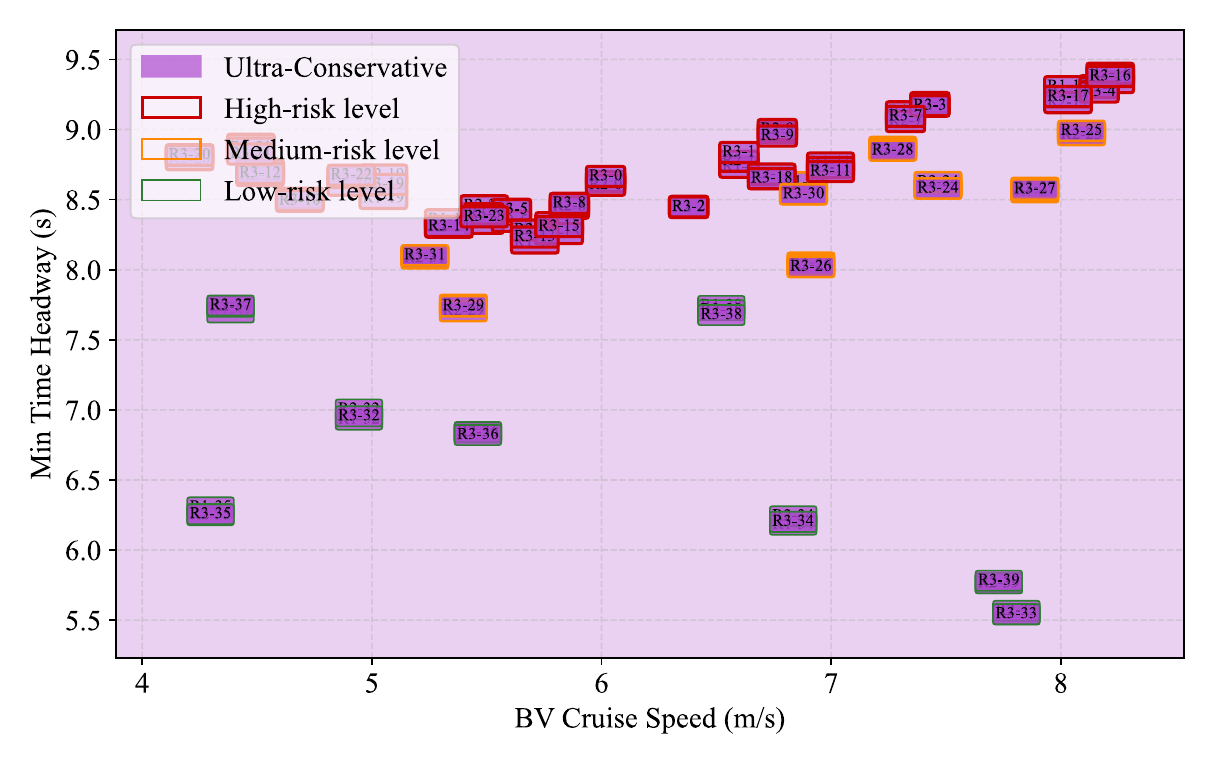}%
    }\\[4pt]
    \subfloat[Gap at AV braking onset vs.\ BV deceleration magnitude.\label{fig:cf_stage3a}]{%
        \includegraphics[width=0.48\textwidth]{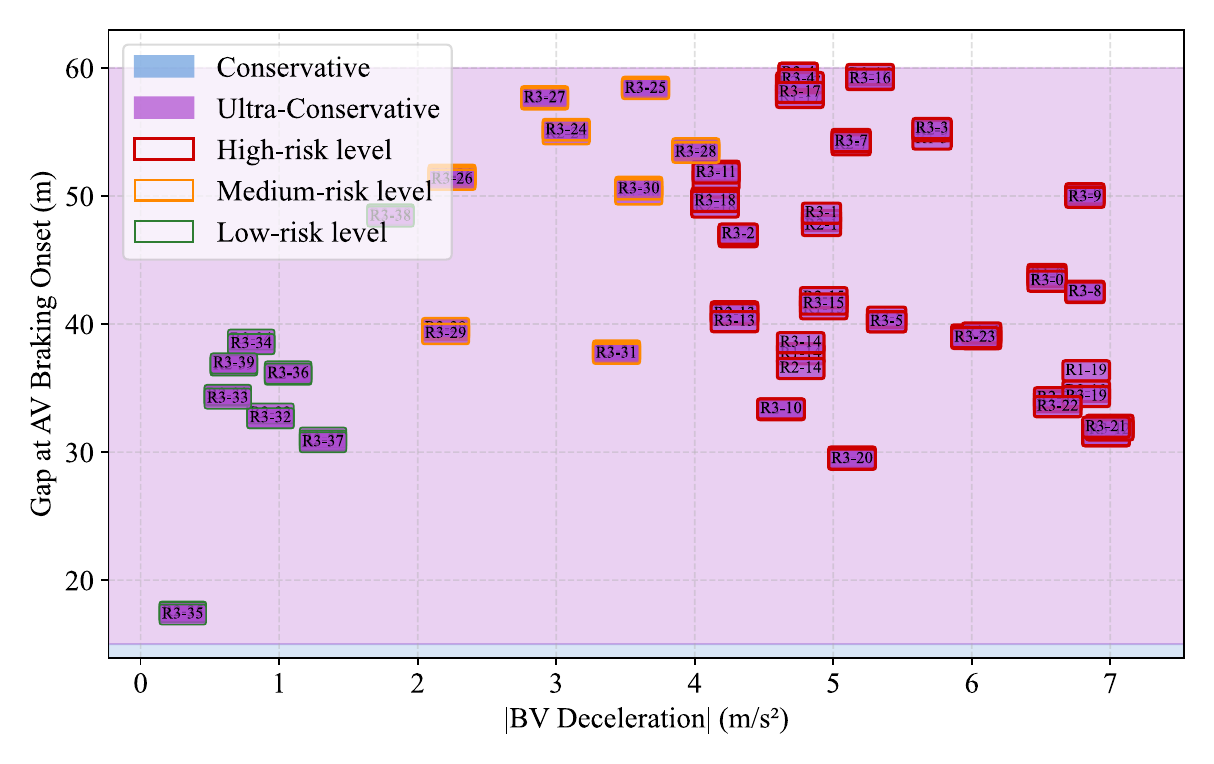}%
    }
    \hfill
    \subfloat[Final stop gap vs.\ BV deceleration magnitude.\label{fig:cf_stage3b}]{%
        \includegraphics[width=0.48\textwidth]{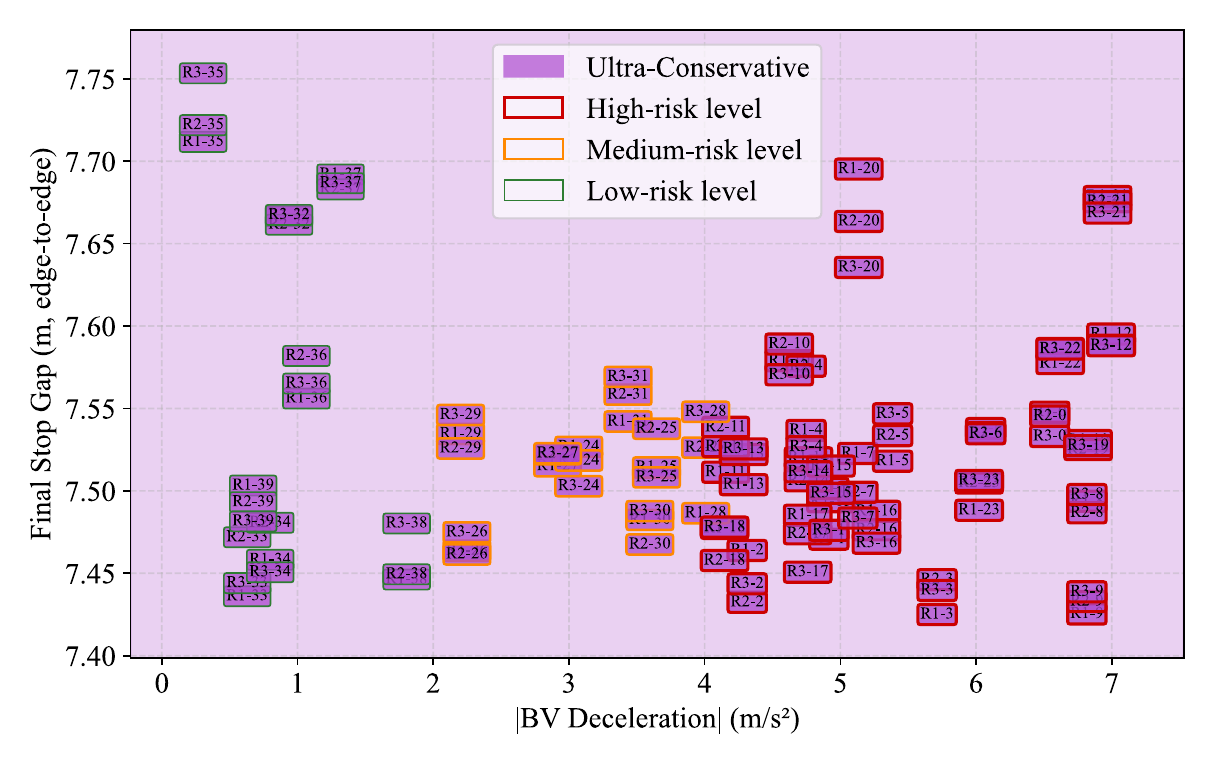}%
    }
    \caption{Analysis results for the detect and follow the leading vehicle competency.}
    \label{fig:cf_all}
\end{figure*}

\subsubsection*{Competency Certification}

Autoware.Universe \textbf{passes} this competency with no collision in any of the $120$ runs ($n = 40$ per round, collision rate $= 0\%$).
The AV follows the leading vehicle through all three phases (startup, cruise, and emergency stop) without any rear-end contact, and its final stop gap is at least $7.42\,\text{m}$ in every run.
Every cruise headway, braking-onset gap, and final stop gap falls in the Ultra-Conservative zone.

\subsubsection*{Behavioral Cautiousness Assessment}

\textbf{Stage 1: Startup Delay.}
When the leading vehicle begins to accelerate from rest, the AV starts to accelerate $1.9$--$2.9\,\text{s}$ later (median $2.0\,\text{s}$) across all $120$ runs (Fig.~\ref{fig:cf_stage1}).
$104$ runs ($86.7\%$) fall in the Conservative zone ($2.0$--$3.5\,\text{s}$) and the remaining $16$ ($13.3\%$) are all at $1.9\,\text{s}$, just inside the Normal zone ($0.8$--$2.0\,\text{s}$).
No run is Assertive or Aggressive.
Startup delay shows no meaningful dependence on BV acceleration ($r = -0.06$).

\textbf{Stage 2: Cruise Time Headway.}
The minimum time headway during cruise is $5.54$--$9.40\,\text{s}$ (median $8.51\,\text{s}$) across all $120$ runs (Fig.~\ref{fig:cf_stage2}), and every run is Ultra-Conservative (${>}3.0\,\text{s}$).
The smallest headway is about $2.5$ times the Normal upper bound of $2.2\,\text{s}$.
Headway shows no clear dependence on BV cruise speed ($r = 0.14$).

\textbf{Stage 3: Emergency Braking Response.}
When the leading vehicle decelerates to a stop, the gap at AV braking onset is $17.3$--$59.6\,\text{m}$ (median $41.1\,\text{m}$, Fig.~\ref{fig:cf_stage3a}), above the $15\,\text{m}$ Ultra-Conservative boundary in every run.
The braking-onset gap shows only a weak relation with BV deceleration ($r = 0.17$).
The smallest gaps ($17.3$--$17.5\,\text{m}$) belong to low-risk case~35, which has the gentlest BV deceleration ($0.3\,\text{m/s}^2$).
The final edge-to-edge stop gap (Fig.~\ref{fig:cf_stage3b}) is tightly clustered at $7.42$--$7.75\,\text{m}$ (median $7.52\,\text{m}$) regardless of BV deceleration ($0.3$--$7.0\,\text{m/s}^2$, $r = -0.12$), placing every run in the Ultra-Conservative zone (${>}5\,\text{m}$).
The largest final gaps ($7.71$--$7.75\,\text{m}$) belong to the low-risk case with the gentlest BV deceleration (case~35, $0.3\,\text{m/s}^2$).

\subsection{Detect and Respond to Same-Direction Lane Encroachment}
\label{subsec:lane_dep_same}

\begin{figure*}[!p]
    \centering
    \subfloat[Scorecard.\label{fig:lds_scorecard}]{%
        \includegraphics[width=0.48\textwidth]{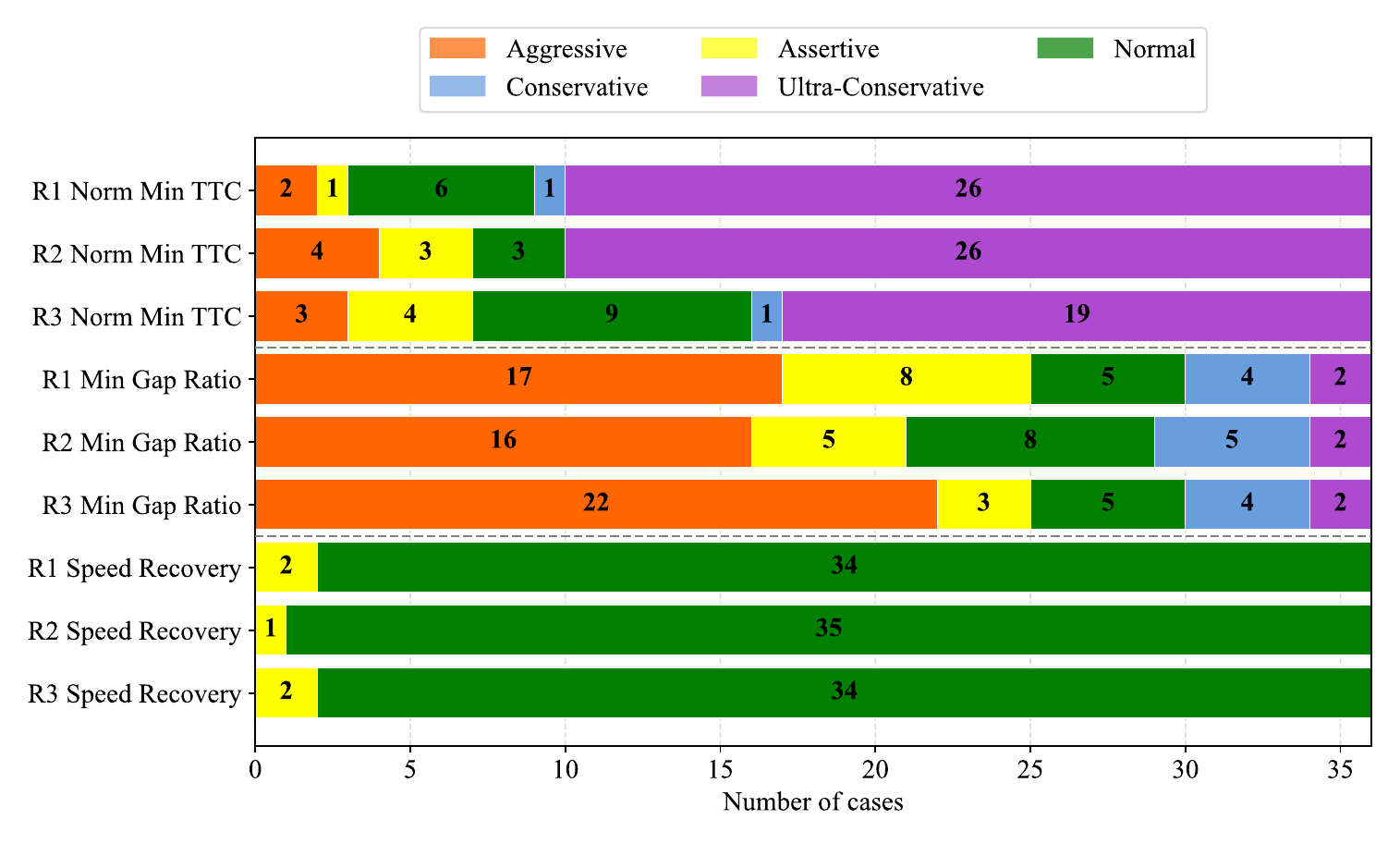}%
    }
    \hfill
    \subfloat[Normalized Minimum TTC vs.\ TTC at case start.\label{fig:lds_stage1}]{%
        \includegraphics[width=0.48\textwidth]{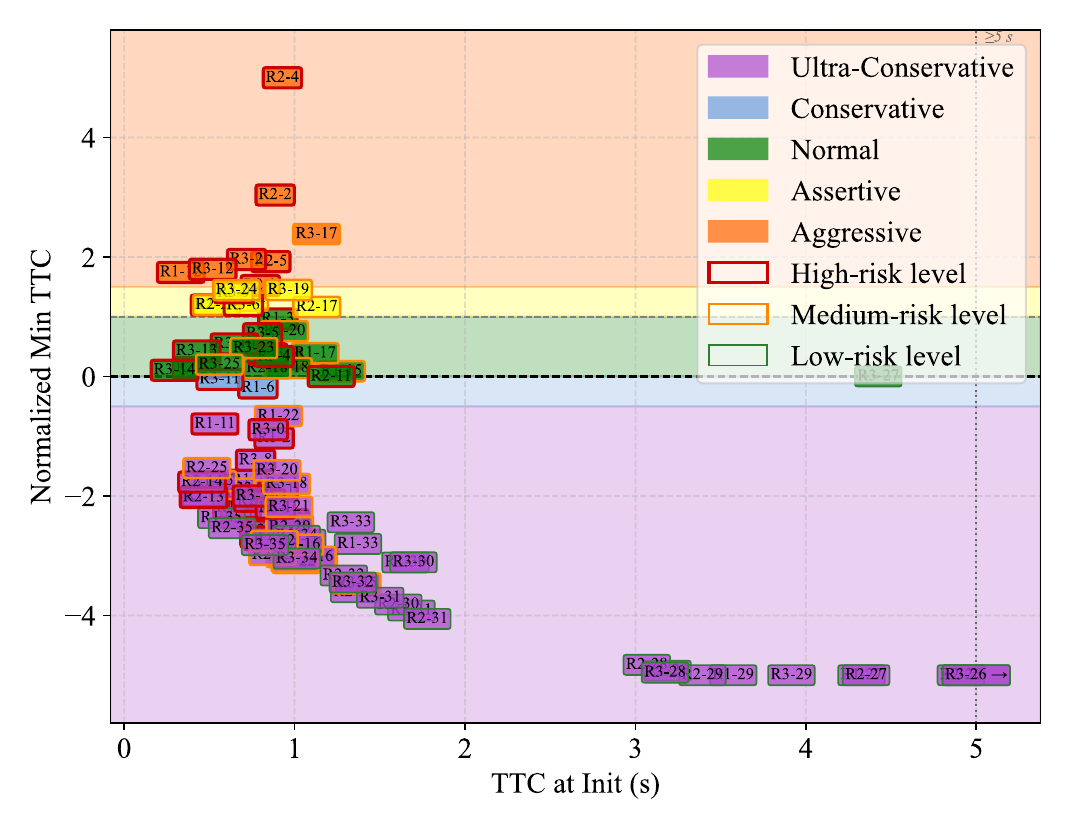}%
    }\\[4pt]
    \subfloat[Minimum Gap Ratio vs.\ Longitudinal Gap at Case Start.\label{fig:lds_stage2}]{%
        \includegraphics[width=0.48\textwidth]{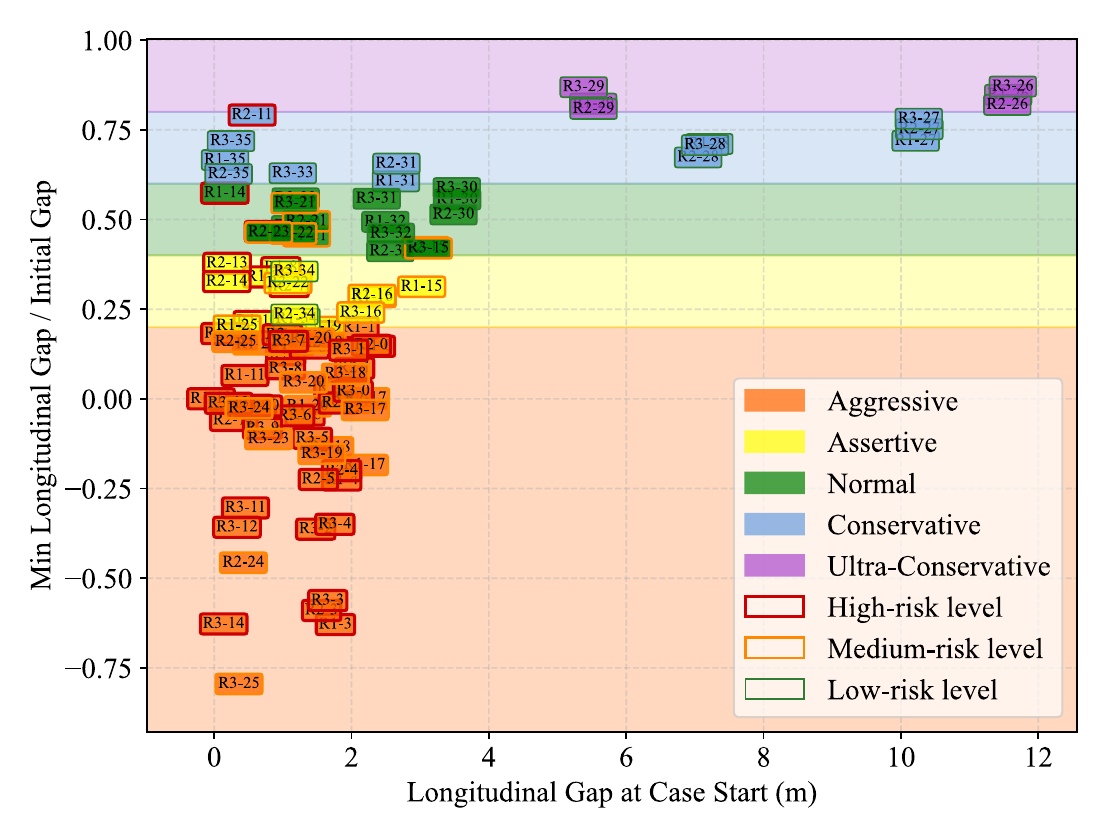}%
    }
    \hfill
    \subfloat[Speed Recovery Time vs.\ Minimum Gap Ratio.\label{fig:lds_stage3}]{%
        \includegraphics[width=0.48\textwidth]{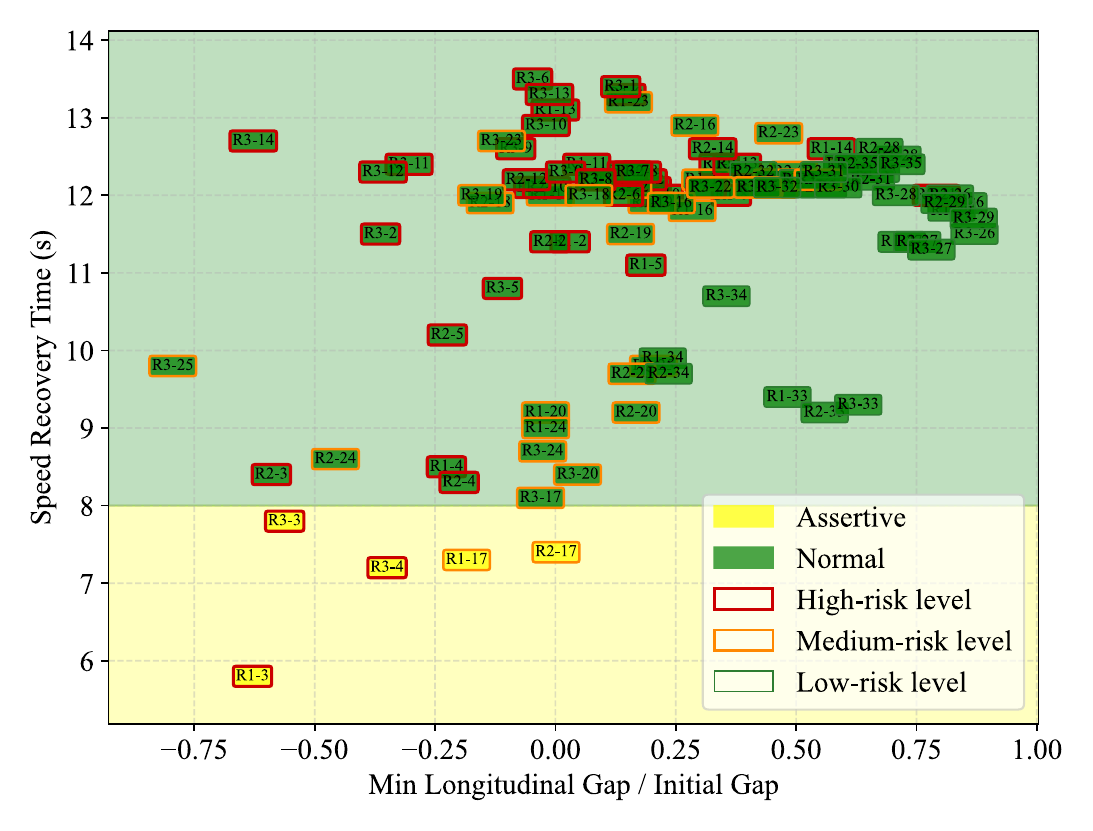}%
    }\\[4pt]
    \subfloat[Reaction time vs.\ TTC at case start.\label{fig:lds_reaction}]{%
        \includegraphics[width=0.48\textwidth]{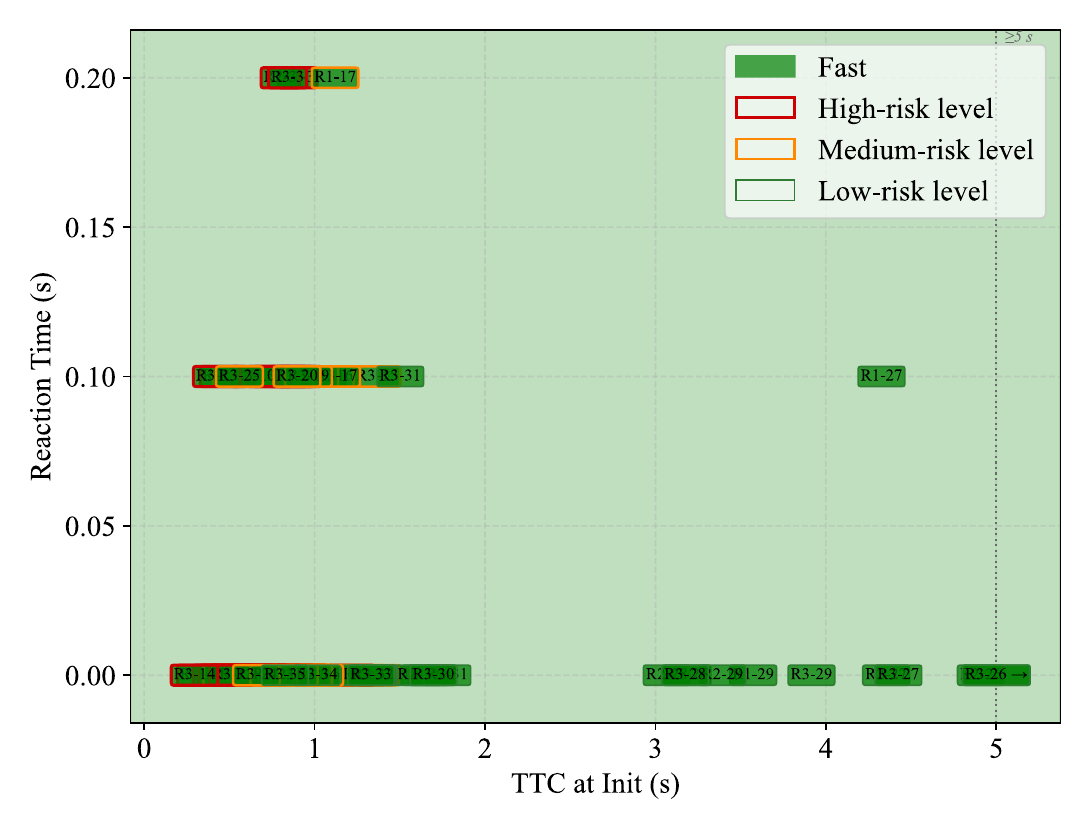}%
    }
    \hfill
    \subfloat[Normalized Deceleration vs.\ required deceleration.\label{fig:lds_decel}]{%
        \includegraphics[width=0.48\textwidth]{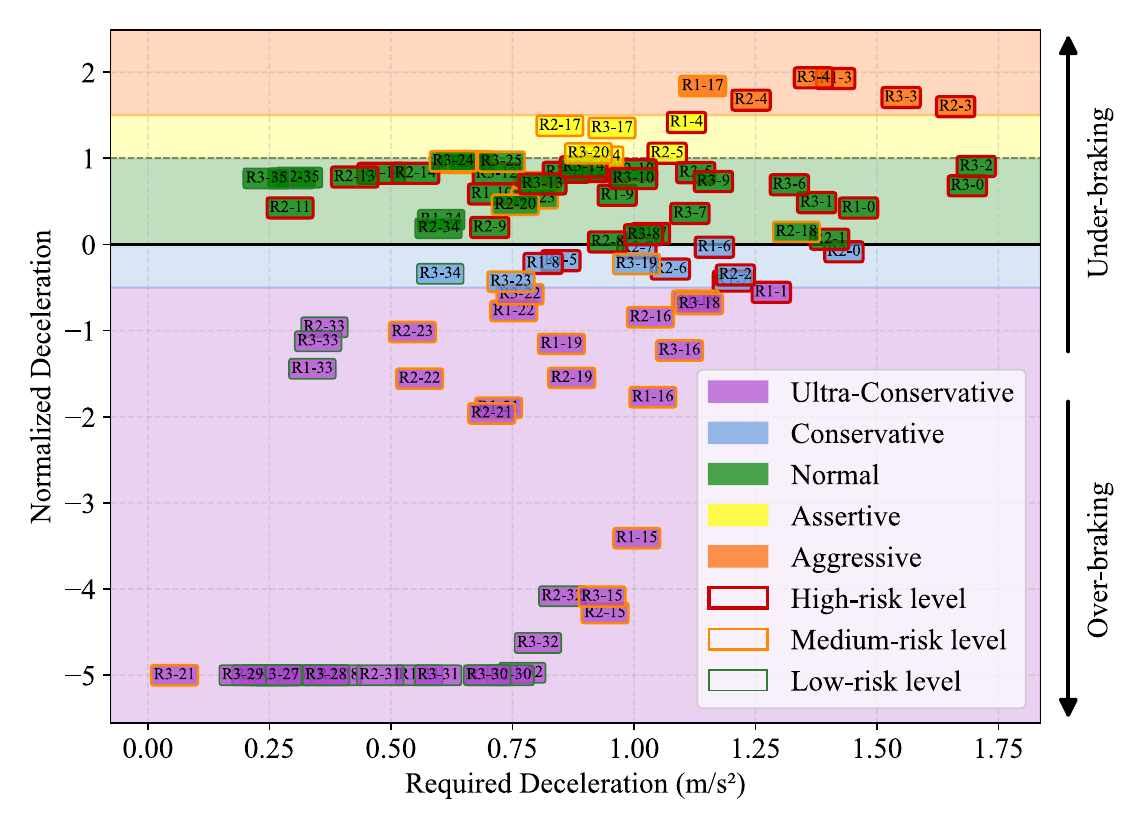}%
    }
    \caption{Analysis results for the detect and respond to same-direction lane encroachment competency.}
    \label{fig:lds_all}
\end{figure*}

\subsubsection*{Competency Certification}

Autoware.Universe \textbf{passes} this competency with no collision in any of the $108$ runs ($n = 36$ per round, collision rate $= 0\%$).
In Min Gap Ratio, $55$ of $108$ runs ($51\%$) are Aggressive.

\subsubsection*{Behavioral Cautiousness Assessment}

\textbf{Stage 1: Normalized Minimum TTC.}
In the scorecard (Fig.~\ref{fig:lds_scorecard}), $71$ of $108$ runs ($66\%$) are Ultra-Conservative.
The counts (Aggressive / Assertive / Normal / Conservative / Ultra-Conservative) are $9/8/18/2/71$.
In the scatter plot (Fig.~\ref{fig:lds_stage1}), $60$ of the $95$ runs with TTC at case start of $0.30$--$1.78\,\text{s}$ are Ultra-Conservative (norm min TTC $-4.06$ to $-0.66$, median $-2.56$), and the rest spread from the Conservative to the Aggressive zone.
R2-11 has a value of exactly $0$ (Normal).
Eight runs are Aggressive: R1-12, R2-2, R2-3, R2-4, R2-5, R3-2, R3-12, and R3-17 (norm min TTC $1.52$--$3.04$, with R2-4 at the clip value of $5$).
The ninth Aggressive run in the scorecard, R1-13, starts with the vehicles overlapping longitudinally (initial gap $-0.05\,\text{m}$).
All Aggressive runs are high-risk except R3-17 (medium-risk).
The $12$ low-risk runs of cases~26--29, with TTC at case start of $3.1$--$5.4\,\text{s}$, lie at or near the $-5$ floor ($-4.83$ to $-5$), except R3-27, which sits at $0.0007$ in the Normal zone.
Norm min TTC decreases with TTC at case start ($r = -0.56$).

\textbf{Stage 2: Minimum Gap Ratio.}
In the scorecard, the counts are $55/16/18/13/6$.
Only $37$ of $108$ runs ($34\%$) reach Normal or a more conservative gap ratio, and the median ratio over all runs is $0.19$.
In the scatter plot (Fig.~\ref{fig:lds_stage2}), $47$ of the $73$ runs with initial longitudinal gaps below $2\,\text{m}$ ($40$ high-, $24$ medium-, and $9$ low-risk) are Aggressive, $12$ Assertive, $9$ Normal, and $5$ Conservative, with a median ratio of $0.14$.
$29$ of these $73$ runs have a negative ratio, with the AV's front passing the rear of the encroaching vehicle so that the two vehicles overlap longitudinally during the encounter.
The $23$ runs with initial gaps of $2$--$4\,\text{m}$ are split between Normal ($9$), Aggressive ($8$), Assertive ($4$), and Conservative ($2$).
The $12$ low-risk runs with initial gaps of $5.4$--$11.6\,\text{m}$ are Conservative ($6$) or Ultra-Conservative ($6$), with ratios of $0.67$--$0.87$.
Across all gap sizes, $31$ runs have a negative ratio ($20$ high- and $11$ medium-risk), with the lowest at $-0.79$ (R3-25).
By risk level, $36$ of $45$ high-risk runs and $19$ of $33$ medium-risk runs are Aggressive, against none of the $30$ low-risk runs.

\textbf{Stage 3: Speed Recovery Time.}
In the scorecard, speed recovery is Normal ($8$--$16\,\text{s}$) in $103$ of $108$ runs and Assertive ($4$--$8\,\text{s}$) in the other $5$.
In the scatter plot (Fig.~\ref{fig:lds_stage3}), recovery time spans $5.8$--$13.5\,\text{s}$, and the five Assertive runs (R1-3, R1-17, R2-17, R3-3, R3-4) recover in $5.8$--$7.8\,\text{s}$.
Recovery time increases with the Min Gap Ratio ($r = 0.40$).
R1-3 (Min Gap Ratio $-0.63$) has the shortest recovery time in the dataset ($5.8\,\text{s}$), while R3-25, the most negative ratio ($-0.79$), recovers in $9.8\,\text{s}$.

\subsubsection*{Behavior Diagnosis}

\textbf{Reaction Time.}
The reaction time plot (Fig.~\ref{fig:lds_reaction}) shows reaction times of at most $0.2\,\text{s}$ in all $108$ runs, all in the Fast zone ($<0.7\,\text{s}$).
Reaction time is $0\,\text{s}$ in $83$ runs, $0.1\,\text{s}$ in $21$, and $0.2\,\text{s}$ in $4$ (R1-3, R1-17, R2-3, R3-3).

\textbf{Deceleration Adequacy.}
In the deceleration adequacy plot (Fig.~\ref{fig:lds_decel}), the counts (Aggressive / Assertive / Normal / Conservative / Ultra-Conservative) are $6/6/41/11/42$.
Across all runs, the average deceleration ranges from $0.65$ to $2.67\,\text{m/s}^2$, with a median of $1.83\,\text{m/s}^2$ and an interquartile range of $1.46$--$1.90\,\text{m/s}^2$.
The $12$ Assertive or Aggressive runs ($7$ high- and $5$ medium-risk) under-brake, with an average deceleration of only $0.65$--$1.04\,\text{m/s}^2$ against $a_{\text{req}}$ of $0.85$--$1.66\,\text{m/s}^2$.
The $42$ Ultra-Conservative runs brake at $0.88$--$2.67\,\text{m/s}^2$ against $a_{\text{req}}$ of only $0.05$--$1.28\,\text{m/s}^2$.
Nineteen of them ($18$ low-risk runs of cases~26--31 and R3-21) sit at the $-5$ floor with $a_{\text{req}}$ of $0.05$--$0.75\,\text{m/s}^2$.

Of the $12$ under-braking runs, ten have a negative Min Gap Ratio and the other two have ratios of $0.001$ and $0.05$.
They account for all five Assertive speed recovery times, and their recovery times overall are $5.8$--$10.2\,\text{s}$.
The $55$ runs with an Aggressive Min Gap Ratio all react within $0.2\,\text{s}$, and most brake at a typical rate (median average deceleration $1.81\,\text{m/s}^2$): $28$ have Normal, $8$ Conservative, and $5$ Ultra-Conservative normalized deceleration, while $12$ are Assertive or Aggressive.
Among the $31$ runs with longitudinal overlap (negative ratio), $16$ have Normal normalized deceleration and $10$ are Assertive or Aggressive.

\subsection{Detect and Respond to Oncoming Lane Encroachment}
\label{subsec:lane_dep_opp}

\begin{figure*}[!t]
    \centering
    \subfloat[Scorecard.\label{fig:ldo_scorecard}]{%
        \includegraphics[width=0.48\textwidth]{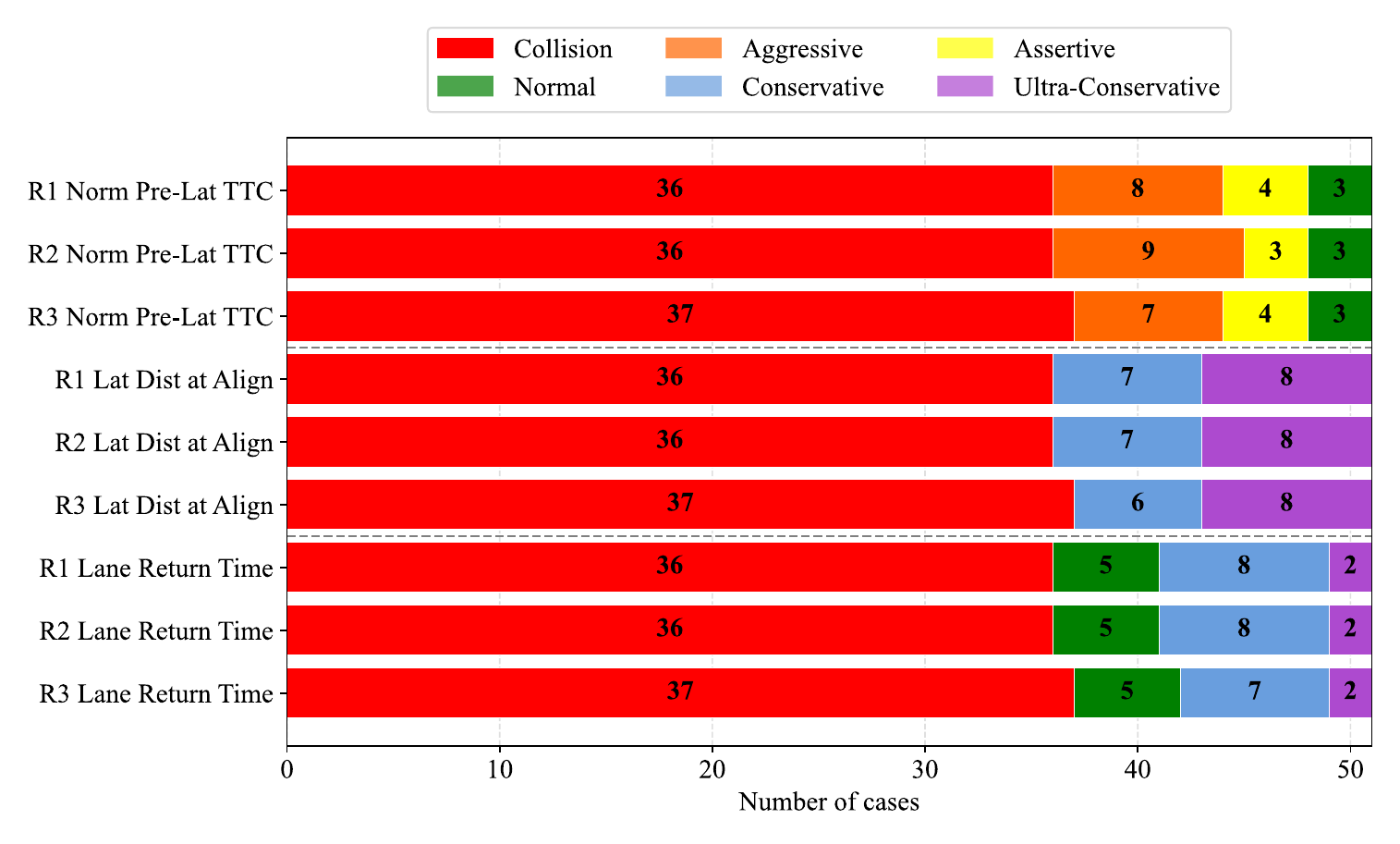}%
    }
    \hfill
    \subfloat[Normalized Pre-Lateral TTC vs.\ TTC at Case Start.\label{fig:ldo_stage1}]{%
        \includegraphics[width=0.48\textwidth]{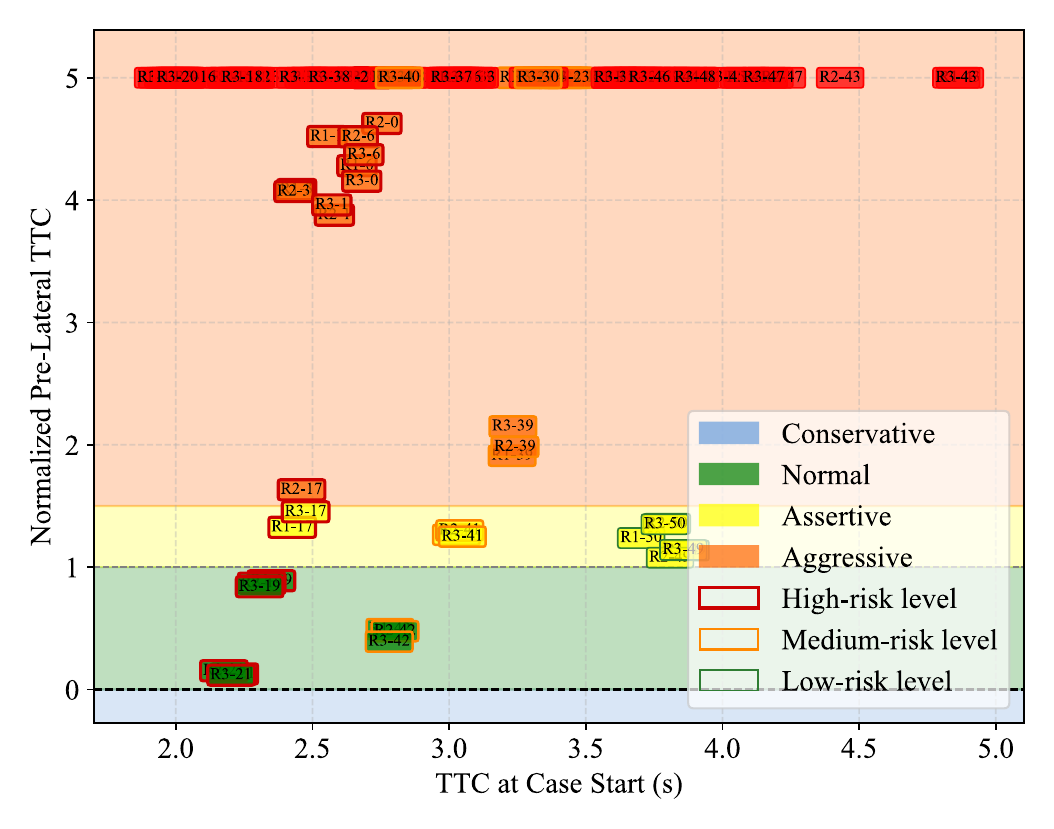}%
    }\\[4pt]
    \subfloat[Lateral Distance at Center Alignment vs.\ TTC at Case Start.\label{fig:ldo_stage2}]{%
        \includegraphics[width=0.48\textwidth]{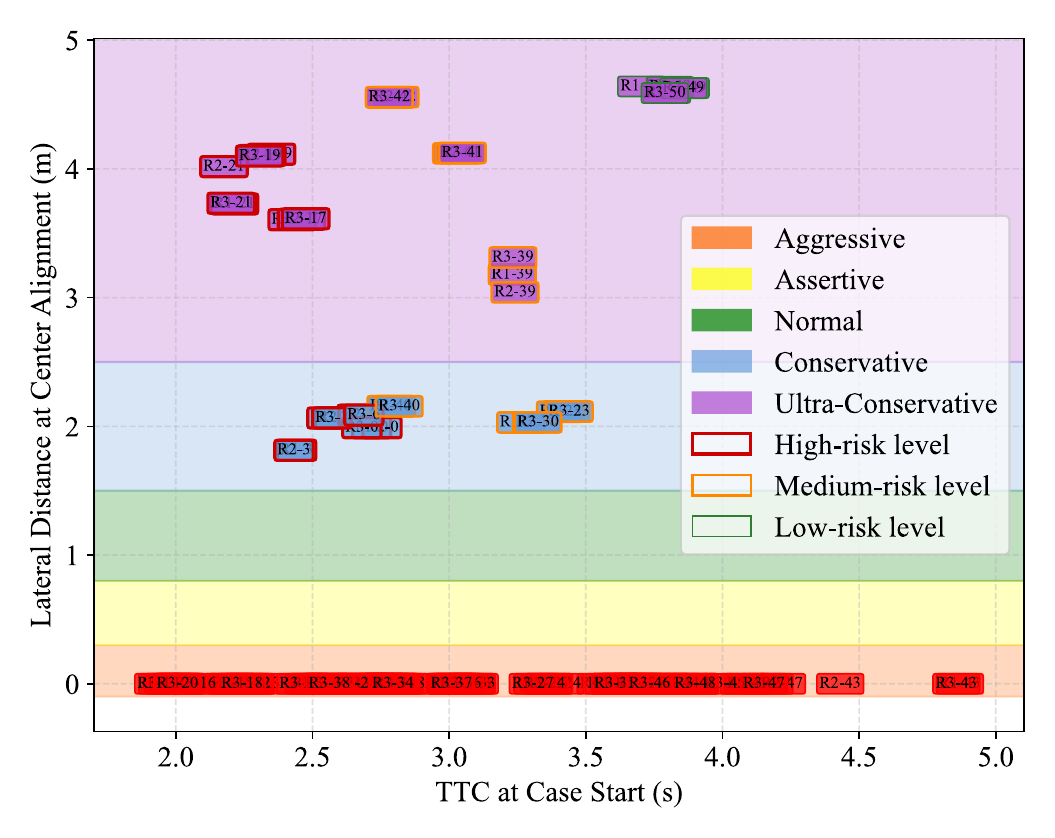}%
    }
    \hfill
    \subfloat[Lane Return Time vs.\ TTC at Case Start.\label{fig:ldo_stage3}]{%
        \includegraphics[width=0.48\textwidth]{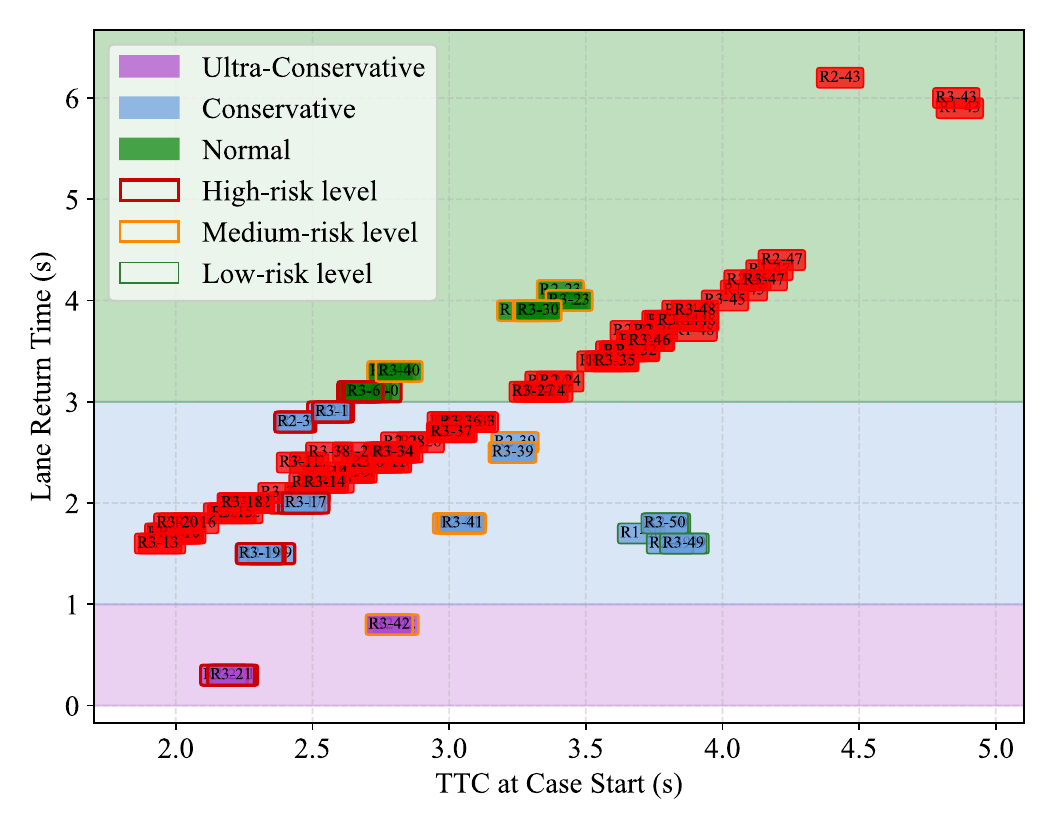}%
    }
    \caption{Analysis results for the detect and respond to oncoming lane encroachment competency.}
    \label{fig:ldo_all}
\end{figure*}

\subsubsection*{Competency Certification}

Autoware.Universe \textbf{fails} this competency with $71.2\%$ of runs resulting in collision ($109$ of $153$ runs), the second-highest collision rate across all 14 competencies after VRU without crosswalk ($78.8\%$).
The 109 collision runs come from 37 of the 51 cases.
The 44 non-collision runs comprise 20 high-risk, 18 medium-risk, and 6 low-risk runs.

\subsubsection*{Behavioral Cautiousness Assessment}

The AV does not move laterally in any of the 153 runs. Its lateral position drifts by at most $3.4\,\text{cm}$ after case start and its peak lateral speed is $0.006\,\text{m/s}$, below the $0.15\,\text{m/s}$ onset threshold.
The three stage metrics describe the encounter geometry and the duration of the test case.
The scorecard (Fig.~\ref{fig:ldo_scorecard}) bins the 44 non-collision runs.

\textbf{Stage 1: Normalized Pre-Lateral TTC.}
In the scorecard, 24 runs are Aggressive, 11 Assertive, and 9 Normal, with no Conservative or Ultra-Conservative runs.
The Normalized Pre-Lateral TTC equals the minimum longitudinal TTC over the whole run.
In the scatter plot (Fig.~\ref{fig:ldo_stage1}), collision cases sit at the assigned value~5 across TTC at case start of $1.94$--$4.87\,\text{s}$.
In the 20 runs of case IDs 0, 1, 3, 6, 23, 30, and 40, the vehicles pass each other, and all 20 are Aggressive ($3.88$--$5.0$, with 10 runs at the clip value of~5). Their minimum longitudinal TTC is $0.002$--$0.09\,\text{s}$, i.e.\ the longitudinal gap closes to nearly zero as the vehicles pass side by side. This TTC does not account for lateral offset.
In the 24 runs of case IDs 17, 19, 21, 39, 41, 42, 49, and 50, the test case ends $0.3$--$2.6\,\text{s}$ after case start while the vehicle centers are still $9.8$--$25.3\,\text{m}$ apart longitudinally, and all three stage metrics are evaluated at the end of the test case, before the vehicles meet.
These runs have minimum TTC of $0.53$--$2.22\,\text{s}$ and normalized values of $0.12$--$2.15$: case IDs 19, 21, and 42 are Normal ($0.12$--$0.89$), case IDs 41, 49, and 50 are Assertive ($1.08$--$1.35$), case ID~17 is Assertive in two runs and Aggressive in one ($1.33$--$1.63$), and case ID~39 is Aggressive ($1.91$--$2.15$).
TTC at case start does not separate outcomes. All 12 runs above $3.87\,\text{s}$ collide, while in the $3.70$--$3.87\,\text{s}$ range some cases survive and others collide. Between $2.18$ and $2.28\,\text{s}$, all 3 runs of one case survive and all 9 runs of three other cases collide.

\textbf{Stage 2: Lateral Distance at Center Alignment.}
In the scorecard, 20 runs are Conservative ($1.5$--$2.5\,\text{m}$) and 24 Ultra-Conservative ($>2.5\,\text{m}$), with no Normal, Assertive, or Aggressive runs.
The Conservative runs are exactly the runs in which the vehicles pass, and the Ultra-Conservative runs are exactly the runs in which the vehicles do not meet.
In the scatter plot (Fig.~\ref{fig:ldo_stage2}), collision cases are plotted at the assigned value of $0\,\text{m}$.
In the passing runs, the BV center comes within $1.8$--$2.1\,\text{m}$ of the AV's lane center, and the center-to-center lateral offset at alignment is $1.81$--$2.16\,\text{m}$, about one vehicle width (the AV is $2.0\,\text{m}$ wide).
In the runs in which the vehicles do not meet, the BV moves at most $0.6\,\text{m}$ toward the AV's lane, and the lateral distance is $3.04$--$4.64\,\text{m}$, which is the BV's lateral position in its own lane at the end of the test case.
No non-collision run falls between $0.3$ and $1.5\,\text{m}$.

\textbf{Stage 3: Lane Return Time.}
In the scorecard, 15 runs are Normal ($3$--$8\,\text{s}$), 23 Conservative ($1$--$3\,\text{s}$), and 6 Ultra-Conservative ($<1\,\text{s}$).
The AV never leaves the $0.2\,\text{m}$ return tolerance, and the Lane Return Time equals the time of center alignment in every run.
In the scatter plot (Fig.~\ref{fig:ldo_stage3}), the passing runs, which last $7.0$--$14.7\,\text{s}$ in total, have a Lane Return Time equal to the time at which the vehicles pass ($2.8$--$4.1\,\text{s}$, Conservative for case IDs 1 and 3 and Normal for the rest).
For the runs in which the vehicles do not meet, it equals the duration of the test case ($0.3$--$2.6\,\text{s}$), and the two Ultra-Conservative cases (case IDs 21 and 42, $0.3$ and $0.8\,\text{s}$) are the two shortest.
Collision cases appear in this plot at their time of contact ($1.6$--$6.2\,\text{s}$), which increases with TTC at case start.

\subsubsection*{Behavior Diagnosis}

\textbf{Lateral Response.}
The AV executes no measurable lateral movement in any of the 153 runs.

\textbf{Longitudinal Response.}
The only measurable AV response is longitudinal deceleration. In the passing runs the AV slows from $8.3\,\text{m/s}$ to a minimum of $1.2$--$4.4\,\text{m/s}$ before the vehicles pass. In the runs in which the vehicles do not meet, the AV speed decreases only from $7.8$--$8.3$ to $7.7\,\text{m/s}$.
In collision runs the minimum AV speed before contact ranges from $0.6$ to $8.3\,\text{m/s}$ (median $7.6\,\text{m/s}$), and the AV slows by more than $1\,\text{m/s}$ in only 47 of the 109 collision runs.

\subsection{Detect and Respond to Left-Turning Vehicle}
\label{subsec:left_turn_straight}

\begin{figure*}[!p]
    \centering
    \subfloat[Scorecard.\label{fig:lts_scorecard}]{%
        \includegraphics[width=0.48\textwidth]{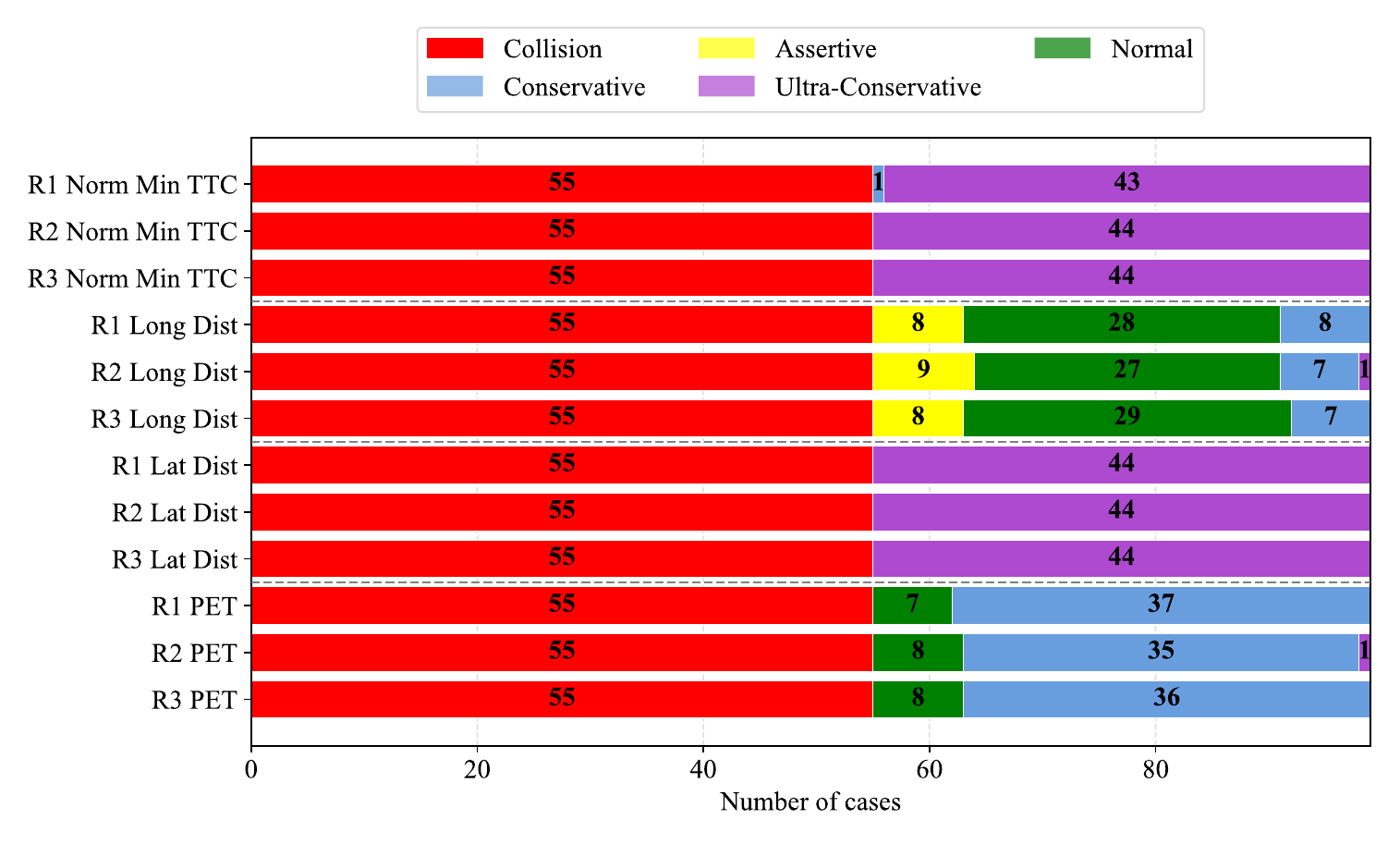}%
    }
    \hfill
    \subfloat[Normalized Minimum TTC vs.\ TTC at Init.\label{fig:lts_stage1}]{%
        \includegraphics[width=0.48\textwidth]{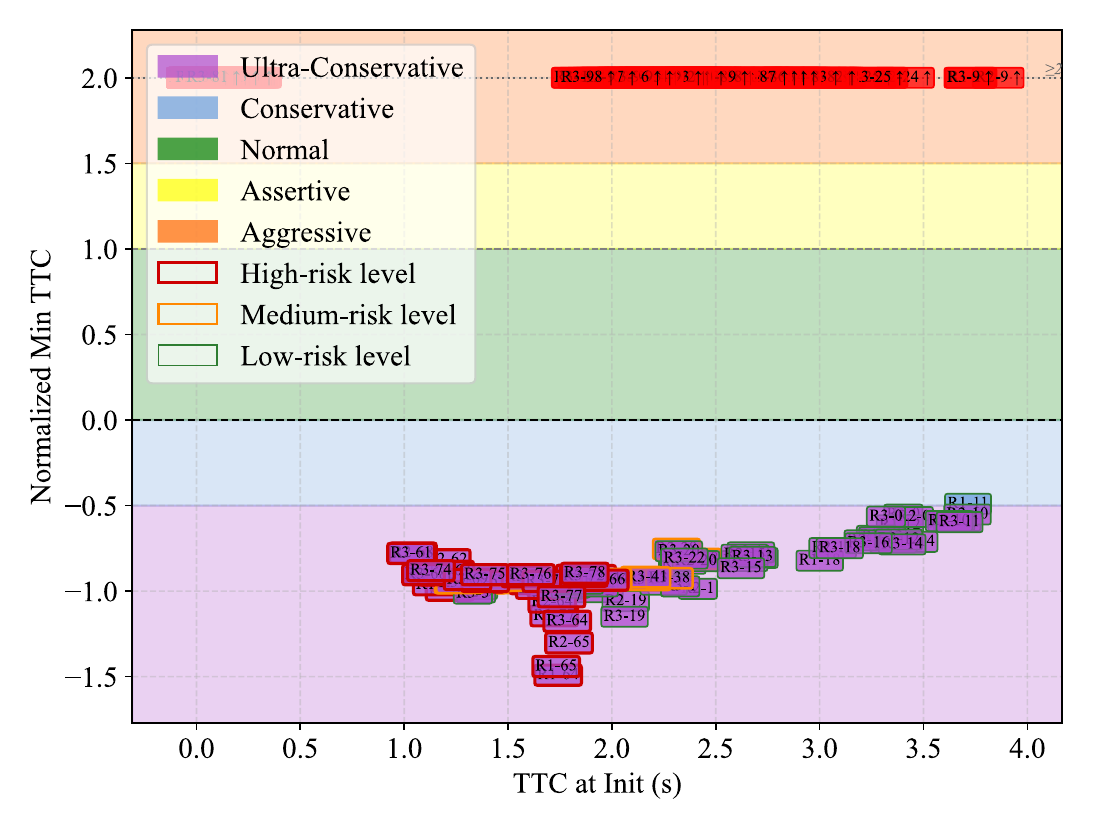}%
    }\\[4pt]
    \subfloat[Minimum Longitudinal Distance vs.\ Longitudinal Distance at Case Start.\label{fig:lts_stage2}]{%
        \includegraphics[width=0.48\textwidth]{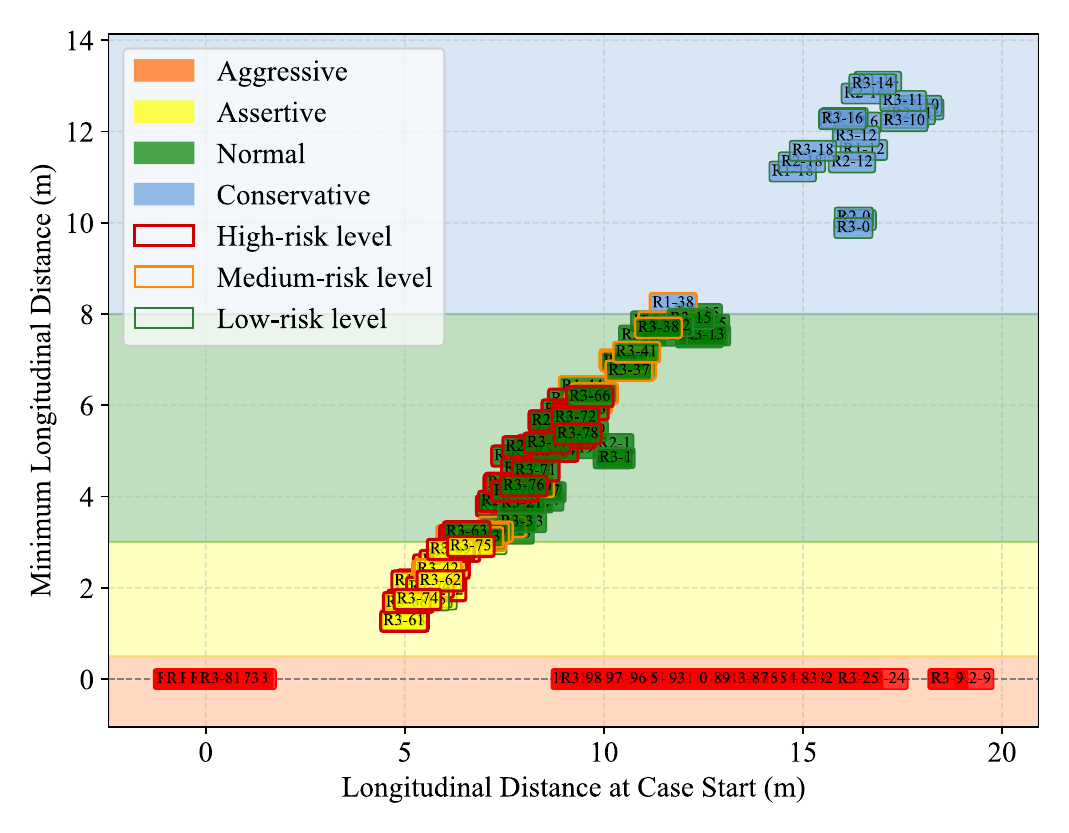}%
    }
    \hfill
    \subfloat[Post-Encroachment Time vs.\ Minimum Longitudinal Distance.\label{fig:lts_stage3}]{%
        \includegraphics[width=0.48\textwidth]{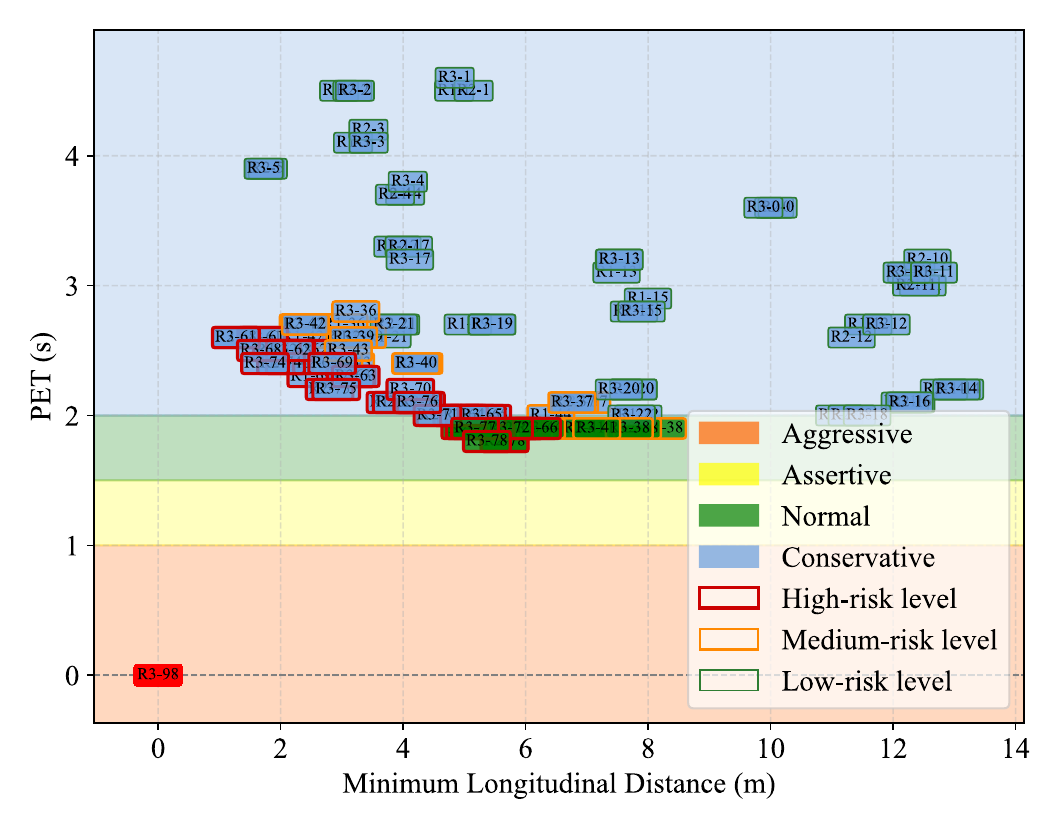}%
    }\\[4pt]
    \subfloat[Reaction Time vs.\ TTC at Init.\label{fig:lts_reaction}]{%
        \includegraphics[width=0.48\textwidth]{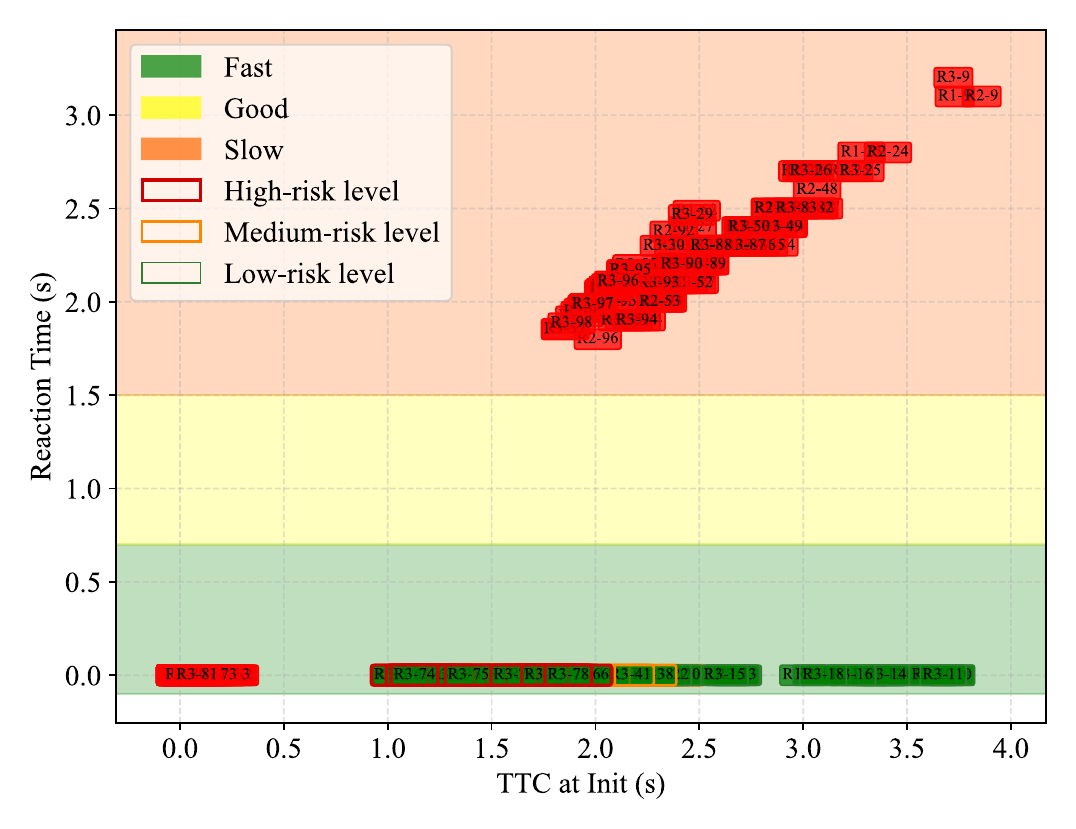}%
    }
    \hfill
    \subfloat[Normalized Deceleration vs.\ Required Deceleration.\label{fig:lts_decel}]{%
        \includegraphics[width=0.48\textwidth]{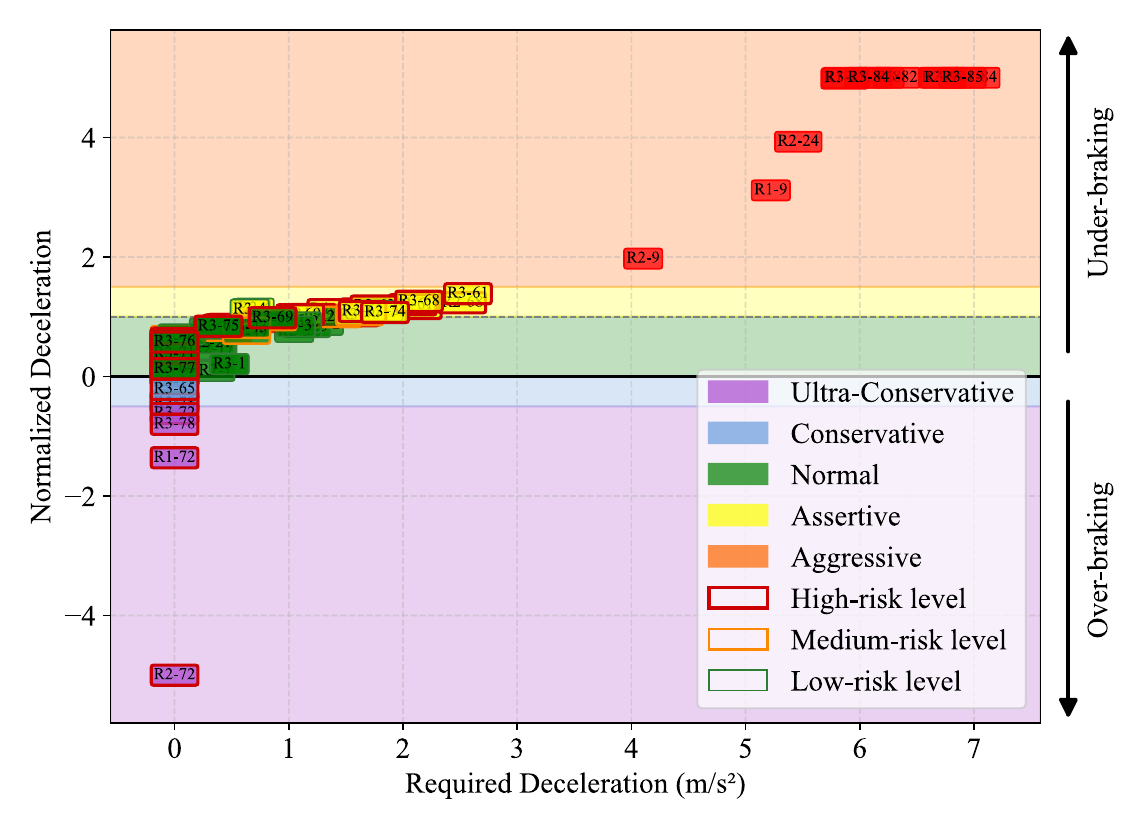}%
    }
    \caption{Analysis results for the detect and respond to left-turning vehicle competency.}
    \label{fig:lts_all}
\end{figure*}

\subsubsection*{Competency Certification}

Autoware.Universe \textbf{fails} this competency with $55.6\%$ of runs resulting in collision ($165$ of $297$ runs), the third-highest collision rate across all 14 competencies.
The 165 collision runs are all runs of 55 of the 99 cases, and the 132 non-collision runs are all runs of the other 44 cases.
Collisions occur at every risk level: 75 of 123 high-risk, 51 of 78 medium-risk, and 39 of 96 low-risk runs collide.

\subsubsection*{Behavioral Cautiousness Assessment}

The scorecard (Fig.~\ref{fig:lts_scorecard}) bins the 132 non-collision runs.

\textbf{Stage 1: Normalized Minimum TTC.}
In the scorecard, 131 runs are Ultra-Conservative and one run (R1-11, $-0.49$) is Conservative. No run is Normal, Assertive, or Aggressive.
The scatter plot (Fig.~\ref{fig:lts_stage1}) shows the 165 collision runs at the display cap of~2 (assigned value~5). Their TTC at case start is $0$--$0.28\,\text{s}$ for 63 runs ($38\%$, short-TTC collisions) and $1.85$--$3.86\,\text{s}$ for 102 runs ($62\%$, long-TTC collisions, median $2.45\,\text{s}$, with 80 of them between $2.0$ and $3.0\,\text{s}$).
The non-collision runs span TTC at case start of $1.03$--$3.71\,\text{s}$ and Normalized Minimum TTC of $-1.49$ to $-0.49$ (median $-0.93$), i.e.\ close to the instant-reaction reference at~$-1$. Twelve runs fall below $-1$ and eight lie between $-0.6$ and $-0.5$.

\textbf{Stage 2: Minimum Longitudinal Clearance.}
In the scorecard, the split of Minimum Longitudinal Distance is 25 Assertive ($1.27$--$2.96\,\text{m}$), 84 Normal ($3.04$--$8.00\,\text{m}$), and 22 Conservative ($8.23$--$13.09\,\text{m}$). The single Ultra-Conservative entry is R2-36, in which the AV is stationary at the record start, $72.3\,\text{m}$ from the conflict zone. No run has a negative or Aggressive value.
The scatter plot (Fig.~\ref{fig:lts_stage2}) shows collision cases at the assigned value of $0\,\text{m}$. Their initial distance is $-0.83$ to $1.28\,\text{m}$ for the 63 short-TTC collisions and $9.3$--$19.3\,\text{m}$ for the 102 long-TTC collisions.
Non-collision runs start $5.0$--$18.0\,\text{m}$ from the conflict zone, and their minimum distance increases with initial distance. Initial distances of $5$--$10\,\text{m}$ yield $1.27$--$6.41\,\text{m}$ (Assertive to Normal), and initial distances of $14.7$--$18.0\,\text{m}$ yield $9.9$--$13.1\,\text{m}$ (Conservative).
No non-collision run starts closer than $5.0\,\text{m}$, while non-collision runs and long-TTC collisions overlap over initial distances of $9.3$--$18.0\,\text{m}$.

\textbf{Stage 3: Post-Encroachment Time and Lateral Distance.}
In the scorecard, all 132 runs are Ultra-Conservative in Lateral Distance.
For PET, the split is 23 Normal ($1.8$--$2.0\,\text{s}$) and 108 Conservative ($2.0$--$4.6\,\text{s}$). The single Ultra-Conservative entry is R2-36. No run is Assertive or Aggressive.
The PET is positive in every non-collision run, i.e.\ the BV clears the conflict point before the AV reaches it and the AV yields.
The scatter plot (Fig.~\ref{fig:lts_stage3}) shows all collision cases at the origin (PET and minimum longitudinal distance both set to~0).
Non-collision PET ranges from $1.8$ to $4.6\,\text{s}$ and is not correlated with minimum longitudinal distance (Pearson $r=-0.13$).
The 23 Normal-zone runs ($1.8$--$2.0\,\text{s}$, case IDs 18, 22, 38, 41, 44, 65, 66, 72, 77, 78) have minimum longitudinal distances of $5.0$--$11.6\,\text{m}$. The other 108 runs are Conservative.

\subsubsection*{Behavior Diagnosis}

\textbf{Reaction Time.}
The reaction time plot (Fig.~\ref{fig:lts_reaction}) separates the outcomes.
The non-collision runs are Fast at $0\,\text{s}$, i.e.\ the AV is already decelerating when the BV begins its turn.
The 63 short-TTC collisions also have reaction time $0\,\text{s}$, at TTC at case start below $0.3\,\text{s}$.
The 102 long-TTC collisions are Slow at $1.8$--$3.2\,\text{s}$ (median $2.2\,\text{s}$). Of these, 79 start braking at $1.8$--$3.2\,\text{s}$ (median $2.3\,\text{s}$), and 23 never brake.
Over all 297 runs, 194 ($65\%$) have reaction time~0, and the median reaction time of the collision runs is $2.1\,\text{s}$.
The AV speed at the record start is $5.0\,\text{m/s}$ in every run except R2-36, where the AV is stationary. The 79 late-braking collisions begin braking at $5.0\,\text{m/s}$, whereas the non-collision runs are already braking at $3.9$--$5.0\,\text{m/s}$ (median $4.8\,\text{m/s}$) when the BV begins its turn.

\textbf{Deceleration Adequacy.}
In the deceleration adequacy plot (Fig.~\ref{fig:lts_decel}), 53 non-collision runs are Normal, 22 are Assertive (mild under-braking, required deceleration $0.66$--$2.57\,\text{m/s}^2$), 4 are Conservative, and 4 are Ultra-Conservative. The Conservative and Ultra-Conservative runs all have a required deceleration of~0.
Among collision runs, the required deceleration given the actual reaction time has a median of $11.7\,\text{m/s}^2$, with 63 runs above $10\,\text{m/s}^2$ and 10 above $50\,\text{m/s}^2$.
Twelve collision runs are Aggressive ($1.97$--$5.0$) at required decelerations of $4.1$--$7.0\,\text{m/s}^2$.
The average deceleration achieved by the 79 late-braking collisions is $0.09$--$1.33\,\text{m/s}^2$ (median $0.40\,\text{m/s}^2$), against a median required deceleration of $10.7\,\text{m/s}^2$ for the long-TTC collisions.

In the 102 long-TTC collisions, Autoware.Universe continues at $5.0\,\text{m/s}$ after the BV begins its turn and either starts braking a median of $2.3\,\text{s}$ later (about $11.5\,\text{m}$ of travel) or does not brake at all. The deceleration it then applies (median $0.40\,\text{m/s}^2$) is far below what is required to stop before the conflict zone.
In the 63 short-TTC collisions, the BV begins its turn when the AV is within $1.3\,\text{m}$ of the conflict zone or up to $0.8\,\text{m}$ past its boundary.
Every non-collision run is one in which the AV is already decelerating at the start of the BV's turn and the BV clears the conflict point first.

\subsection{Execute Unprotected Left Turn}
\label{subsec:left_turn_turn}

\begin{figure*}[!p]
    \centering
    \subfloat[Scorecard.\label{fig:ltt_scorecard}]{%
        \includegraphics[width=0.48\textwidth]{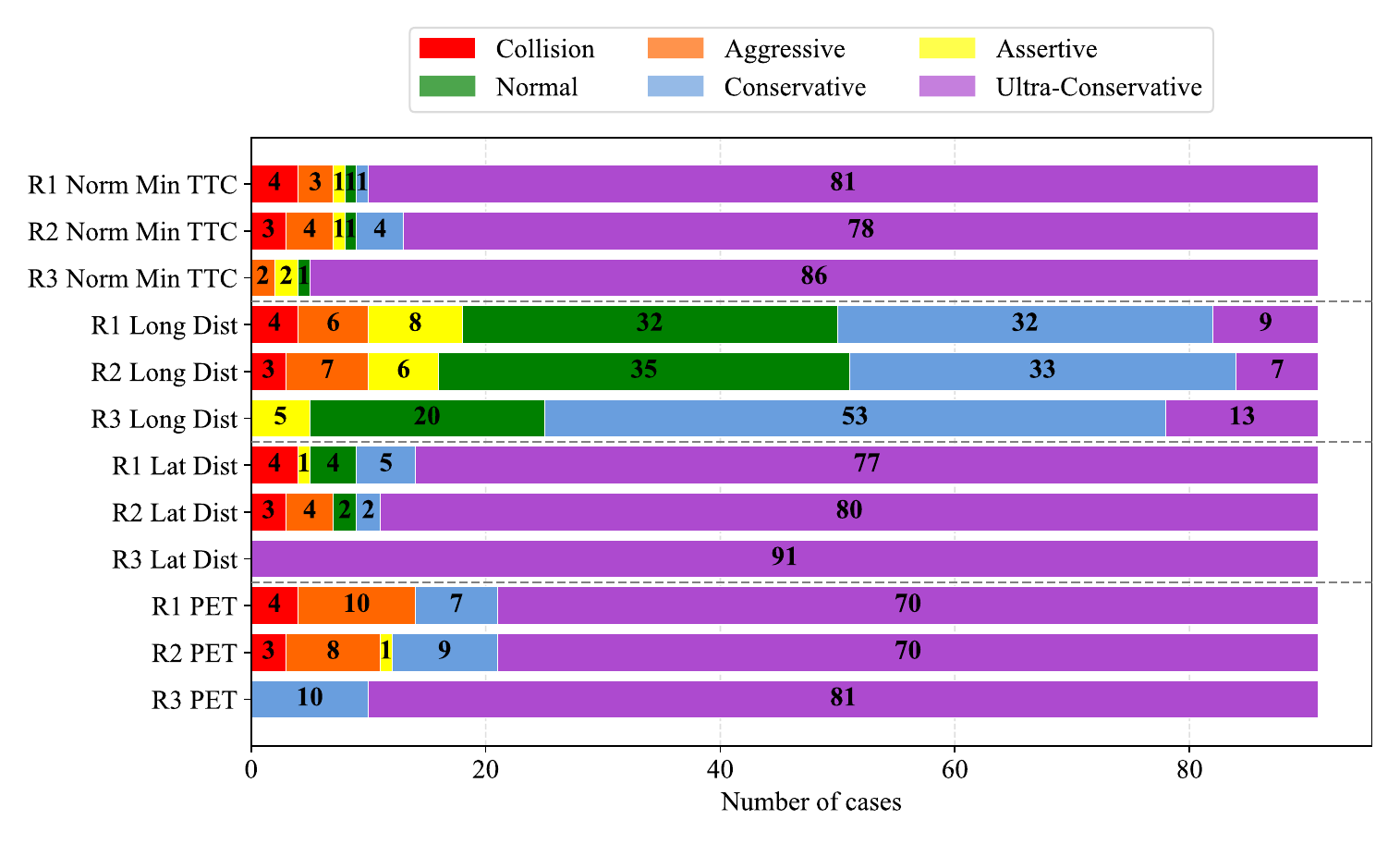}%
    }
    \hfill
    \subfloat[Normalized Minimum TTC vs.\ TTC at case start.\label{fig:ltt_stage1}]{%
        \includegraphics[width=0.48\textwidth]{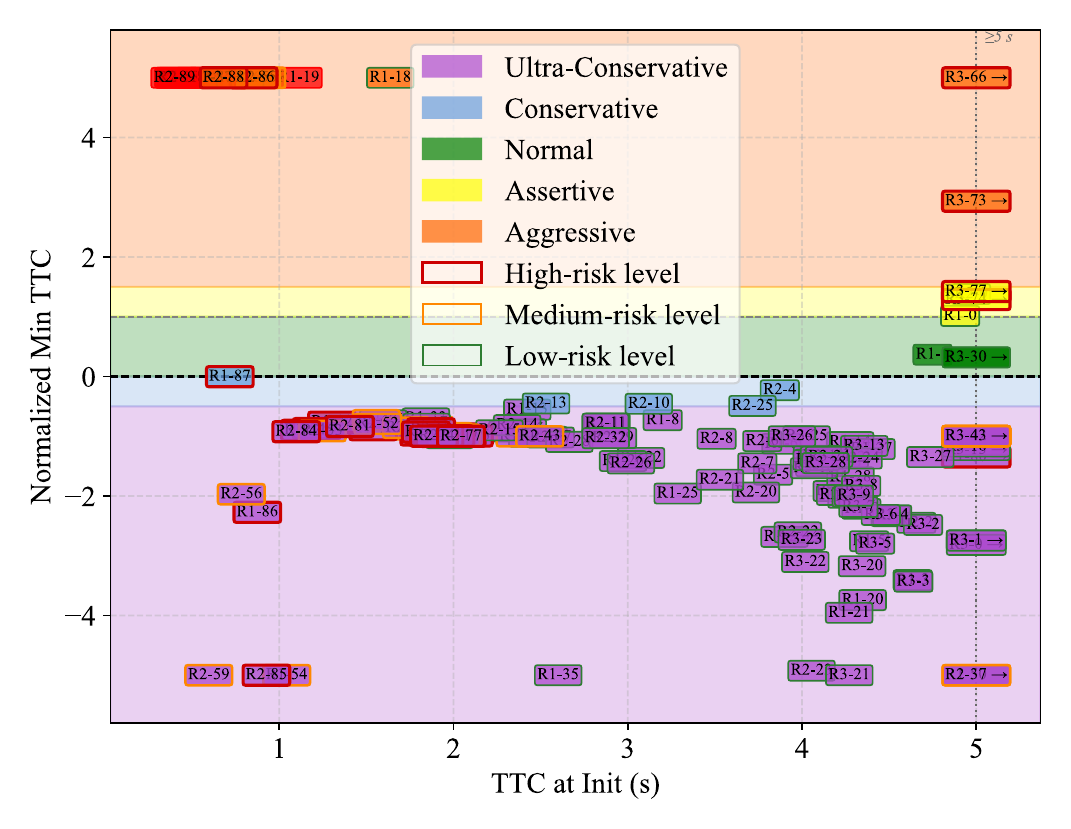}%
    }\\[4pt]
    \subfloat[Minimum Longitudinal Distance vs.\ Longitudinal Distance at case start.\label{fig:ltt_stage2}]{%
        \includegraphics[width=0.48\textwidth]{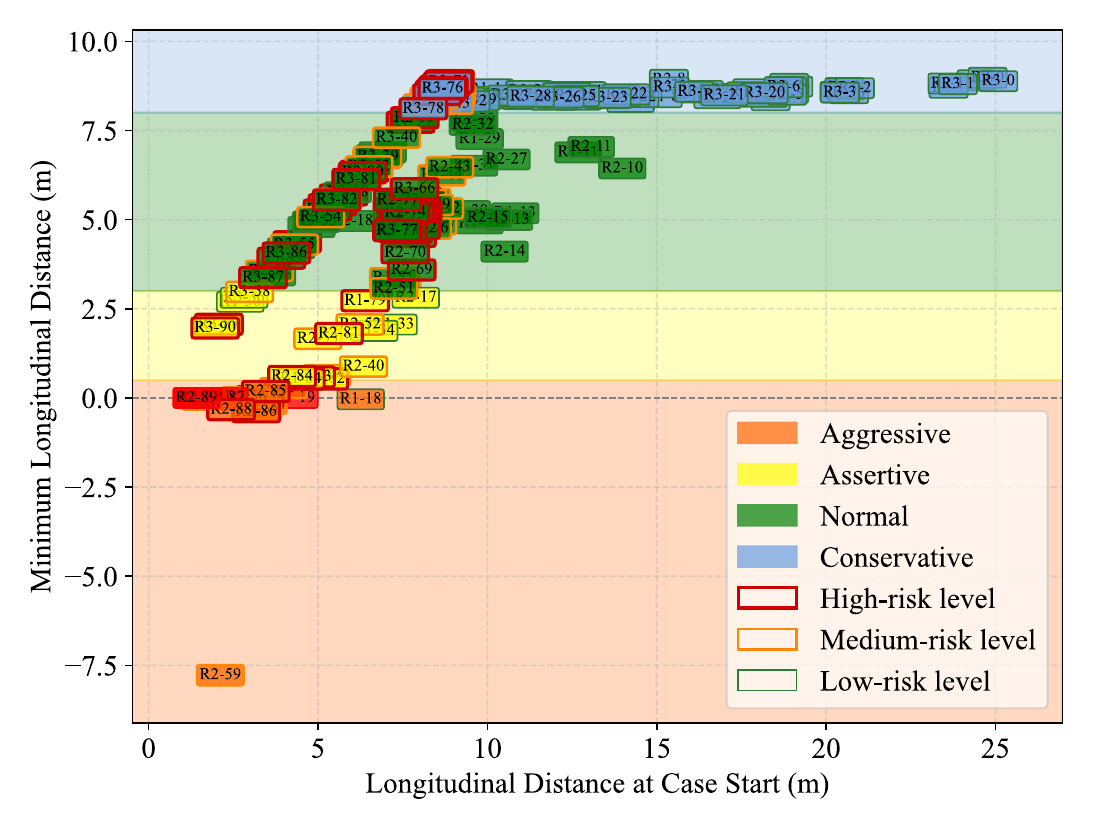}%
    }
    \hfill
    \subfloat[Post-Encroachment Time vs.\ Minimum Longitudinal Distance.\label{fig:ltt_stage3}]{%
        \includegraphics[width=0.48\textwidth]{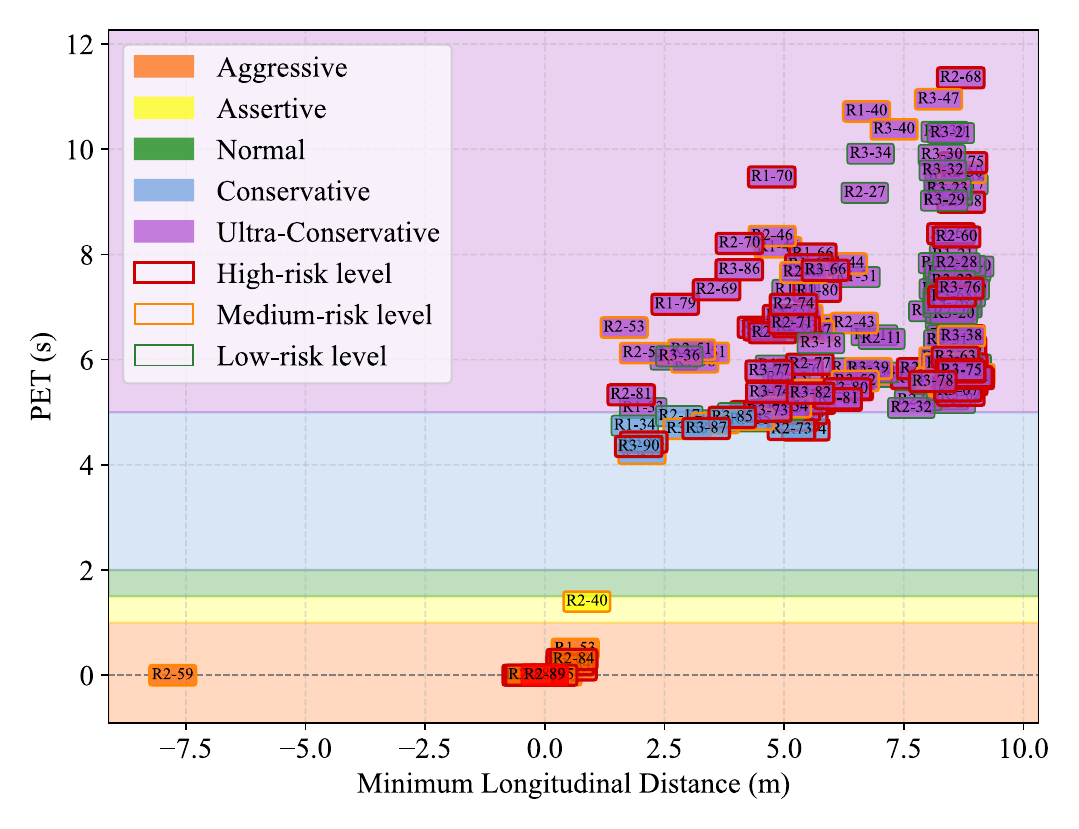}%
    }\\[4pt]
    \subfloat[Reaction time vs.\ TTC at case start.\label{fig:ltt_reaction}]{%
        \includegraphics[width=0.48\textwidth]{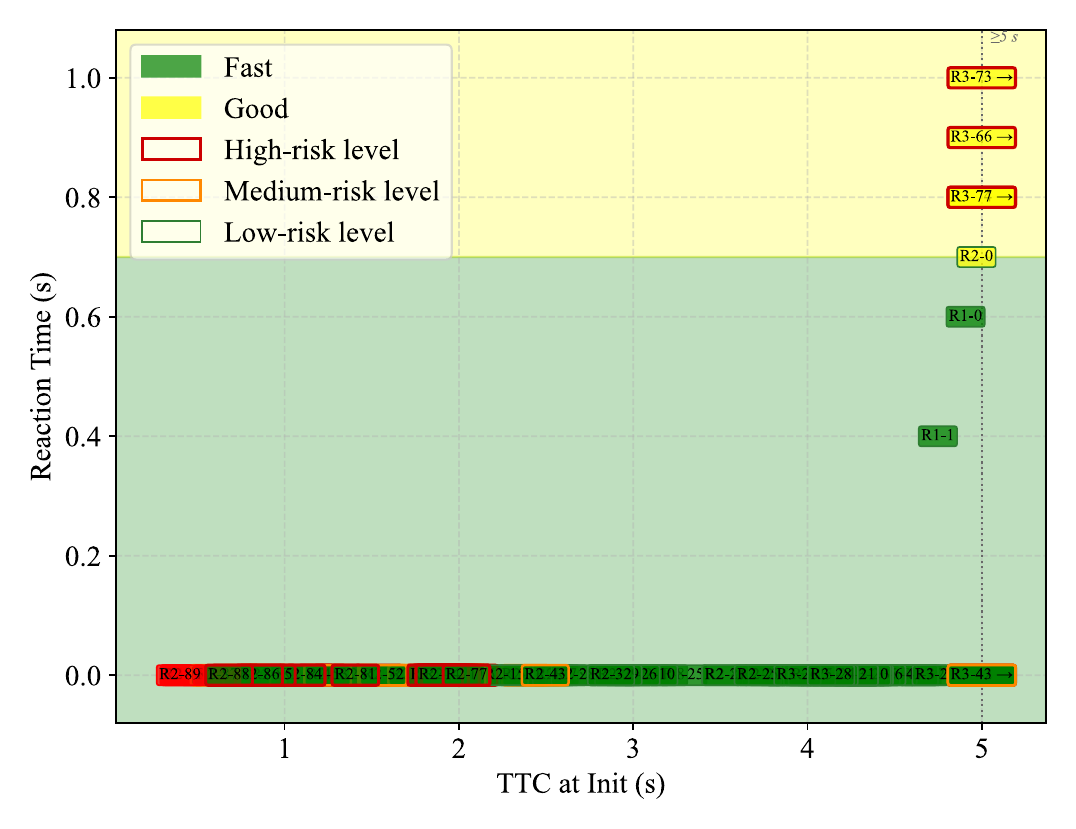}%
    }
    \hfill
    \subfloat[Normalized Deceleration vs.\ required deceleration.\label{fig:ltt_decel}]{%
        \includegraphics[width=0.48\textwidth]{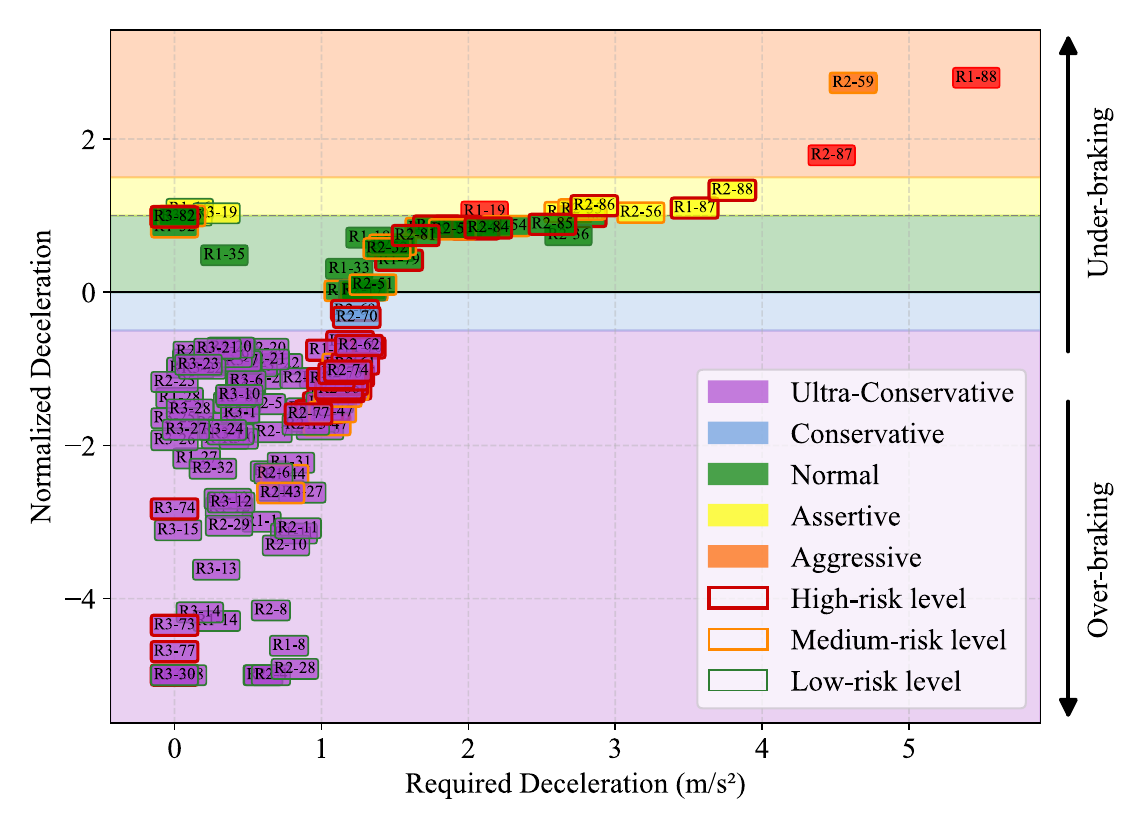}%
    }
    \caption{Analysis results for the execute unprotected left turn competency.}
    \label{fig:ltt_all}
\end{figure*}

\subsubsection*{Competency Certification}

Autoware.Universe \textbf{fails} this competency with an overall collision rate of $2.6\%$ (7 of 273 runs, $n = 91$ per round).
Five distinct cases collide (IDs 19, 58, 87, 88, 89), and cases 58 and 89 collide in two runs each.

\subsubsection*{Behavioral Cautiousness Assessment}

\textbf{Stage 1: Normalized Minimum TTC.}
In the scorecard (Fig.~\ref{fig:ltt_scorecard}), 245 of 266 non-collision runs ($92\%$) are Ultra-Conservative.
The remaining non-collision runs are Aggressive (9), Assertive (4), Normal (3), and Conservative (5).
In the scatter (Fig.~\ref{fig:ltt_stage1}), most non-collision runs lie between $-0.55$ and the $-5$ floor, across initial TTCs from $0.6\,\text{s}$ to above $5\,\text{s}$.
All seven collisions sit at the $+5$ ceiling at initial TTCs of $0.40$--$1.11\,\text{s}$.
Seven Aggressive non-collision runs (R1-18, R1-55, R1-59, R2-36, R2-55, R2-86, R2-88) also sit at the $+5$ ceiling with a minimum TTC of zero, a PET of 0, and negative minimum longitudinal distance, and six of them have initial TTCs of $0.68$--$1.64\,\text{s}$.
Four other Aggressive and Assertive runs (R3-66, R3-73, R3-74, R3-77) have initial TTCs of $5.2$--$9.1\,\text{s}$ and keep $4.7$--$5.9\,\text{m}$ of longitudinal distance and $5.0$--$7.7\,\text{s}$ of PET.

\textbf{Stage 2: Minimum Longitudinal Clearance.}
In the scorecard (Fig.~\ref{fig:ltt_scorecard}), 87 non-collision runs are Normal ($3$--$8\,\text{m}$) and 118 are Conservative ($8$--$15\,\text{m}$), together $77\%$ of non-collision runs.
Another 29 runs are Ultra-Conservative.
Aggressive and Assertive distances (${<}3\,\text{m}$) account for 32 runs, 13 of them Aggressive.
The scatter (Fig.~\ref{fig:ltt_stage2}) shows minimum distance rising with initial distance up to about $9\,\text{m}$ of initial distance and then leveling off (overall $r \approx 0.65$).
Of the 73 non-collision runs that start $10\,\text{m}$ or more away, 60 end at $8.4$--$9.0\,\text{m}$ (Conservative).
Collision runs end at $0\,\text{m}$ with initial distances of $1.4$--$4.3\,\text{m}$.
Nine non-collision runs have negative minimum distance, i.e., the AV passes the conflict-zone boundary without a collision.
Eight of them lie between $-0.003$ and $-0.39\,\text{m}$, and R2-59 reaches $-7.8\,\text{m}$.
These nine runs are Aggressive in both the scatter and the scorecard.

\textbf{Stage 3: Post-Encroachment Time and Lateral Distance.}
In the scorecard (Fig.~\ref{fig:ltt_scorecard}), the lateral distance row is Ultra-Conservative for 248 non-collision runs, i.e., the BV is more than $15\,\text{m}$ beyond the conflict zone when the AV enters it.
For PET, 221 non-collision runs ($83\%$) are Ultra-Conservative ($\geq 5\,\text{s}$) and 26 are Conservative ($2$--$5\,\text{s}$).
The other 19 non-collision runs are Aggressive (18, PET $<1\,\text{s}$) or Assertive (1, PET $1.4\,\text{s}$).
The scatter (Fig.~\ref{fig:ltt_stage3}) shows that most non-collision runs yield to the BV (PET $>0$), with over $90\%$ of non-collision runs above $1.5\,\text{s}$, spread between $4.2$ and $11.4\,\text{s}$.
The Aggressive non-collision runs have PET between $0$ and $0.5\,\text{s}$, and 13 of them are at exactly $\text{PET}=0$ (overlapping conflict-point occupancy without a collision).
The seven collision runs have a PET of 0.
All 19 non-collision runs with PET below $1.5\,\text{s}$ have initial TTCs of up to $1.64\,\text{s}$, and every non-collision run with an initial TTC above $1.64\,\text{s}$ has a PET of at least $4.2\,\text{s}$.

\subsubsection*{Behavior Diagnosis}

\textbf{Reaction Time.}
The reaction time plot (Fig.~\ref{fig:ltt_reaction}) shows that most runs are already braking at case start.
For non-collision runs, the median reaction time is $0\,\text{s}$.
The Good-zone runs (R2-0, R3-66, R3-73, R3-74, R3-77, $0.7$--$1.0\,\text{s}$) have initial TTCs near or above $5\,\text{s}$.
All collision runs also have a reaction time of $0\,\text{s}$, with the AV braking from case start at an average of $1.85$--$2.68\,\text{m/s}^2$ (peak $3.40$--$3.62\,\text{m/s}^2$).

\textbf{Deceleration Adequacy.}
In the deceleration adequacy plot (Fig.~\ref{fig:ltt_decel}), the largest group of non-collision runs is Ultra-Conservative (119 runs), i.e., braking harder than the conservative anchor, at required decelerations of at most $1.3\,\text{m/s}^2$.
Normal runs (33) extend up to $a_{\text{req}} \approx 2.8\,\text{m/s}^2$.
Of the 10 Assertive and Aggressive non-collision runs, 7 are near-miss runs at or just past the conflict-zone boundary (R1-55, R1-87, R2-55, R2-56, R2-59, R2-86, R2-88), at $a_{\text{req}} \approx 2.7$--$4.6\,\text{m/s}^2$ and actual average deceleration of $2.3$--$3.2\,\text{m/s}^2$ (R2-59 brakes at only $0.39\,\text{m/s}^2$).
The other three (R1-16, R2-19, R3-19) have $a_{\text{req}}$ below $0.3\,\text{m/s}^2$.
Collision runs require $a_{\text{req}} = 2.1$--$9.4\,\text{m/s}^2$, above their actual $1.85$--$2.68\,\text{m/s}^2$.
R1-19 is Assertive ($1.05$), and R2-87 and R1-88 are Aggressive ($1.79$ and $2.80$).

All seven collisions and all 19 non-collision runs with PET below $1.5\,\text{s}$ start at initial TTCs of $0.40$--$1.64\,\text{s}$.
In this range the AV brakes from case start, but its average deceleration ($1.85$--$2.68\,\text{m/s}^2$ in collisions, mostly $2.3$--$3.8\,\text{m/s}^2$ in near-misses) is below the required deceleration in every collision.
Runs with larger initial TTC yield with a PET of at least $4.2\,\text{s}$ and minimum longitudinal distance mostly in the Normal-to-Conservative range ($3$--$9\,\text{m}$).

\subsection{Detect and Respond to Right-Turning Vehicle}
\label{subsec:right_turn_straight}

\begin{figure*}[!p]
    \centering
    \subfloat[Scorecard.\label{fig:rts_scorecard}]{%
        \includegraphics[width=0.48\textwidth]{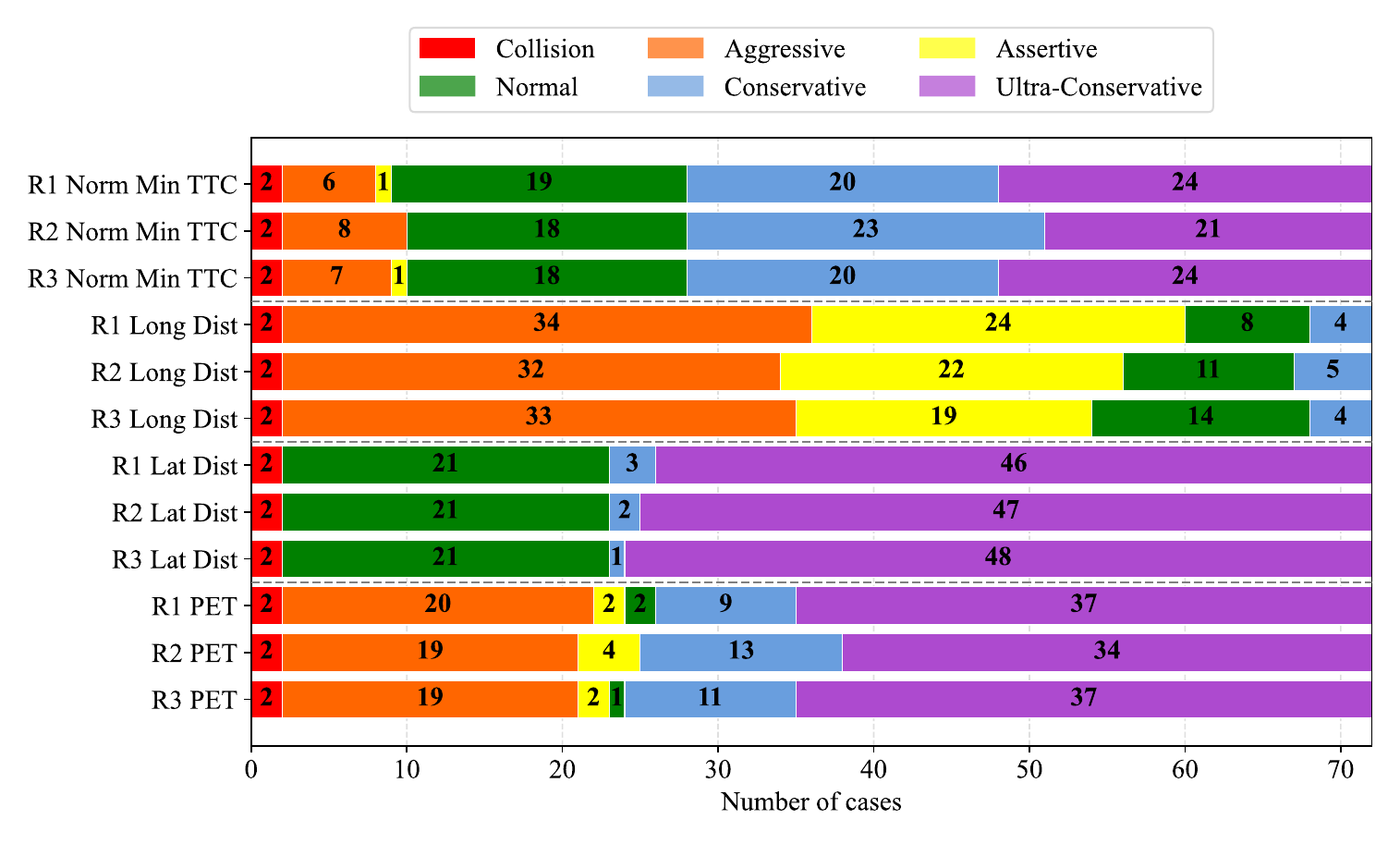}%
    }
    \hfill
    \subfloat[Normalized Minimum TTC vs.\ TTC at case start.\label{fig:rts_stage1}]{%
        \includegraphics[width=0.48\textwidth]{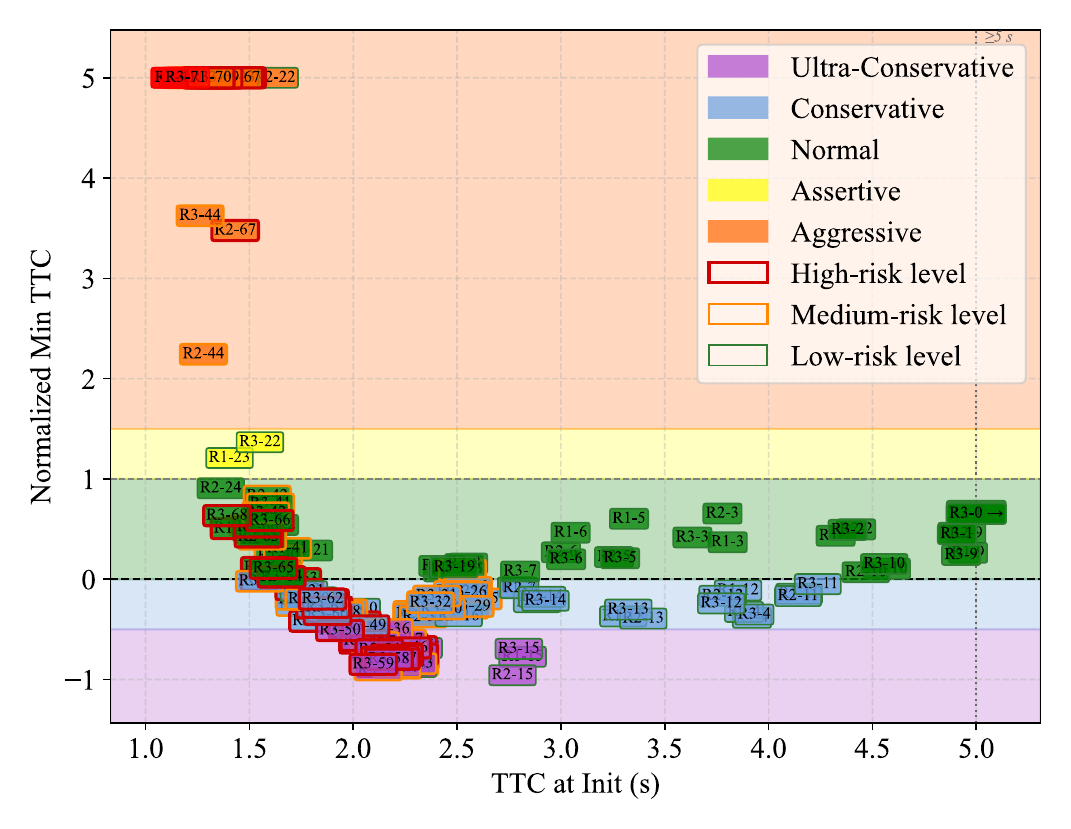}%
    }\\[4pt]
    \subfloat[Minimum Longitudinal Distance vs.\ Longitudinal Distance at case start.\label{fig:rts_stage2}]{%
        \includegraphics[width=0.48\textwidth]{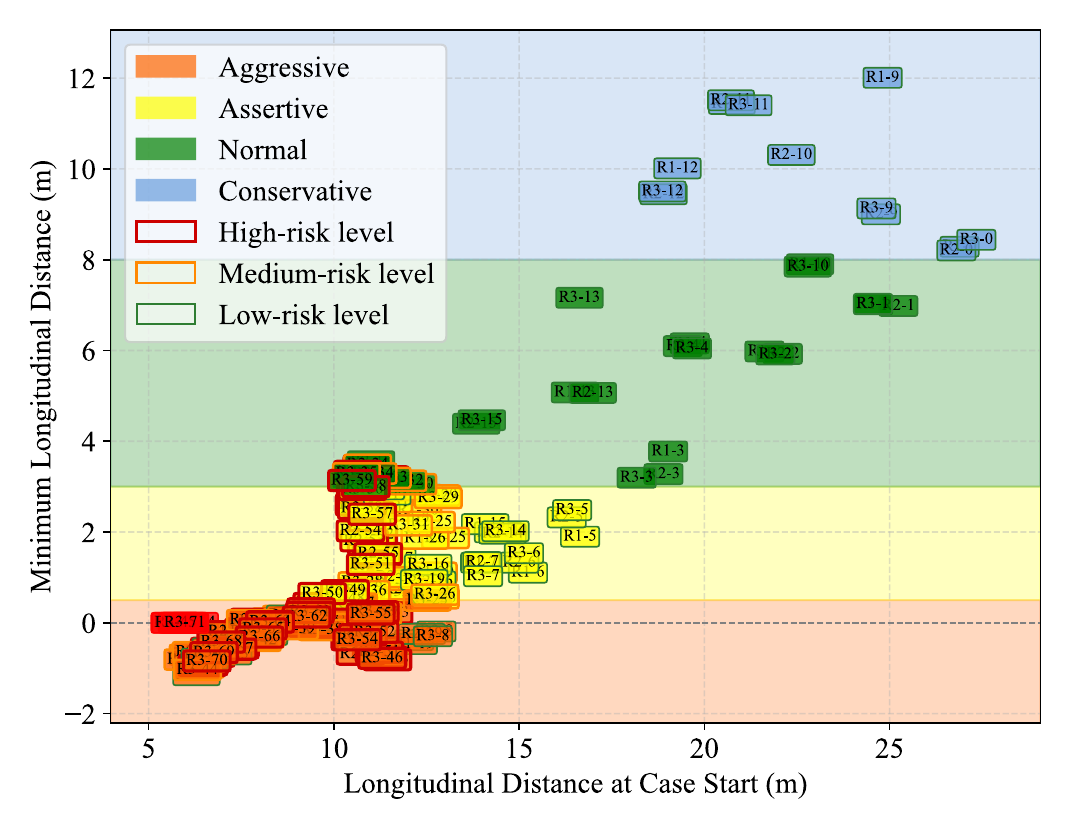}%
    }
    \hfill
    \subfloat[Post-Encroachment Time vs.\ Minimum Longitudinal Distance.\label{fig:rts_stage3}]{%
        \includegraphics[width=0.48\textwidth]{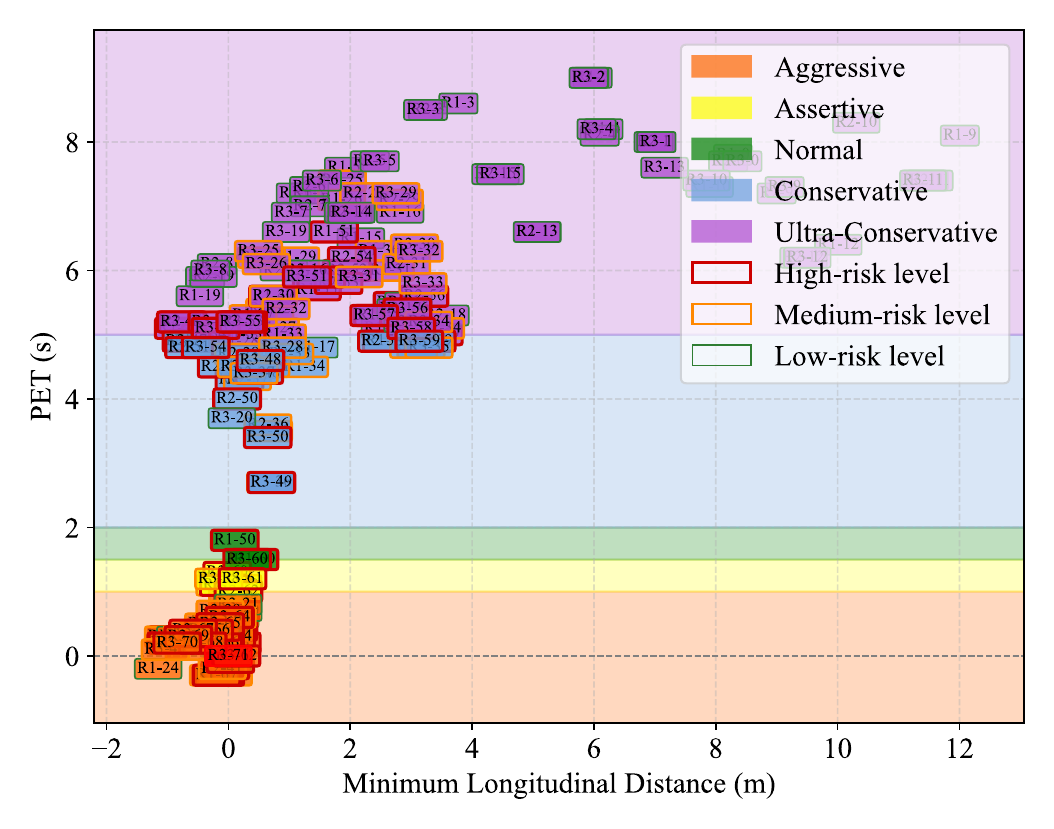}%
    }\\[4pt]
    \subfloat[Reaction time vs.\ TTC at case start.\label{fig:rts_reaction}]{%
        \includegraphics[width=0.48\textwidth]{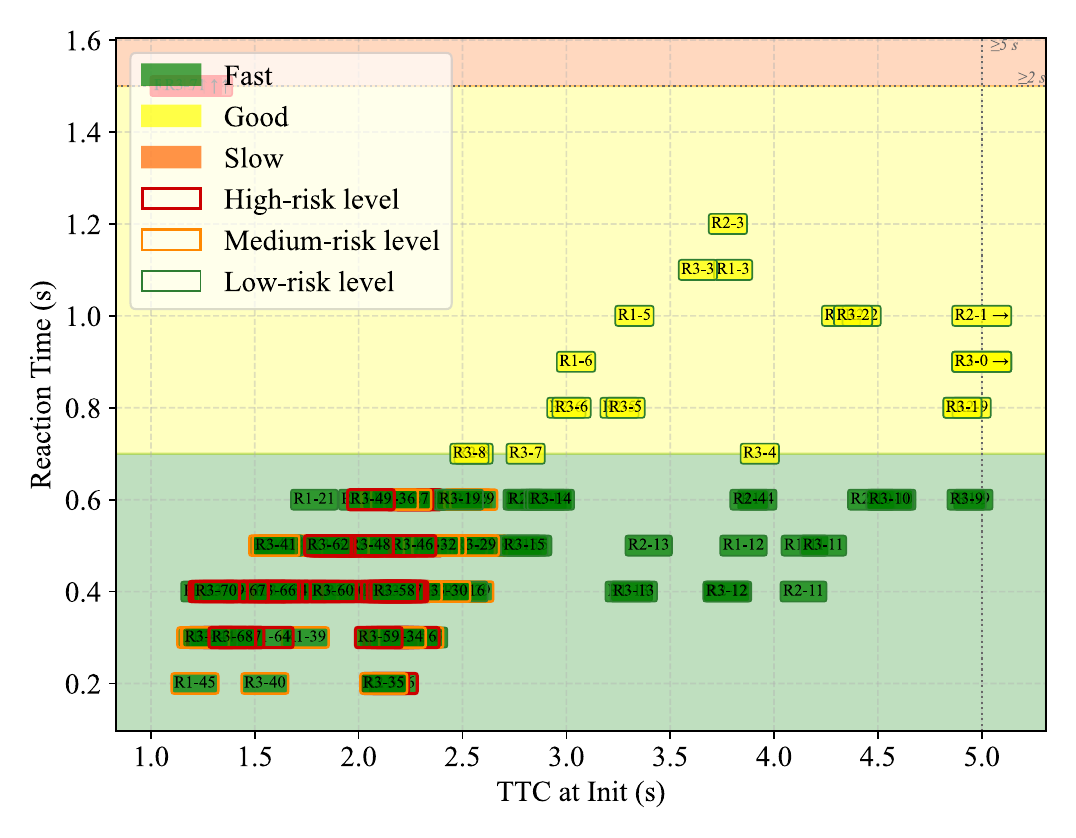}%
    }
    \hfill
    \subfloat[Normalized Deceleration vs.\ required deceleration.\label{fig:rts_decel}]{%
        \includegraphics[width=0.48\textwidth]{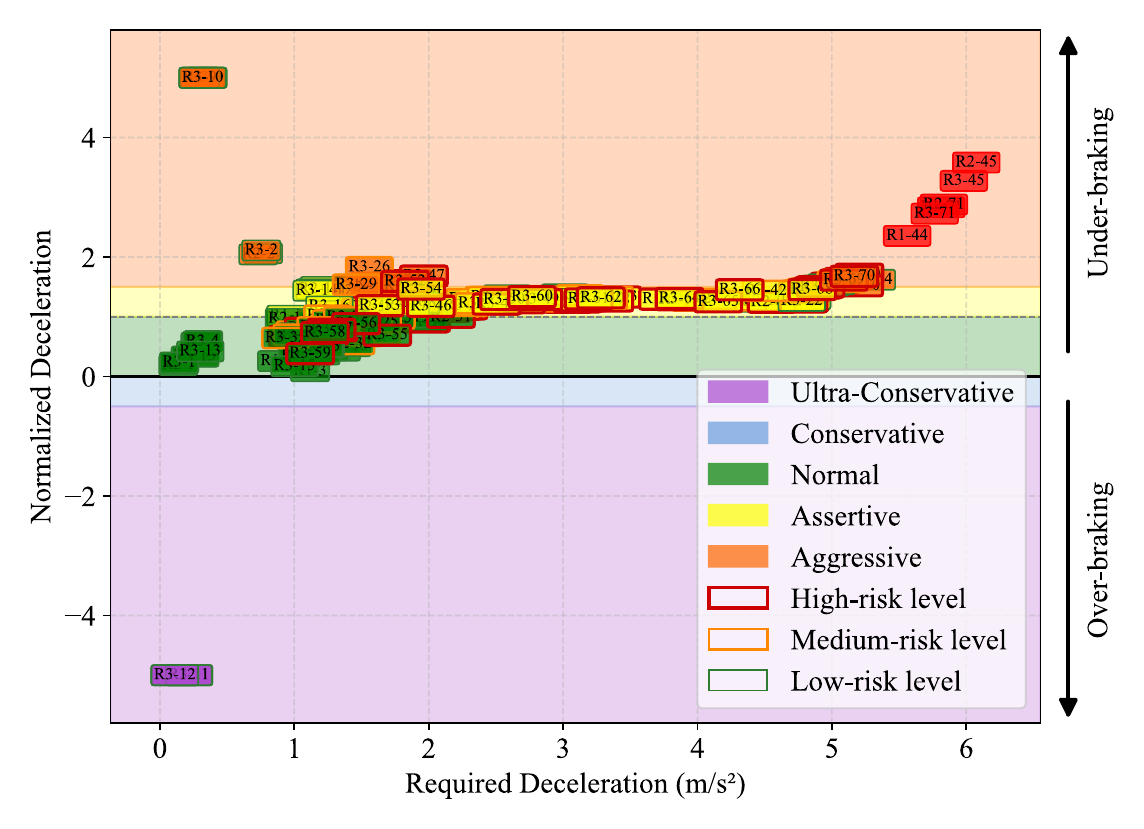}%
    }
    \caption{Analysis results for the detect and respond to right-turning vehicle competency.}
    \label{fig:rts_all}
\end{figure*}

\subsubsection*{Competency Certification}

Autoware.Universe \textbf{fails} this competency with a collision rate of $2.8\%$ (6 of 216 runs, $n = 72$ per round).
Case 71 collides in all three runs, case 45 in two (R2-45, R3-45), and case 44 in one (R1-44).

\subsubsection*{Behavioral Cautiousness Assessment}

\textbf{Stage 1: Normalized Minimum TTC.}
In the scorecard (Fig.~\ref{fig:rts_scorecard}), 132 of 210 non-collision runs ($63\%$) are Conservative or Ultra-Conservative, 55 ($26\%$) are Normal, 2 are Assertive, and 21 ($10\%$) are Aggressive.
In the scatter (Fig.~\ref{fig:rts_stage1}), the non-collision runs spread across the zones.
Runs with negative Normalized Minimum TTC (Conservative and Ultra-Conservative) have initial TTCs of $1.55$--$4.24\,\text{s}$, and all Ultra-Conservative runs (down to $-0.96$) lie at $1.94$--$2.82\,\text{s}$.
Normal runs appear both at low initial TTC ($1.36$--$1.79\,\text{s}$) and at $2.4$--$5.5\,\text{s}$.
The Aggressive and Assertive runs all lie at initial TTCs of $1.21$--$1.62\,\text{s}$, most at the $+5$ ceiling with a minimum TTC of zero, and all have negative minimum longitudinal distance ($-0.20$ to $-1.15\,\text{m}$).
The collision runs sit at the $+5$ ceiling at initial TTCs of $1.14$--$1.25\,\text{s}$, next to these near-miss runs.

\textbf{Stage 2: Minimum Longitudinal Clearance.}
In the scorecard (Fig.~\ref{fig:rts_scorecard}), 164 of 210 non-collision runs ($78\%$) are Aggressive or Assertive (${<}3\,\text{m}$).
The Aggressive zone alone holds 99 runs, of which 64 have negative minimum distance (down to $-1.15\,\text{m}$), i.e., the AV passes the conflict-zone boundary without a collision.
Normal runs number 33 and Conservative runs 13, and no run exceeds $15\,\text{m}$.
The scatter (Fig.~\ref{fig:rts_stage2}) shows minimum distance rising with initial distance ($r \approx 0.87$).
All 174 non-collision runs that start within $15\,\text{m}$ end with at most $4.5\,\text{m}$ (from $-1.15\,\text{m}$), while the 18 runs that start beyond $20\,\text{m}$ keep $5.9$--$12.0\,\text{m}$.

\textbf{Stage 3: Post-Encroachment Time and Lateral Distance.}
In the scorecard (Fig.~\ref{fig:rts_scorecard}), the lateral distance row is Ultra-Conservative for 141 non-collision runs and Normal for 63.
For PET, 108 non-collision runs ($51\%$) are Ultra-Conservative ($5$--$9\,\text{s}$) and 58 ($28\%$) are Aggressive (PET $<1\,\text{s}$).
The remaining 44 runs are Conservative (33), Assertive (8), or Normal (3).
Aggressive PET ranges from $-0.3$ to $0.9\,\text{s}$, and only 10 runs are negative (AV clears the conflict point first).
Most Aggressive runs have small positive PET, i.e., the BV clears the conflict point $0.1$--$0.9\,\text{s}$ before the AV reaches it.
The collision runs have a PET of 0.
The scatter (Fig.~\ref{fig:rts_stage3}) shows most yielding runs at PET of $5$--$9\,\text{s}$ (Ultra-Conservative) and $2.7$--$5\,\text{s}$ (Conservative), $79\%$ of them with minimum distances below $4\,\text{m}$.
The Aggressive runs cluster at minimum distances of $-1.15$ to $0.16\,\text{m}$ and PET of $-0.3$ to $0.9\,\text{s}$.

\subsubsection*{Behavior Diagnosis}

\textbf{Reaction Time.}
The reaction time plot (Fig.~\ref{fig:rts_reaction}) shows all non-collision runs braking, with reaction times of $0.2$--$1.2\,\text{s}$ (Fast in 191 runs, Good in 19) across initial TTCs of $1.2$--$5.5\,\text{s}$.
None of the six collision runs brakes, and their peak deceleration is $0.0001$--$0.0002\,\text{m/s}^2$, compared with average decelerations of $0.2$--$1.4\,\text{m/s}^2$ for non-collision runs.
Case 44 collides without braking in R1-44 (initial TTC $1.25\,\text{s}$), and its other two runs brake (average $0.71$--$0.73\,\text{m/s}^2$, peak $2.59$--$2.61\,\text{m/s}^2$) at initial TTCs of $1.26$--$1.28\,\text{s}$ and avoid collision with minimum distance of $-0.98$ to $-1.04\,\text{m}$.
Case 45 brakes and avoids collision in R1-45 (initial TTC $1.21\,\text{s}$, minimum distance $-0.81\,\text{m}$) and collides without braking in its other two runs (initial TTC $1.14$--$1.15\,\text{s}$).
Case 71 collides without braking in all three runs (initial TTC $1.18$--$1.20\,\text{s}$).
All runs without braking start at initial TTCs of $1.14$--$1.25\,\text{s}$.

\textbf{Deceleration Adequacy.}
The deceleration adequacy plot (Fig.~\ref{fig:rts_decel}) plots each run's required deceleration $a_{\text{req}}$ against Normalized Deceleration (0 matches the conservative anchor, 1 matches the aggressive anchor).
Of the 114 non-collision runs with $a_{\text{req}} < 2\,\text{m/s}^2$, 65 ($57\%$) are Normal.
The rest include 24 Assertive runs, 19 Aggressive runs (cases 0, 9, and 10 at the $+5$ ceiling with $a_{\text{req}} \approx 0.3\,\text{m/s}^2$), and 6 Ultra-Conservative runs (cases 11 and 12 at the $-5$ floor).
The 96 non-collision runs needing $2.0$--$5.3\,\text{m/s}^2$ sit at Normalized Deceleration $0.99$--$1.71$, mostly Assertive (79) with 16 Aggressive, i.e., they brake slightly softer than the aggressive anchor and avoid collision.
The collision runs require $a_{\text{req}} = 5.56$--$6.07\,\text{m/s}^2$, higher than every non-collision run (maximum $5.30\,\text{m/s}^2$) and below $A_{\text{MAX}} = 7.06\,\text{m/s}^2$.
With near-zero average deceleration, they lie at Normalized Deceleration $2.35$--$3.58$ (Aggressive).
The non-collision near-miss runs at similar initial TTC ($1.21$--$1.62\,\text{s}$) brake at an average of only $0.61$--$1.42\,\text{m/s}^2$ and end with negative minimum distance.

The collisions in this competency coincide with the absence of any braking response in runs whose required deceleration lies within $A_{\text{MAX}}$.

\subsection{Execute Right Turn}
\label{subsec:right_turn_turn}

\begin{figure*}[!p]
    \centering
    \subfloat[Scorecard.\label{fig:rtt_scorecard}]{%
        \includegraphics[width=0.48\textwidth]{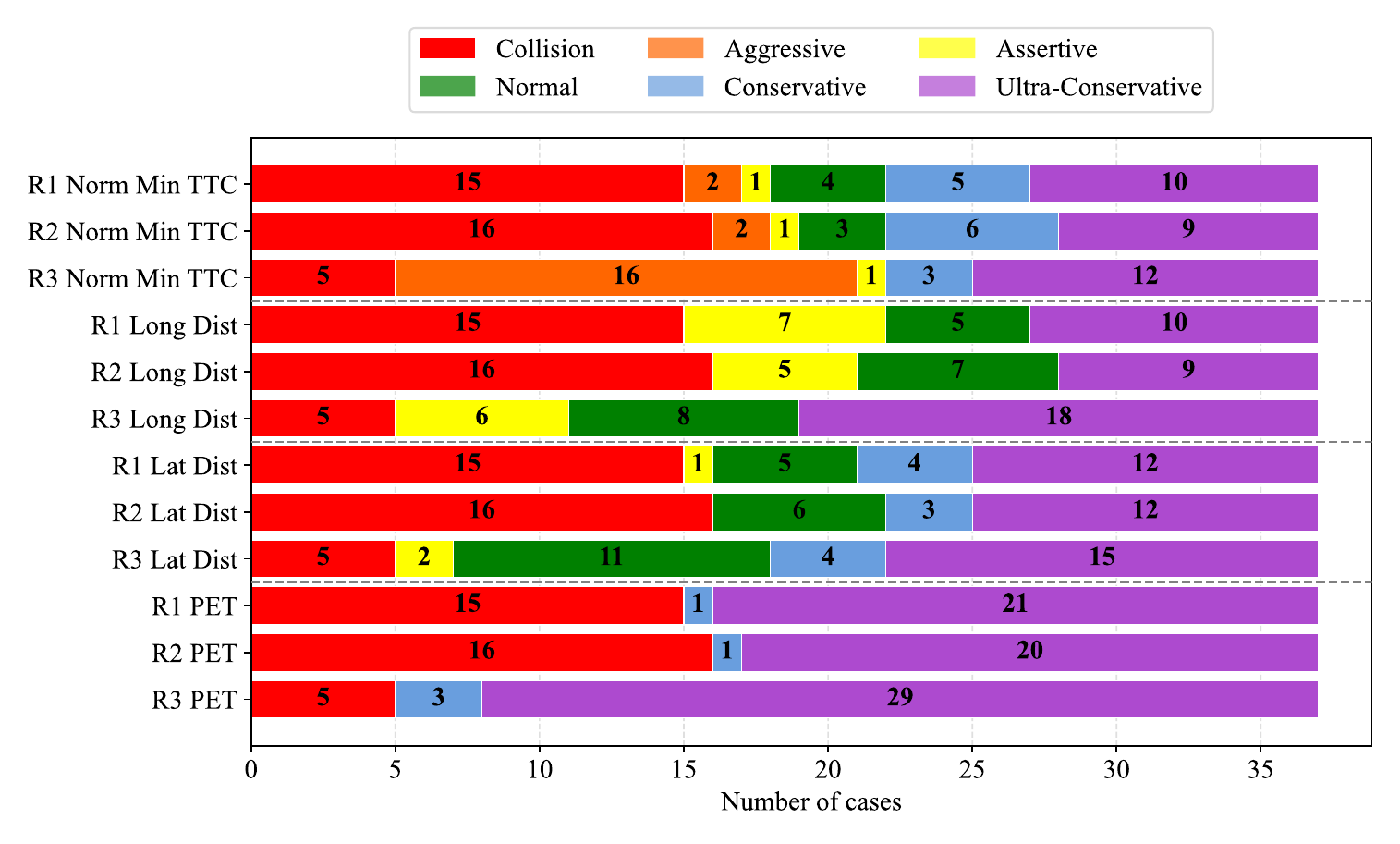}%
    }
    \hfill
    \subfloat[Normalized Minimum TTC vs.\ Initial TTC.\label{fig:rtt_stage1}]{%
        \includegraphics[width=0.48\textwidth]{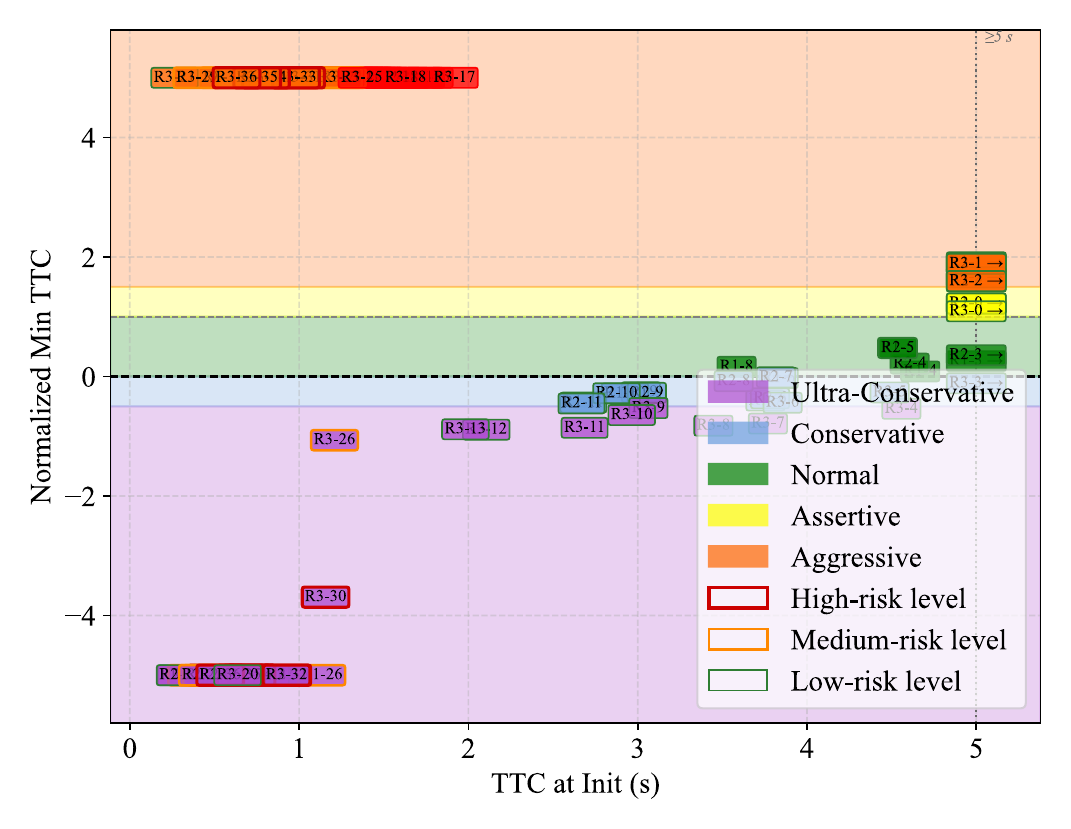}%
    }\\[4pt]
    \subfloat[Minimum Longitudinal Clearance vs.\ Initial Longitudinal Clearance.\label{fig:rtt_stage2}]{%
        \includegraphics[width=0.48\textwidth]{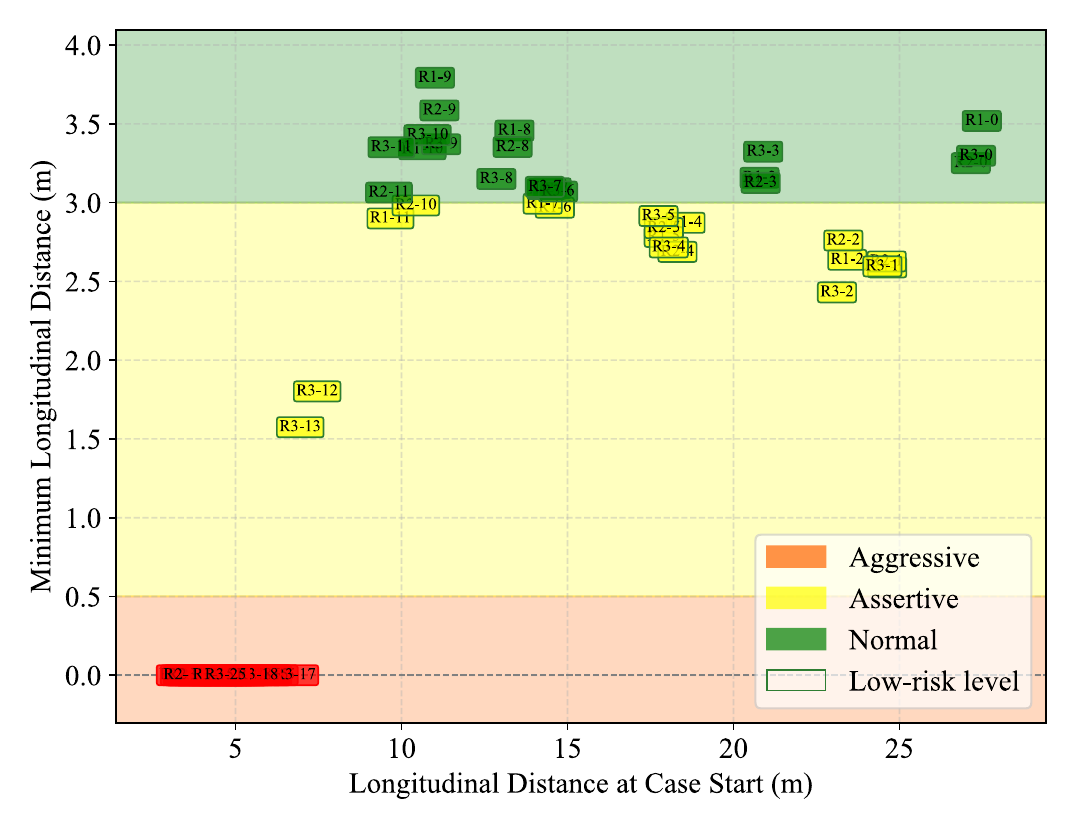}%
    }
    \hfill
    \subfloat[Post-Encroachment Time vs.\ Minimum Longitudinal Clearance.\label{fig:rtt_stage3}]{%
        \includegraphics[width=0.48\textwidth]{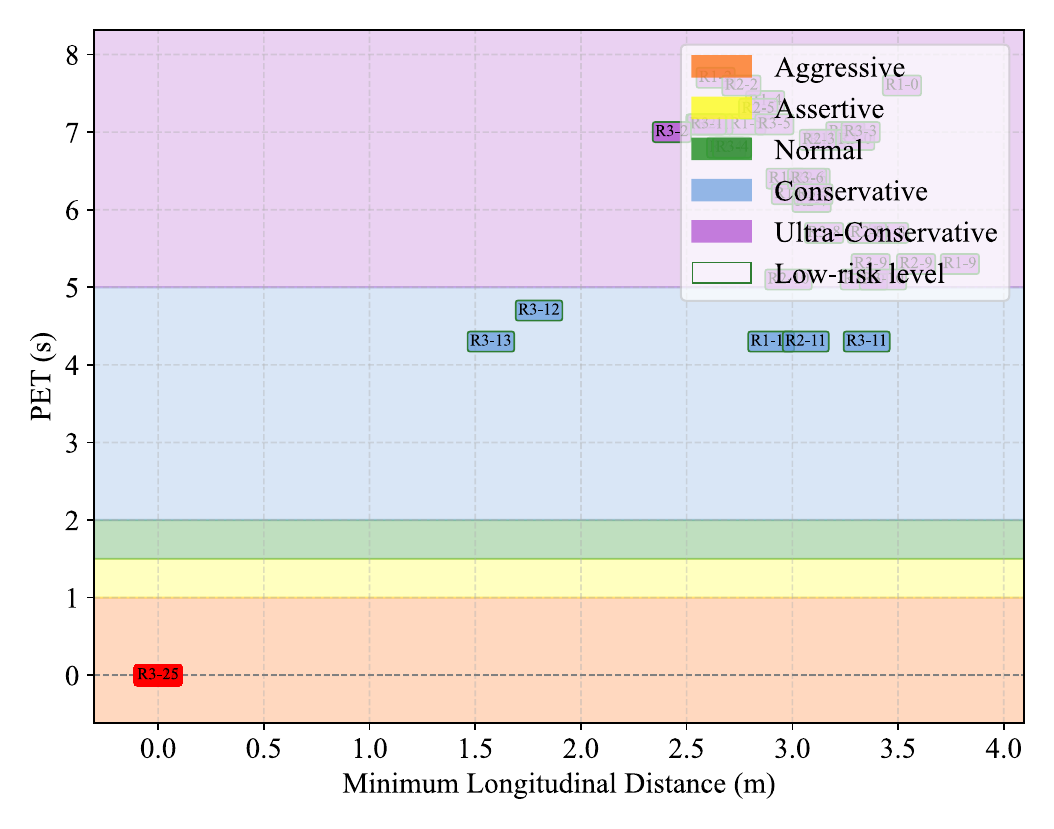}%
    }\\[4pt]
    \subfloat[Reaction time vs.\ Initial TTC.\label{fig:rtt_reaction}]{%
        \includegraphics[width=0.48\textwidth]{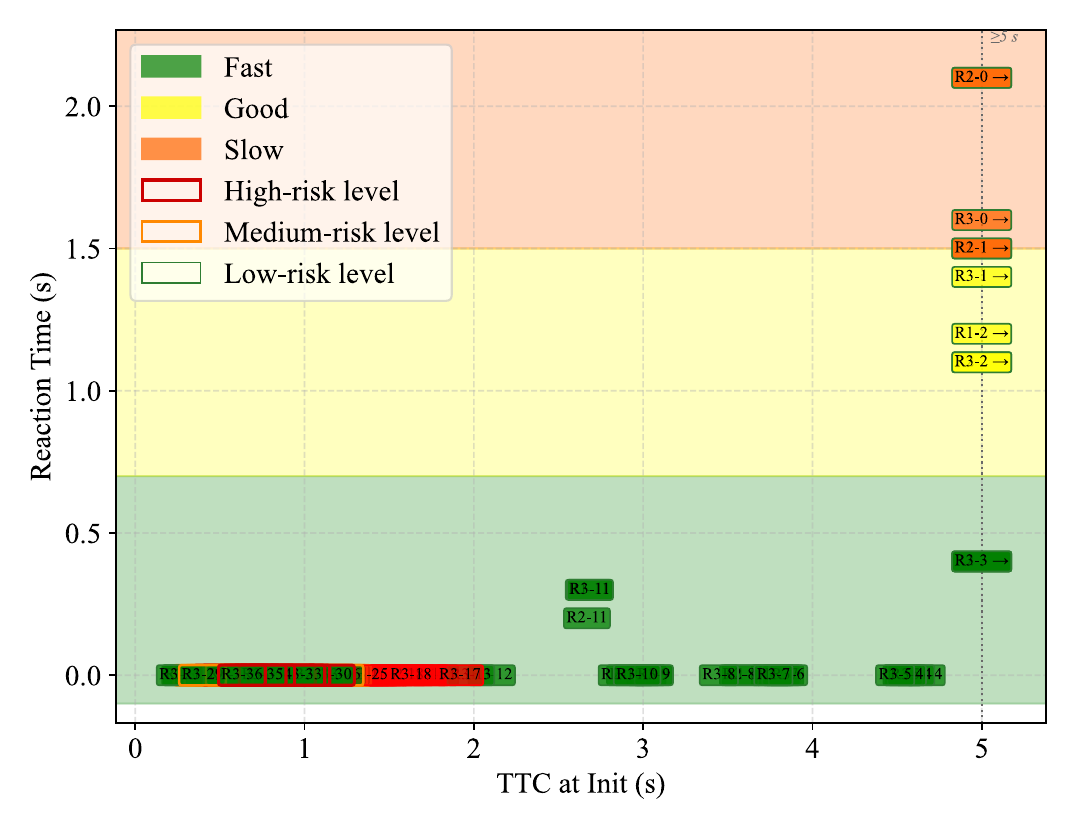}%
    }
    \hfill
    \subfloat[Normalized Deceleration vs.\ required deceleration.\label{fig:rtt_decel}]{%
        \includegraphics[width=0.48\textwidth]{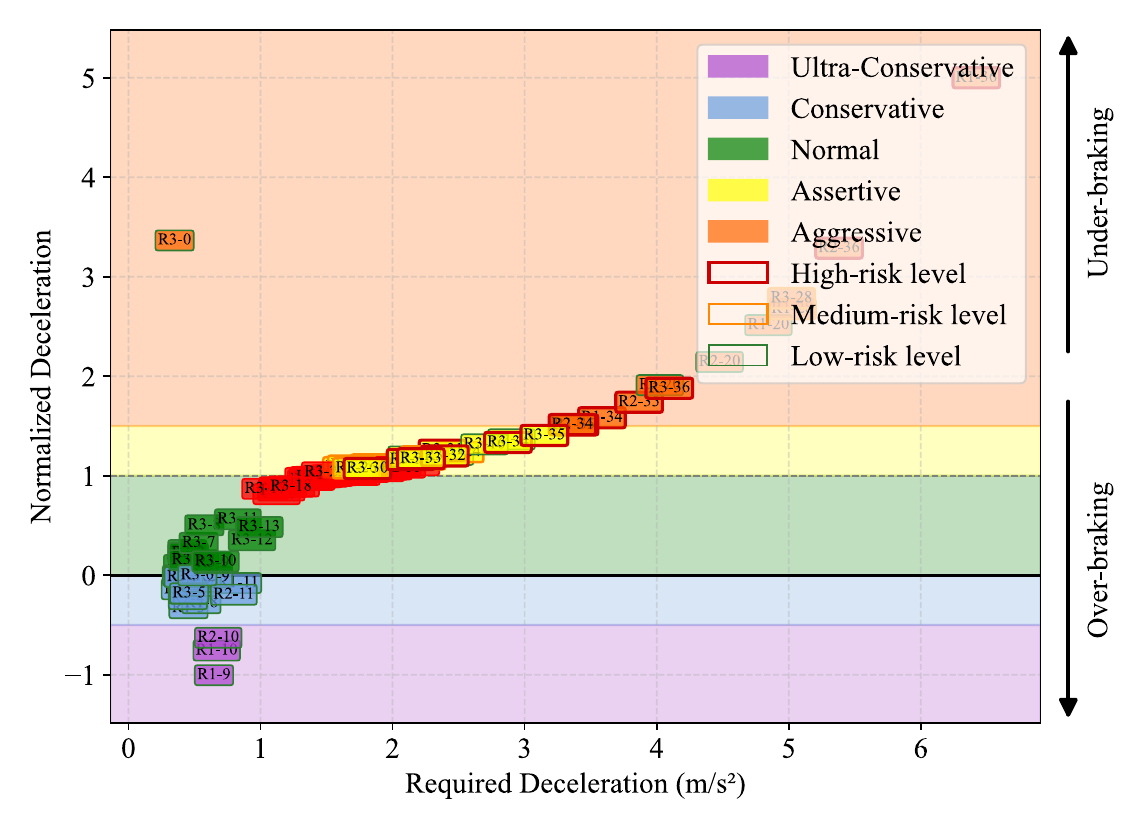}%
    }
    \caption{Analysis results for the execute right turn competency.}
    \label{fig:rtt_all}
\end{figure*}

\subsubsection*{Competency Certification}

Autoware.Universe \textbf{fails} this competency.
Collisions occur in 36 of 111 runs ($32.4\%$).

\subsubsection*{Behavioral Cautiousness Assessment}

\textbf{Stage 1: Normalized Minimum TTC.}
In the scorecard (Fig.~\ref{fig:rtt_scorecard}), Normalized Minimum TTC places 20 of the 75 survivors in the Aggressive zone, 3 in Assertive, 7 in Normal, 14 in Conservative, and 31 in Ultra-Conservative.
In the scatter (Fig.~\ref{fig:rtt_stage1}), all 36 collisions sit at the assigned value of 5 (Aggressive), at Initial TTC $0.99$--$1.92\,\text{s}$.
Survivors with Initial TTC of $2.0$--$5.2\,\text{s}$ fall in the Normal, Conservative, or Ultra-Conservative zones ($-0.89$ to $0.48$), while cases 0--2 (Initial TTC $5.7$--$6.9\,\text{s}$) are Assertive or Aggressive ($1.09$--$1.91$) in all nine runs.
Of the 37 survivors with low Initial TTC ($0.26$--$1.25\,\text{s}$), 21 sit at the Ultra-Conservative floor ($-5$, minimum TTC $1.24$--$2.13\,\text{s}$) and 14 sit at the Aggressive ceiling, with the AV's front at the conflict point and a minimum TTC of $0\,\text{s}$.
These 14 runs account for 14 of the 20 Aggressive survivors.

\textbf{Stage 2: Minimum Longitudinal Clearance.}
In the scorecard (Fig.~\ref{fig:rtt_scorecard}), no survivor is Aggressive. Minimum Longitudinal Clearance places 18 survivors in the Assertive zone and 20 in the Normal zone (all between $1.57$ and $3.79\,\text{m}$), and 37 in the Ultra-Conservative zone.
In the scatter (Fig.~\ref{fig:rtt_stage2}), the cases in which the BV passes the conflict point first end with clearances in a narrow band of $2.4$--$3.8\,\text{m}$ for initial clearances of $9.6$--$27.5\,\text{m}$, i.e., the final clearance does not grow with the initial clearance.
R3-12 and R3-13 (initial clearance $7.0$--$7.5\,\text{m}$) end lower, at $1.57$--$1.80\,\text{m}$ (Assertive).
Collision cases start at $3.3$--$6.8\,\text{m}$.

\textbf{Stage 3: Post-Encroachment Time and Lateral Distance.}
In the scorecard (Fig.~\ref{fig:rtt_scorecard}), every survivor is Conservative or Ultra-Conservative for PET, with 70 of the 75 survivors Ultra-Conservative and no PET in the Aggressive, Assertive, or Normal zones.
The Lateral Distance row (BV's distance from the conflict point while the AV occupies it) is Ultra-Conservative for 39 survivors.
In the scatter (Fig.~\ref{fig:rtt_stage3}), 5 runs of cases 11--13 have PET $4.3$--$4.7\,\text{s}$ (Conservative), and the other cases in which the BV passes first have PET $5.1$--$7.7\,\text{s}$ (Ultra-Conservative).
Collision cases are assigned a PET of $0$.
Lateral Distance ranges from $1.0$ to $10.8\,\text{m}$ when the AV passes the conflict point first and is $31$--$37\,\text{m}$ when the BV passes first.

\subsubsection*{Behavior Diagnosis}

\textbf{Reaction Time.}
The reaction time plot (Fig.~\ref{fig:rtt_reaction}) shows braking onset at $0\,\text{s}$ (i.e., the AV is already decelerating at case start) in 96 of 111 runs, including all 36 collisions.
Overall, 102 runs are Fast, 5 Good, and 4 Slow. All Good and Slow runs are low-risk cases 0--2 with Initial TTC $5.7$--$6.9\,\text{s}$ (reaction time $1.1$--$2.1\,\text{s}$).
None of the collisions has a Slow reaction time.

\textbf{Deceleration Adequacy.}
Required deceleration ranges from $0.35$ to $6.4\,\text{m/s}^2$ in the deceleration adequacy plot (Fig.~\ref{fig:rtt_decel}), and actual average deceleration is $0.5$--$1.7\,\text{m/s}^2$ in every run except R3-0 ($0.10\,\text{m/s}^2$).
The 36 collisions require $1.04$--$2.18\,\text{m/s}^2$, receive $1.20$--$1.68\,\text{m/s}^2$, and have Normalized Deceleration $0.82$--$1.11$ (18 Normal, 18 Assertive, none Aggressive). That is, the AV braked approximately at the level needed to stop with the $0.75\,\text{m}$ minimum margin.
In 34 of the 36 collisions the AV was stationary at the end of the test (R3-17 and R3-18: $0.43$ and $0.14\,\text{m/s}$), with its center $2.6$--$5.1\,\text{m}$ short of the conflict point and the BV center $6.2$--$9.2\,\text{m}$ before it.
Among the survivors with low Initial TTC, 12 runs are Aggressive (required $3.4$--$6.4\,\text{m/s}^2$, actual $1.3$--$1.6\,\text{m/s}^2$), and in 10 of them the AV passes the conflict point first.
Survivors with Initial TTC of at least $2\,\text{s}$ mostly fall in the Normal or Conservative zones, with three Ultra-Conservative (R1-9, R1-10, R2-10, i.e., over-braking) and one Aggressive (R3-0).
The deceleration of the collision runs is not uniform over time.
In the first half of each collision run the AV decelerates at only $0.20$--$1.07\,\text{m/s}^2$ (median $0.47\,\text{m/s}^2$).
The peak deceleration of $4.23$--$4.68\,\text{m/s}^2$ occurs in the last $1.4\,\text{s}$ before contact in all 36 collisions (median $0.7\,\text{s}$), and a median of $61\%$ of the speed reduction takes place in the last second.

Although the average deceleration of the collision runs is close to the requirement, the AV brakes too little in the early part of the encounter and brakes hard only just before contact.
The average deceleration does not reveal this pattern and alone is not sufficient to assess braking adequacy.

\subsection{Detect and Respond to VRU Crossing Without Crosswalk}
\label{subsec:vru_no_crosswalk}

\begin{figure*}[!p]
    \centering
    \subfloat[Scorecard.\label{fig:vwc_scorecard}]{%
        \includegraphics[width=0.48\textwidth]{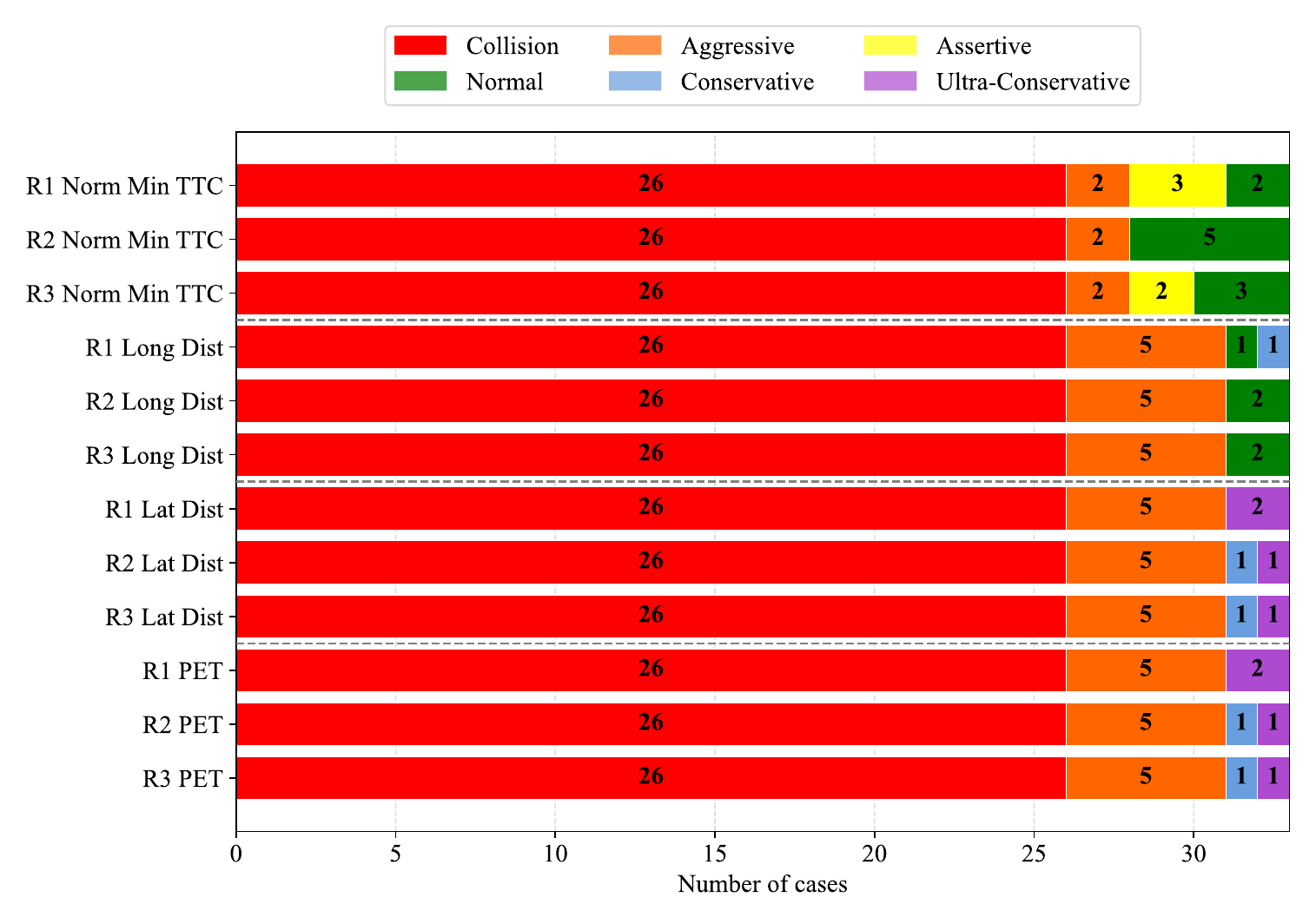}%
    }
    \hfill
    \subfloat[Normalized Minimum TTC vs.\ Initial TTC.\label{fig:vwc_stage1}]{%
        \includegraphics[width=0.48\textwidth]{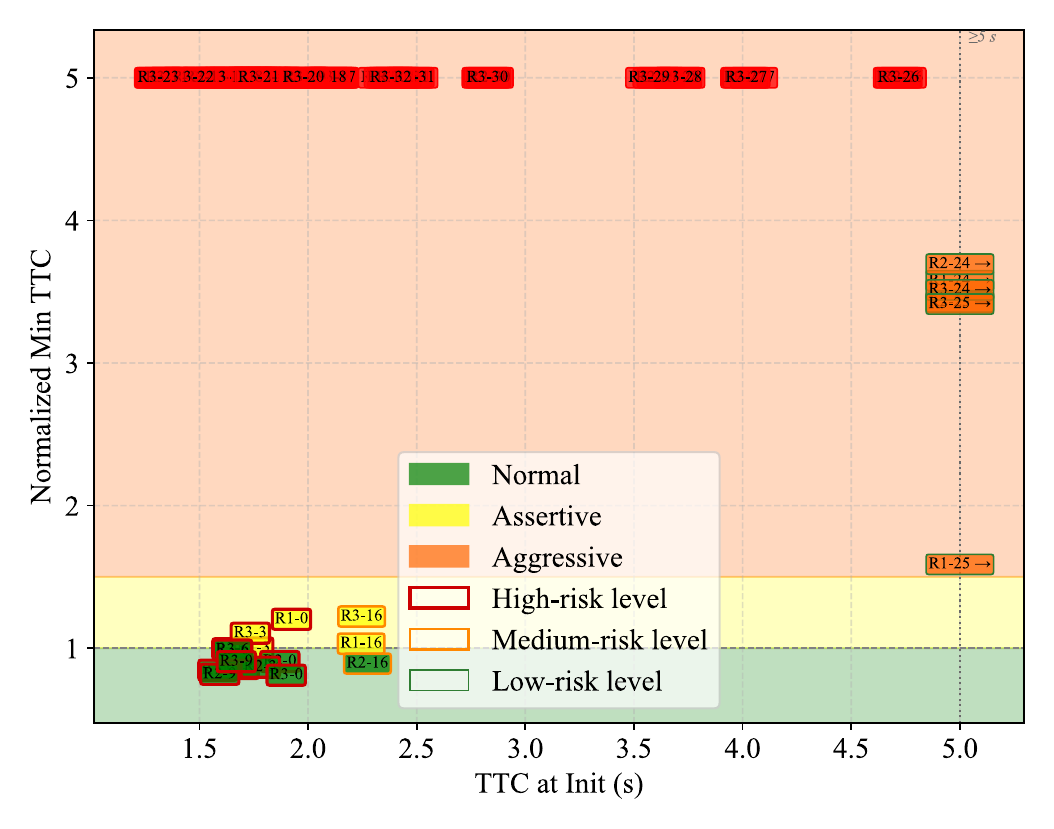}%
    }\\[4pt]
    \subfloat[Minimum Longitudinal Clearance vs.\ Initial Longitudinal Clearance.\label{fig:vwc_stage2}]{%
        \includegraphics[width=0.48\textwidth]{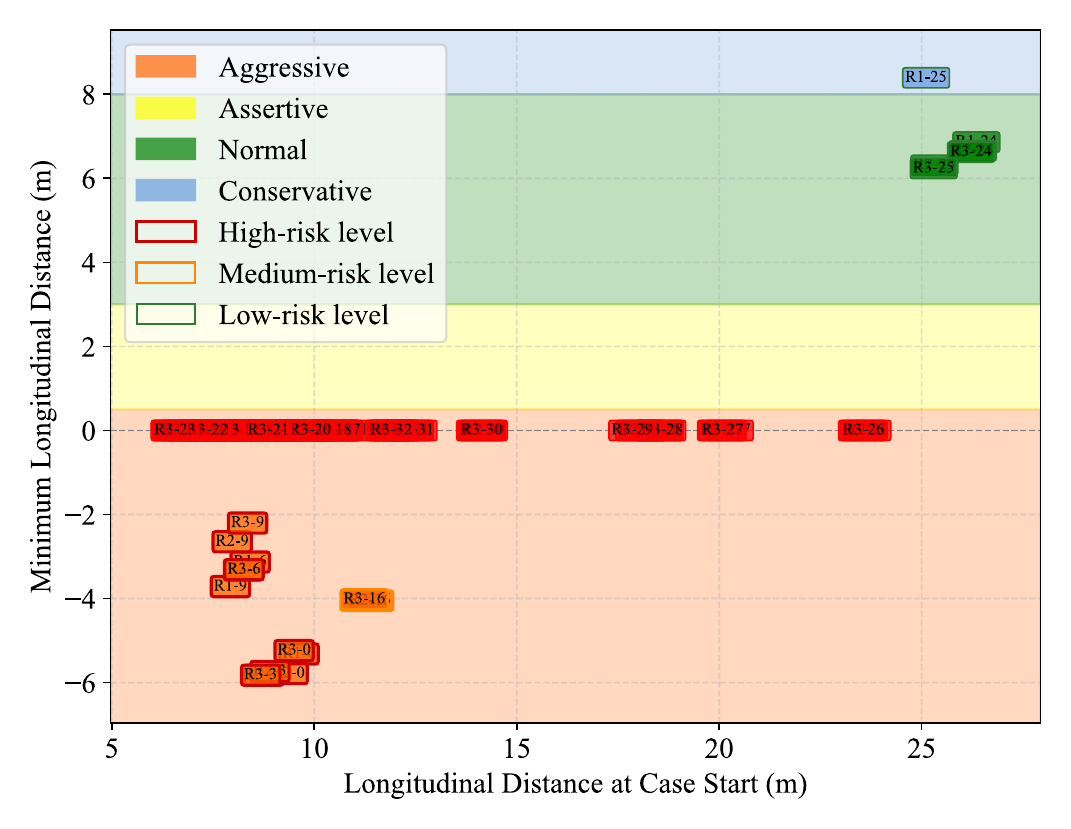}%
    }
    \hfill
    \subfloat[Post-Encroachment Time vs.\ Lateral Distance.\label{fig:vwc_stage3}]{%
        \includegraphics[width=0.48\textwidth]{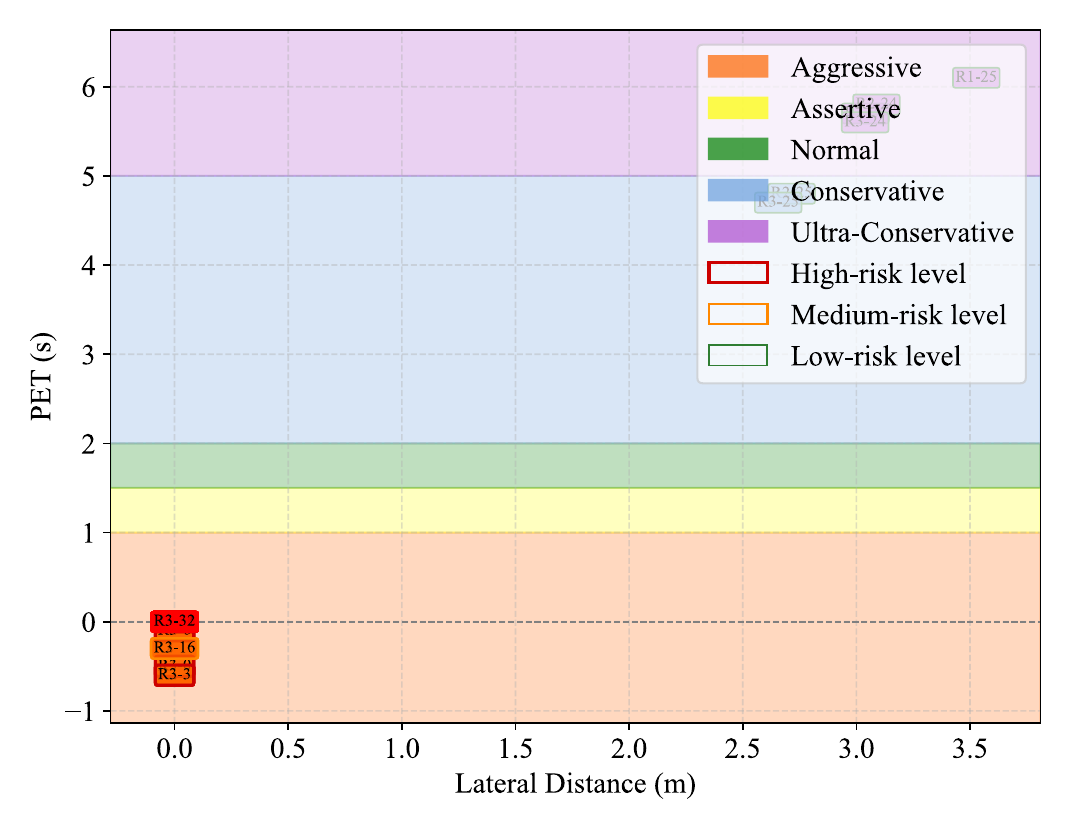}%
    }\\[4pt]
    \subfloat[Reaction time vs.\ Initial TTC.\label{fig:vwc_reaction}]{%
        \includegraphics[width=0.48\textwidth]{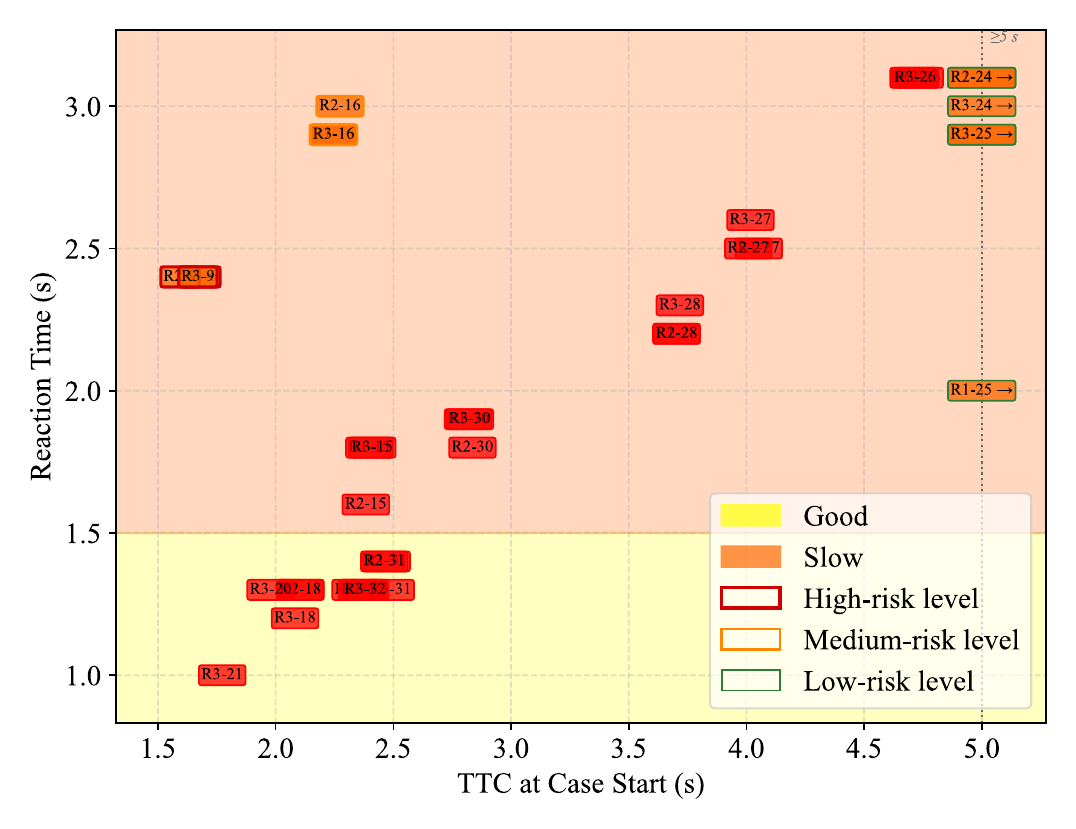}%
    }
    \hfill
    \subfloat[Normalized Deceleration vs.\ required deceleration.\label{fig:vwc_decel}]{%
        \includegraphics[width=0.48\textwidth]{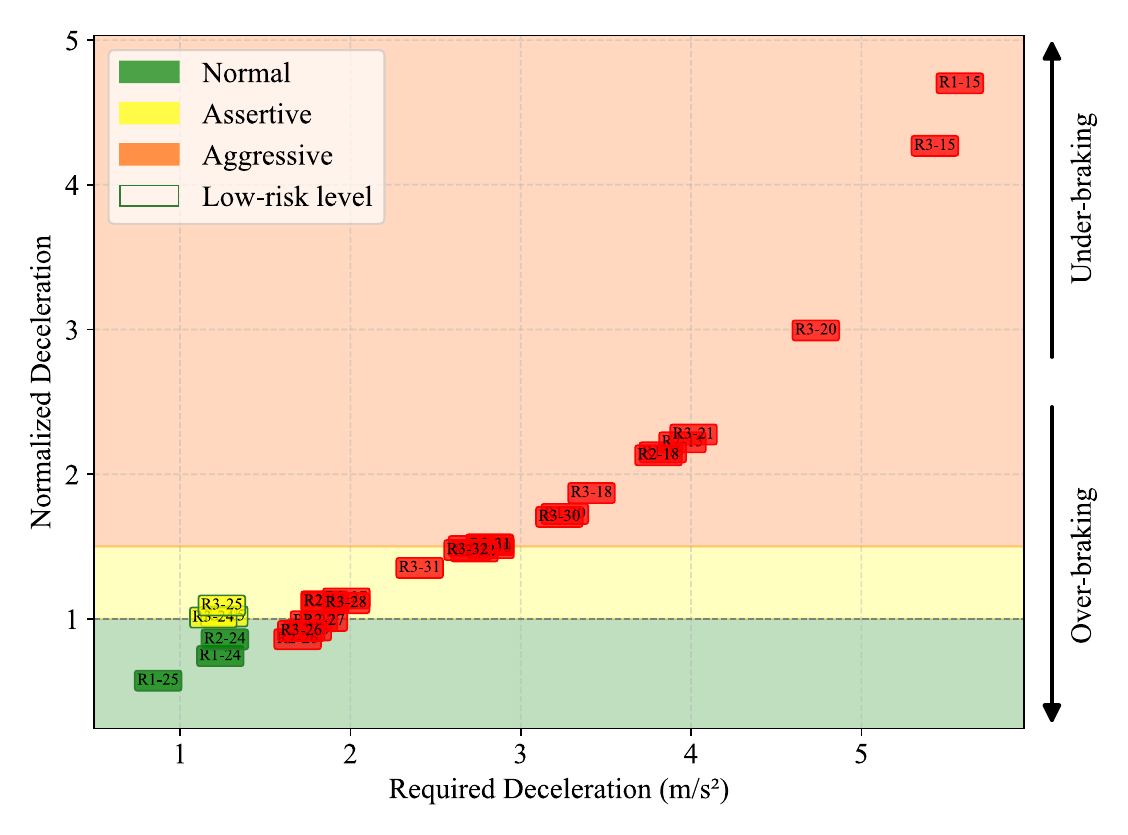}%
    }
    \caption{Analysis results for the VRU crossing without crosswalk competency.}
    \label{fig:vwc_all}
\end{figure*}

\subsubsection*{Competency Certification}

Autoware.Universe \textbf{fails} this competency: 78 of 99 runs ($78.8\%$) result in collision ($n = 33$).
Collisions occur in 33 of 45 high-risk, 24 of 27 medium-risk, and 21 of 27 low-risk runs.
The 21 surviving runs come from seven cases: high-risk cases 0, 3, 6, and 9, medium-risk case 16, and low-risk cases 24 and 25.
In the first five the AV reaches and passes the crossing point ahead of the VRU. In cases 24 and 25 the AV yields to the VRU.

\subsubsection*{Behavioral Cautiousness Assessment}

\textbf{Stage 1: Normalized Minimum TTC.}
The scorecard (Fig.~\ref{fig:vwc_scorecard}) shows the 78 collisions and the 21 survivors.
For Normalized Minimum TTC, 6 survivors are Aggressive, 5 Assertive, and 10 Normal.
In the scatter (Fig.~\ref{fig:vwc_stage1}), the five AV-first cases (four high-risk, one medium-risk, Initial TTC $1.6$--$2.3\,\text{s}$) lie at $0.81$--$1.22$, around the boundary between Normal and Assertive at 1.
The two low-risk yielding cases 24 and 25 (Initial TTC $5.0$--$5.3\,\text{s}$) lie in the Aggressive zone at $3.41$--$3.69$, except R1-25 at $1.59$, just above the Aggressive threshold of $1.5$.
All 78 collision runs are set to the ceiling value of 5.

\textbf{Stage 2: Minimum Longitudinal Clearance.}
In the scorecard (Fig.~\ref{fig:vwc_scorecard}), all 15 AV-first survivor runs are Aggressive, and the runs of cases 24 and 25 are Normal or Conservative.
In the scatter (Fig.~\ref{fig:vwc_stage2}), the Minimum Longitudinal Clearance is $0$ for all collisions, with initial clearances spanning $6.6$--$23.7\,\text{m}$.
The five AV-first survivors have negative minimum clearance ($-2.2$ to $-5.8\,\text{m}$, Aggressive) in every run, i.e., the AV had already passed the conflict point when the VRU entered the conflict zone.
Their initial clearances ($7.9$--$11.4\,\text{m}$) lie inside the range of the high- and medium-risk collision cases ($6.6$--$12.1\,\text{m}$).
Case 24 ends at $6.6$--$6.9\,\text{m}$ (Normal). Case 25 ends at $6.2$--$6.3\,\text{m}$ (Normal) in two runs and at $8.4\,\text{m}$ (Conservative) in R1-25.

\textbf{Stage 3: Post-Encroachment Time and Lateral Distance.}
In the scorecard (Fig.~\ref{fig:vwc_scorecard}), all 15 AV-first survivor runs are Aggressive for both PET and Lateral Distance, and the runs of cases 24 and 25 are Conservative or Ultra-Conservative.
Fig.~\ref{fig:vwc_stage3} plots PET against the edge-to-edge Lateral Distance.
The five AV-first survivors have PET between $-0.6$ and $0\,\text{s}$ and an edge-to-edge Lateral Distance of $0\,\text{m}$. Negative PET means the AV cleared the conflict zone before the VRU entered it. R1-6, R2-9, and R3-9 have PET $=0$ and end without collision.
Case 24 has PET $5.6$--$5.8\,\text{s}$ and Lateral Distance $3.04$--$3.09\,\text{m}$ (both Ultra-Conservative) in every run.
Case 25 has PET $4.7$--$4.8\,\text{s}$ with Lateral Distance $2.66$--$2.72\,\text{m}$ (Conservative) in two runs, and PET $6.1\,\text{s}$ with Lateral Distance $3.53\,\text{m}$ (Ultra-Conservative) in R1-25.

\subsubsection*{Behavior Diagnosis}

\textbf{Reaction Time.}
Reaction times (Fig.~\ref{fig:vwc_reaction}) range from $1.0$ to $3.1\,\text{s}$, with 27 runs in the Slow zone ($\geq 1.5\,\text{s}$) and 11 in the Good zone ($0.7$--$1.5\,\text{s}$).
All 11 Good-zone runs end in collision: R1-18, R2-18, R3-18, R3-20, and R3-21 (medium-risk, $1.0$--$1.3\,\text{s}$) and R1-31, R1-32, R2-31, R2-32, R3-31, and R3-32 (low-risk, $1.3$--$1.4\,\text{s}$).
In AV-first survivors 6, 9, and 16, braking onset occurs $2.4$--$3.0\,\text{s}$ after case start, when the AV front is already $3.3$--$4.0\,\text{m}$ past the conflict-zone boundary.
In every run the AV's maximum speed is $5.0\,\text{m/s}$, and the AV is still at $5.0\,\text{m/s}$ at braking onset.

\textbf{Deceleration Adequacy.}
In the deceleration adequacy plot (Fig.~\ref{fig:vwc_decel}), Normalized Deceleration ranges from $0.57$ to $4.70$. No run is Conservative or Ultra-Conservative, 8 are Normal, 11 Assertive, and 13 Aggressive, so 24 of 32 ($75\%$) brake less than the minimum-margin requirement.
Required deceleration spans $0.87$--$5.58\,\text{m/s}^2$ and actual average deceleration $0.09$--$1.88\,\text{m/s}^2$.
Medium-risk collisions 15, 18, 20, and 21 average only $0.09$--$0.25\,\text{m/s}^2$ (Normalized Deceleration $1.87$--$4.70$, Aggressive).
Low-risk collisions 30--32 average $0.50$--$1.00\,\text{m/s}^2$ ($1.35$--$1.73$).
Low-risk collisions 26--28 brake at $1.64$--$1.88\,\text{m/s}^2$ ($0.86$--$1.14$) but start braking $2.2$--$3.1\,\text{s}$ after case start.

In this competency, Autoware.Universe brakes late and too softly for a VRU crossing outside a designated crosswalk.
Braking starts $1.0$--$3.1\,\text{s}$ after case start from $5.0\,\text{m/s}$ and is below the minimum-margin requirement in $75\%$ of runs.
The AV survives only when it reaches the crossing point before the VRU (5 cases, PET $\leq 0\,\text{s}$, Lateral Distance $0\,\text{m}$) or when the initial clearance is large ($25$--$26\,\text{m}$, cases 24 and 25).
By contrast, in the VRU crossing at crosswalk competency (Sect.~\ref{subsec:vru_crosswalk}) the AV's reaction times are $0.6$--$1.5\,\text{s}$ and no collision occurs.

\subsection{Detect and Respond to VRU Crossing at Crosswalk}
\label{subsec:vru_crosswalk}

\begin{figure*}[!p]
    \centering
    \subfloat[Scorecard.\label{fig:vac_scorecard}]{%
        \includegraphics[width=0.48\textwidth]{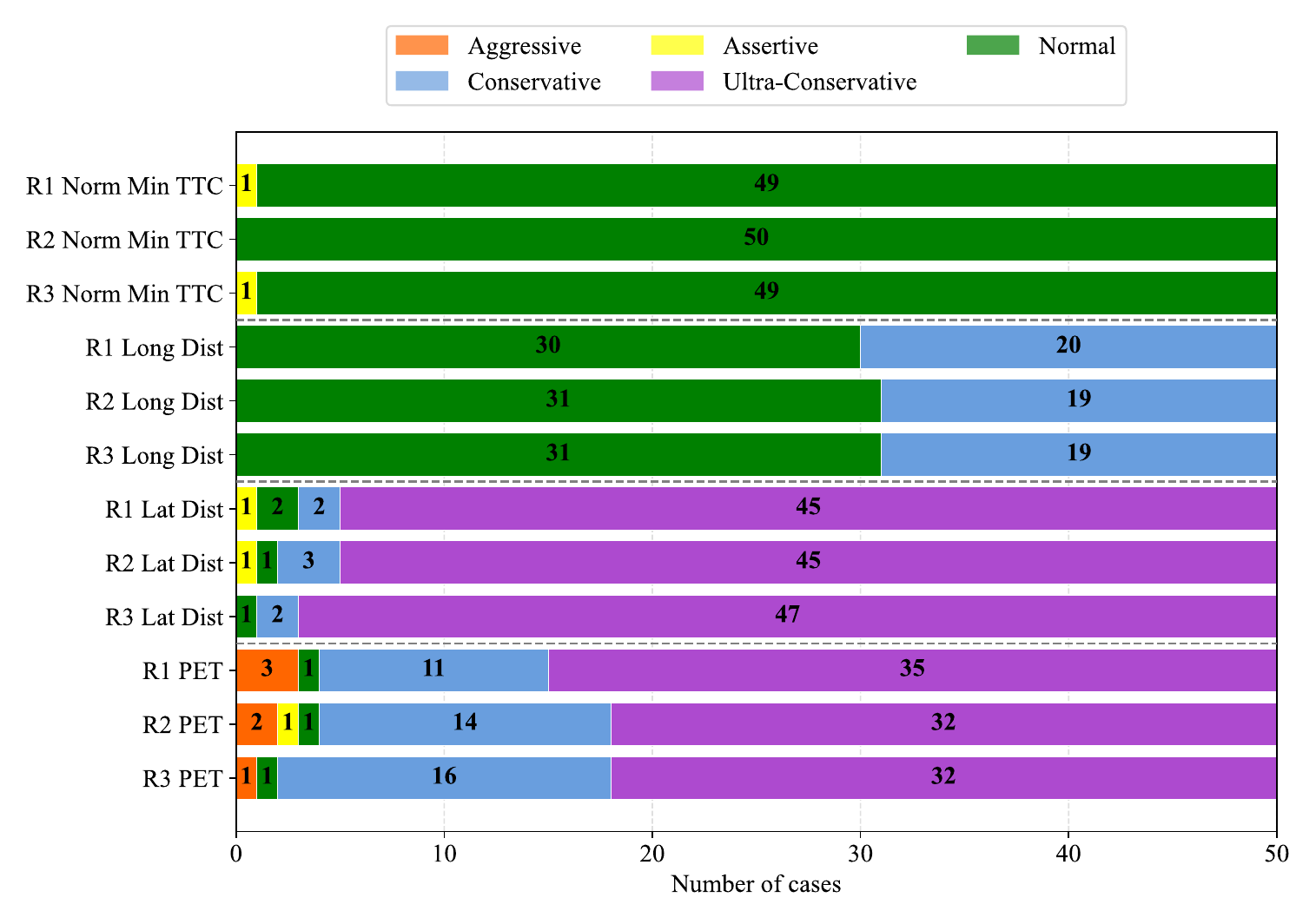}%
    }
    \hfill
    \subfloat[Normalized Minimum TTC vs.\ Initial TTC.\label{fig:vac_stage1}]{%
        \includegraphics[width=0.48\textwidth]{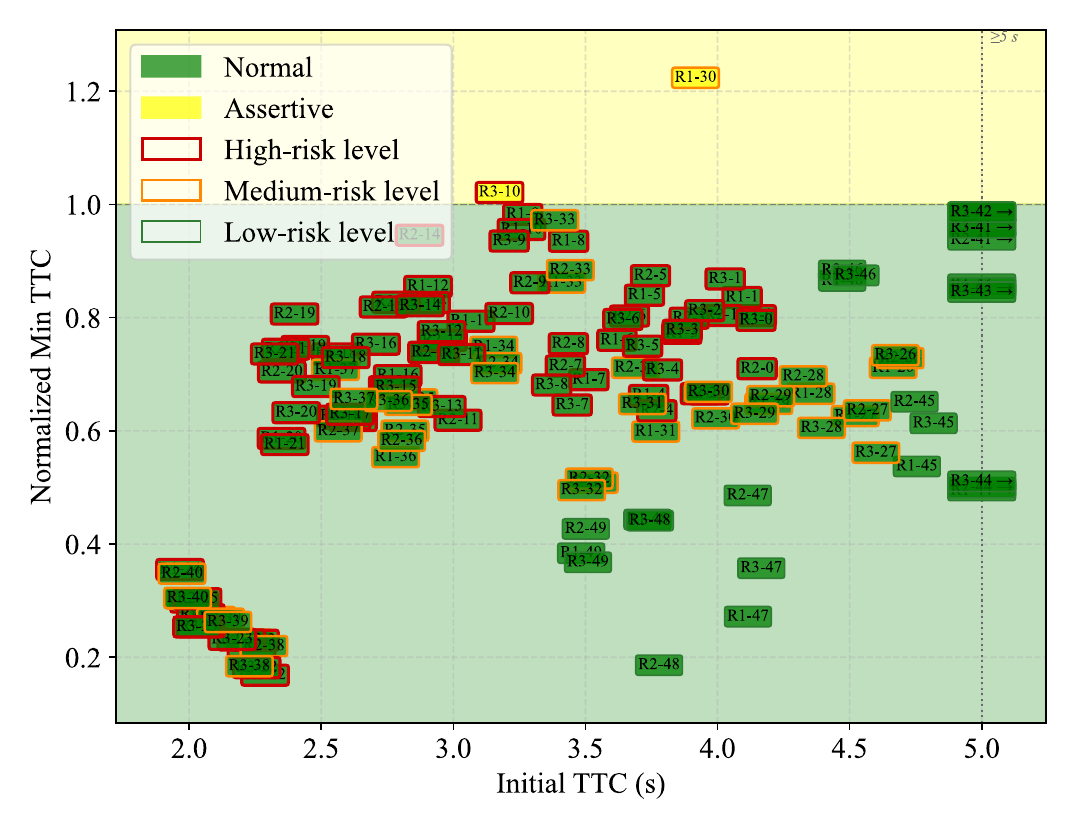}%
    }\\[4pt]
    \subfloat[Minimum Longitudinal Clearance vs.\ Initial Longitudinal Clearance.\label{fig:vac_stage2}]{%
        \includegraphics[width=0.48\textwidth]{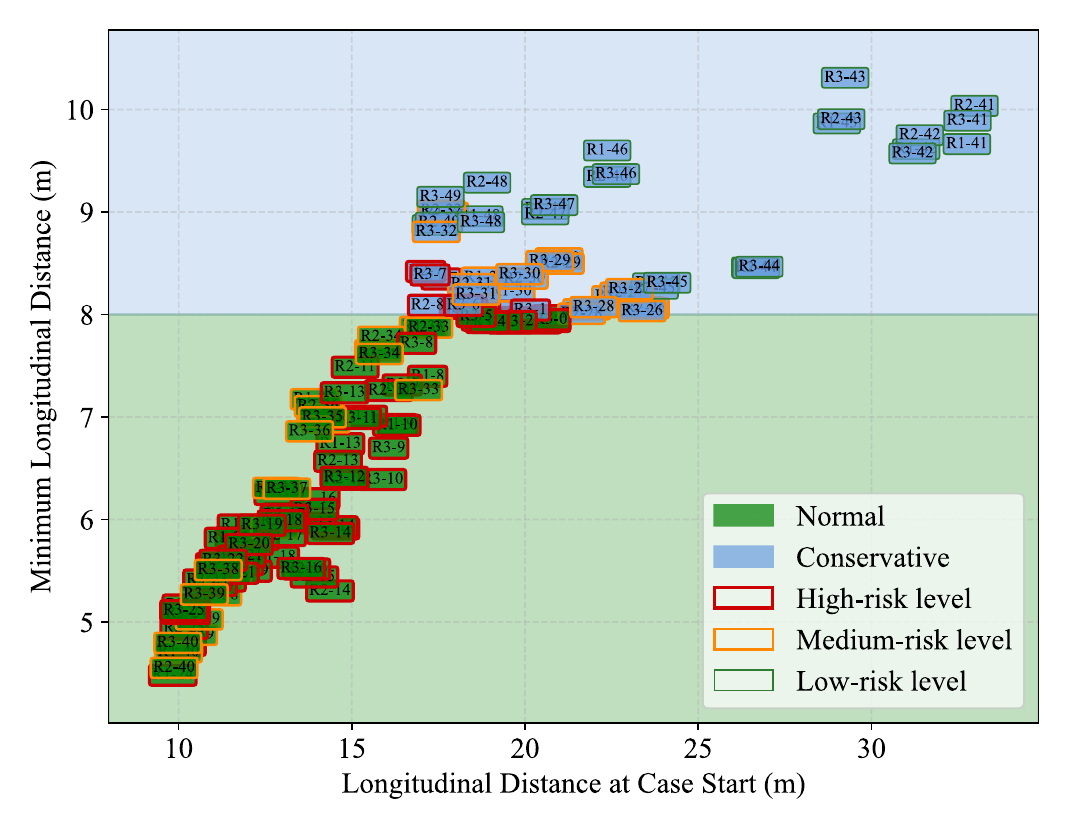}%
    }
    \hfill
    \subfloat[Post-Encroachment Time vs.\ Lateral Distance.\label{fig:vac_stage3}]{%
        \includegraphics[width=0.48\textwidth]{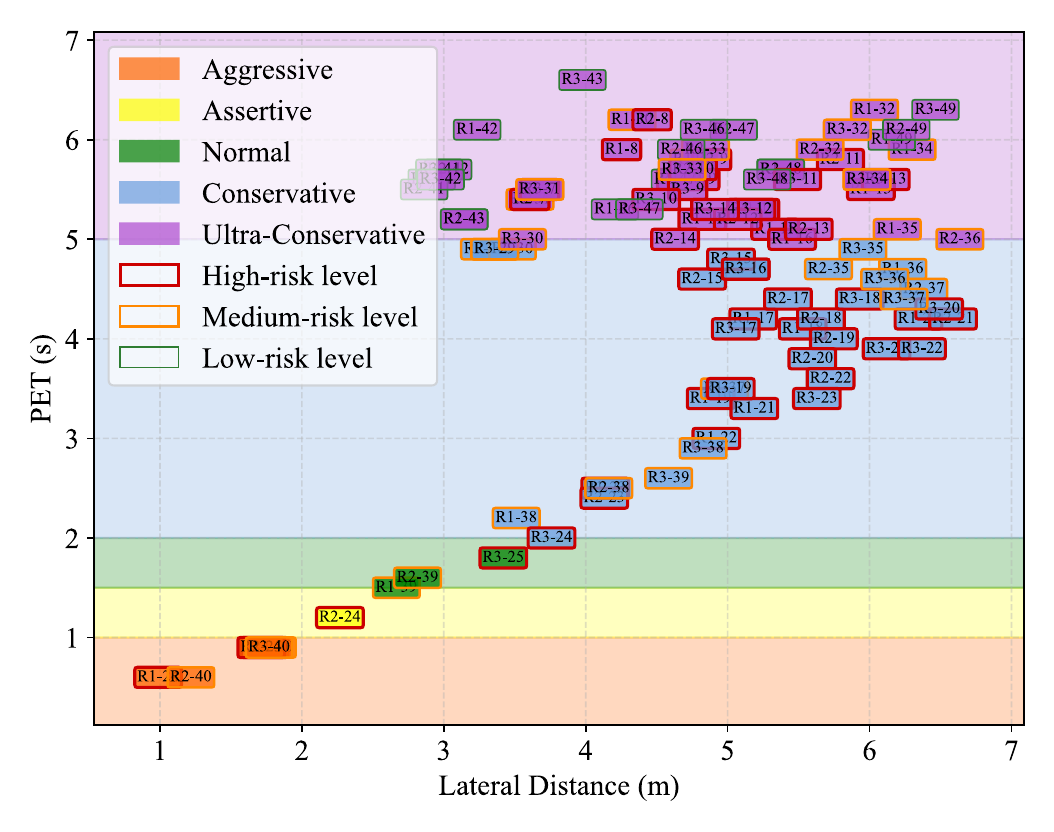}%
    }\\[4pt]
    \subfloat[Reaction time vs.\ Initial TTC.\label{fig:vac_reaction}]{%
        \includegraphics[width=0.48\textwidth]{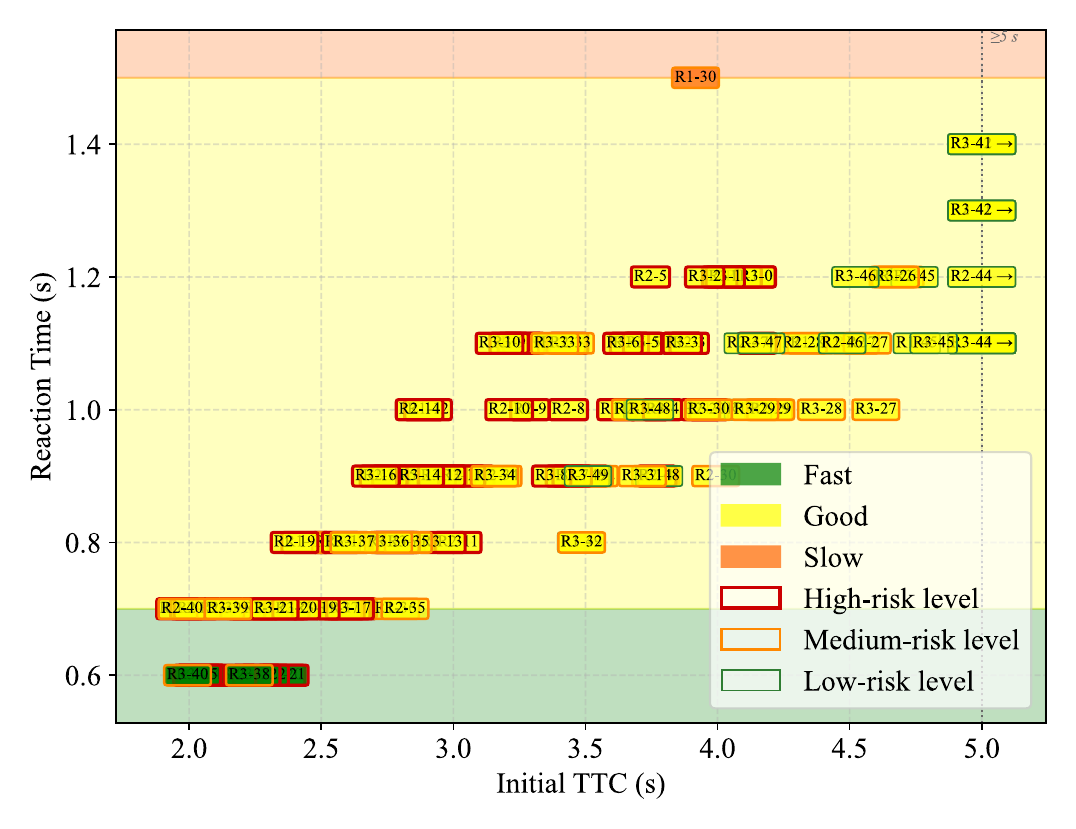}%
    }
    \hfill
    \subfloat[Normalized Deceleration vs.\ required deceleration.\label{fig:vac_decel}]{%
        \includegraphics[width=0.48\textwidth]{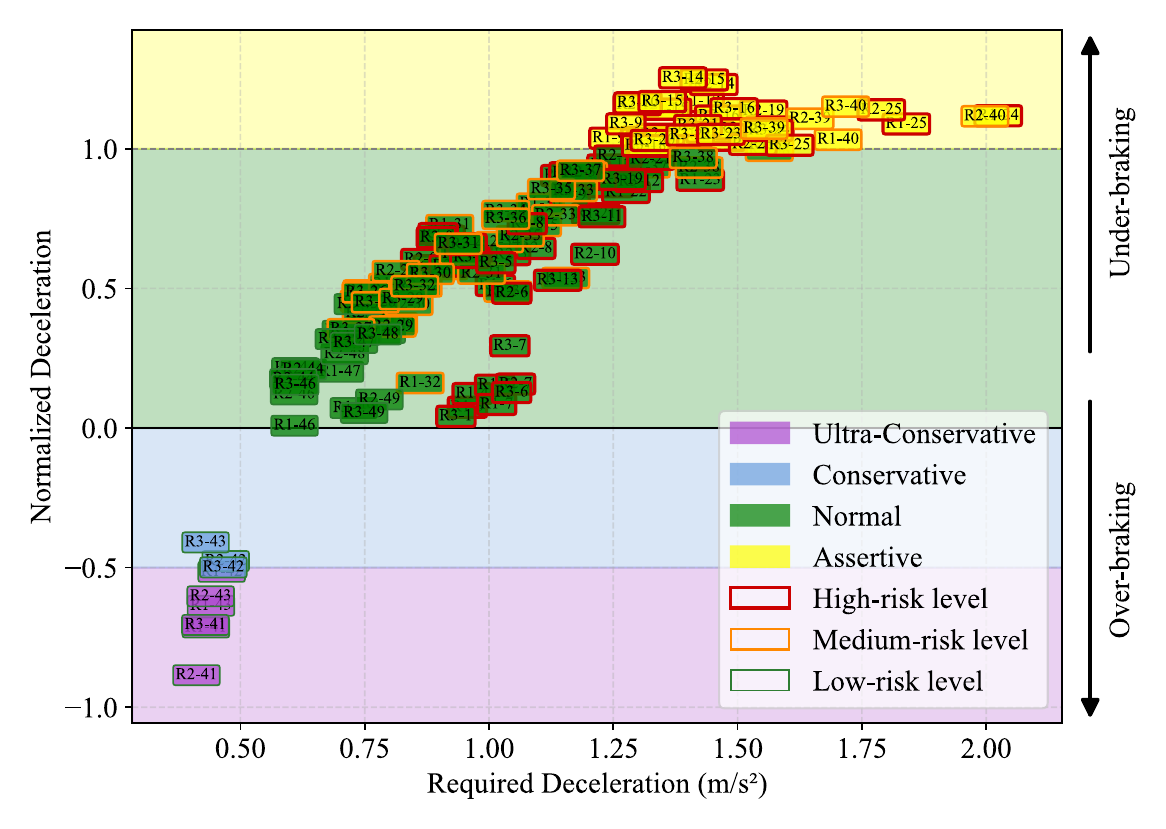}%
    }
    \caption{Analysis results for the VRU crossing at crosswalk competency.}
    \label{fig:vac_all}
\end{figure*}

\subsubsection*{Competency Certification}

Autoware.Universe \textbf{passes} this competency: no collision occurs in any of the 150 runs ($n = 50$ per round, collision rate $= 0\%$).
Avoiding collision does not imply adequate behavior, however.
As detailed below, the AV under-brakes relative to the minimum-margin requirement in $27\%$ of runs (mostly high-risk), and its PET is Ultra-Conservative ($5\,\text{s}$ or more) in 99 of the 150 runs.

\subsubsection*{Behavioral Cautiousness Assessment}

\textbf{Stage 1: Normalized Minimum TTC.}
In the scorecard (Fig.~\ref{fig:vac_scorecard}), Normalized Minimum TTC is concentrated in the Normal zone: 148 of 150 runs are Normal and two are Assertive (R1-30 at $1.22$ and R3-10 at $1.02$). None is Aggressive.
The Normalized Minimum TTC (Fig.~\ref{fig:vac_stage1}) quantifies braking timeliness: a value of~0 corresponds to the minimum TTC of an instantaneous braking response, and a value of~1 to that of a response delayed by a human-typical $1.5\,\text{s}$.
Values between 0 and~1 lie in the Normal zone. Values between 1 and~1.5 are Assertive (worse than the $1.5\,\text{s}$ reference), and values above~1.5 are Aggressive.
In the scatter, all values lie between $0.17$ and $1.22$.

\textbf{Stage 2: Minimum Longitudinal Clearance.}
In the scorecard (Fig.~\ref{fig:vac_scorecard}), no run falls in the Aggressive ($<0.5\,\text{m}$) or Assertive ($0.5$--$3\,\text{m}$) zone.
92 of 150 runs ($61\%$) are Normal ($3$--$8\,\text{m}$) and 58 ($39\%$) Conservative ($8$--$15\,\text{m}$), with none Ultra-Conservative.
In the scatter (Fig.~\ref{fig:vac_stage2}), the Minimum Longitudinal Clearance ranges from $4.48\,\text{m}$ (R1-24) to $10.31\,\text{m}$.
By risk level, high-risk runs are mostly Normal (68 of 78), all 27 low-risk runs are Conservative, and medium-risk runs split 24 Normal to 21 Conservative.
The 11 runs with the lowest clearances ($4.48$--$5.17\,\text{m}$) all under-brake (see Behavior Diagnosis).

\textbf{Stage 3: Post-Encroachment Time and Lateral Distance.}
In the scorecard (Fig.~\ref{fig:vac_scorecard}), PET and Lateral Distance are dominated by the Ultra-Conservative zone, with Ultra-Conservative PET in 99 of 150 runs.
Fig.~\ref{fig:vac_stage3} plots PET against the edge-to-edge Lateral Distance.
In 37 runs, the AV does not enter the conflict zone before the test ends, and these runs are Ultra-Conservative in the PET and Lateral Distance rows.
No PET is negative, i.e., the AV never entered the conflict zone before the VRU had left it.
Among the 113 runs in which the AV enters the conflict zone, 62 are Ultra-Conservative ($5.0$--$6.6\,\text{s}$), 41 Conservative ($2.0$--$4.9\,\text{s}$), 3 Normal (R1-39, R2-39, R3-25), 1 Assertive (R2-24, $1.2\,\text{s}$), and 6 Aggressive ($0.6$--$0.9\,\text{s}$: R1-24, R1-25, R1-40, R2-25, R2-40, R3-40).
In these runs, Lateral Distance ranges from $0.99$ to $6.64\,\text{m}$, and only R1-24 ($0.99\,\text{m}$) and R2-40 ($1.22\,\text{m}$) are below $1.5\,\text{m}$ (Assertive).

\subsubsection*{Behavior Diagnosis}

\textbf{Reaction Time.}
The reaction time plot (Fig.~\ref{fig:vac_reaction}) shows the time from case start to braking onset against the Initial TTC.
The AV brakes in all 150 runs, with reaction times of $0.6$--$1.5\,\text{s}$: 29 runs are Fast ($0.6$--$0.7\,\text{s}$, all high- or medium-risk, Initial TTC $2.0$--$2.8\,\text{s}$), 120 Good, and 1 Slow (R1-30 at $1.5\,\text{s}$).
Reaction time increases with Initial TTC ($r=0.86$).
The AV is at $5.0\,\text{m/s}$ at braking onset in every run.

\textbf{Deceleration Adequacy.}
In the deceleration adequacy plot (Fig.~\ref{fig:vac_decel}), the x-axis shows the deceleration required to stop with a $0.75\,\text{m}$ margin (the minimum-margin, aggressive anchor), computed from the AV's speed and clearance at braking onset. The y-axis shows the actual average deceleration during braking, normalized against the case's own conservative and aggressive anchors.
41 of 150 runs ($27\%$, 35 high-risk and 6 medium-risk) are Assertive, i.e., they brake less than the minimum-margin requirement (Normalized Deceleration $1.0$--$1.25$). 100 runs ($67\%$) are Normal, and none is Aggressive.
The 9 over-braking runs (3 Conservative and 6 Ultra-Conservative) are all low-risk cases 41--43.
Required and actual deceleration are strongly correlated ($r=0.86$, required $0.41$--$2.02\,\text{m/s}^2$, actual $0.64$--$1.65\,\text{m/s}^2$). Braking effort scales with the requirement but stays below it in the highest-requirement runs.
\textbf{R1-24} (high-risk, initial longitudinal clearance $\approx 9.8\,\text{m}$) has the highest required deceleration in the dataset ($a_{\text{req}}\approx2.02\,\text{m/s}^2$). The AV's average deceleration ($\approx1.63\,\text{m/s}^2$) falls short of it, and the run has the smallest minimum longitudinal clearance ($4.48\,\text{m}$, Normal), the lowest PET ($0.6\,\text{s}$, Aggressive), and the smallest Lateral Distance ($0.99\,\text{m}$, Assertive) of all 150 runs. R2-40 is almost identical ($a_{\text{req}}\approx2.00\,\text{m/s}^2$, clearance $4.55\,\text{m}$, PET $0.6\,\text{s}$, Lateral Distance $1.22\,\text{m}$).
In contrast, \textbf{R2-41} (low-risk, initial longitudinal clearance $\approx 33.0\,\text{m}$) has the smallest required deceleration ($a_{\text{req}}\approx0.41\,\text{m/s}^2$). The AV's average deceleration ($\approx0.64\,\text{m/s}^2$) exceeds it (Normalized Deceleration $-0.89$, Ultra-Conservative), and the run ends with a minimum clearance of $10.0\,\text{m}$ (Conservative) and PET $5.5\,\text{s}$.

The AV exceeds the deceleration requirement in low-risk cases and falls short of it in the highest-requirement cases.

\subsection{Merge into Roundabout with Circulating Traffic}
\label{subsec:roundabout_av_outside}

\begin{figure*}[!p]
    \centering
    \subfloat[Scorecard.\label{fig:rdo_scorecard}]{%
        \includegraphics[width=0.48\textwidth]{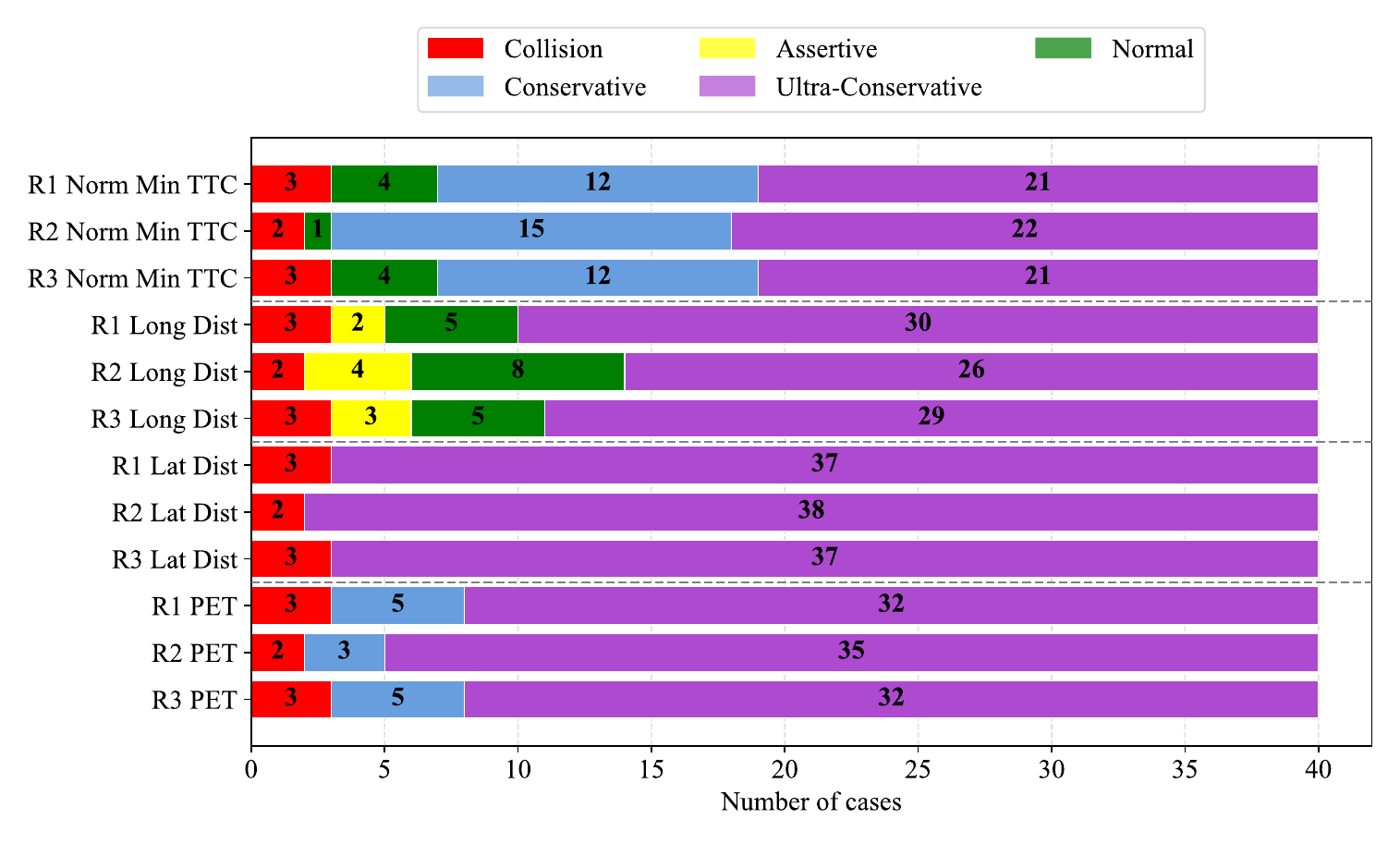}%
    }
    \hfill
    \subfloat[Normalized Minimum TTC vs.\ TTC at Init.\label{fig:rdo_stage1}]{%
        \includegraphics[width=0.48\textwidth]{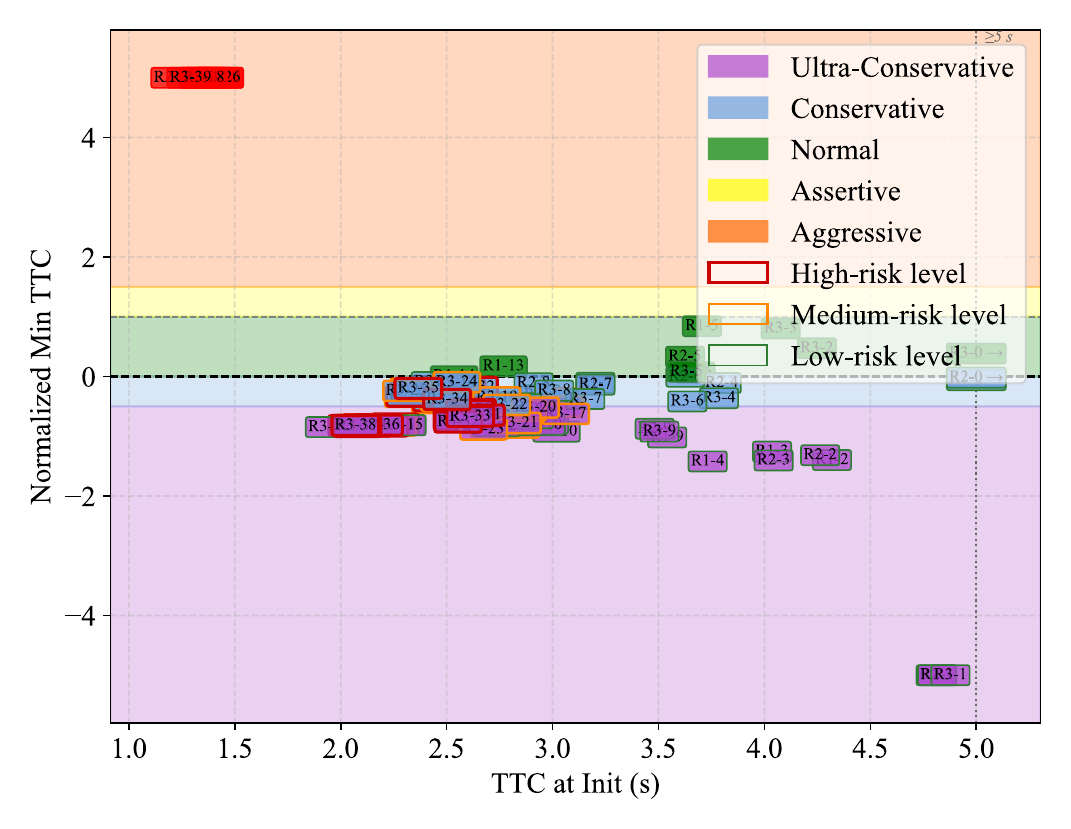}%
    }\\[4pt]
    \subfloat[Minimum Longitudinal Distance vs.\ Longitudinal Distance at Case Start.\label{fig:rdo_stage2}]{%
        \includegraphics[width=0.48\textwidth]{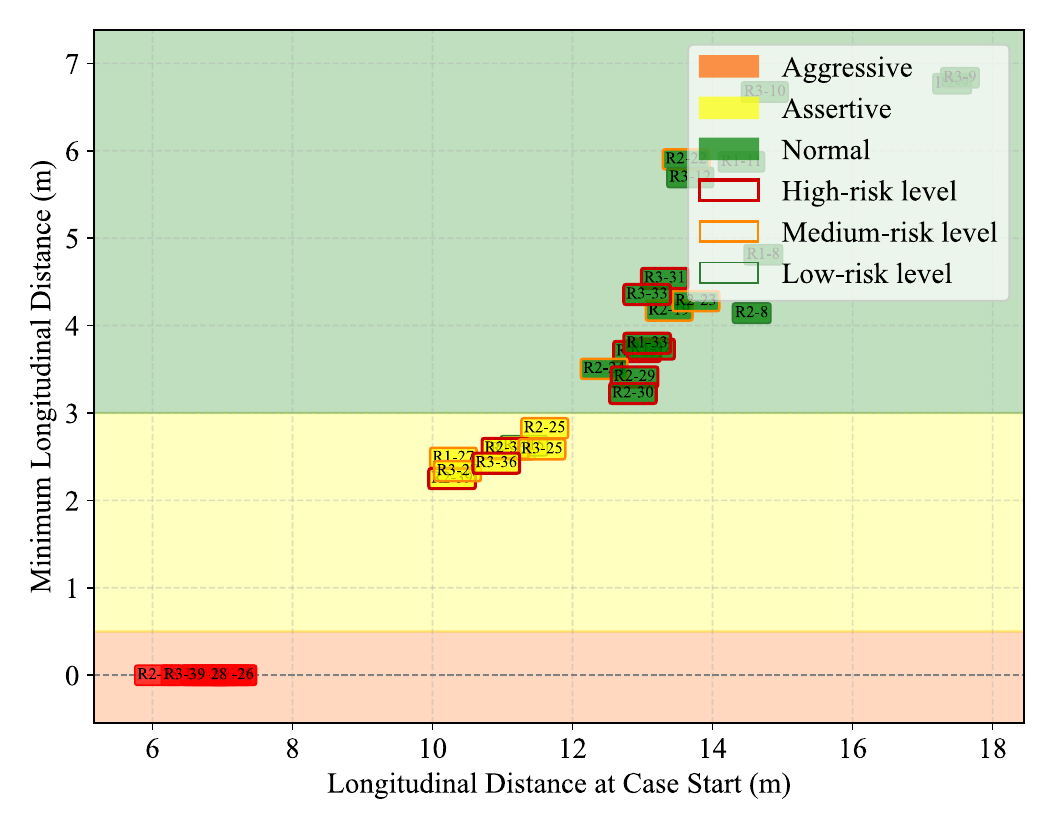}%
    }
    \hfill
    \subfloat[Post-Encroachment Time vs.\ Minimum Longitudinal Distance.\label{fig:rdo_stage3}]{%
        \includegraphics[width=0.48\textwidth]{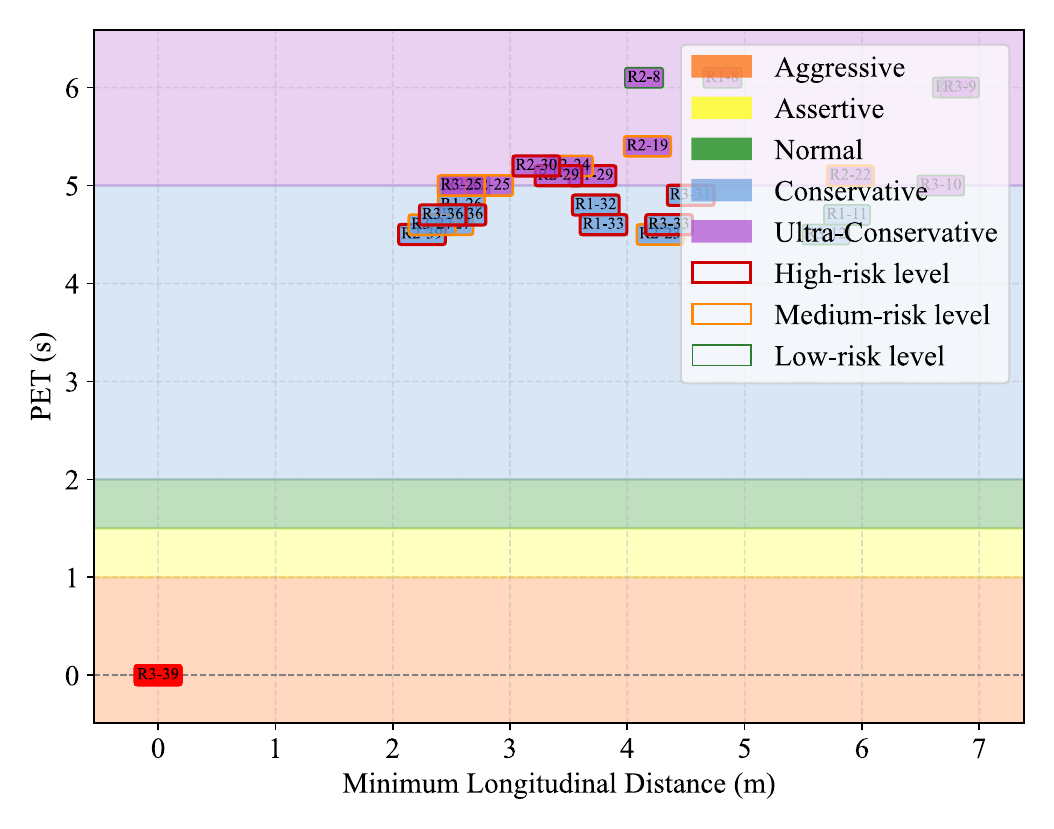}%
    }\\[4pt]
    \subfloat[Reaction Time vs.\ TTC at Init.\label{fig:rdo_reaction}]{%
        \includegraphics[width=0.48\textwidth]{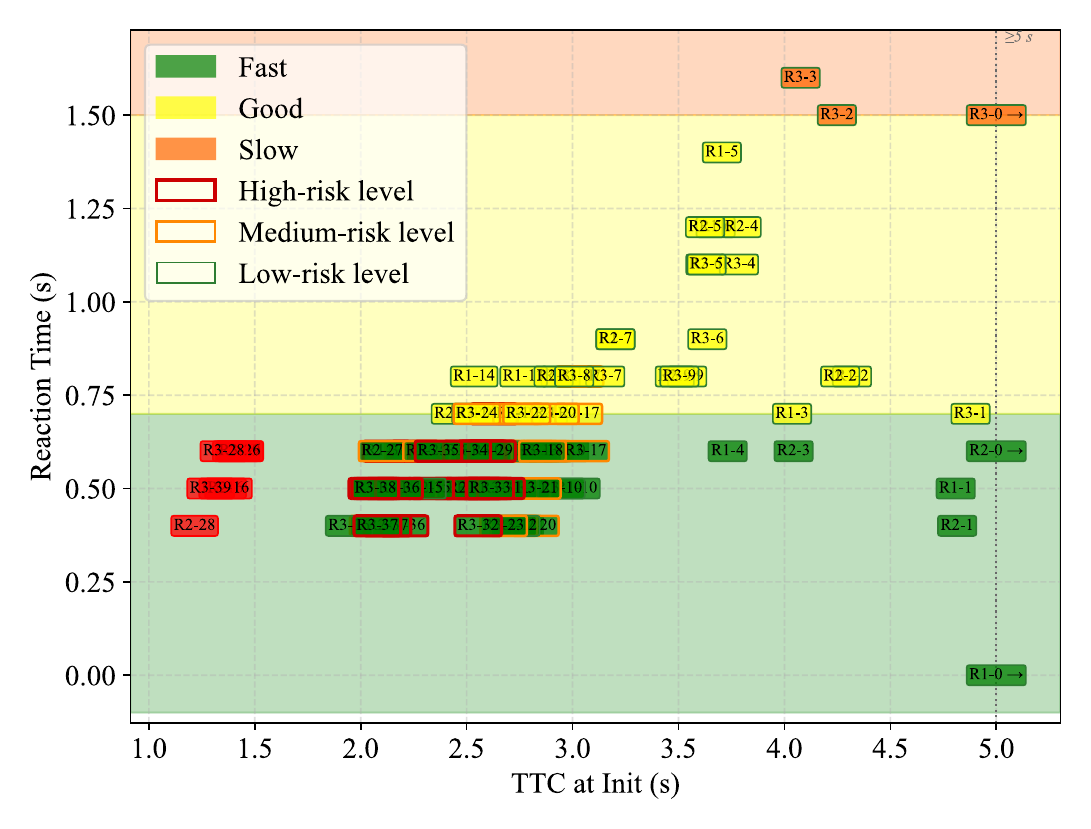}%
    }
    \hfill
    \subfloat[Normalized Deceleration vs.\ Required Deceleration.\label{fig:rdo_decel}]{%
        \includegraphics[width=0.48\textwidth]{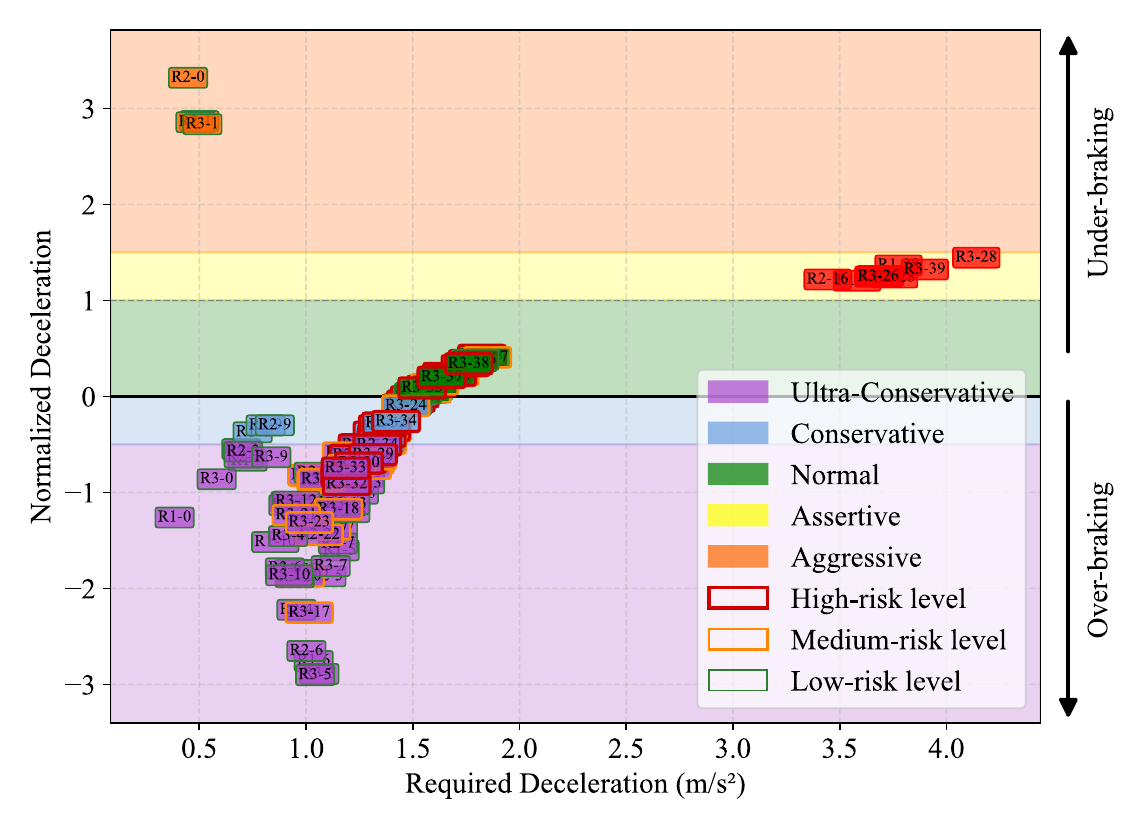}%
    }
    \caption{Analysis results for the merge into roundabout with circulating traffic competency.}
    \label{fig:rdo_all}
\end{figure*}

\subsubsection*{Competency Certification}

Autoware.Universe \textbf{fails} this competency with a collision rate of $6.7\%$ ($8$ of $120$ runs, $n = 40$ per round).
The $8$ collision runs come from designed cases 28 ($3$ runs), 16 and 39 ($2$ runs each), and 26 ($1$ run).

\subsubsection*{Behavioral Cautiousness Assessment}

\textbf{Stage 1: Normalized Minimum TTC.}
In the scorecard (Fig.~\ref{fig:rdo_scorecard}), Normalized Minimum TTC of the $112$ non-collision runs is Ultra-Conservative in $64$ ($57\%$) and Conservative in $39$ ($35\%$), with only $9$ Normal runs and no Assertive or Aggressive run.
The scatter plot (Fig.~\ref{fig:rdo_stage1}) shows the non-collision runs at TTC at Init $1.95$--$5.49\,\text{s}$, mostly between $-1.4$ and $0$ (Ultra-Conservative and Conservative zones).
The $9$ Normal runs lie at $0$--$0.84$.
Designed case 1 sits at the lower clip value ($-5$) in all three of its runs at TTC at Init $4.8$--$4.9\,\text{s}$, with the AV moving at $5\,\text{m/s}$ at case start.
The collision runs start at TTC at Init $1.22$--$1.43\,\text{s}$, below every non-collision run, and are plotted at the value $+5$ (Aggressive) assigned to collisions.

\textbf{Stage 2: Minimum Longitudinal Clearance.}
In $85$ non-collision runs the BV never reaches the conflict zone during the test, and these runs are Ultra-Conservative for Minimum Longitudinal Distance (Fig.~\ref{fig:rdo_scorecard}).
Of the other $27$ non-collision runs, $9$ are Assertive ($2.25$--$2.83\,\text{m}$) and $18$ are Normal ($3.22$--$6.83\,\text{m}$). None is Aggressive or Conservative.
In the scatter plot (Fig.~\ref{fig:rdo_stage2}), these Assertive and Normal runs start $10.3$--$17.5\,\text{m}$ from the conflict zone.
Non-collision runs in which the BV never reaches the conflict zone start $9.7$--$27.5\,\text{m}$ away, and the collision runs start $6.1$--$7.2\,\text{m}$ away.

\textbf{Stage 3: Post-Encroachment Time and Lateral Distance.}
Lateral Distance is Ultra-Conservative for every non-collision run (Fig.~\ref{fig:rdo_scorecard}).
Post-Encroachment Time is Ultra-Conservative for $99$ non-collision runs. This count consists of the $85$ runs in which the BV never covers the conflict point during the test and $14$ runs with PET of $5.0$--$6.1\,\text{s}$.
The remaining $13$ runs are Conservative (PET $4.5$--$4.9\,\text{s}$), and no non-collision run falls in the Normal, Assertive, or Aggressive PET zones.
Every non-collision PET value is positive, i.e.\ the BV always clears the conflict point before the AV reaches it.
The scatter plot (Fig.~\ref{fig:rdo_stage3}) places all non-collision PET values between $4.5$ and $6.1\,\text{s}$, on both sides of the $5\,\text{s}$ Conservative/Ultra-Conservative boundary, across minimum distances of $2.25$--$6.83\,\text{m}$.

\subsubsection*{Behavior Diagnosis}

\textbf{Reaction Time.}
In the reaction time plot (Fig.~\ref{fig:rdo_reaction}), $82$ non-collision runs fall in the Fast zone (${<}0.7\,\text{s}$) and $27$ in the Good zone ($0.7$--$1.5\,\text{s}$).
The other $3$ runs (R3-0, R3-2, and R3-3) are Slow ($1.5$--$1.6\,\text{s}$).
All collision runs react within $0.4$--$0.6\,\text{s}$ (Fast zone).

\textbf{Deceleration Adequacy.}
The deceleration adequacy plot (Fig.~\ref{fig:rdo_decel}) separates the collision runs from the rest.
For non-collision runs the required deceleration is $0.38$--$1.85\,\text{m/s}^2$, and $85$ of $112$ ($76\%$) have a normalized deceleration below $0$ ($66$ Ultra-Conservative and $19$ Conservative), i.e.\ the AV brakes harder than the conservative anchor requires.
The $23$ Normal runs all have a required deceleration of $1.52$--$1.85\,\text{m/s}^2$ (normalized deceleration $0.03$--$0.43$).
The only Aggressive non-collision runs are the three runs of designed case 1 and R2-0 (normalized deceleration $2.83$--$3.32$, average deceleration $0.12$--$0.15\,\text{m/s}^2$ against a required $0.45$--$0.51\,\text{m/s}^2$).
Across non-collision runs the AV's average deceleration has a median of $2.47\,\text{m/s}^2$ (interquartile range $2.37$--$2.57\,\text{m/s}^2$).
The collision runs require $3.44$--$4.14\,\text{m/s}^2$, more than any non-collision run, while their average deceleration is $2.58$--$2.97\,\text{m/s}^2$.
Their normalized deceleration is $1.20$--$1.44$ (Assertive), i.e.\ the AV brakes less than the $0.75\,\text{m}$ minimum-margin anchor requires.

\subsection{Detect and Respond to Vehicle Merging into Roundabout}
\label{subsec:roundabout_av_inside}

\begin{figure*}[!p]
    \centering
    \subfloat[Scorecard.\label{fig:rdi_scorecard}]{%
        \includegraphics[width=0.48\textwidth]{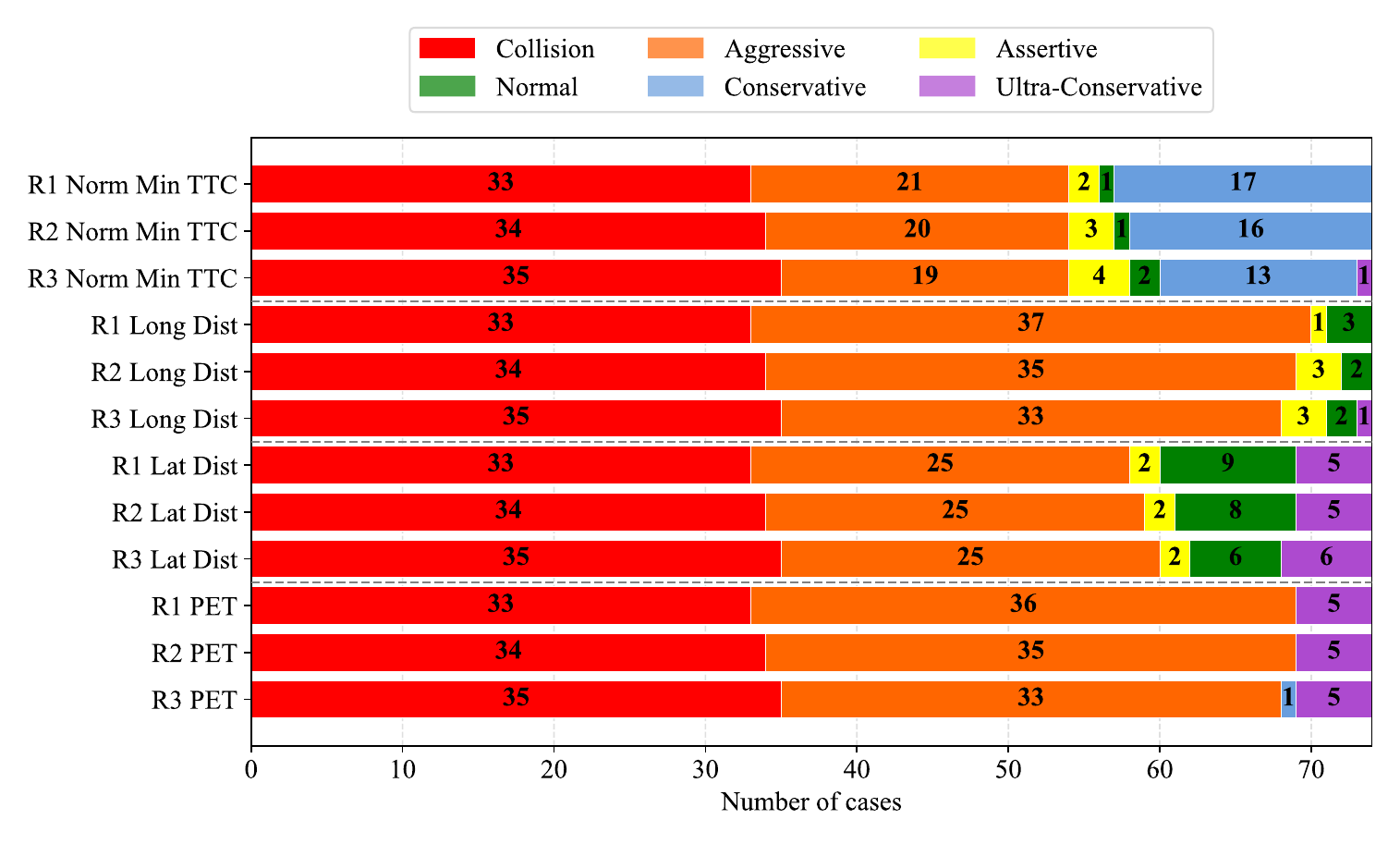}%
    }
    \hfill
    \subfloat[Normalized Minimum TTC vs.\ TTC at Init.\label{fig:rdi_stage1}]{%
        \includegraphics[width=0.48\textwidth]{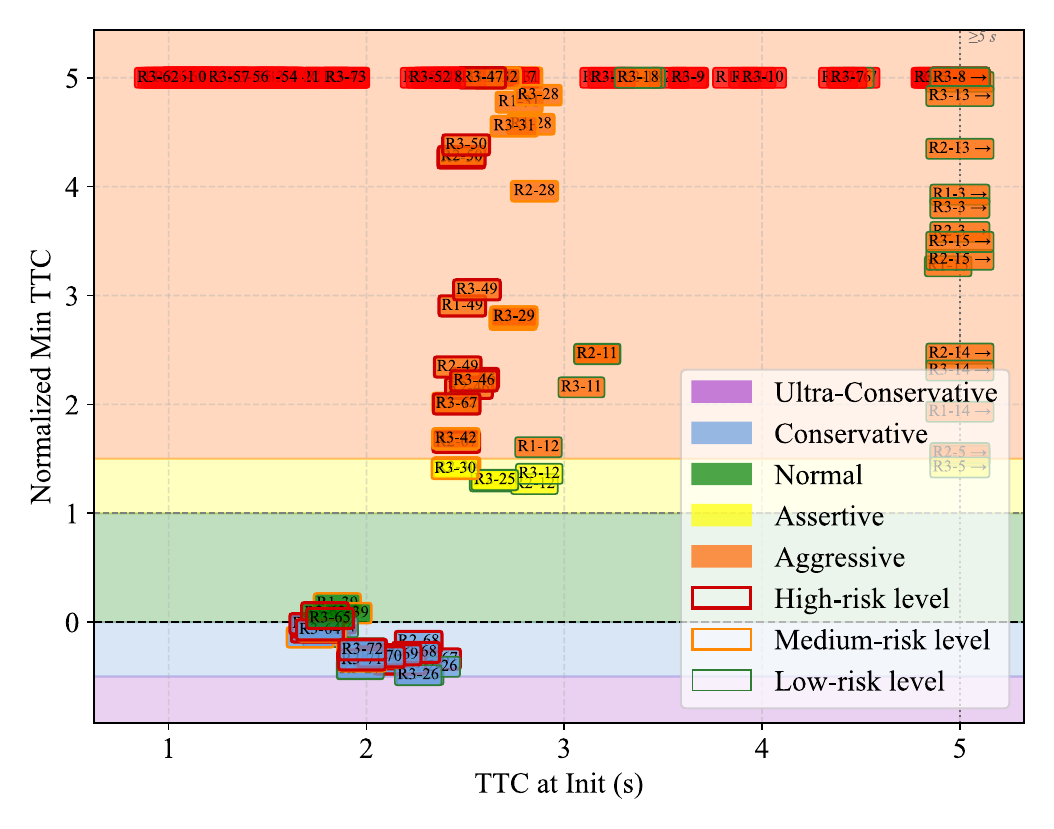}%
    }\\[4pt]
    \subfloat[Minimum Longitudinal Distance vs.\ Longitudinal Distance at Case Start.\label{fig:rdi_stage2}]{%
        \includegraphics[width=0.48\textwidth]{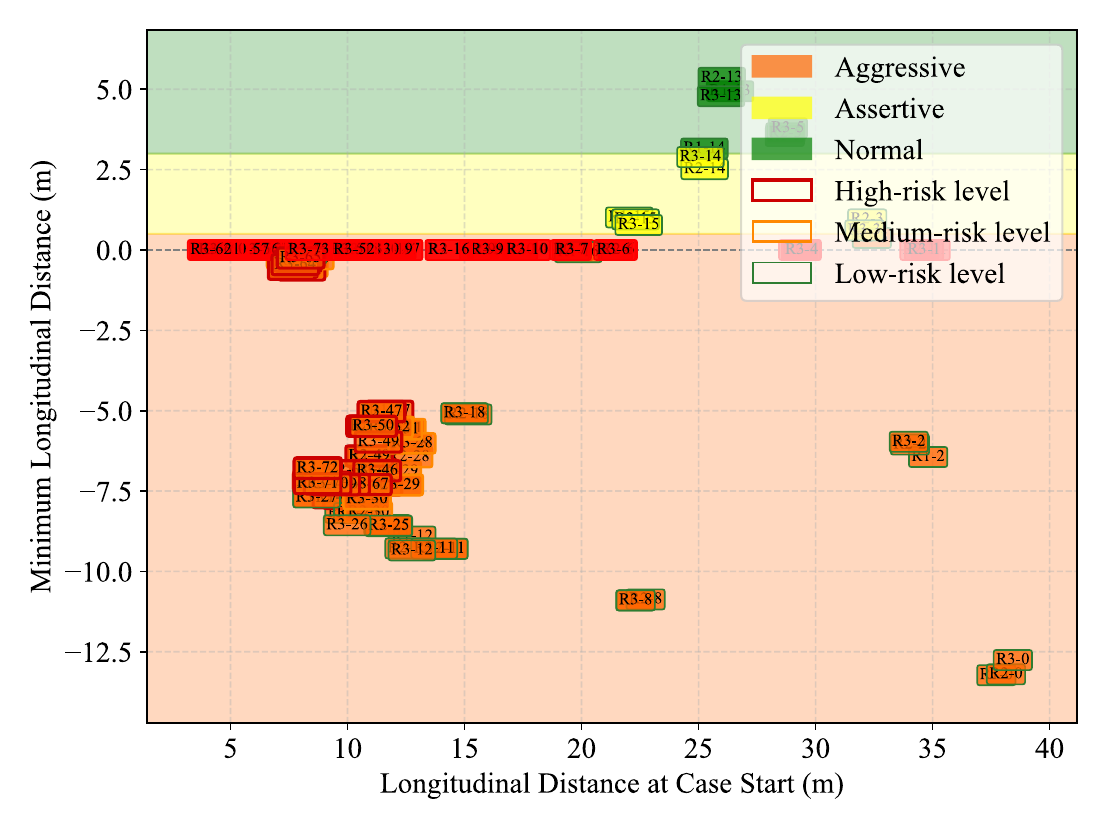}%
    }
    \hfill
    \subfloat[Post-Encroachment Time vs.\ Minimum Longitudinal Distance.\label{fig:rdi_stage3}]{%
        \includegraphics[width=0.48\textwidth]{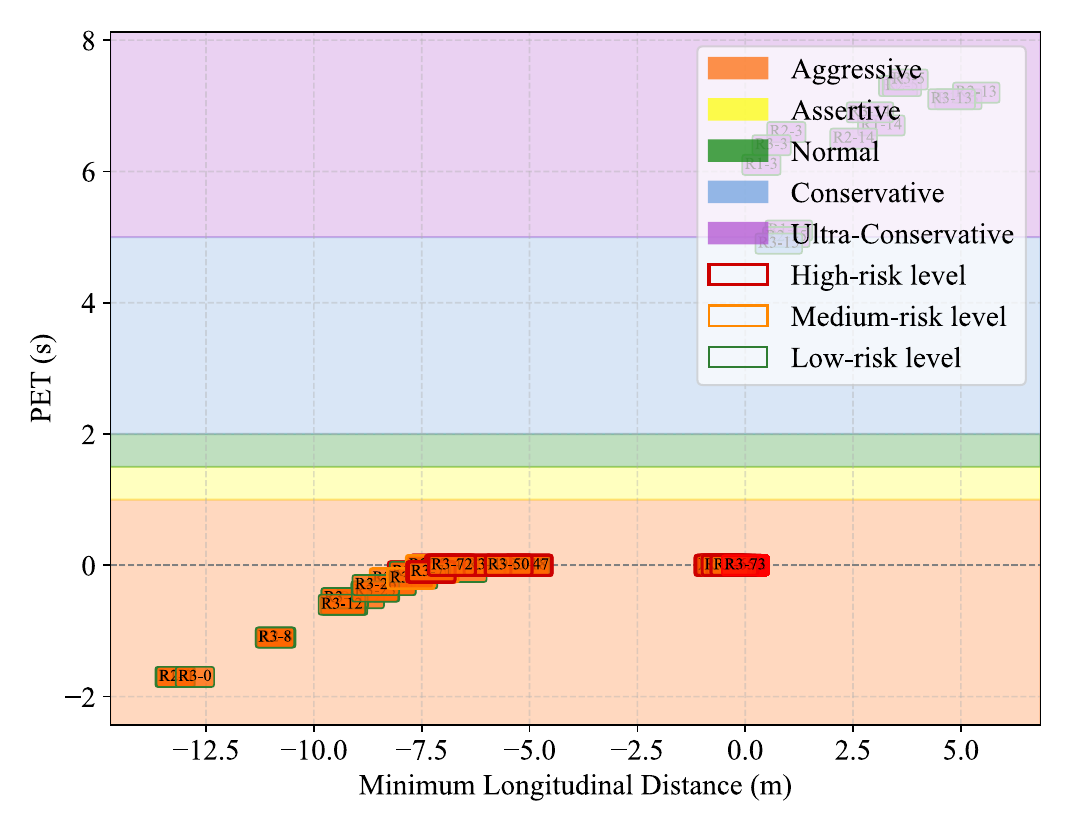}%
    }\\[4pt]
    \subfloat[Reaction Time vs.\ TTC at Init.\label{fig:rdi_reaction}]{%
        \includegraphics[width=0.48\textwidth]{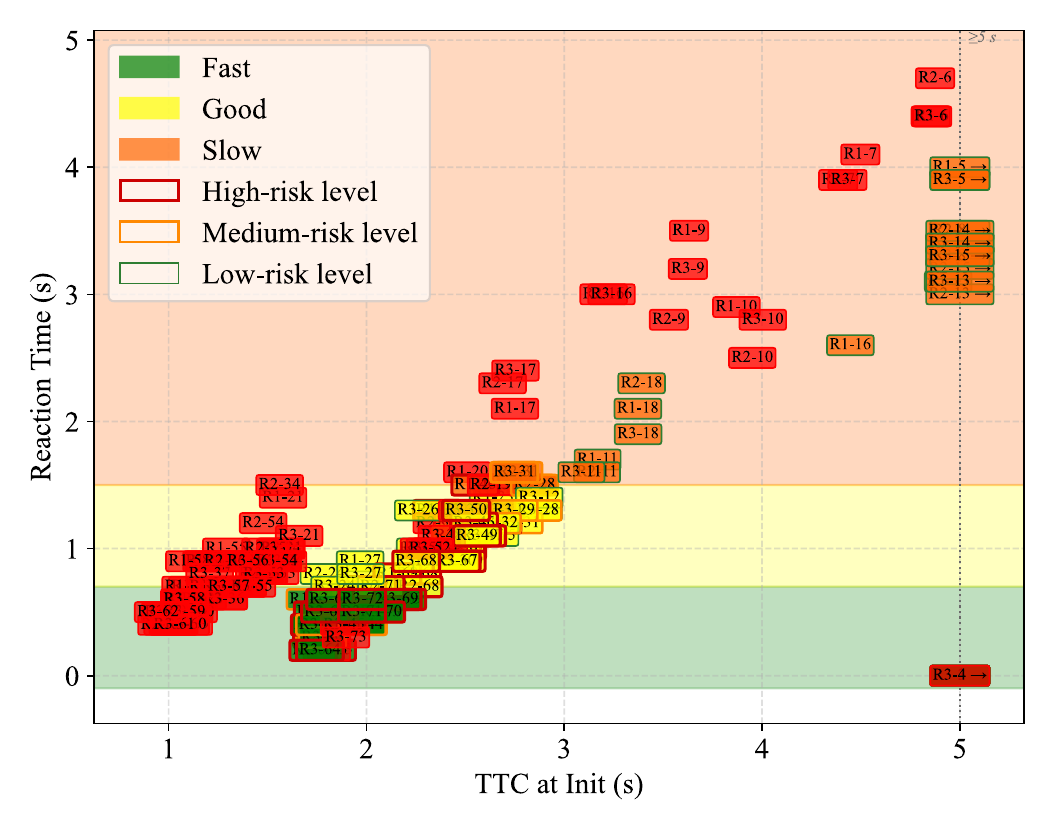}%
    }
    \hfill
    \subfloat[Normalized Deceleration vs.\ Required Deceleration.\label{fig:rdi_decel}]{%
        \includegraphics[width=0.48\textwidth]{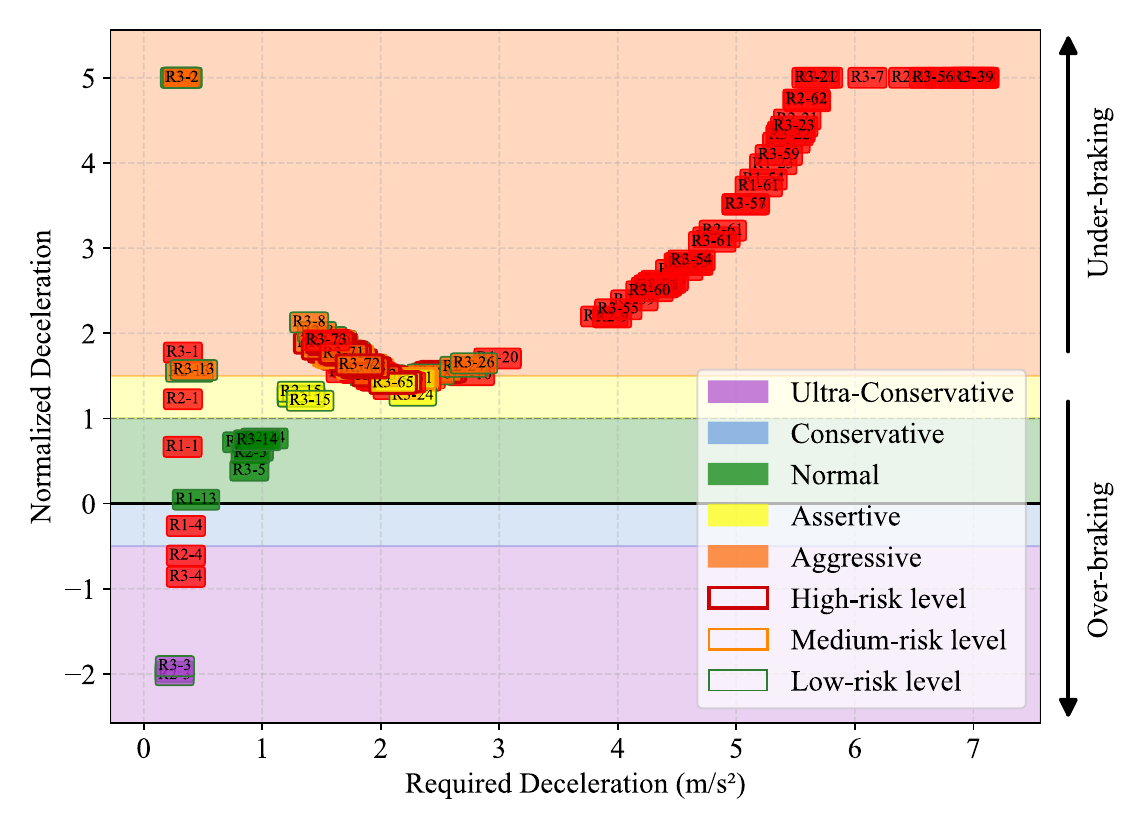}%
    }
    \caption{Analysis results for the detect and respond to vehicle merging into roundabout competency.}
    \label{fig:rdi_all}
\end{figure*}

\subsubsection*{Competency Certification}

Autoware.Universe \textbf{fails} this competency with $45.9\%$ of runs resulting in collision ($102$ of $222$, $n = 74$ per round), the fourth-highest collision rate across all 14 competencies.

\subsubsection*{Behavioral Cautiousness Assessment}

\textbf{Stage 1: Normalized Minimum TTC.}
In the scorecard (Fig.~\ref{fig:rdi_scorecard}), Normalized Minimum TTC of the $120$ non-collision runs is Aggressive in $60$ ($50\%$) and Conservative in $46$ ($38\%$), with $9$ Assertive, $4$ Normal, and $1$ Ultra-Conservative run.
The $60$ Aggressive non-collision runs involve almost no braking: their average deceleration has a median of $0.03\,\text{m/s}^2$, and only $13$ of the $60$ exceed $0.1\,\text{m/s}^2$.
In the scatter plot (Fig.~\ref{fig:rdi_stage1}), the collision runs sit at the clip value $+5$, which is assigned to all collisions, across TTC at Init $0.95$--$7.4\,\text{s}$.
The Aggressive non-collision runs spread from $1.53$ to the same clip value ($22$ of $60$ runs in $1.5$--$3$, and $20$ at the clip) across TTC at Init $2.45$--$8.0\,\text{s}$.
The Conservative non-collision runs have the shortest TTC at Init among non-collision runs ($1.72$--$2.36\,\text{s}$), and in all of them the AV clears the conflict point first or simultaneously (PET $-0.3$ to $0\,\text{s}$, minimum longitudinal distance $-8.6$ to $-0.06\,\text{m}$).
The $15$ runs in which the AV yields (designed cases 3, 5, 13, 14, and 15, PET $4.9$--$7.4\,\text{s}$) have a Normalized Minimum TTC of $1.42$--$4.96$ (Assertive or Aggressive).
The only Ultra-Conservative run is R3-66, in which the AV is stationary at case start.

\textbf{Stage 2: Minimum Longitudinal Clearance.}
Minimum Longitudinal Distance is Aggressive for $105$ non-collision runs (Fig.~\ref{fig:rdi_scorecard}). In $104$ of these the distance is negative ($-13.2$ to $-0.06\,\text{m}$), i.e.\ the AV is already past the conflict-zone boundary when the minimum is reached.
The other non-collision runs are Assertive ($7$), Normal ($7$), or Ultra-Conservative (R3-66). No run is Conservative.
The scatter plot (Fig.~\ref{fig:rdi_stage2}) shows $104$ non-collision runs with negative minimum distance. Of these, $75$ lie between $-10$ and $-5\,\text{m}$, $23$ between $-5$ and $0\,\text{m}$, and $6$ (designed cases 0 and 8) below $-10\,\text{m}$, down to $-13.2\,\text{m}$.
Positive minimum distances ($0.38$--$5.36\,\text{m}$, Aggressive to Normal) occur only for the $15$ yielding runs, which start $22$--$32\,\text{m}$ from the conflict zone.

\textbf{Stage 3: Post-Encroachment Time.}
Post-Encroachment Time is Aggressive for $104$ non-collision runs (Fig.~\ref{fig:rdi_scorecard}). Of these, $47$ have negative PET ($-1.7$ to $-0.1\,\text{s}$, AV clears the conflict point first) and $57$ have PET $=0$ without collision (both vehicles cover the conflict point at overlapping times).
PET is Ultra-Conservative for $15$ runs, the $14$ yielding runs with PET of $5.0\,\text{s}$ or more and R3-66. The fifteenth yielding run (R3-15) is Conservative (PET $=4.9\,\text{s}$).
No non-collision run falls in the Normal or Assertive PET zone. The PET of collision runs is set to $0$.
The scatter plot (Fig.~\ref{fig:rdi_stage3}) shows the non-collision PET values in two ranges, $-1.7$ to $0\,\text{s}$ (Aggressive, at negative minimum distance) and $4.9$ to $7.4\,\text{s}$ (Conservative or Ultra-Conservative, at positive minimum distance), with no run in between.

\subsubsection*{Behavior Diagnosis}

\textbf{Reaction Time.}
The reaction time plot (Fig.~\ref{fig:rdi_reaction}) shows only a weak relationship with initial TTC ($r = 0.37$).
At TTC at Init $1.8$--$2.2\,\text{s}$ the braking onset is $0.2$--$0.9\,\text{s}$ after case start.
Of the $24$ runs with TTC at Init above $5.5\,\text{s}$, $15$ (designed cases 0--4) brake at case start and $9$ (designed cases 5, 13, and 14) at $3.0$--$4.0\,\text{s}$.
Overall, $38\%$ of runs fall in the Fast zone, $39\%$ in the Good zone, and $23\%$ in the Slow zone.
Of the $102$ collision runs, $39$ are Fast, $43$ Good, and $20$ Slow.

\textbf{Deceleration Adequacy.}
Required deceleration among collision runs spans $0.33$ to $446\,\text{m/s}^2$ (median $4.8\,\text{m/s}^2$, $14$ runs above $10\,\text{m/s}^2$, $26\%$ between $1$ and $3\,\text{m/s}^2$), while the average deceleration is ${\leq}0.5\,\text{m/s}^2$ in $96$ of $102$ collision runs ($94\%$).
In the deceleration adequacy plot (Fig.~\ref{fig:rdi_decel}), $65$ collision runs are Aggressive and $8$ Assertive. The remaining four (one Normal, one Conservative, and two Ultra-Conservative) all belong to designed cases 1 and 4, described below.
Among non-collision runs, $90$ are Aggressive and $19$ Assertive.
Normal braking occurs in only $7$ non-collision runs (the three runs each of designed cases 5 and 14, plus R1-13), with required deceleration $0.44$--$1.02\,\text{m/s}^2$ and average deceleration $1.05$--$1.20\,\text{m/s}^2$.
Designed case 3 is Ultra-Conservative in all three of its runs (required $0.26\,\text{m/s}^2$, average $0.48\,\text{m/s}^2$).

\textbf{Cases 1 and 4.}
Designed cases 1 and 4 (both low-risk) collide in all six of their runs by a different mechanism, in which the AV takes priority and passes the conflict point ahead of the challenger but brakes instead of accelerating away, and the challenger hits it from behind.
The two vehicles approach the merge point nearly side by side (center distance $3.5$--$3.9\,\text{m}$), with the challenger at $1.8\,\text{m/s}$ (case 1) or $2.1\,\text{m/s}$ (case 4).
The AV holds $4.2$--$4.5\,\text{m/s}$ until it is $2.7$--$3.4\,\text{m}$ (arc length) short of the conflict point, then brakes with a peak deceleration of $2.5$--$2.6\,\text{m/s}^2$ and drops to $0.5$--$0.6\,\text{m/s}$ within about $2\,\text{s}$, just past the conflict point.
At the time of contact the AV travels at $0.5$--$1.2\,\text{m/s}$, slower than the challenger, whose center is $2.1$--$2.5\,\text{m}$ behind the AV's center along the AV's heading (heading difference $14$--$19^{\circ}$).
At contact, the AV's center is already $0.9$--$2.3\,\text{m}$ past the conflict point.
Normalized deceleration ranges from $0.67$ to $1.77$ for case 1 (Normal, Assertive, and Aggressive) and from $-0.86$ to $-0.26$ for case 4 (Conservative and Ultra-Conservative). The average-based deceleration metric does not flag either case as under-braking in a consistent way, although both end in collision.

In most collisions the AV reacts within $1.5\,\text{s}$ ($82$ of $102$ runs) but brakes at $0.5\,\text{m/s}^2$ or less, regardless of the required deceleration.

\subsection{Detect and Respond to Stopped Vehicle Partially Blocking Lane}
\label{subsec:vehicle_encroachment}

\begin{figure*}[!p]
    \centering
    \subfloat[Stop distance vs.\ encroachment angle.\label{fig:ve_angle}]{%
        \includegraphics[width=0.48\textwidth]{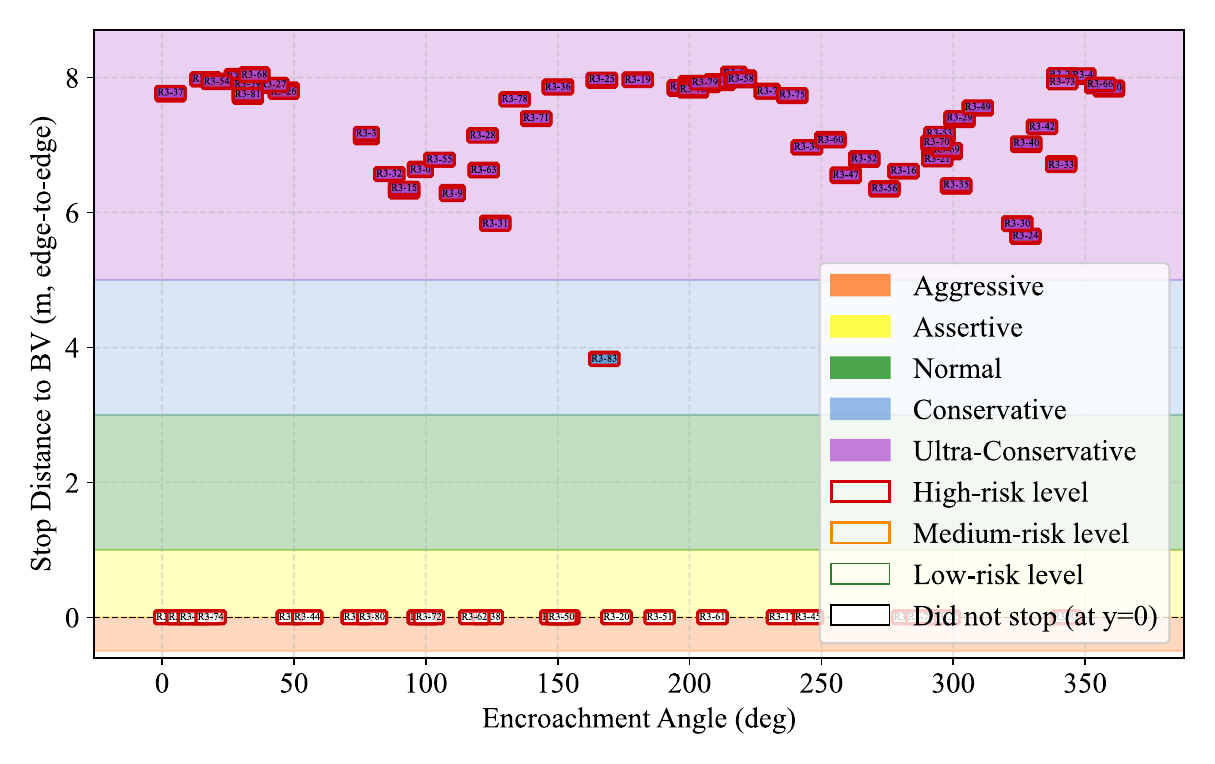}%
    }
    \hfill
    \subfloat[Stop distance vs.\ lateral offset of BV center.\label{fig:ve_offset}]{%
        \includegraphics[width=0.48\textwidth]{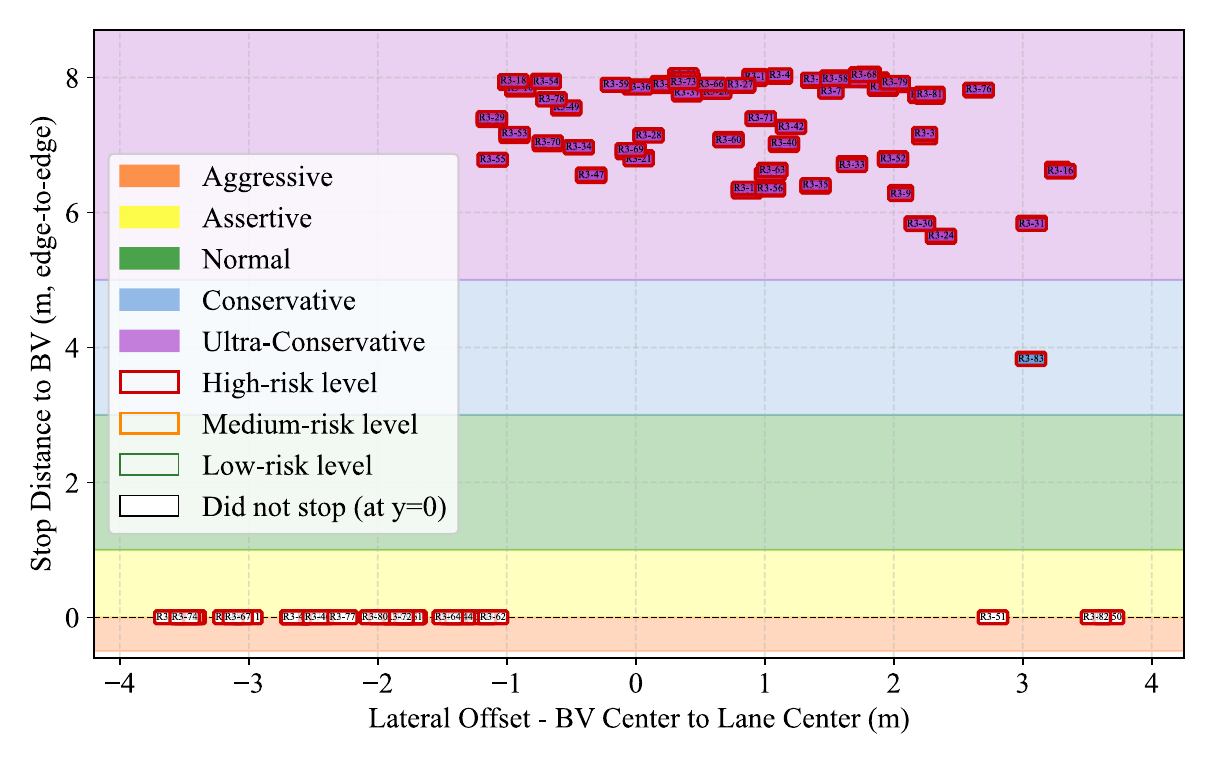}%
    }
    \caption{Analysis results for the detect and respond to stopped vehicle partially blocking lane competency.}
    \label{fig:ve_all}
\end{figure*}

\subsubsection*{Competency Certification}

Autoware.Universe \textbf{passes} this competency with no collision, but its behavior is inconsistent.
In $84$ of the $252$ runs ($33\%$) the AV never stops and passes the vehicle. These runs come from $28$ of the $84$ designed cases, and each of these cases fails to stop in all three of its runs.
None of these runs makes contact. The minimum edge-to-edge clearance during the pass ranges from $0.24$ to $2.75\,\text{m}$, and the tightest is designed case 51 ($0.24$--$0.25\,\text{m}$).
Of the $168$ stopping runs, $165$ ($98\%$) stop in the Ultra-Conservative zone ($5.64$--$8.06\,\text{m}$), and $142$ of them stop at $6$--$8\,\text{m}$.
The other $3$ runs (designed case 83) stop in the Conservative zone at $3.82$--$3.84\,\text{m}$.
No run stops in the Normal ($1$--$3\,\text{m}$), Assertive ($0$--$1\,\text{m}$), or Aggressive zone.
The stop distance of each designed case varies by at most $0.11\,\text{m}$ across its three runs.

\subsubsection*{Behavioral Analysis}

\textbf{Stop Distance and Encroachment Angle.}
The stop-distance versus encroachment angle scatter (Fig.~\ref{fig:ve_angle}) shows two populations: cases that do not stop ($y=0$) and cases that stop at $3.8$--$8.1\,\text{m}$, concentrated at $6$--$8\,\text{m}$, with a sub-cluster at $5.6$--$5.9\,\text{m}$ ($9$ runs) and the single case at $3.8\,\text{m}$.
Both populations span nearly the full $0$--$360^{\circ}$ range of encroachment angles. Per-$45^{\circ}$-octant stop rates vary between $43\%$ and $90\%$ without a monotonic trend.

\textbf{Stop Distance and Lateral Offset.}
The stop-distance versus lateral offset scatter (Fig.~\ref{fig:ve_offset}) covers offsets from $-3.62$ to $+3.67\,\text{m}$ (BV center relative to lane center, negative to the left and positive to the right). The stop decision depends on both the size and the sign of the offset.
The AV stops in all $147$ runs with offsets between $-1.0$ and $+2.5\,\text{m}$.
For offsets to the left beyond $-1.0\,\text{m}$ the stop rate falls to $12\%$ ($-2.5$ to $-1.0\,\text{m}$, $51$ runs) and $0\%$ (below $-2.5\,\text{m}$, $30$ runs).
For offsets to the right beyond $+2.5\,\text{m}$ the stop rate is $62\%$ ($24$ runs).
All stopping runs have offsets between $-1.12$ and $+3.29\,\text{m}$.
The no-stop runs with the BV to the left ($75$ runs) keep a minimum clearance of $1.05$--$2.75\,\text{m}$, while the $9$ no-stop runs with the BV to the right (designed cases at $+2.77$, $+3.57$, and $+3.67\,\text{m}$) keep only $0.24$--$1.46\,\text{m}$.

The lateral offset separates stopping from non-stopping runs much more sharply than the encroachment angle does, and every designed case has the same stop/no-stop outcome in all three of its runs.
When the AV stops, it stops at least $3.8\,\text{m}$ from the BV, and in $98\%$ of stopping runs more than $5\,\text{m}$ away.

\subsection{Detect and Respond to Traffic Light}
\label{subsec:traffic_signal}

\begin{figure*}[!p]
    \centering
    \subfloat[Stop/go decision vs.\ THW at yellow onset.\label{fig:ts_decision}]{%
        \includegraphics[width=0.48\textwidth]{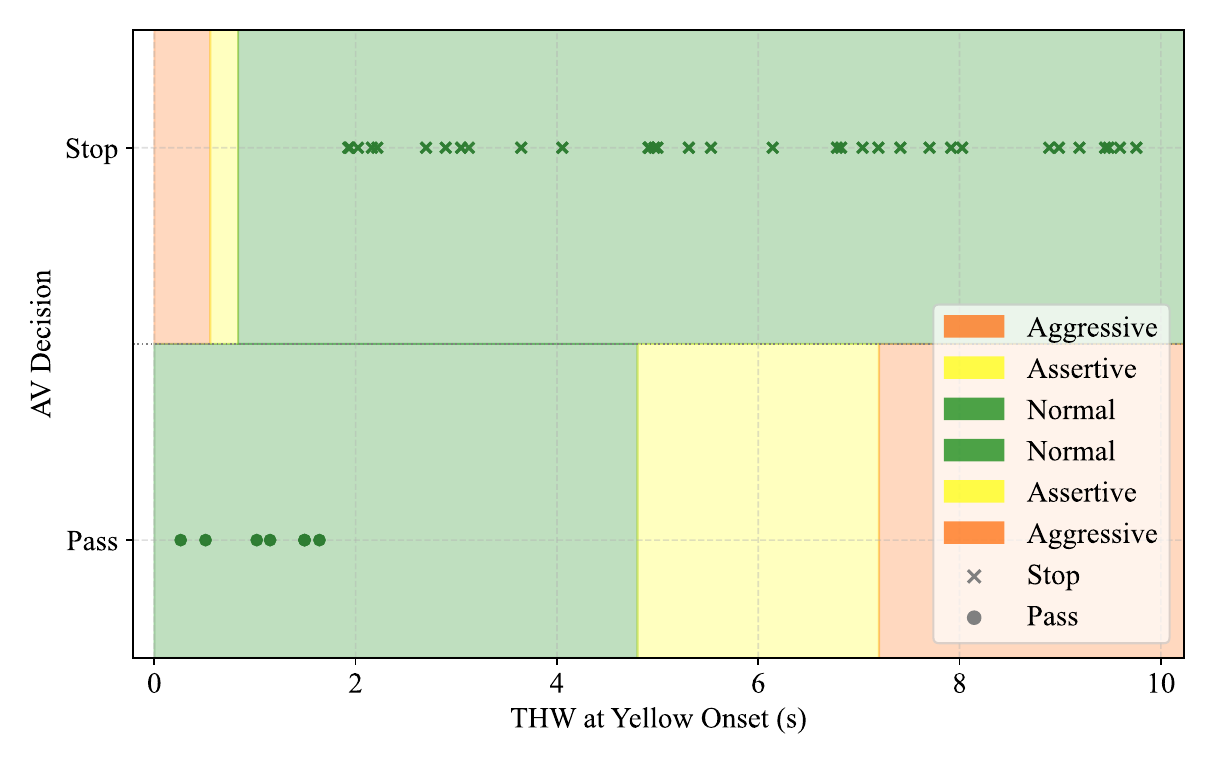}%
    }
    \hfill
    \subfloat[Stop position relative to stop line vs.\ THW at yellow onset.\label{fig:ts_stopdist}]{%
        \includegraphics[width=0.48\textwidth]{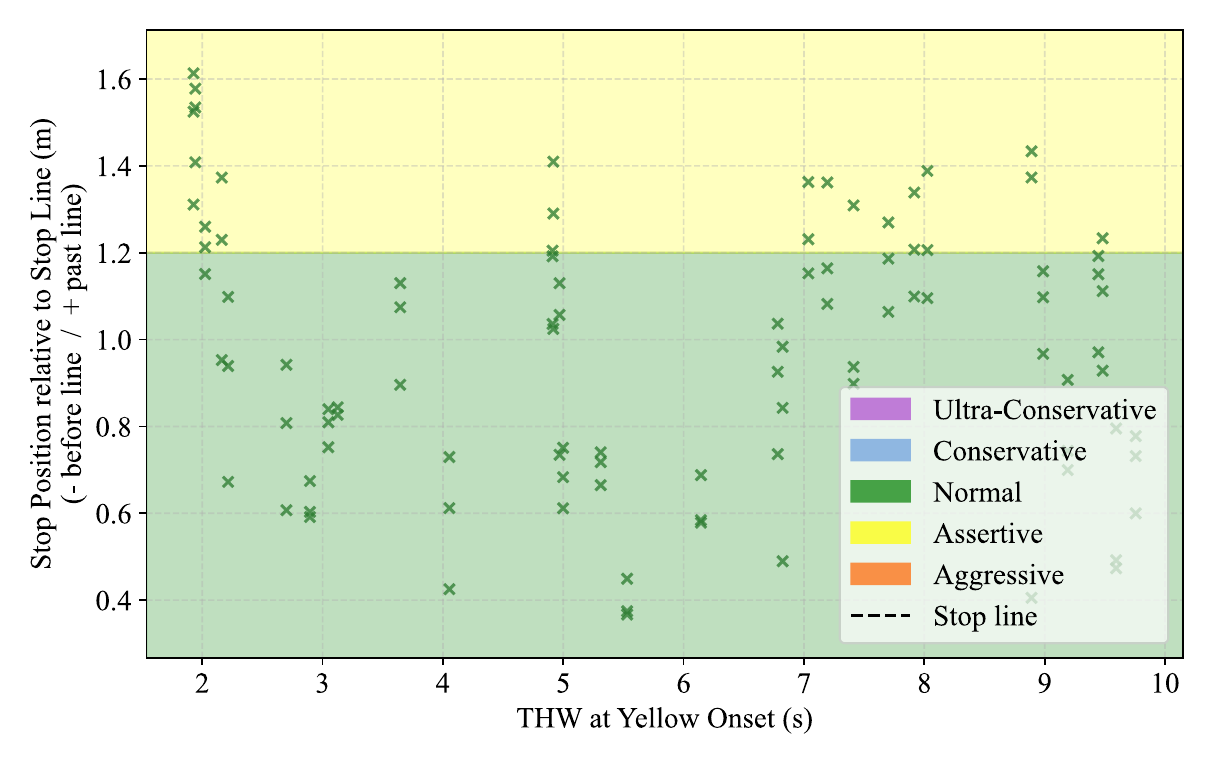}%
    }\\[4pt]
    \subfloat[THW at braking onset vs.\ THW at yellow onset.\label{fig:ts_thw}]{%
        \includegraphics[width=0.48\textwidth]{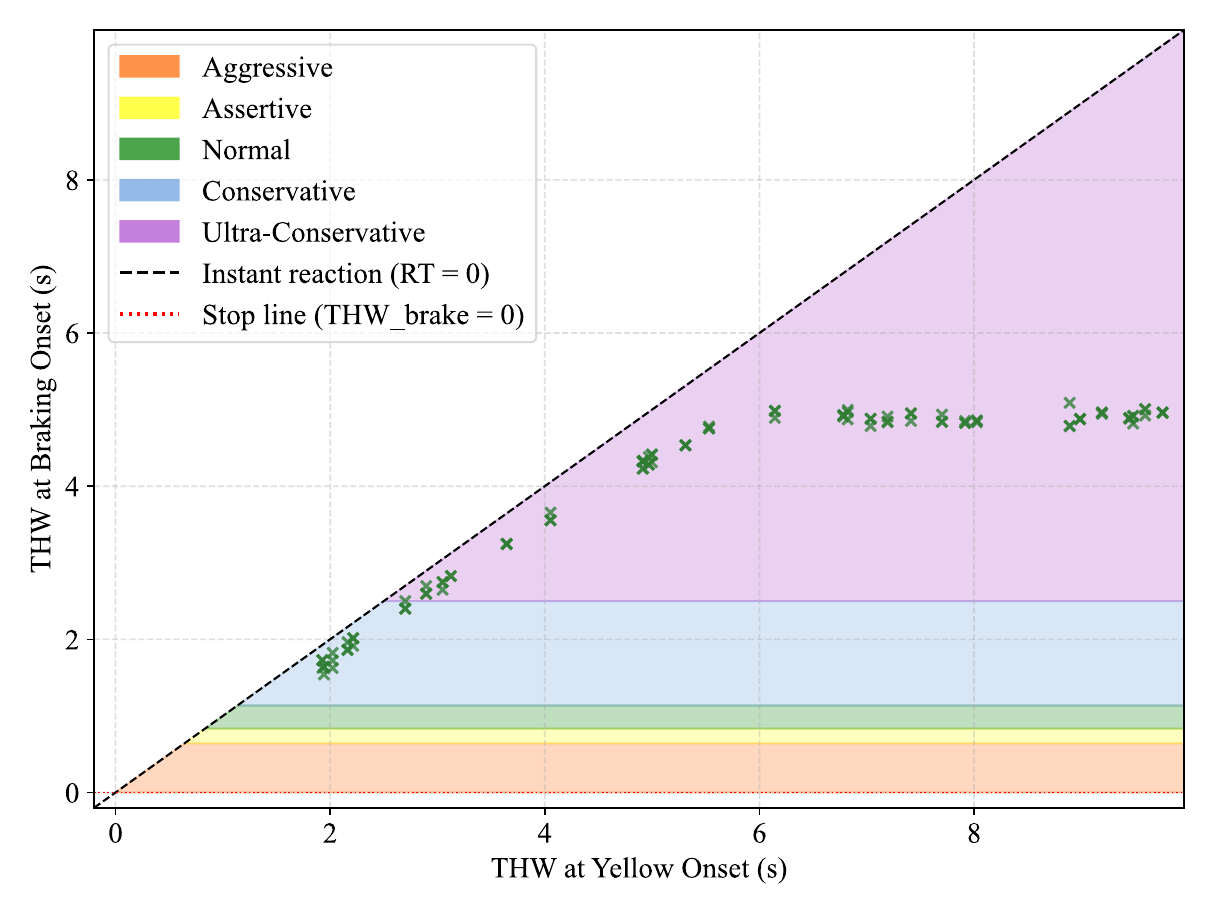}%
    }
    \caption{Analysis results for the detect and respond to traffic light competency.}
    \label{fig:ts_all}
\end{figure*}

\subsubsection*{Competency Certification}

Autoware.Universe \textbf{passes} this competency with no collision in $119$ runs.
All $98$ stop decisions occur at THW $1.93$--$9.76\,\text{s}$ and all $21$ pass decisions at THW $0.26$--$1.64\,\text{s}$, and both sets lie entirely in their Normal decision zones.

\subsubsection*{Behavioral Analysis}

\textbf{Stop/Go Decision.}
The stop/go decision plot (Fig.~\ref{fig:ts_decision}) shows every stop case in the Normal zone (THW ${\geq}0.83\,\text{s}$), i.e.\ stopping required less than $3.0\,\text{m/s}^2$.
Every pass case lies in the Normal pass zone (THW ${<}4.8\,\text{s}$).
The largest pass THW is $1.64\,\text{s}$ (the three runs of one designed case), and the smallest stop THW is $1.93\,\text{s}$. The AV's stop/go decisions are separated at a THW between $1.64$ and $1.93\,\text{s}$.
No decision falls in an Assertive or Aggressive zone.

\textbf{Stop Position.}
The stop position plot (Fig.~\ref{fig:ts_stopdist}) shows that every stop ends past the stop line, by $0.37$--$1.61\,\text{m}$ (median $0.97\,\text{m}$).
Of the $98$ stops, $73$ ($74\%$) are in the Normal zone ($0$--$1.2\,\text{m}$) and $25$ ($26\%$) in the Assertive zone ($1.2$--$3.0\,\text{m}$). None reaches the Aggressive zone (${\geq}3.0\,\text{m}$), and none stops before the line.
Assertive overshoots occur across the THW range ($1.93$--$9.48\,\text{s}$). The six stops with the shortest THW ($1.93$--$1.94\,\text{s}$) are all Assertive ($1.31$--$1.61\,\text{m}$).

\textbf{Braking Onset.}
The braking onset plot (Fig.~\ref{fig:ts_thw}) shows a saturating relationship between THW at yellow onset and THW at braking onset.
For THW at yellow onset of $6.1\,\text{s}$ and above, braking begins at THW $4.78$--$5.09\,\text{s}$ regardless of the initial THW.
Below $6\,\text{s}$, THW at braking onset follows THW at yellow onset, $0.2$--$0.8\,\text{s}$ lower.
In all $98$ stops, braking starts in the Conservative or Ultra-Conservative zone. Of these, $80$ ($82\%$) are Ultra-Conservative (THW at braking onset $2.60$--$5.09\,\text{s}$), i.e.\ the AV starts braking when stopping requires less than $1.0\,\text{m/s}^2$.
The other $18$ ($18\%$), at THW at yellow onset $1.93$--$2.70\,\text{s}$, are Conservative (THW at braking onset $1.54$--$2.50\,\text{s}$).
No braking onset falls in the Normal, Assertive, or Aggressive zone.

\section*{Acknowledgments}
This research was partially funded by the U.S. Department of Transportation (USDOT) Automated Driving System Demonstration Grant (\#693JJ319NF00001), USDOT Region 5 University Transportation Center: Center for Connected and Automated Transportation (CCAT) of the University of Michigan (\#69A3551747105), and the National Science Foundation (CMMI \#2223517). Any opinions, findings, conclusions, or recommendations expressed in this material are those of the authors and do not necessarily reflect the official policy or position of the U.S. government. 

We thank the late Prof. Huei Peng of the University of Michigan and Dr. Tony Geara of the City of Detroit for insightful discussions and thank Jim Lollar and Alex Rozelle of the University of Michigan for supporting field tests at the Mcity Test Facility.

\bibliographystyle{IEEEtran}
\bibliography{ref}

\vfill

\end{document}